%% file: dissertation.tex
\documentclass[12pt]{report}	

\usepackage{utdiss2}  		

\usepackage{amsmath,amsthm,amsfonts,amscd,amssymb} 
\usepackage{eucal} 	 	
\usepackage{verbatim}      	
\usepackage{makeidx}       	
\usepackage{url}		
\usepackage{graphicx}
\usepackage{cancel}
\usepackage{subfigure}
\usepackage{algpseudocode}
\usepackage{algorithm}
\usepackage{multicol}
\graphicspath{{./lidar/}{./auro/}{./appendix/}{./ecmr/}{./iros/}{./tro/}}

\DeclareMathOperator*{\argmin}{argmin}

\author{Kwan Suk Kim}  	

\address{\url{kskim@utexas.edu}}

\title{ Intelligent Collision Management in Dynamic Environments for Human-Centered Robots }

\supervisor
	{Luis Sentis}

\committeemembers
	[Benito Fernandez]
	[Dongmei Chen]
	{Peter Stone}

\previousdegrees{}
\graduationmonth{December}      
\graduationyear{2017}   
\oneandonehalfspacequote

\theoremstyle{definition}

\theoremstyle{remark}

\newcommand{\latexe}{{\LaTeX\kern.125em2%
                      \lower.5ex\hbox{$\varepsilon$}}}

\chardef\bslash=`\\	

\makeatletter		
\def\square{\RIfM@\bgroup\else$\bgroup\aftergroup$\fi
  \vcenter{\hrule\hbox{\vrule\@height.6em\kern.6em\vrule}%
                                              \hrule}\egroup}
\makeatother		

\makeatletter
\def\blfootnote{\xdef\@thefnmark{}\@footnotetext}
\makeatother

\makeindex    

\begin{document}

\copyrightpage          

%
%
%
\commcertpage           

\titlepage              

%


%
\utabstract
\index{Abstract}%
\indent
Thanks to rapid breakthroughs on robotics, their historical deployment in industrial setups, their current extensions to warehouses, and the highly anticipated deployment of autonomous vehicles on our streets, self-guided mobile robots equiped with manipulators or varied payloads are paving their way into unstructured and dynamic environments such as cities, hospitals, human-populated work areas, and facilities of all types. As a result, intentional or unintentional contact between humans, objects and these robots is bound to occur increasingly more often. In this context, a major focus of this thesis is on unintentional collisions, where a straight goal is to eliminate injury from users and passerby's via realtime sensing and control systems. A less obvious focus is to combine collision response with tools from motion planning in order to produce intelligent safety behaviors that ensure the safety of multiple people or objects. Yet, an even more challenging problem is to anticipate future collisions between objects external to the robot and have the robot intervene to prevent imminent accidents. In this dissertation, we study all of these sophisticated flavors of collision reaction and intervention. The posit here is that no matter how hard we try, collisions will always happen, and therefore we need to confront and study them as a central topic both during navigation or dexterous manipulation. We investigate in-depth multiple key and interesting topics related to collisions and safety of mobile robots and robotic manipulators operating in human environments. We show that simple sensor architectures can reconstruct sophisticated external force information including location, direction, and magnitude over the whole body of some types of robots. We devise technologies to quickly sense and react to unexpected collisions as fast as possible during navigation. We investigate robots colliding against walls in difficult tilted terrains and quickly figuring out new practical motion plans. We study methods to recognize intentional contacts from humans and use them as a non-verbal communicaton medium. We study fusing sensor data from contact sensors and time of flight laser sensors to reason about the multiplicity of contacts on a robot from human users. We investigate statistical problems like the probability that an externally moving object collides against a human. We then devise novel motion planning and control algorithms to stop the impending collisions using any part of a robot's upper humanoid body. Such behaviors constitutes some of the most advanced collision mitigation and intervention techniques we have seen in the academic communities. Overall we deeply investigate collisions from many perspectives and develop techniques that borrow and contribute to the areas of mechatronic design, sensor processing, feedback controls, motion planning, and probabilistic reasoning methods. The result of this study is a set of key experiments and guidelines to deal with collisions in mobile robots and robotic manipulators. This study aims at influencing future studies on field operations of robots and accelerate the employment of advanced robots in our daily environments without compromising our safety.

\begin{figure}[t]\centering
\includegraphics[width=\linewidth] {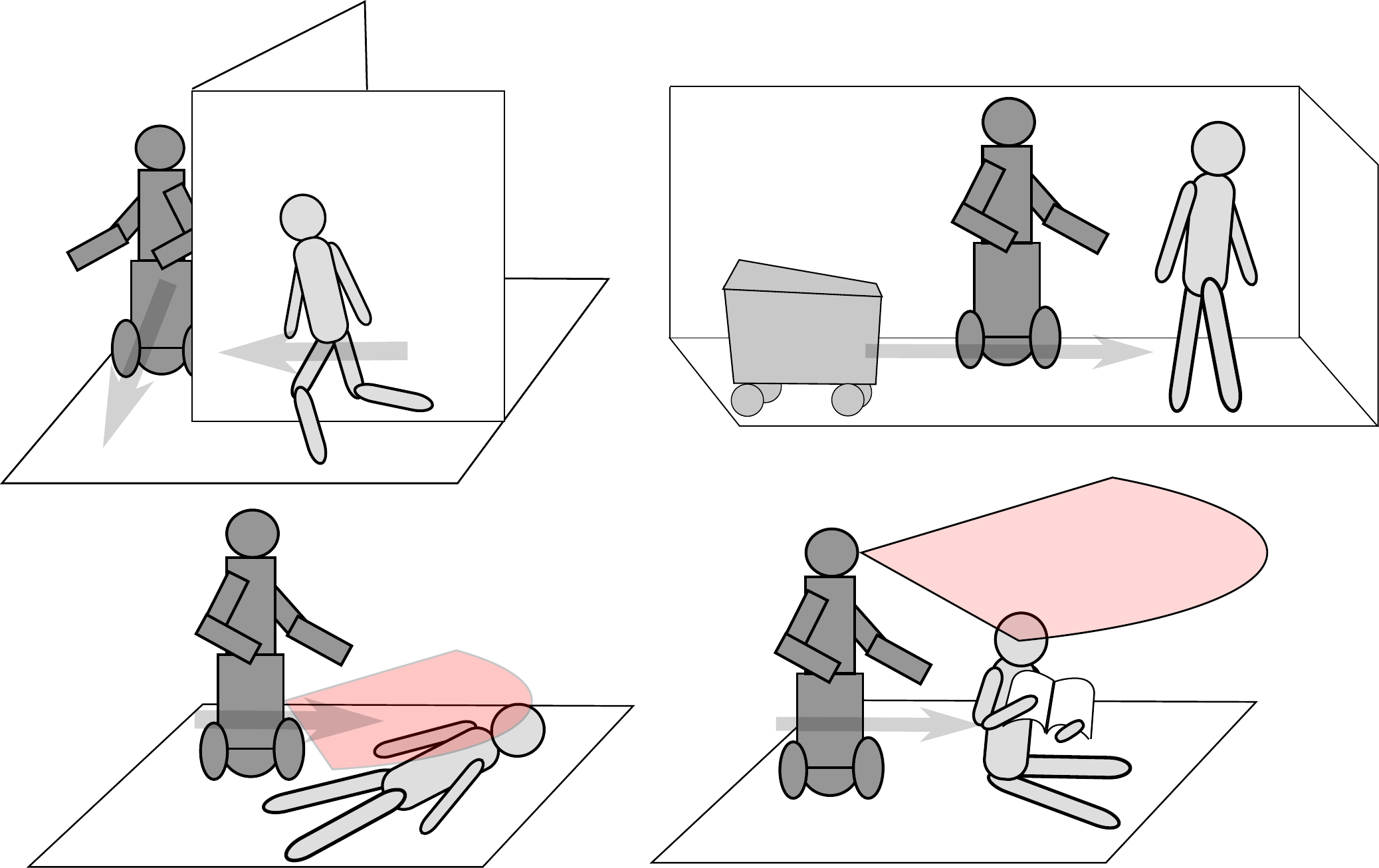}
\caption
[ Various types of accidents addressed in this dissertation] 
{{\bf Various types of accidents addressed in this dissertation.} 
	The top-left image shows the case where a mobile robot encounters all of a sudden a person coming out of a corner. The top-right image depicts the case where a cart is moving towards the back of a person and a robot is around to intervene if it would be ethically correct. The lower-left image shows a robot unable to detect a person on the floor due to blind spots on its visual sensing. Finally, the bottom-right image shows a similar case of a person reading a book on a park for instance and a robot unable to see the person for whatever reason.} \label{fig:types_accidents} 
\end{figure}

\tableofcontents   

\listoftables      
\listoffigures     

%
%
\include{chapter-ecmr}
\include{chapter-iros}

\include{chapter-auro}

\include{chapter-lidar}

\include{chapter-tro}
\include{chapter-conclusion}
\appendices
\index{Appendices@\emph{Appendices}}%

\include{chapter-appendix-lower}

\include{chapter-appendix1}

\include{chapter-appendix2}

\include{chapter-appendix3}

\include{chapter-appendix5}

\nocite{*}      
\bibliographystyle{plain}  
\bibliography{diss}        
\index{Bibliography@\emph{Bibliography}}

\printindex


\begin{vita}
{\bf Kwan Suk Kim} received B.S., and M.S. degrees in electrical engineering from Sogang university in Seoul, Korea, in 1998 and 2000, where he implemented a real-time localization algorithm for an aircraft using a vision sensor. He worked for LG Electronics Inc. as an embedded systems software engineer. He is currently working towards his Ph.D. in Mechanical Engineering at The University of Texas at Austin. His research interests include safe mobility and safe whole-body behaviors for humanoid robots and an agile skill generation for human-centered robots.
\end{vita}

\end{document}

%% file: chapter-ecmr.tex
\chapter{Contact Sensing and Mobility in Rough and Cluttered Environments } 

One of the main purposes of mobile robots is to manipulate objects with high accuracy. This goal requires the robot's mobile base to execute subtle maneuvers, i.e. precise and slow movements that can, for instance, enable the assembly of objects. When operating outdoors, robots must implement these subtle movements in the presence of nearby obstacles that can cause accidental collisions, and at the same time while dealing with rough terrains. With these issues in mind, this chapter aims to provide new methods and their experimental validation for the support of future mobile robotic systems in human-centered or outdoor scenarios.
\blfootnote{This chapter has been published in 2013 European Conference on Mobile Robots (ECMR) \cite{Kim2013}. Alan S. Kwok contributes the manufacture of the mobile robot, Trikey.}

In particular, our everyday world is filled with clutter, obstacles, moving people, and different topographies. There is a need to deal with dynamic obstacles, perhaps by safely handling collisions or using them as supports instead of avoiding them. 
On the other hand, mobile robots in general avoid contacts. However, our hypothesis is that by occasionally establishing contacts, robots will become more effective, for instance, in moving among crowded environments, recovering from falls, or anchoring their bodies against objects to engage in subtle manipulations.

We present here various methods and experiments to endow the following capabilities: (1) to move skillfully in uneven terrains without falling, (2) to respond reactively to unknown dynamic collisions using force compliance, (3) to estimate at runtime the geometry of the collision surfaces, and (4) to safely maneuver along the contour of obstacles until they have been cleared. 
To demonstrate the effectiveness of our methods, we have built a compliant holonomic base equipped with omni wheels and passive rollers on the robot's side panels. These rollers are distributed such that the base can smoothly slide along various types of surfaces during contact. We also have built an experimental setup containing an inclined terrain and a movable wall to test the adaptive interactions.

To move on uneven terrains without sliding down, we model gravity effects on the wheels as a function of both the terrain's slope and the base's heading. We obtain robot heading and position using a motion capture sensing system, incorporate the sensor data on the gravity model, and add the resulting gravity disturbances to a floating body dynamic model of the robot. We then compensate for the estimated gravity disturbances using a compliant control strategy. 

To respond reactively to collisions, we first limit the forces applied to the obstacles by using compliant control. We achieve this capability by estimating an inverse dynamic model of the robot and using it to control the effective mechanical impedance (i.e. the force generated when constraining the motion). 

To estimate the geometry of the collision surfaces, we first detect the contact when the measured distance between the desired path and the current position is larger than the motion uncertainty bounds. Then we estimate the surface normal of the object using a least squares fitting function based on the movement data points. Finally, we project the compliant control model into a mathematical constraint defined as a function of the estimated surface normal. This projection, effectively removes the direction of the wall from the motion controller, leading to a motion that is tangent to the wall while serving as a support.

Overall, although the concept of mobile robots and autonomous ground vehicles bumping into obstacles has been out for some time, this study is unique on its focus on sensing and adapting to dynamic collisions in the rough terrains. 
\begin{figure}[t]\centering
\vspace{0.1in}
\includegraphics[width=0.95\linewidth] {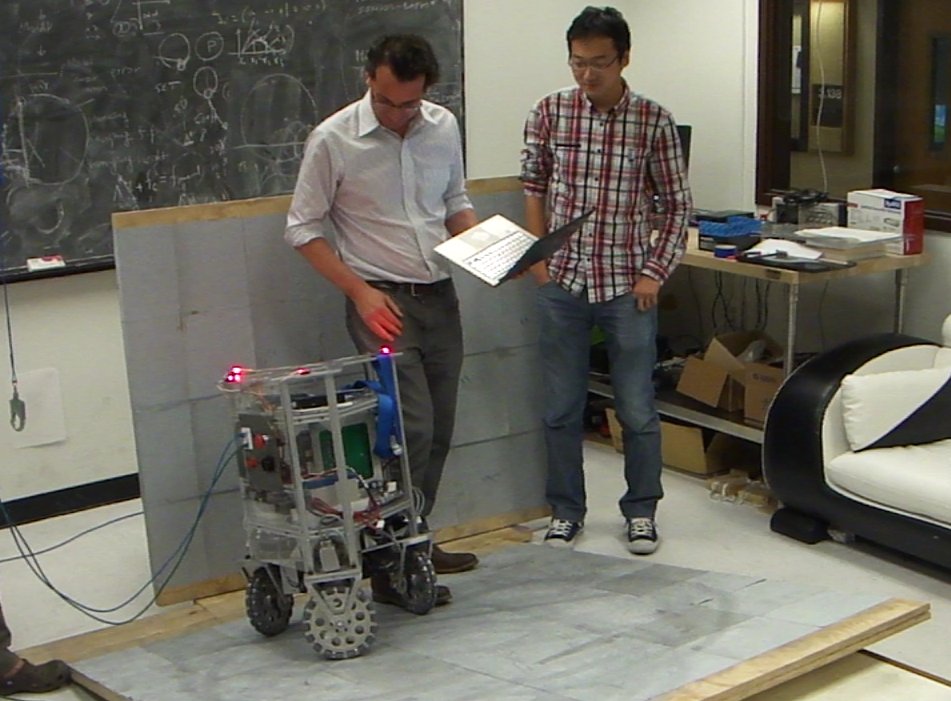}
\caption[Colliding Against Humans]{{\bf Colliding Against Humans:} Trikey safely bumps into people and corrects its heading based on the estimated contact direction.}\label{test_env}
\end{figure}



\section{Related Work}


The concept of moving around by bumping into objects is well known, for instance in bumper cars, but most recently in home robotics such as iRobot's Roomba \cite{forlizzi2006service}, which changes direction upon colliding with walls and obstacles. Bumping into objects and humans is a major area of concern in the area of free-roaming vehicles and trucks, such as AGV-Forklift systems. It has normally been dealt by using passive bumpers \cite{muselli1993anti} \cite{he2006intelligent}
or intelligent sensing using a variety techniques such as impact management accelerometers \cite{rubey1980shockswitch}
 or indoor positioning systems \cite{michel2008novell}. However, encouraging bumping into things has been largely discouraged \cite{fraichard2007short} in large autonomous systems. This is true in most cases except for a few studies, such as manipulating by pushing \cite{gerkey2002pusher} for instance. Overall, sensing and adapting to obstacles by bumping into them is a fairly unexplored area, which combined with rough terrain mobility seems to be even more unique.

Navigation using visual landmarks \cite{beinhoferlandmark}, in rough terrains using optimal control \cite{5513197}, and also in rough terrains while avoiding obstacles \cite{spenko2006hazard, sermanet2009multirange} have been recently explored in the context of accurate mobile navigation. Although we are related to those studies, none of them addresses the issue on bumping and adapting to dynamic obstacles.

Robots with spring loaded casters \cite{Kwon2011} or trunks \cite{lim2000human} have been developed for contact interactions in rough terrains but used in a very limited, even conceptual level. A step climbing robot based on omnidirectional wheels was presented in \cite{yamashita2001development}, however the focus was on stepping over obstacles rather than colliding on them and moving along their contour. An area where robots in contact have made a large impact is in physical human robot interaction using robot manipulators. Some of these pioneering studies include \cite{Zinn2004,kemp2007challenges,de2006collision}. In contrast, our study focuses on mobile bases and on performing the adaptations on rough terrains.

In the area of contact sensing, there exists pioneering work focused on localization using force \cite{gadeyne2001markov} or tactile sensing \cite{grimson1984model,petrovskaya2011global}. Our methods for detecting contact are much simpler, based on comparing wheel or visual odometry with respect to the desired trajectories. However, in contrast with those studies we focus on mobile base contact mobility instead of manipulation. Our methods are related to those discussed in \cite{iagnemma2003control}, which explores using articulated suspension to enhance tipover stability. One major difference is that we focus on colliding with objects which the previous work does not address.


\section{Omni-directional Holonomic Mobile Robot Suited for Contact}

\begin{figure}[t]\centering
\begin{subfigure}[Model View]{
\includegraphics[scale=.65, clip=true ] {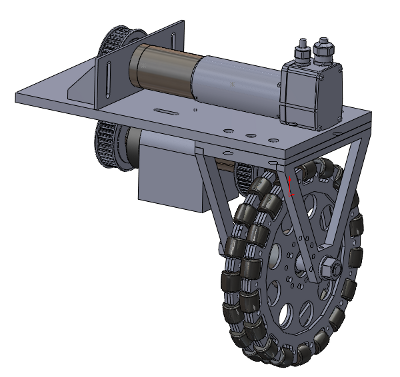}
}
\end{subfigure}
\begin{subfigure}[Front View]
{
\includegraphics[scale=.55, clip=true ] {drivetrain_02.png}
}
\end{subfigure}
\caption
[Drive Train of Trikey's Wheels] {{\bf Drive Train of Trikey's Wheels:} The drive train consists of 3 axes (motor, sensor, and wheel), which are connected by timing belts to minimize mechanical backlash. \label{drive}}
\end{figure}

\begin{figure}\centering
\includegraphics[scale=.5, clip=true ] {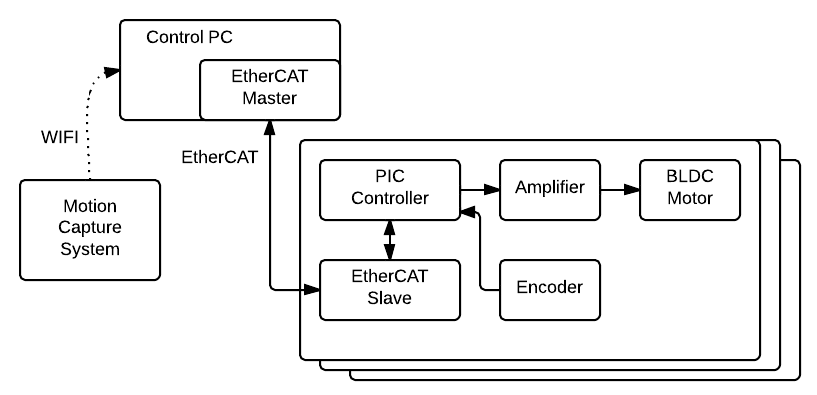}
\caption
[ Embedded System ]
{{\bf Embedded System:} 
	Each actuation module consists of a motor, an
amplifier, an encoder, a PIC processor, and a EtherCAT module. They are
connected to a small desktop PC through a daisy chained
EtherCAT bus.\label{elec_sys}}
\end{figure}

Our base, Trikey (Fig \ref{test_env}), is an omni-directional mobile base designed for compliant interactions. The 40kg robot includes three 250W Maxon brushless DC motors with omni wheels in a triangular configuration (axes separated by $120\,^{\circ}$). At 1:43 gear reduction, each wheel can achieve a maximum speed of 100 rpm and stall torque of 100N-m. The drive train (Fig \ref{drive}) is specially designed to minimize friction, and it includes a rotary strain gauge torque sensor to enable torque feedback control of the wheels. It comprises motors, encoders and torque sensors, and omni wheels connected through a timing belt transmission.

As shown in the electrical diagram of Fig. \ref{elec_sys}, an on-board desktop PC communicates with embedded PIC motor controllers via EtherCAT serial ports. This communication architecture allows the robot to daisy chain its signals into one channel, resulting in more compact and efficient electrical wiring. The PIC controllers generate an analog signal every 500 $\mu$s to control the Elmo motor amplifiers for current control \-- the torque sensors where not enabled yet for this study. Four 12V lead acid batteries are embedded inside the base, allowing the robot to operate untethered for half an hour, while supplying average power of 150W.


\section{Control Approach}

\subsection{Constrained Dynamics}

\begin{figure}\centering
\includegraphics[scale=.5, clip=true ] {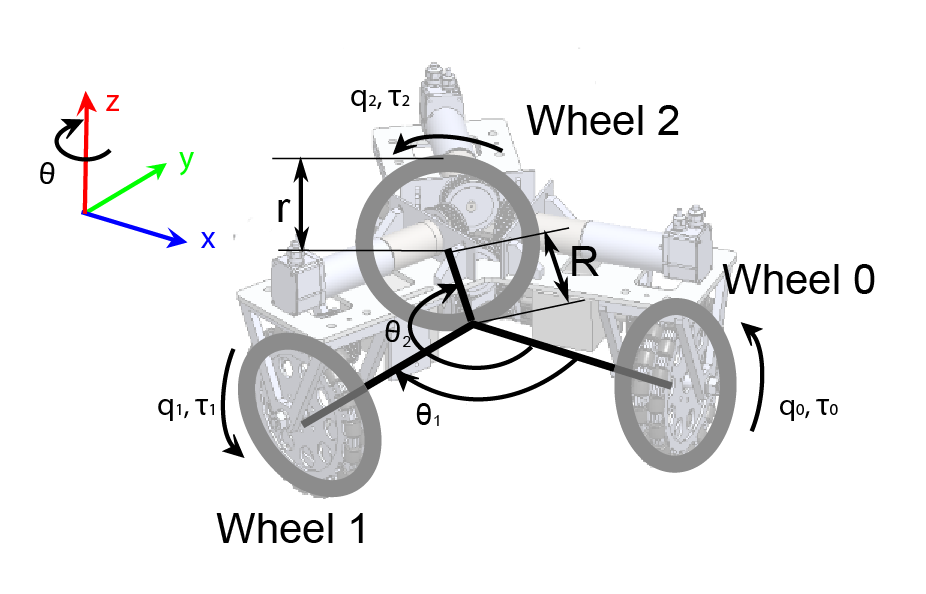}
\caption
[Trikey Structure]
{Trikey Structure shows that the wheel axes are configured at $120\,^{\circ}$. \label{tri_str}}
\end{figure}

Our robot consists of a 3-dof wheeled system (shown in Fig. \ref{tri_str}). To avoid directly deriving the kinematics between the robot's Cartesian frame and the wheel angles, we develop a planar floating body dynamic model of the base consisting of 6-dof generalized coordinates and kinematic constraints which will be later described in detail. This model includes the position and orientation of the center of the base on a plane, and the joint positions of the wheels:
\begin{equation}\label{q}
{\mathbf q} = 
\left[ 
\begin{array}{cccccc} 
{x} & {y} &  {\theta} & {q_0} & {q_1} & {q_2}  
\end{array}
\right] ^T
\in \mathbf{R}^6.
\end{equation}
To derive the floating base dynamic equation, we used Lagrange formalism as follows
\begin{multline}
\label{lag01}
L = T - V ={\frac{1}{2}} M \left( \dot{x}^2 + \dot{y}^2 \right) + {\frac{1}{2}} I_M \dot{\theta}^2 
+ {\frac{1}{2}} I_m \left( \dot{q_0}^2 + \dot{q_1}^2 + \dot{q_2}^2 \right),
\end{multline}
where $L$ is the Lagrangian, $T$ is the kinetic energy, and $V$ is the potential energy. Moreover, $M$, $I_M$ and $I_m$ are the total mass of the mobile robot, the total inertia around the vertical axis, and the inertia of a wheel around its rotating axis, respectfully. Using the Lagrangian,
\begin{equation} \label{lag02}
{\frac{\partial}{\partial t}} {\frac{\partial L}{\partial \dot{q}}} - {\frac{\partial L}{\partial q}} = \tau, 
\end{equation}
we obtain the unconstrained system dynamics
\begin{equation}\label{dyn01}
\begin{bmatrix}
M  \ddot{x} \\
M  \ddot{y} \\
I_M \ddot{\theta} \\
I_m \ddot{q_0} \\
I_m \ddot{q_1} \\
I_m \ddot{q_2} \\
\end{bmatrix}
=
\begin{bmatrix}
0 \\ 0 \\ 0 \\ \tau_0 \\ \tau_1 \\ \tau_2
\end{bmatrix},
\end{equation}
which can be expressed in matrix form as
\begin{equation}\label{dyn02}
\mathbf{A} \ \ddot{\mathbf{q}} = \mathbf{U}^T \ \mathbf{T},
\end{equation}
where
\begin{equation}
\begin{aligned}
\mathbf{A} =& 
\left[
\begin{array}{ccc}
M \ \mathbf{I}_{2\times 2}   & \mathbf{0}_{2 \times 1}   & \mathbf{0}_{3\times 3} \\
\mathbf{0}_{1\times 2}             & I_{M}                & \mathbf{0}_{1\times 3}  \\
\mathbf{0}_{3\times 2}             & \mathbf{0}_{3 \times 1}  & I_m \ \mathbf{I}_{3\times 3}
\end{array}
\right],\\
\mathbf{U} =& 
\left[
\begin{array}{cc}
\mathbf{0}_{3\times 3} & \mathbf{I}_{3\times3}
\end{array}
\right],\\
\mathbf{T} =& 
\left[
\begin{array}{ccc}
\tau_0 & \tau_1 &  \tau_2
\end{array}
\right] ^ T.
\end{aligned}
\end{equation}

In the above model, because there is no direct dependency between the wheel and the base motion, we cannot control the position and the orientation of the robot using the wheel torques. As such, we introduce a new dependency based on rolling constraints on the wheels with respect to the terrain. We consider the velocity of the wheel center with respect to the terrain in direction of the wheel rotation, $\mathbf{v}_w$, and the velocity of the wheel center in the perpendicular plane $\mathbf{v}_r$. Here $r$ stands for roller, as in the side rollers of the omni wheels. Then we can write the velocity of the i-th wheel, $\mathbf{v}_i$ as
\begin{equation}\label{const1}
\begin{aligned}
\mathbf {v}_i\ = &\ \mathbf{v}_{w,i} + \mathbf{v}_{r,i} \\
= & \ \left[-v_{w,i} \sin \left( \theta + \theta_i \right) + v_{r,i} \cos \left( \theta + \theta_i \right) \right] \mathbf{i}_x \\
& \ + \left[v_{w,i} \cos \left( \theta + \theta_i \right) + v_{r,i} \sin \left( \theta + \theta_i \right) \right] \mathbf{i}_y,
\end{aligned}
\end{equation}
where $\theta_i$ is the heading of the i-th wheel shown in Fig. \ref{tri_str}. Additionally,
\begin{equation}\label{vel_w}
v_{w,i} =\left| \mathbf{v}_{w,i} \right| = r \ \dot{q}_i.
\end{equation}
The velocity of each wheel can be described as the function of the position and the orientation of the center of the base from the rigid body kinematics.
\begin{equation}
\begin{aligned}
\label{const2}
\mathbf{v}_i &= \mathbf{v}_{CoM} + \mathbf{\omega} \times \mathbf{p}_i \\
& = \dot{x} \  \mathbf{i}_x + \dot{y} \  \mathbf{i}_y + \dot{\theta}\ \mathbf{i}_z \times \mathbf{p}_i,
\end{aligned}
\end{equation}
where $\mathbf{p_i}$ is the vector from the center of the base to the i-th wheel:
\begin{equation}
\mathbf{p}_i = R \cos \left( \theta + \theta_i \right) \mathbf{i}_x + R \sin \left( \theta + \theta_i \right) \mathbf{i}_y.
\end{equation}
When we combine Eqs. (\ref{const1}) and (\ref{const2}), we can derive the following kinematic equality.
\begin{equation}\label{const3}
\begin{aligned}
\left[
\begin{array}{c}
-v_{w,i} \sin \left( \theta + \theta_i \right) + v_{r,i} \cos \left( \theta + \theta_i \right) \\
v_{w,i} \cos \left( \theta + \theta_i \right) + v_{r,i} \sin \left( \theta + \theta_i \right)
\end{array}
\right] \\
=
\left[
\begin{array}{c}
\dot{x} \\
\dot{y}
\end{array}
\right]
+
R\ \dot{\theta}
\left[
\begin{array}{c}
-\sin \left( \theta + \theta_i \right) \\
\cos \left( \theta + \theta_i \right) 
\end{array}
\right].
\end{aligned}
\end{equation}
From Eqs. (\ref{vel_w}) and (\ref{const3}), we eliminate $v_{r,i}$ and derive the relationship between joint positions of the wheels and the position and orientation of the center of the base,
\begin{equation}\label{const4}
r\ \dot{q}_i = - \dot{x} \sin \left( \theta + \theta_i \right)  + \dot{y} \cos \left( \theta + \theta_i \right) + R \  \dot{\theta}.
\end{equation}
The above equation expressed for each wheel can be combined into matrix form as
\begin{equation}\label{eq:something}
r \left[ 
\begin{array}{c}
\dot{q}_0 \\ \dot{q}_1 \\ \dot{q}_2
\end{array}
\right]	
=
\mathbf{J}_{c,x} \left[
\begin{array}{c}
\dot{x} \\ \dot{y} \\ \dot{\theta}
\end{array}
\right],
\end{equation}
with,
\begin{equation}\label{j_cx}
\mathbf{J}_{c,x} \triangleq 
\left[
\begin{array}{ccc}
-\sin \left( \theta \right) & \cos \left( \theta \right) & R \\
-\sin \left( \theta - {\frac{2}{3}} \pi \right) & \cos \left( \theta - {\frac{2}{3}} \pi  \right) & R\\
-\sin \left( \theta + {\frac{2}{3}} \pi \right) & \cos \left( \theta + {\frac{2}{3}} \pi \right) & R\\
\end{array}
\right].
\end{equation}
By re-arranging Eq. (\ref{eq:something}) to relate to generalized velocities $\mathbf{q}$ we get
\begin{equation}\label{jc0}
\mathbf{J}_c \  \dot{\mathbf{q}} = \mathbf{0},
\end{equation}
which uses a constrained Jacobian defined as
\begin{equation}\label{j_c}
\mathbf{J}_c = \mathbf{J}_{c,wheel} \triangleq 
\left[
\begin{array}{cc}
\mathbf{J}_{c,x}
& -r \mathbf{I}_{3\times3} 
\end{array}
\right]
\in \mathbf{R}^{3 \times 6}.
\end{equation}

The above rolling constraints, can be incorporated into the robot's dynamics by extending Eq. (\ref{dyn02}) to
\begin{equation}\label{dyn03}
\mathbf{A} \ \ddot{\mathbf{q}} + {\mathbf{J}_c}^T \lambda_c = \mathbf{U}^T \ \mathbf{T}.
\end{equation}
where $\lambda_c$ are Lagrangian multipliers that model the constrained forces generated due to rolling friction between the wheels and the terrain.

Using \cite{sentis2012}, we derive a constraint null space $\mathbf{N}_c$ matrix and a constrained mass matrix $\Lambda_c$
\begin{equation}\label{L_c}
\Lambda_c = \left( \mathbf{J}_c \ \mathbf{A}^{-1} \ \mathbf{J}_c ^T \right) ^+,
\end{equation}
\begin{equation}\label{N_c}
\mathbf{N}_c = \mathbf{I} - \overline{\mathbf{J}_c} \mathbf{J}_c,  
\end{equation}
where $\overline{\mathbf{J}_c} \triangleq \mathbf{A}^{-1} \mathbf{J}_c ^T \Lambda_c$ is the dynamically consistent generalized inverse of $\mathbf{J}_c$ and $(\cdot)^+$ is the pseudo-inverse operator. By left-multiplying Eq. (\ref{dyn03}) by $\mathbf{J}_c \Lambda_c$ and merging it with the time derivative of Eq. (\ref{jc0}), we solve for the Lagrangian multipliers $\lambda_c$ as
\begin{equation}\label{lc}
\lambda_c = \overline{\mathbf{J}_c}^T \mathbf{U}^T \mathbf{T} - \Lambda_c \dot{\mathbf{J}_c} \dot{\mathbf{q}}.
\end{equation}
Finally, by the above Equation into Eq. (\ref{dyn03}), we derive the robot constrained dynamics as
\begin{equation}\label{dyn04}
\ddot{\mathbf{q}} = \mathbf{A}^{-1} \ \mathbf{N}_c ^T \ \mathbf{U}^T \ \mathbf{T}
- \mathbf{J}_c^T \Lambda_c \dot{\mathbf{J}_c} \dot{\mathbf{q}}.
\end{equation}
Once more, we highlight that we did not need to explicitly derive complex wheel kinematics.
\begin{figure}\centering
\centering
\includegraphics[scale=.70, clip=true ] {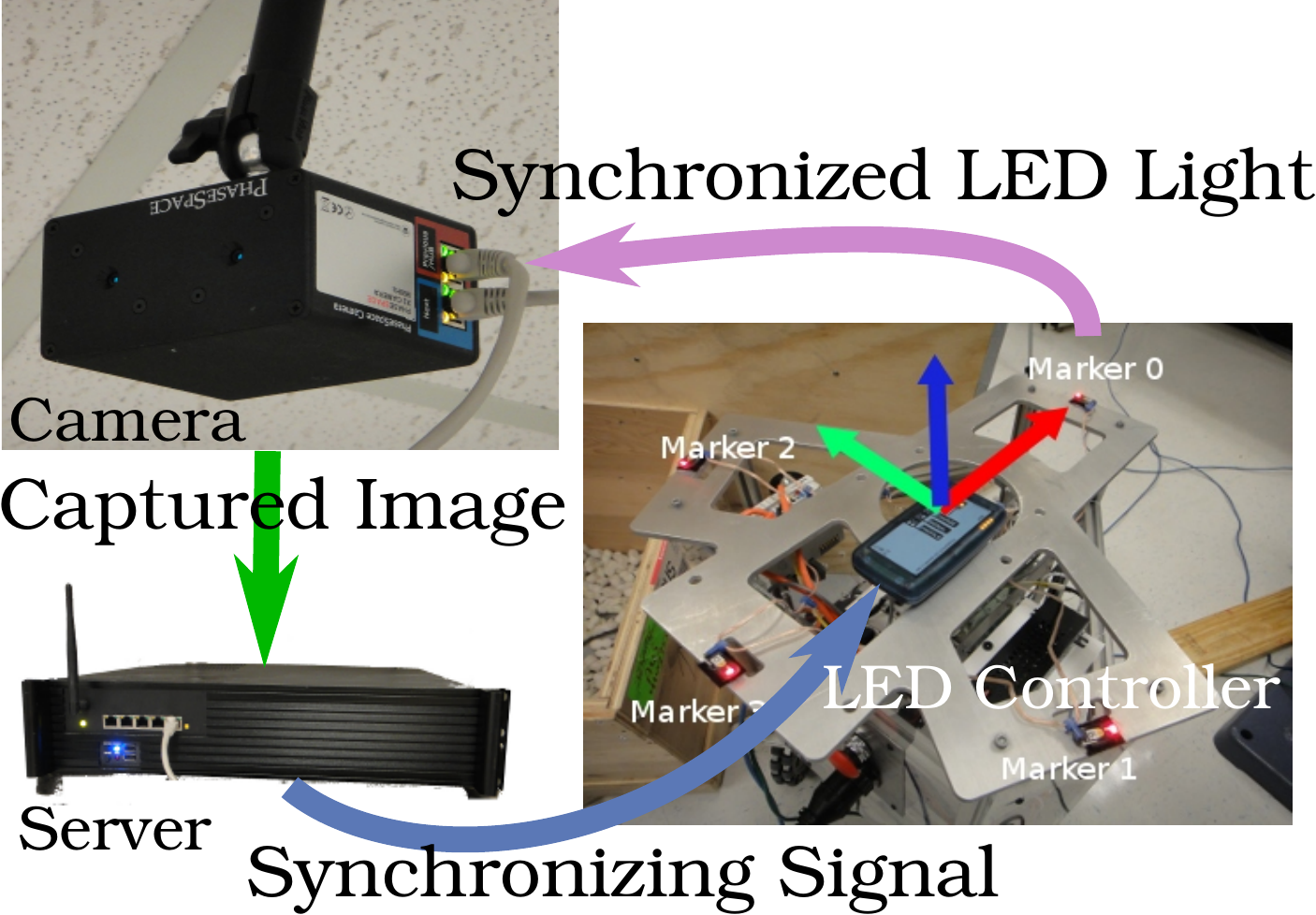}
\caption
[Motion Capture System]
{{\bf Motion Capture System:} four markers on Trikey allow for heading and position detection.} \centering
\label{marker}
\end{figure}

\subsection{Operational Space Controller}

We define an operational space controller based on the robot's Cartesian coordinates $x$, $y$, and $\theta$. In turn, the task kinematics can be defined as
\begin{equation}\label{xdef}
\begin{aligned}
\mathbf{x} & \triangleq 
\left[
\begin{array}{ccc}
x & y & \theta
\end{array}
\right]^T, \\
\mathbf{J} & \triangleq 
\left[
\begin{array}{ccc}
\mathbf{I}_{3\times3} & \mathbf{0}_{3\times3}
\end{array}
\right], \\
\dot{\mathbf{x}} &= \mathbf{J} \ \dot{\mathbf{q}}.
\end{aligned}
\end{equation}
As discussed in \cite{sentis2012}, constrained kinematics can be expressed as a function of wheel velocities alone, 
\begin{equation}
\dot {\mathbf x} = \mathbf{J}^* \begin{bmatrix} \dot q_0&\dot q_1 &\dot q_2 \end{bmatrix}^T,
\end{equation}
where
\begin{align}
&\mathbf{J}^* = \mathbf{J}_{t|s}\ \overline{\mathbf{UN}_c},\\
&\mathbf{J}_{t|s} \triangleq \mathbf{J} \ \mathbf{N}_c,
\end{align}
are projections of the Jacobian into the space consistent with the rolling constraints and $\overline{(\cdot)}$ is a dynamically consistent inverse operator.
\begin{figure}[t]
\begin{center}
\includegraphics[scale=.70, clip=true ] {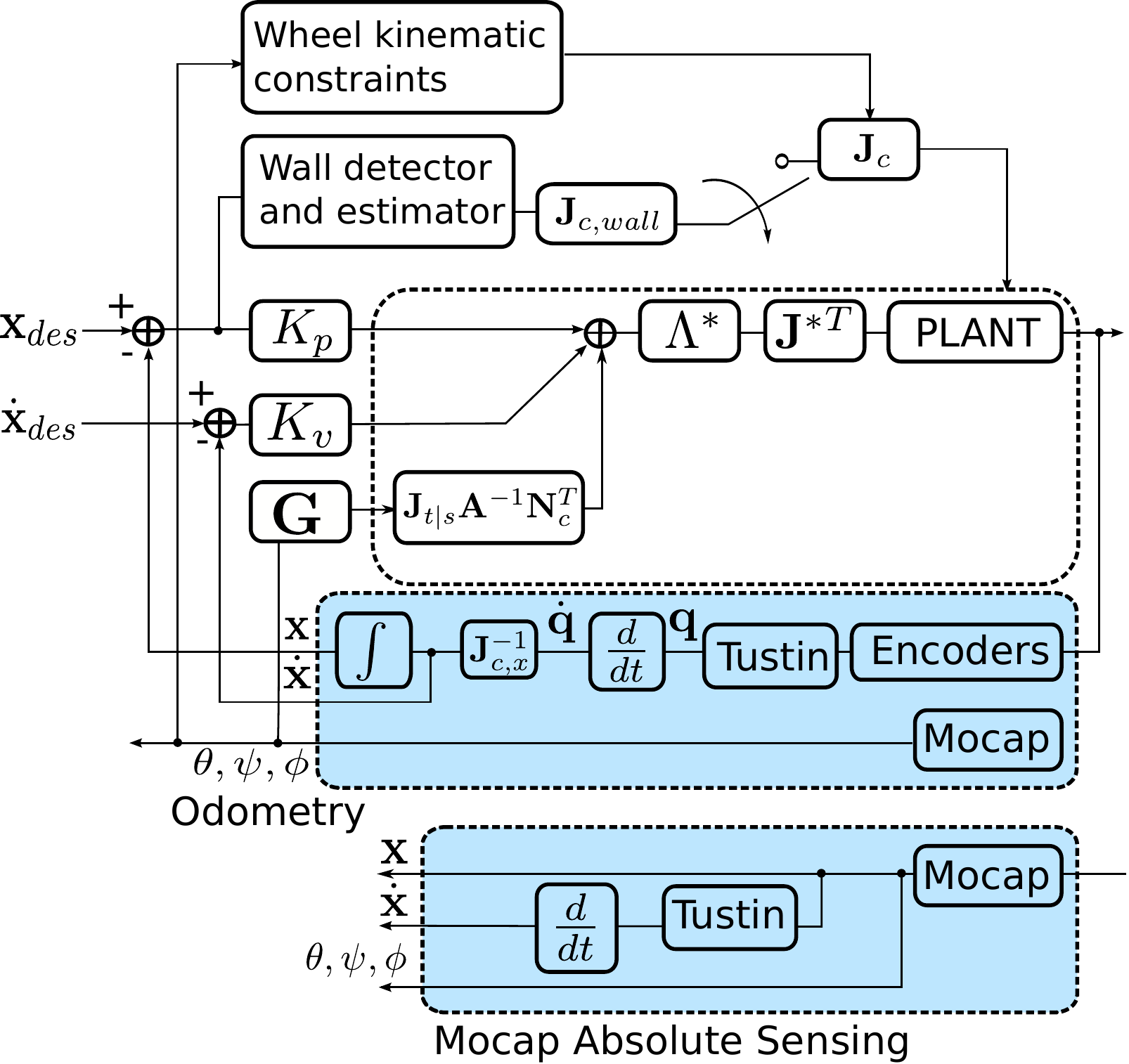}
\caption
[Control Diagram]
{{\bf Control Diagram:} includes PD gains, an operational space controller, a constraint generator, and a gravity compensator. The output of the constraint generator, $\mathbf{J}_c$ is used to compute $\Lambda^*$, $\mathbf{J}_{t|s}$, $\mathbf{N}_c$, and $\mathbf{J}^*$.\label{controller}}
\end{center}
\end{figure}

A dual expression of the above kinematics exists, providing the torque control strategy in operational space (see \cite{Sentis-thesis:07}) as
\begin{align}
\label{op03}
\mathbf{T} &= \mathbf{J}^{*T} \ \mathbf{F},\\
\label{op01}
\ddot{\mathbf{x}} &= \mathbf{J}_{t|s} \ \mathbf{A}^{-1} \ \mathbf{N}_c ^T \ \mathbf{U}^T \mathbf{T}
- \mathbf{J}_{t|s} \mathbf{J}_c^T \Lambda_c \dot{\mathbf{J}_c} \dot{\mathbf{q}}, \\
\label{op02}
\mathbf{F} &= \Lambda^{*} \ \mathbf{a}^{ref}
+ \Lambda^{*} \mathbf{J}_{t|s} \mathbf{J}_c^T \Lambda_c \dot{\mathbf{J}_c} \dot{\mathbf{q}},
\end{align}
where $\mathbf{a}^{ref}$ is the acceleration command applied to the system in the operational space, as described in Eq. (\ref{xdef}), and 
\begin{equation} 
\mathbf{\Lambda}^{*} \triangleq \left(\mathbf{J}_{t|s} \ \mathbf{A}^{-1} \ \mathbf{J}_{t|s}^T\right)^{-1},
\end{equation}
is the effective mass of the robot in the constrained space.

\subsection{Gravity Compensation}
\begin{figure}\centering
\begin{subfigure}
{
\includegraphics[width=\linewidth, clip=true ] {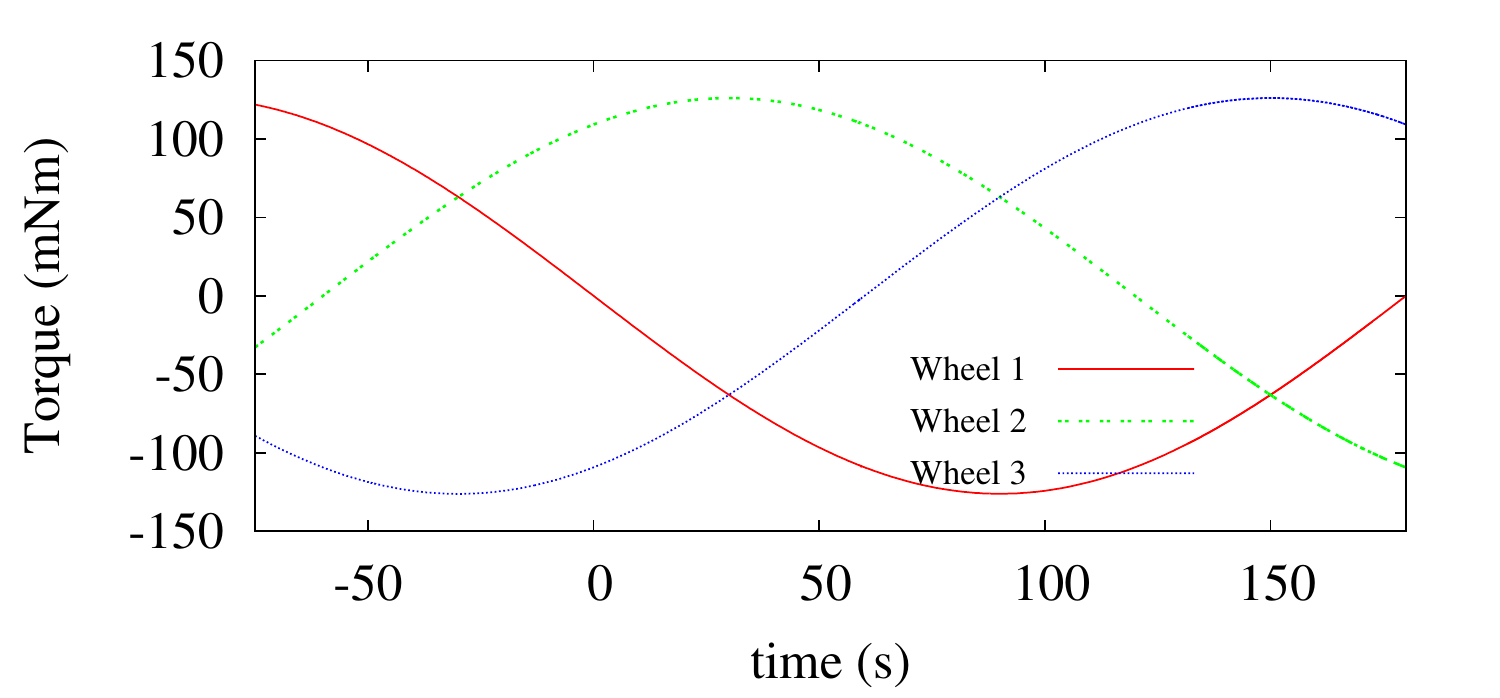}
}
\end{subfigure}
\begin{subfigure}
{
\includegraphics[scale=0.40, clip=true ] {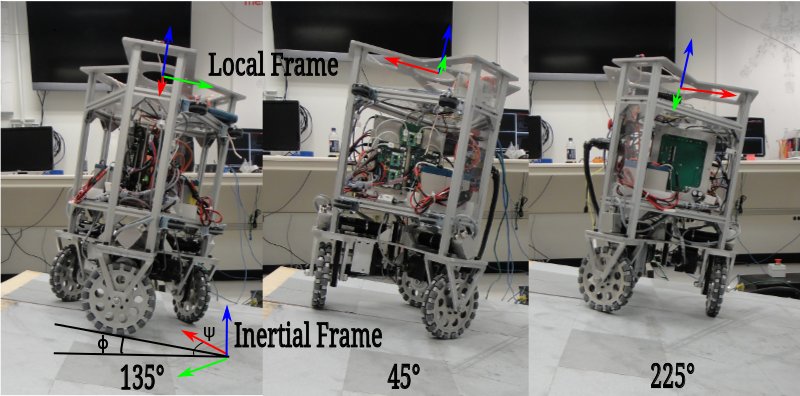}
}
\end{subfigure}
\caption
[Gravity compensation]
{{\bf Gravity compensation:} Trikey stays up on the surface while a user turns around the base. Notice that no controllers where used to maintained the robot on the fixed position. Instead, gravity compensation prevented the robot from falling down. The graph shows the torques applied to the wheels.}\label{rotate}
\end{figure}

In the case of a sloped surface, every point on the X-Y plane has a certain potential energy. If the slope's degree is $\phi$ and the heading is $\psi$ (see Fig. \ref{rotate}), the potential energy can be defined as
\begin{equation}\label{grav}
V = m g \cos \phi \left( x \cos \psi + y \sin \psi \right).
\end{equation}
We can now extend Eq. (\ref{dyn03}) to include gravity as
\begin{equation}\label{dyn05}
\mathbf{A} \ \ddot{\mathbf{q}} + {\mathbf{J}_c}^T \lambda_c + \mathbf{G}= \mathbf{U}^T \ \mathbf{T},
\end{equation}
where
\begin{equation}\label{dyn06}
\mathbf{G} = m g \cos \phi 
\left[ 
\begin{array}{cccccc} \cos \psi & \sin \psi & 0 & 0 & 0 & 0 
\end{array} 
\right] ^T.
\end{equation}
Again, using \cite{sentis2012}, $\lambda_c$ is eliminated and the above dynamic equation can be further evolved to yield
\begin{equation}\label{dyn07}
\ddot{\mathbf{q}} = \mathbf{A}^{-1} \ \mathbf{N}_c^T \ \mathbf{U}^T \ \mathbf{T} 
- \mathbf{A}^{-1} \mathbf{N}_c^T \mathbf{G}
- \mathbf{J}_c^T \Lambda_c \dot{\mathbf{J}_c} \dot{\mathbf{q}}.
\end{equation}
Finally, Eqs. (\ref{op01}) and (\ref{op02}) can be extended to include gravity effects,
\begin{equation}\label{op04}
\ddot{\mathbf{x}} = \mathbf{J}_{t|s} \ \mathbf{A}^{-1} \ \mathbf{N}_c ^T \ \left( \mathbf{U}^T \mathbf{T}
- \mathbf{G} \right)
- \mathbf{J}_{t|s} \mathbf{J}_c^T \Lambda_c \dot{\mathbf{J}_c} \dot{\mathbf{q}},
\end{equation}
with
\begin{equation}\label{op05}
\mathbf{F} = \Lambda^{*} \ \mathbf{a}^{ref}
+ \Lambda^{*} \mathbf{J}_{t|s} \mathbf{A}^{-1} {\mathbf{N}_c}^T \mathbf{G}
+ \Lambda^{*} \mathbf{J}_{t|s} \mathbf{J}_c^T \Lambda_c \dot{\mathbf{J}_c} \dot{\mathbf{q}}.
\end{equation}
From Eqs. (\ref{op03}) and (\ref{op05}), we can obtain motor torque commands while compensating for gravity disturbances (i.e without falling along the slope).

\subsection{Contact Sensing and Adaptation}

We focus for now on collisions and in particular on collisions against walls. A wall can be easily modeled as a first order inequality constraint. However, for simplicity we use an equality constraint defined as
\begin{equation}\label{wall}
y = a x + b.
\end{equation}
A Jacobian describing such constraint can be formulated as
\begin{equation} \label{jc_wall}
\begin{aligned}
&\mathbf{J}_{c,wall} \triangleq \ 
\left[
\begin{array}{cccccc}
a & -1 & 0 & 0 & 0 & 0
\end{array}
\right], \\
&\mathbf{J}_{c,wall} \  \dot{\mathbf{q}} = \  \mathbf{0}.
\end{aligned}
\end{equation}
When a new collision is encountered, such constraint can be added to the previous rolling constraints described in Eq. (\ref{j_c}) extending the constrained Jacobian to
\begin{equation}\label{j_c_new}
\mathbf{J}_{c} \triangleq 
\left[
\begin{array}{c}
\mathbf{J}_{c,wheel}
\\
\mathbf{J}_{c,wall}
\end{array}
\right] 
\in \mathbf{R}^{4 \times 6}.
\end{equation}
We can now re-use the control equations defined in Eqs. (\ref{op03}) and (\ref{op05}) but using the above augmented Jacobian. As described in \cite{sentis2012} the robot will minimize the distance to the desired trajectory if there exists a contact constraint in between.
\begin{figure}\centering
\vspace{0.1in}
\begin{subfigure}[Side View]
{
\includegraphics[scale=.35, clip=true ] {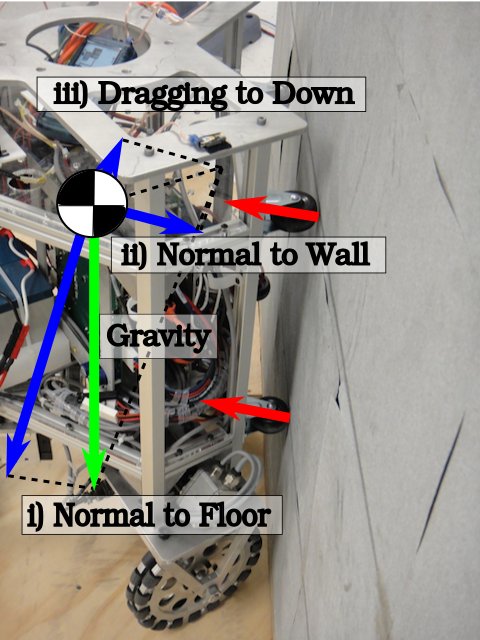}
}
\end{subfigure}
\begin{subfigure}[Top View]
{
\includegraphics[scale=.35, clip=true ] {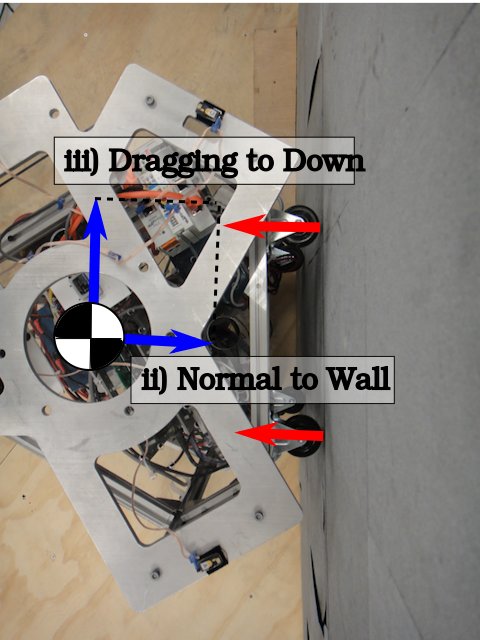}
}
\end{subfigure}
\caption
[ Details on contact with the wall] 
{{\bf Details on contact with the wall} 
	The rollers attached on the side of the robot
  reduce the friction while in contact.}\label{contact}
\end{figure}

To effectively modify the desired trajectory with respect to the contact constraint we need to detect the contact and estimate its direction. Although there are many sensing options, such as using onboard visual or LIDAR information, we choose to implement a simple sensing strategy based on a motion capture system. We measure the distance to the desired trajectory based on the motion captured data, and if the error is greater than a threshold we determine that a contact exists, i.e.
\begin{equation}\label{detect}
\left( \mathbf{Wall\ Detection} \right) = 
\left\{
\begin{array}{cc}
\mathbf{On},& \left| \mathbf{x} - \mathbf{x}_{des} \right| \geq \left(\mathbf{Err Th} \right) \\
\mathbf{Off},& \left| \mathbf{x} - \mathbf{x}_{des} \right| < \left(\mathbf{Err Th} \right) \\
\end{array}
\right.
\end{equation}
where $\mathbf{x}$, $\mathbf{x}_{des}$, and $\left(\mathbf{Err Th}\right)$ are the current and desired operational space coordinates, and the error threshold, respectively. We further estimate the wall heading by using a least square fitting formula defined as
\begin{align}\label{lms}
a_n &= \frac{
n \left( \sum x'_i \, y'_i \right) - \left( \sum x'_i \right) \left( \sum y'_i \right)}
{ n \left( \sum {x'}_i^2 \right) - \left( \sum x'_i \right)^2},\\
x'_i & \triangleq x_i + x_d,\\
y'_i & \triangleq y_i + y_d.
\end{align}
where $x_i$ and $y_i$ are the X-Y data points of the center of the base while the base, $x_d$ and $y_d$ are the distances from the center of the base to its edge contacting the wall, and $a_n$ is the estimated direction of the wall for iteration $n$.
If $x_d$ and $y_d$ are constant, $a_n$ depends only on $x_i$ and $y_i$,
\begin{equation}\label{lms}
a_n = \frac{
n \left( \sum x_i \, y_i \right) - \left( \sum x_i \right) \left( \sum y_i \right)}
{ n \left( \sum x_i^2 \right) - \left( \sum x_i \right)^2}.
\end{equation}

\section{Experiments}
\begin{figure*}[p]\centering
\begin{subfigure}[Circular Trajectory]
{
\includegraphics[scale=1., trim=50 0 80 0, clip=true ] {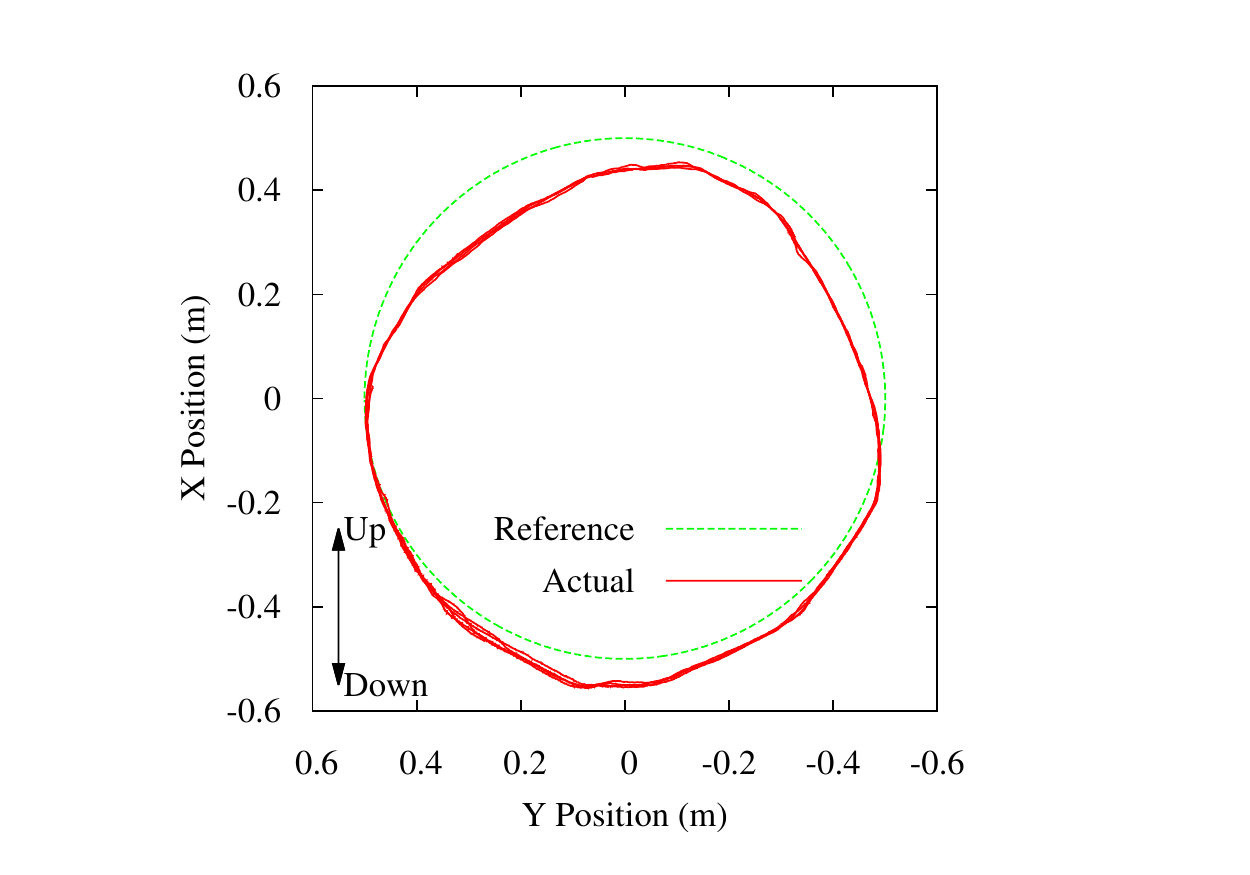}
}
\end{subfigure}
\begin{subfigure}[Torque, Position, and Accuracy]
{
\includegraphics[scale=.65 ] {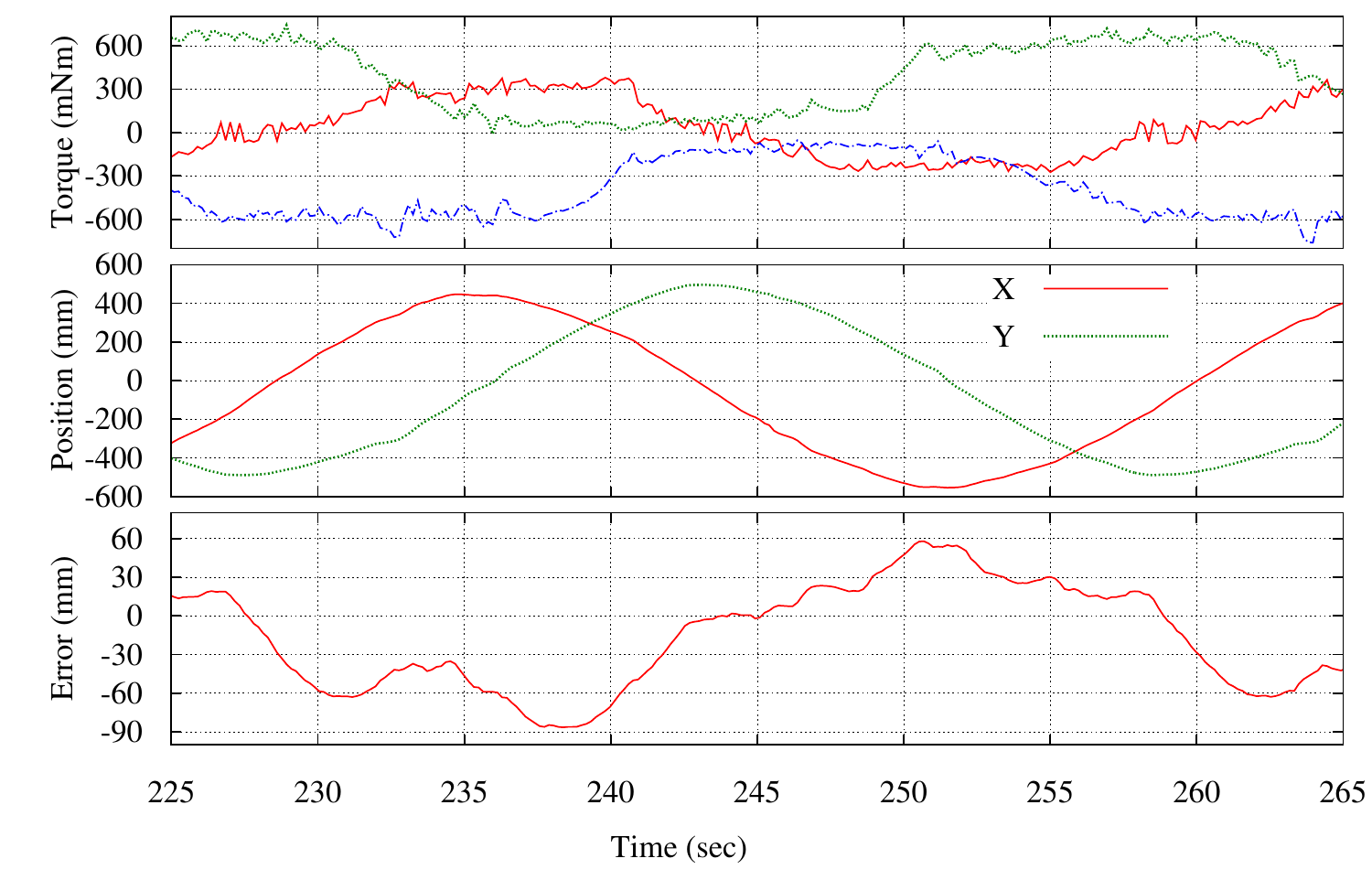}
}
\end{subfigure}
\caption
[ Circular Motion on Inclined Surface]  
{{\bf Circular Motion on Inclined Surface:} 
	This figure shows the accuracy of tracking the desire circular path on a 10$^\circ$ terrain and aided by a MoCap system.}\label{ex02}
\end{figure*}
We have built an adjustable inclined platform with a movable wall
(Fig. \ref{test_env}) and attached sand papers to the terrain to increase surface traction. The trajectory of the robot is measured by a motion capture system from Phase Space that provides about 1mm precision. By attaching four active markers on the mobile robot (Fig. \ref{marker}),
the motion capture server processes and detects a single rigid body in real time. We use the captured data in the feedback control loop and the wall detection algorithm.
Fig.~\ref{controller} depicts the controller used in the experiments. The feedback input can be based on odometry or motion captured data. Velocity is derived by differentiation of the data using the Tustin Z transform. The state variables are delivered to the PD control loop, then fed to the wall detector and estimator module, to the wheel constraints, and to the gravity compensator. To verify the controller, we command the robot follow a circular trajectory in the inclined terrain. Trikey turns circles of various radii on both a flat surface and inclined terrains. The desired angular velocities are 1 rad/s for a 1m radius, and 0.5 rad/s for a 0.5m radius. The gains of the PD controller are adjusted to limit the forces upon collision through the compliant controllers. Additionally, we limit the velocity gain to avoid feedback chattering due to gear backlash. 

\subsection{Tracking Experiment on an Inclined Surface}
We placed the robot on a $10^{\circ}$ slope. Obviously, when the robot is not powered it falls quickly down the surface because of the effect of gravity. We then turn on our Cartesian controller of Eq.~\ref{op05}, with zero desired accelerations, i.e. $a^{ref}=0$, and the base stays at its initial position without falling down because of gravity compensation (see Fig.~\ref{rotate}). Notice, that we rotate the base by hand to various orientations and it remains on the same location. For further validation, we now command the base to follow a circular trajectory on the inclined terrain as shown in in Fig. \ref{ex02}. As we can see, the tracking accuracy is about 10cm for the relatively moderate gains that we have chosen for our implementation.

\subsection{Collision Experiment Against a Wall}

\begin{figure*}[ht]\centering
\includegraphics[scale=0.4 ] {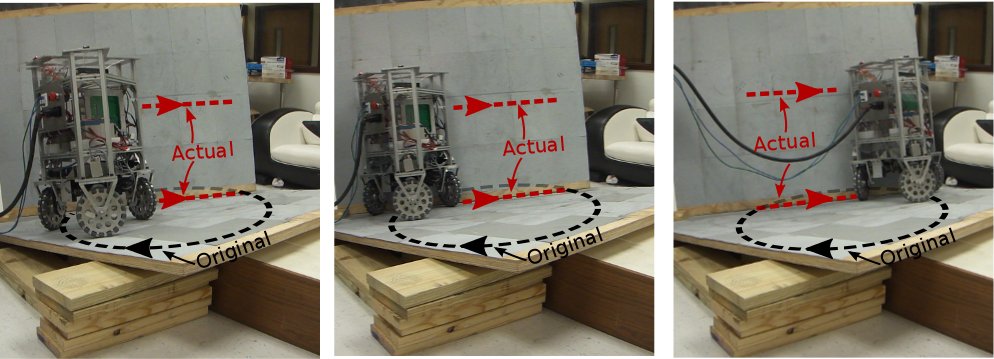}
\caption
[ Circular Motion During Contact] 
{{\bf Circular Motion During Contact:} 
	Trikey tries to move on a circle until it encounters a wall. 
After safely colliding against it using the proposed compliant controller, it estimates the wall's direction based on the algorithm of Eq.~(\ref{lms}).
	The robot finally merges with the planned trajectory when the obstacle is cleared.}\label{ex03}
\end{figure*}

Our second experiment consists on colliding with an unknown wall. As shown in Fig. \ref{ex03}, the robot tries to follow the desired circular trajectory but the contact constraint kicks in prompting the robot to move alongside the wall. Notice that to enable a smooth rolling of the base against the wall we attach 4 roller wheels to the side of the robot's base.

\begin{figure*}[p]\centering
\begin{subfigure}[Trajectory]
{
\includegraphics[scale=1.0, trim=50 0 80 0, clip=true ] {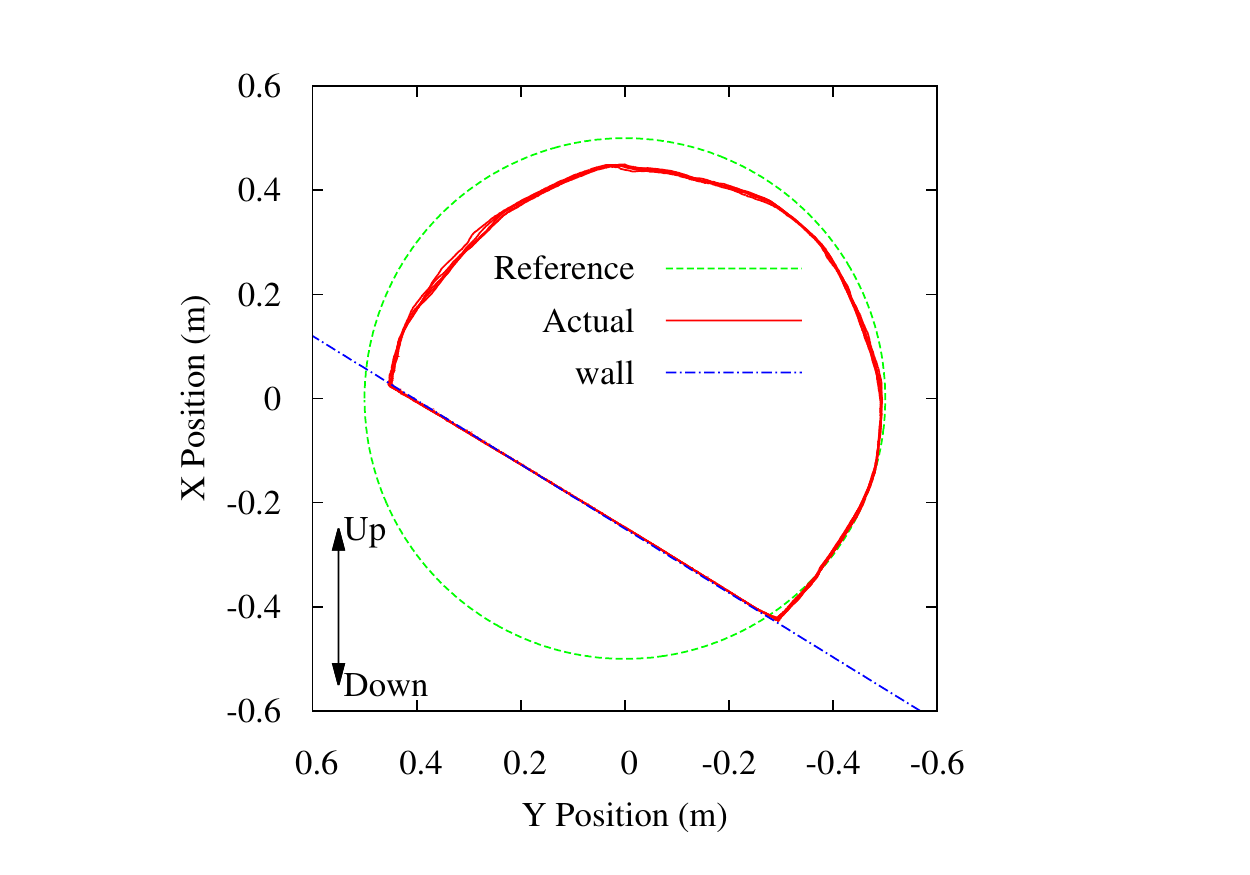}
}
\end{subfigure}
\begin{subfigure}[Torque, Position, and Error]
{
\includegraphics[scale=0.65 ] {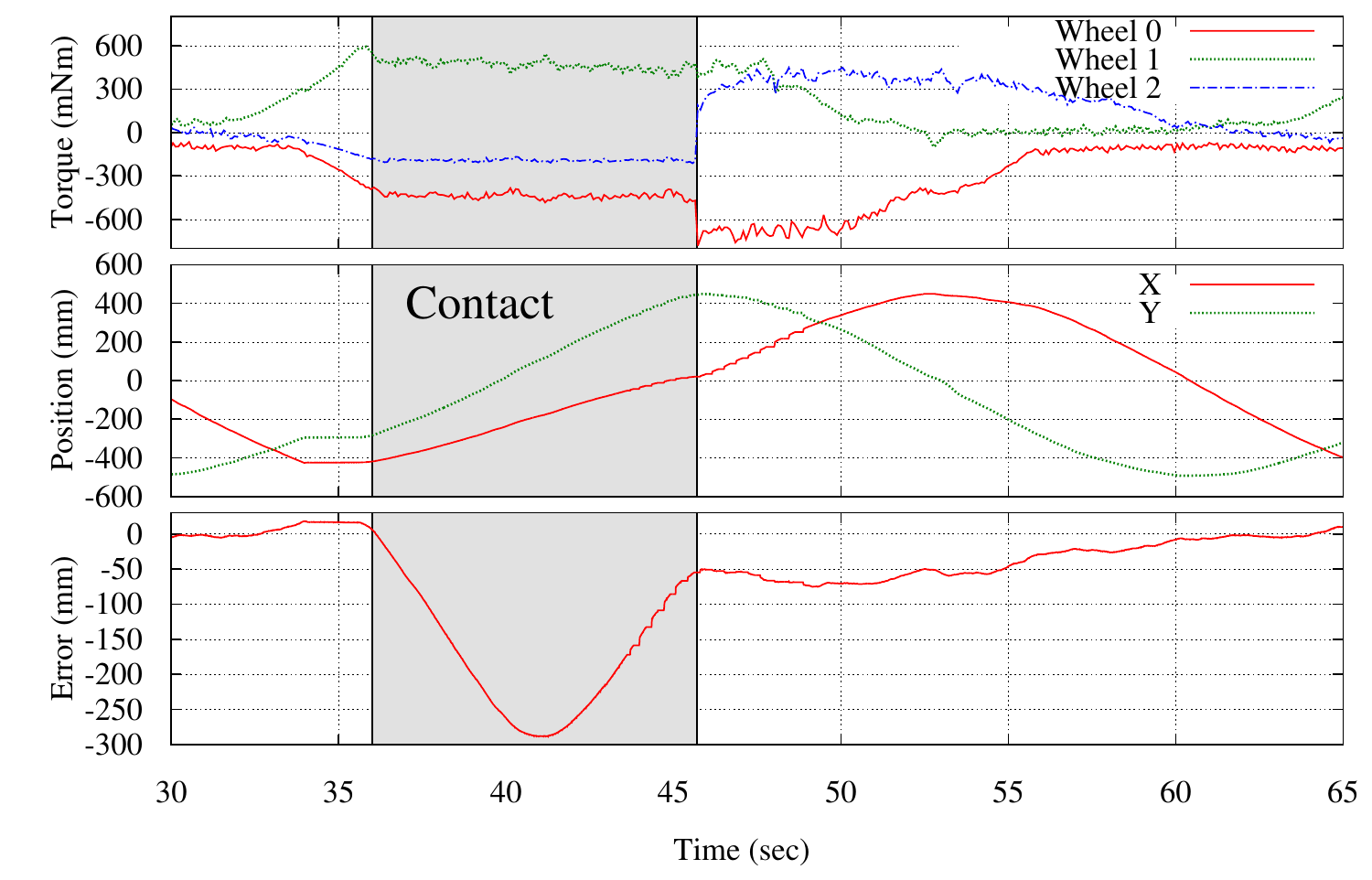}
}
\end{subfigure}
\caption
[ Circular Motion During Contact ] 
{{\bf Circular Motion During Contact:} 
	Aided by the switching controller of Fig.~\ref{controller}, it proceeds to move along the wall without applying excessive forces on the normal direction. Such behavior is enabled by projecting the acceleration on the null space of the wall constraint \-- refer to discussion around Eq.~(\ref{j_c_new}). 
}\label{ex03-1}
\end{figure*}

Fig. \ref{ex03-1} shows the torque commands applied to the wheels. We can see that the torque values remain constant during the contact phase, and increase right after the contact. That might be the result of applying less torques to compensate for gravity because the wall supports the base. 

\begin{figure*}[p]\centering
\begin{subfigure}[Trajectory and Estimated Wall]
{
\includegraphics[scale=0.9, trim=50 0 60 0, clip=true ] {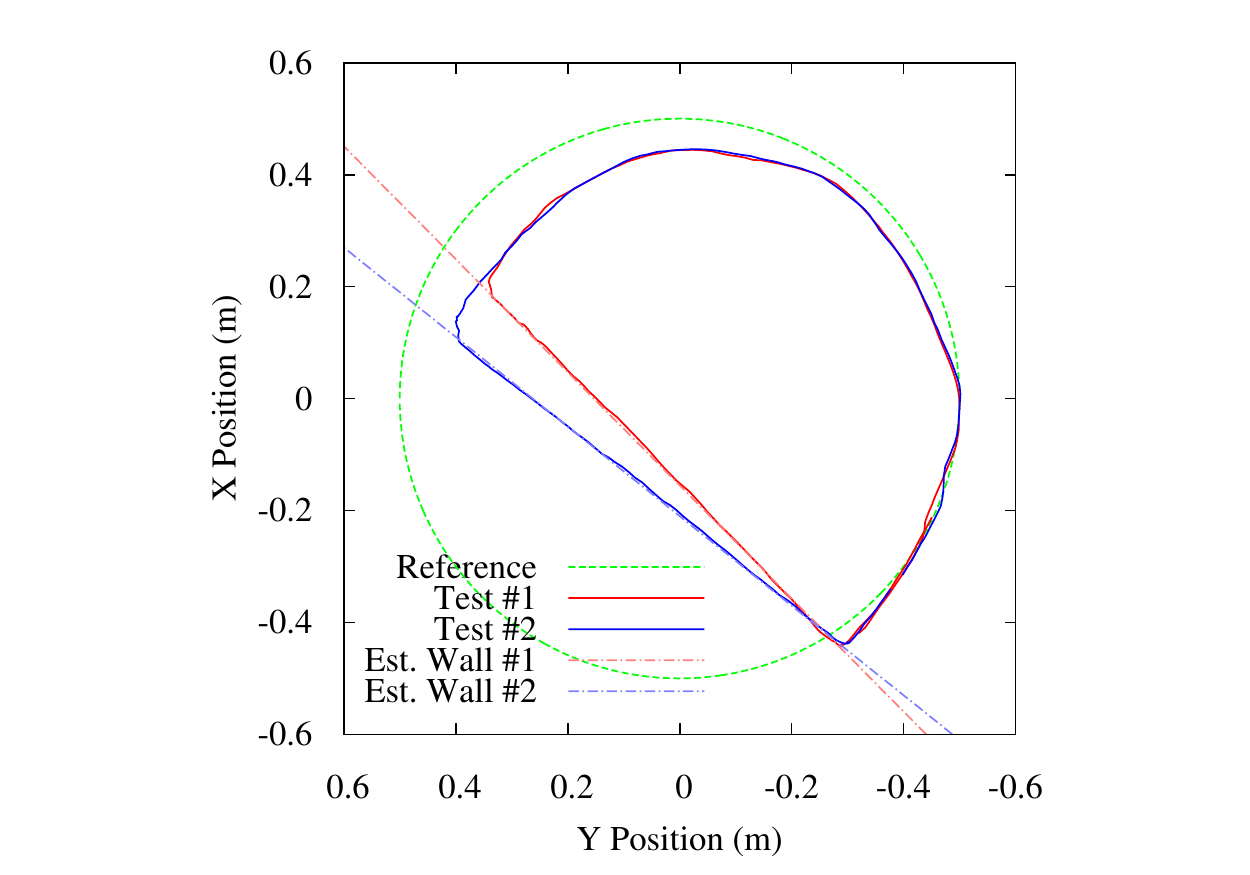}
}
\end{subfigure}
\begin{subfigure}[Estimated Wall Existence and Wall Direction]
{
\includegraphics[scale=0.65 ] {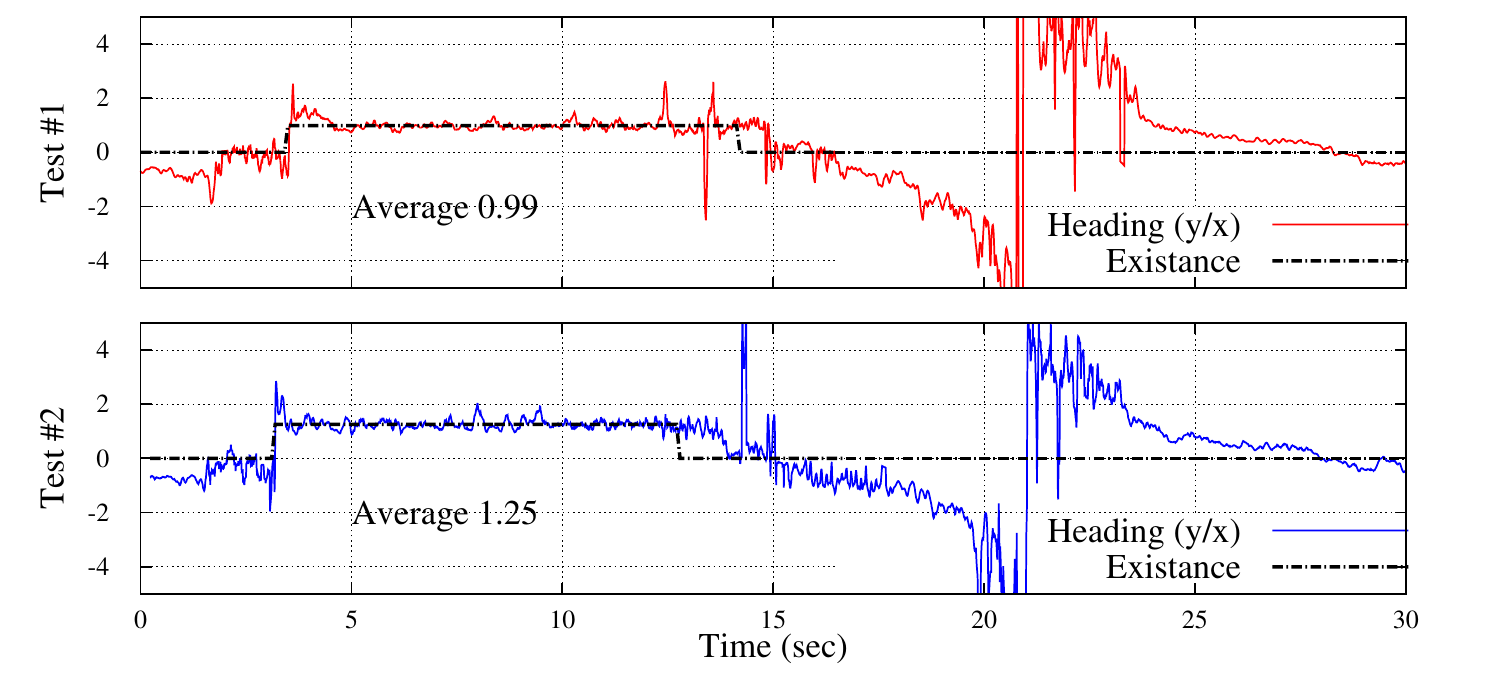}
}
\end{subfigure}
\caption
[ Detection of Wall at Various Orientations] 
{{\bf Detection of Wall at Various Orientations:} 
	In these graphs obtained from experimentation, the wall is randomly oriented to various angles by an operator. Using the procedure discussed in Eq. (\ref{lms}), the robot estimates the direction of the wall while moving and by means of our motion capture system. As we can see in the right graphs, for the slope at 45$^\circ$ the robot estimates the slope to be of value 0.99 which is very close to the expected value of 1. For the slope at 50$\circ$ the robot estimates a slope of 1.25 which is also close to the expected value. The time period where the robot is in contact is between the times of 4 to 12.5 seconds.}\label{ex04}
\end{figure*}
In Fig. \ref{ex04}, we change the orientation of the wall to various angles and observe that it is correctly detected and the estimated orientation is used to activate at runtime the contact consistent controller described in Fig.~\ref{controller}.


\section{Conclusions}
We have developed unique methods to maneuver in rough terrains with unknown contacts. We have also conducted various experiments to show the accuracy of our methods and the ability to reactively respond and adapt to the contact disturbances. We believe that these capabilities will be very important to operate in human-centered and outdoor environments. Our sensing strategy currently relies on position error detection and linear fitting to the data points. Although it is a simplistic approach it has shown to be effective for wall detection.

For our controller, we have relied on sensor-based constraints given the obstacle estimations. An optional strategy would be to implement hybrid position / force control strategies that do not rely on constraints. However, in doing so we require to estimate the contact forces in order to compensate for them in operational space. In contrast, by accounting for reaction forces as contact constraints, direct estimation is not needed. We have observed that the force on the wall can help the robot to support its body effectively, reducing the amount of effort to compensate for gravity. This kind of behavior might be beneficial to recover from certain types of adverse scenarios or to deal with wheel slippage. 

Compliant control not only allows the robot to create practical impedance behaviors for accidental collisions but also enables the base to compensate for gravity disturbances. Compensating for gravity allows the controller to employ smaller PD gains, as there are less disturbances unaccounted for. Using the same methods, we have also conducted successful experiments on safely colliding against humans (see Fig.~\ref{test_env}).

In the near future we will seek to limit the amount of force applied to the wall to be safer and improve the interaction with the colliding objects. Also, we plan to use wheel odometry and the torque sensors on the wheels to help detecting the wall and providing better compliance control. As a natural extension of this research, we would like to handle collisions with more sophisticated object shapes and also with objects in motion, such as walking humans.

Further along, we plan to incorporate an upper body humanoid robot and perform mobile manipulation tasks while dealing with unexpected contacts. We also plan to incorporate visual sensors such as LIDAR and cameras to gain navigation autonomy. 

Overall, we find that the area of human-centered robotics and outdoor navigation will benefit by incorporating compliant and online data-driven controllers to effectively deal with accidental or intentional contacts.

%% file: chapter-iros.tex
\chapter{Fully Omnidirectional Compliance in Mobile Robots Via Drive-Torque Sensor Feedback}

\section{Introduction}
As humanoid robots move out of the laboratory and into the homes and workplaces of the world, their safe physical interaction with humans will be of utmost importance. Robots, with their fallible sensors and perception will need to tread gently in a world they can barely comprehend to avoid catastrophe. We study the compliance of omnidirectional wheeled bases for humanoid robots to facilitate this transition. 
\blfootnote{This chapter has been published in IEEE International Conference on Intelligent Robots and Systems (IROS) 2014 \cite{Kim2014}. Alan S. Kwok contributes the manufacture of the mobile robot, Trikey. Gray C. Thomas contributes the collision experiments.}

Compliance has long been a common feature in robots designed to support humans as they are walking. And these robots are very successful at responding to physical human interaction. Using a force sensitive joystick or handle, these robots are pushed and pulled at the discretion of their often elderly users while providing vertical support and sometimes lateral support in one direction \cite{sabatini2002mobility, Hirata2003, Chuy2007, spenko2006robotic, Kwon2011}. Such robots have used position or velocity controlled wheeled bases to accomplish their admittance control strategy.
A similar, but more complex, strategy employed by the manipulation community uses all the joints of a mobile manipulator in an impedance control scheme, including current controlled 2-DOF caster wheels, to achieve precise impedance control of an end effector \cite{khatib1987unified}. This approach produces a very compliant end effector position, but will not notice a collision occurring on any other portion of its body.

Another design, in the field of medical robots allowed a bed/wheelchair robot, RHOMBUS \cite{mascaro1998docking}, to dock into tight spaces feedback admittance control of the force experienced by sensitive bumpers on the sides of the robot's base. The robot maneuvered with a fully omnidirectional sphere tire drive system, allowing it full planar motion freedom to control its bumper forces. This strategy has great potential for robot safety in part due to the addition of bumpers, which due to their low inertia provide a much safer contact surface for high speed impacts. However this robot was designed for reliable docking and not for unexpected collisions; the bumpers do not extend all the way to the ground, and a human operator is constantly controlling the robot via joystick.

A more intentionally safe mobile robot, DLR's Rollin' Justin \cite{fuchs2009rollin}, uses force sensitive joints in its upper body to lower the Cartesian impedance of every rigid link rather than just one end effector or pair of force sensitive bumpers. While Rollin' Justin's base itself is limited to admittance control. This means that while the upper body is quite compliant, the lower body cannot sense contact itself. 

The Rollin' Justin robot benefited from a deep field of literature on the design of compliant actuators and their safety benefits.
Collision detection and safe reaction through proprioceptive sensor feedback had been studied in \cite{de2006collision} using the the DLR-III Lightweight Manipulator Arm in a shared human-robot workspace. To ensure safety, the controller used the joint torques to identify sudden changes in energy dissipation and treated them as collision events. 
Depending on the system properties, collisions can also be detected by much simpler methods such as spikes in torque \cite{cho2012collision}.

But detecting collision is not the entirety of a strategy for human robot collision safety. As shown by \cite{haddadin2008collision}, the head an neck of a person are much harder to protect against robot bludgeoning injury compared to the softer arms and chest. This result is due to the speed at which the robot can decelerate its joints, and the high joint impedance which leads to a high impact collision.

The low impedance behavior of actuators in human collisions has been an important topic in the field of robot actuator design. A brief review of the various methods of adding compliance to actuators in order to limit their high frequency impedance is presented in
\cite{ham2009compliant}. This high frequency impedance, which is essentially the open loop impedance of the system above the controller bandwidth, can also be improved by increasing the control bandwidth by clever use of two very different actuators in parallel \cite{Zinn2004}. These strategies have the potential to make collisions with lightweight robot limbs much less dangerous by decoupling the inertia of the rest of the robot from that of the colliding limb. However, using rigid actuation for the base of a wheeled humanoid sacrifices little, since wheeled robot masses in the range of 90 to 250 Kg with stiff metal frames will dwarf the impedance of most human body parts.

Perhaps the most effectively safe humanoid base seen thus far in the literature is the pseudo-omnidirectional Azimut-3 \cite{fremy2010pushing} robot, which is equipped with four 2-DOF wheel systems, for a total of 8-DOF. Each wheel system consists of a velocity controlled wheel and a force sensitive steering arm. It uses its force sensitive steering joints to estimate the external force being applied to it, with the caveat that the force sensing cannot detect forces pointing towards its instantaneous pivot point, and has low sensor fidelity for forces pointed close to that point.
The estimated external force is used to control the steering angle and velocity set-points such that the robot is responsive to even very gentle pushes. However, such a robot could, moving nearly parallel to a wall, crush a person without sensing the extra actuator effort required to crush them. While clever and effective, force sensing of this type still has a critical safety flaw in its singularity and the region of extremely low signal fidelity surrounding it.

In designing our force controller for Trikey, we have bypassed this complexity by using a fully holonomic, omnidirectional-wheel based design without singularities. It is our understanding that this is the first time an omnidirectional robot has been outfitted with sensitive force control based on direct measurement of torque instead of motor current. We explain our strategy for simple, high bandwidth force feedback using noisy load cells, and present experiments demonstrating the effectiveness of the design.

\section{Designing a base with torque sensors}
\subsection{Trikey: An Omni-directional mobile robot}
Trikey is an omni-directional, holonomic mobile robot designed for precise mobile manipulation on uneven terrain, while serving as an attachment to an upperbody humanoid.\cite{sentis2012} \cite{Kim2013}

\subsubsection{Electronics} 

\begin{figure}\centering
\includegraphics[width=0.7\linewidth, clip=true ] {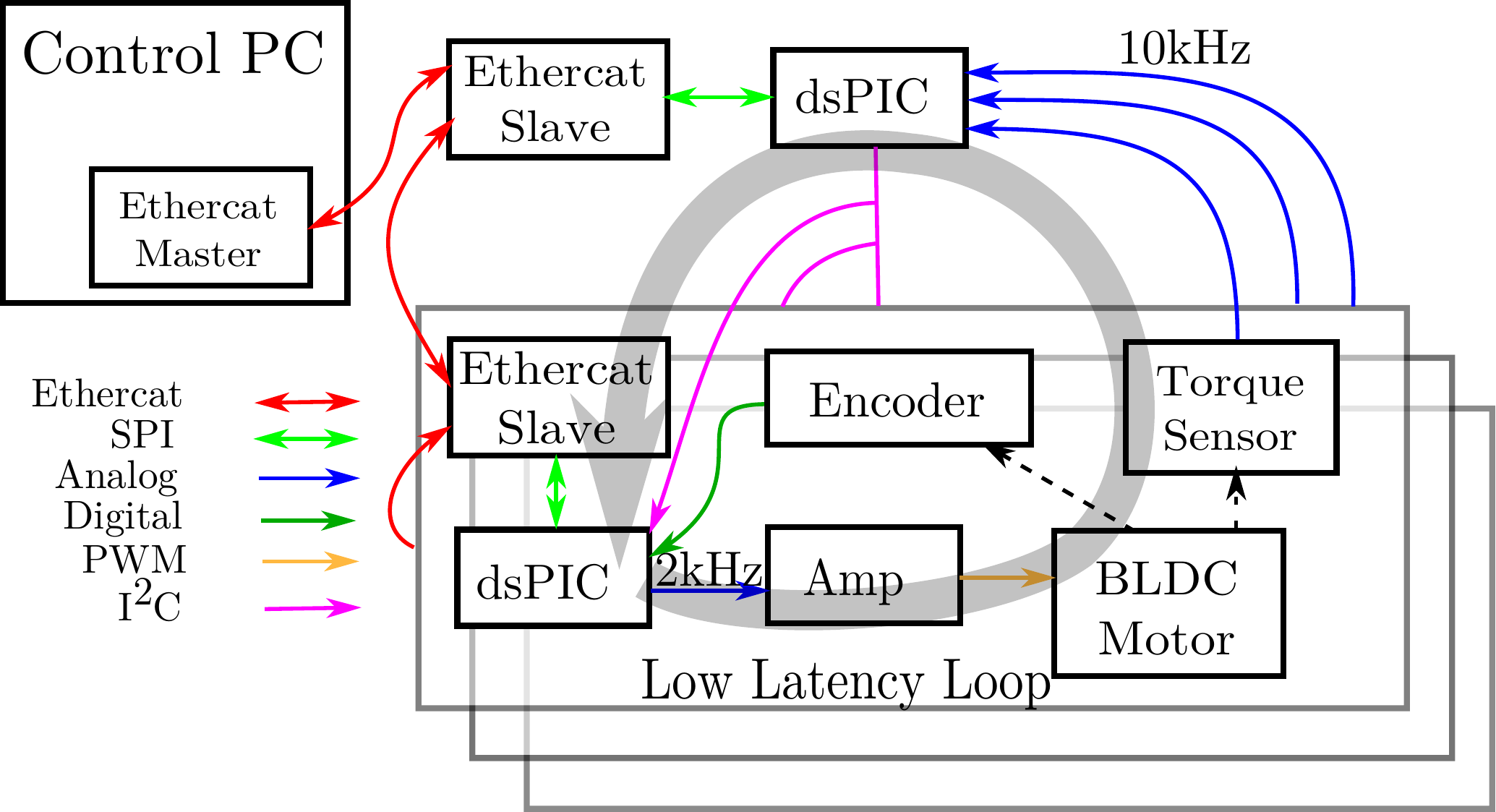}
\caption
[ The Electronic system for the actuators]
{{\bf The Electronic system for the actuators}
	are connected to the control PC throught EtherCAT. Each actuator has a 16-bit microcontroller, which can run a simple control loop. There is a 1kHz $\rm I^2 C$ bus channel between the torque sensor controller and the motor controllers, so a low latency force feedback loop can be completed throught the channel (Gray Loop).\label{elec_sys}}
\end{figure}

The main controller runs on a small form factor computer (Control PC in Fig. \ref{elec_sys}) with a mobile I7 processor. All the actuators in the mobile base and humanoid upperbody are connected to the main controller via an EtherCAT communication bus. Each of these EtherCAT slaves communicates with a 16bit dsPIC microcontroller through an SPI bus. (Fig. \ref{elec_sys}).

\subsubsection{Actuations} 
The wheels are driven by brush-less DC motors with a 50:1-ratio Harmonic drive. A torque sensor is placed in each drivetrain between the motor and the wheel, so the motor torque or external input can be measured and  used by the force controller. 


\subsubsection{Communication Delay} 
The motor command from the main controller is sent to the actuators every 1 ms, but it takes 5 ms for the command to be implemented at the microcontroller level. On the other hand, the microcontroller has a an internal $500\mu \rm sec$ control loop with no time delay. In short, there are two-layers of control loops: the high latency main controller and the low latency microcontrollers. Since torque feedback is sensitive to time delay in the system, the measured torque data is directly sent from the sensor microcontroller to the actuator microcontrollers via $\rm I^2 C$ bus every 1 msec. The torque sensors output their measurements as  analog signals, and 12-bit AD converters digitize and send these signals to the sensor microcontroller every $100 \mu \rm sec$. This sensor microcontroller implements a 1kHz 2nd-order Butterworth low-pass filter for each torque signal.

\subsection{The Actuator Testbed}
\begin{figure}\centering
\includegraphics[width=0.75\linewidth, clip=true ] {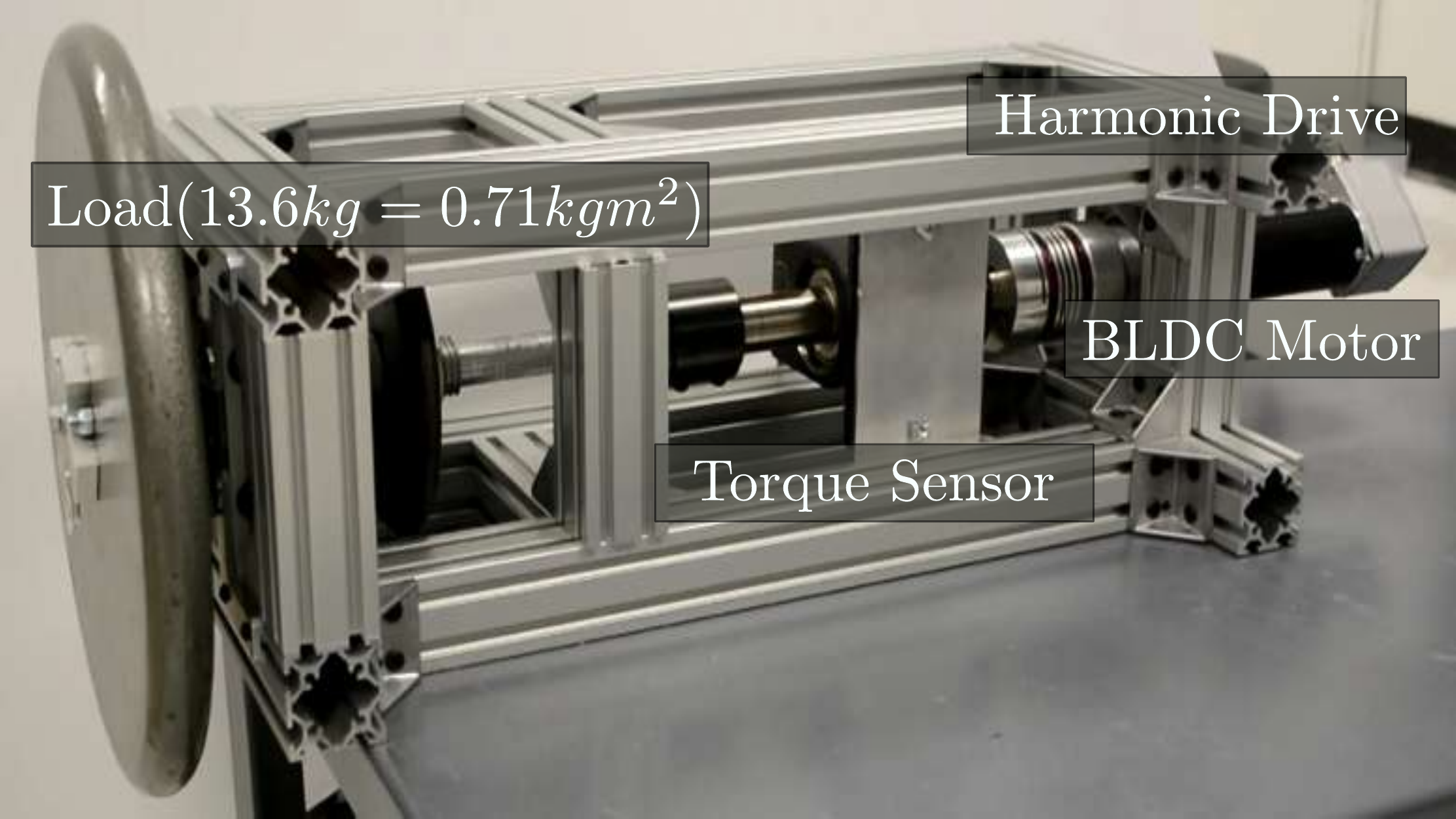}
\caption
[Force Control Testbed]
{{\bf Force Control Testbed:}
	This alternate configuration for Trikey's drivetrain components facilitates testing with various load inertias\label{testbed}}
\end{figure}

To investigate the system dynamics and time delay in detail, we built a testbed which has a similar configuration to the actuators in the mobile base (Fig. \ref{testbed}). The output shaft of the testbed can be connected to various loads with known inertia values. The shaft can also be rigidly affixed to the frame to measure stall torque and to observe the system response to a nearly infinite inertia. 

When the load is fixed, the open-loop step torque is measured in Fig. \ref{approx_ctrl}. 
From the graph, The torque settles down in 40msec, 
implying that the system dynamics have a relatively long time constant compared to the servo rate.

\section{Actuator control with time delays}

The bandwidth of compliant behavior in rigid actuators is limited by the delay between the control action and the sensing of that action's effect \cite{wu1997digital}. In the past the Smith predictor algorithm \cite{smith1959controller} has been used to push out the bandwidth of force and impedance controllers under similar conditions \cite{freund1998high}, \cite{hagglund1992predictive}. Observing the destabilizing nature of finite difference feedback in discrete time controllers of continuous  
delay systems with fast time constants relative to the latency \cite{hagglund1992predictive}, Trikey's controller uses only proportional force feedback, and a smith predictor that assumes a constant gain delayed plant. This controller is run on the embedded digital signal processors to minimize latency.
\subsection{Ideal torque controller with a disturbance}
A simple torque command $u_{(n)}$ with a torque feedback $\tau_s$ can be implemented as follows:
\begin{equation} \label{simple_ctrl}
u_{(n)}= K_{ff} \tau_{d(n)} + K_p [ \tau_{d(n)}- \tau_{s(n)}],
\end{equation}
Where $\tau_d$, $K_{ff}$ and $K_p$ are the desired torque, the gain for the feedforward term, and the gain for the proportional feedback.

The idealized plant model for each actuator, neglecting force sensor compliance, can be represented as a constant gain $G$ and a disturbance $\tau_{ext}$ as:
\begin{equation} \label{ideal_plant}
\tau_s(n) = Gu_{(n)} + \tau_{ext} \left( n \right) 
\end{equation}
Merging Eqs. (\ref{simple_ctrl}) and (\ref{ideal_plant}), and defining $K_{ff}$ as $1/G$, the torque command and the torque feedback can be expressed as follows:
\begin{align} 
u\left(n\right) &= \frac{1}{G} \tau_{d} - \frac{K_p}{1+K_p G} \tau_{ext} \label{simple_ctrl_detail}\\
\tau_s \left(n\right) &= \tau_{d} + \frac{1}{1+K_p G} \tau_{ext}  \label{ideal_plant_detail}
\end{align}
From Eq. (\ref{ideal_plant_detail}), as the gain $K_p$ increases, the effect of the disturbance $\tau_{ext}$ decreases. If a digitized system is not considered \cite{wu1997digital}, ever higher gains appears to  guarantee ever improving performance.
\begin{figure}\centering
\includegraphics[width=0.75\linewidth, clip=true ] {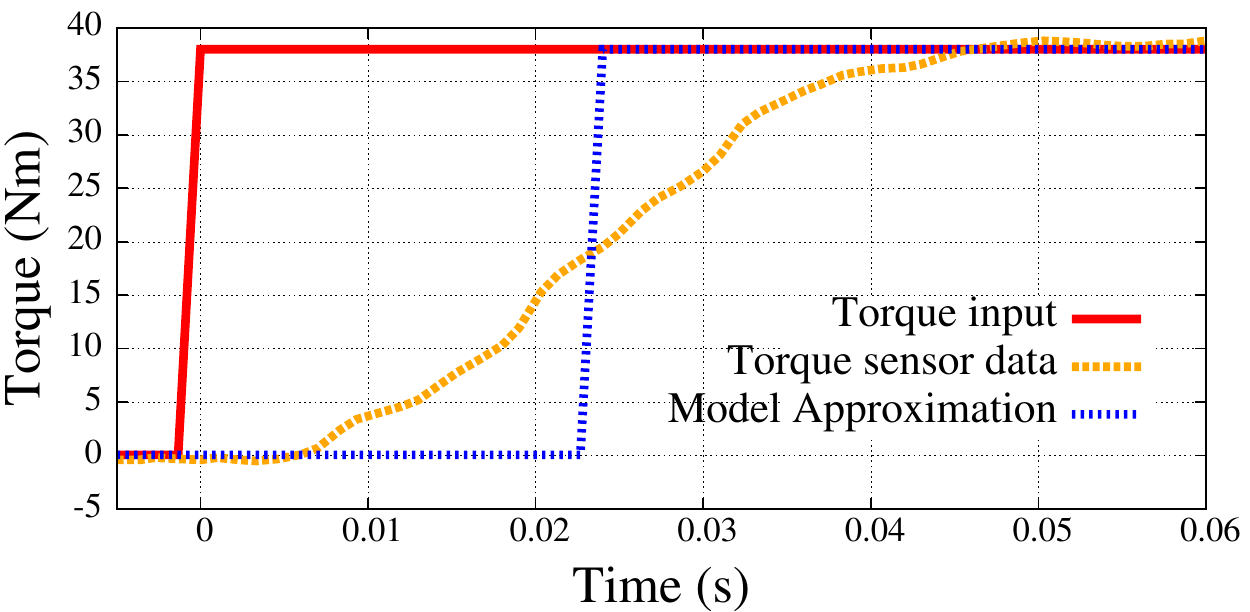}
\caption
[Approximate system plant as constant gain and delay]
{{\bf Approximate system plant as constant gain and delay:} the pure delay and continuous dynamics of the system are modeled using only a constant gain, $G$, with a time delay $d$\label{approx_ctrl}}
\end{figure}

\subsection{Time Delayed Feedback Torque Controller}
In real robots, mechanical or electrical impedance effects tend to act like a low pass filter at high frequencies, causing the simple proportional torque controller in Eqs. (\ref{simple_ctrl_detail}) to fail at high gains. Time delays behave in much the same way. To accurately model the effect of feedback delay and higher order system dynamics we would need to find the roots of a quasi-polynomial expression. To simplify the problem, we approximate the system response as a constant gain, $G$, and the constant time delay, $d$ as Fig. \ref{approx_ctrl}. Then, the torque sensor data can be estimated as follows:
\begin{align}
\tau_s \left( n \right) &= G \ u \left( n - d \right) + \tau_{ext} \left( n \right) \label{real_plant},
\end{align}
\begin{equation}
\begin{aligned}
u\left(n\right) &= 
K_{ff} \tau_d \left( n\right) + K_p \left[ \tau_{d} \left( n \right) - G \ u \left( n-d \right) - \tau_{ext} \left( n \right) \right] \\
& = \frac{1}{G} \tau_d \left( n\right) - \frac{K_p}{1+K_p G} \tau_{ext} \left( n \right) + \frac{K_p}{1+K_p G}\Delta u \left( n \right) \label{real_ctrl_detail02}
\end{aligned}
\end{equation}
where $\Delta u \left( n \right) = u \left(n \right) - u \left( n-d \right)$ and $K_{ff} = 1/G$. 
If the control loop is fast enough and we can assume that $\tau_{ext} \left( n \right) \approx \tau_{ext} \left( n - d \right)$, then the torque sensor data will be expressed as follows:
\begin{equation}
\tau_s = \tau_d \left( n - d \right) - \frac{1}{1+K_p G} \tau_{ext} \left( n \right) + \frac{K_p G}{1+K_p G}\Delta u \left( n \right) \label{real_plant_detail}
\end{equation}
The unexpected term with $\Delta u$ approaches to 1 as the proportional gain $K_p$ increases. Moreover, the term will accumulate in every control loop, so the system can be unstable.

\subsection{Smith predictor}
\begin{figure}\centering
\includegraphics[width=0.85\linewidth, clip=true ] {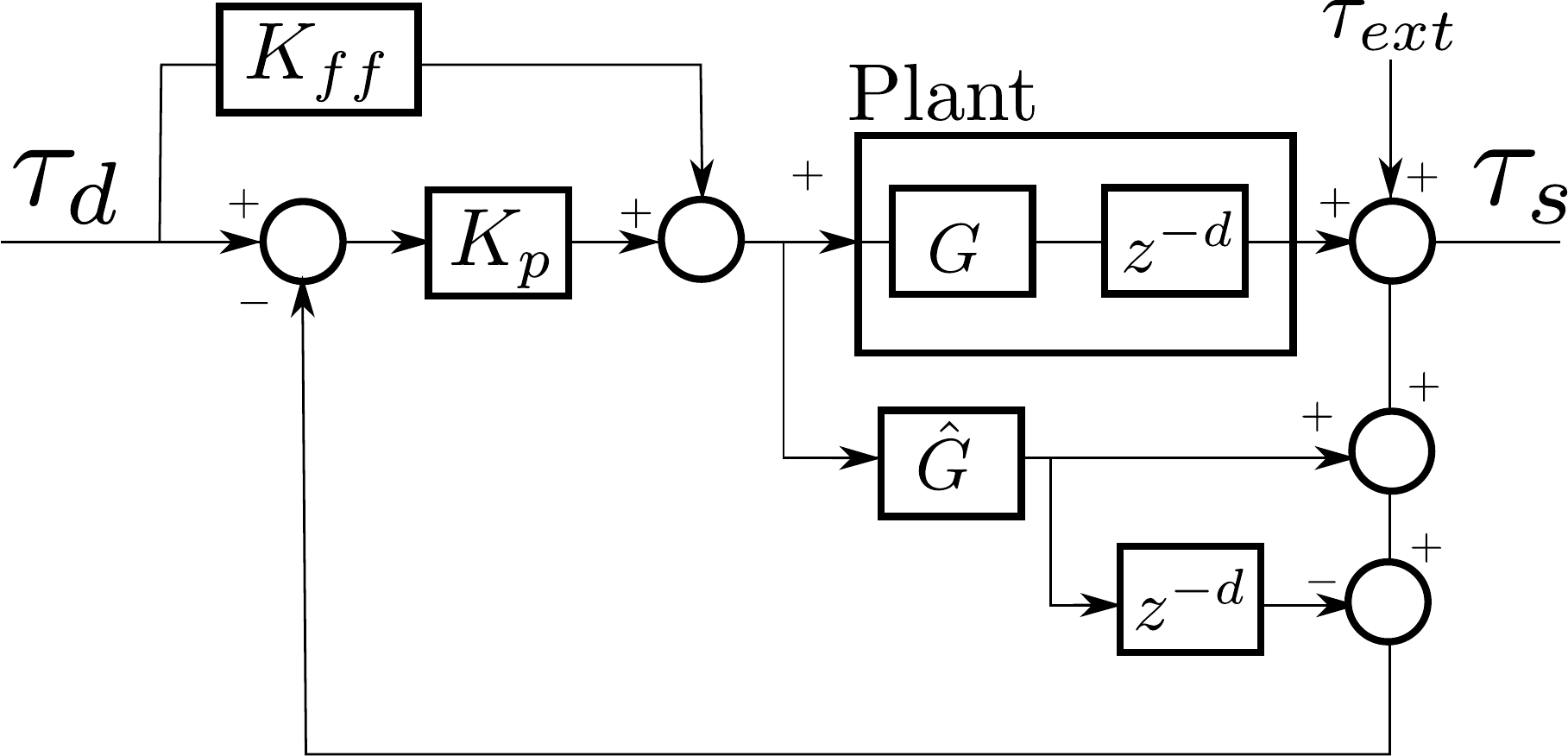}
\caption
[ Torque controller with Smith predictor]
{{\bf Torque controller with Smith predictor}
	removes the effect of the previous control input, which causes the system unstable\label{sp_ctrl}}
\end{figure}

To remove the effect of the unexpected term, compensating terms are needed. 
Fortunately, the previous $u\left(n-d\right)$ is known to the controller, so the compensator can be implemented by estimating the coefficient of the $u\left(n-d\right)$ term. 
The constant gain of the plant, $G$ is estimated as $\hat{G}$. 
To remove $u\left( n -d \right)$, the Smith predictor adds $\hat{G} u\left(n\right) - \hat{G} u\left(n-d\right)$ to $u\left(n\right)$ as in Fig. \ref{sp_ctrl}.
Then, the torque command $u$ can be described as follows:
\begin{equation}
u = \frac{1}{1 + K_p \hat{G}} \left[ \left( K_{ff} + K_p \right) \tau_d + K_p ( \hat{G} - G ) u_d - K_p \tau_{ext} \right]\label{sp_cmd}
\end{equation}
where $u_d = u \left( n - d \right)$.
If $G \approx \hat{G}$, then Eq. (\ref{sp_cmd}) approximates Eq. (\ref{simple_ctrl_detail}).

\section{Experiments}
\subsection{Experimental Setup}
Typically, systems with long time delays uses PI controllers rather than PID controllers because the derivative term becomes destabilizing in such systems \cite{hagglund1992predictive}. Therefore, we use a PI controller in the force controller. The I-term was added to the control system shown in Fig. \ref{sp_ctrl}.
This force controller runs on the microcontroller every $2 \rm kHz$, and in the test bed experiments it receives its force setpoints from a simple impedance controller simulating a very soft spring. This impedance controller runs on the control PC at an update rate of $1 \rm kHz$. 

We attached a $13.6 \rm kg$ weight with an inertia of $0.71 \rm  kgm^2$ to approximate the mobile robot inertia. To analyse the compliant performance of the mobile robot, we used a motion capture system as a global position sensor \cite{Kim2013}. The precision of the motion capture system is $1mm$. 

\subsection{Zero Force Control in the Testbed} \label{zfe}
\begin{figure}[h]
\centering
\includegraphics[width=0.72\linewidth] {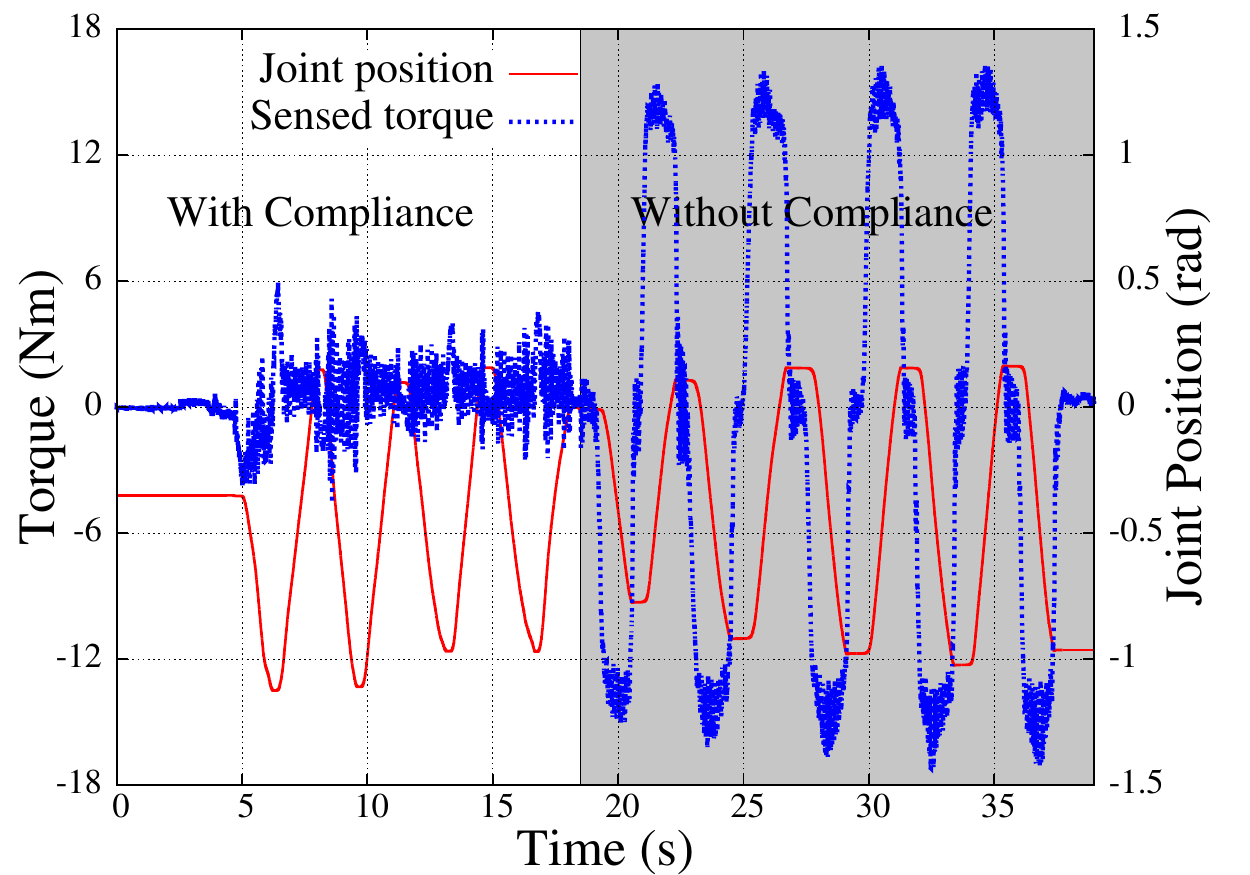}
\caption
[ Zero Torque Control]
{ { \bf Zero Torque Control}
	is implemented by sending $\tau_d = 0$ to the force controller. Then, the controller tries to follow the external force. }\label{zero_torque_01}
\end{figure}

To identify the compliance of the testbed, we set the desired force to zero. We applied an arbitrary force trajectory manually to the weight, while measuring the joint angle and torque. The result is shown in Fig. \ref{zero_torque_01}. 
Without the force controller, around $15 \rm Nm$ is needed to rotate the load by $\pm 1 \rm rad$. Even though the drivetrain is considered backdrivable, there is a large and considerable friction in the high gear-ratio Harmonic drive. With the controller active, only $3 - 5 \rm Nm$ is needed to rotate the load, which is $3 - 5\ \%$ of the actuator maximum torque, $100 \rm Nm$. 

\begin{figure}[p!]
\centering
\includegraphics[width=0.95\linewidth] {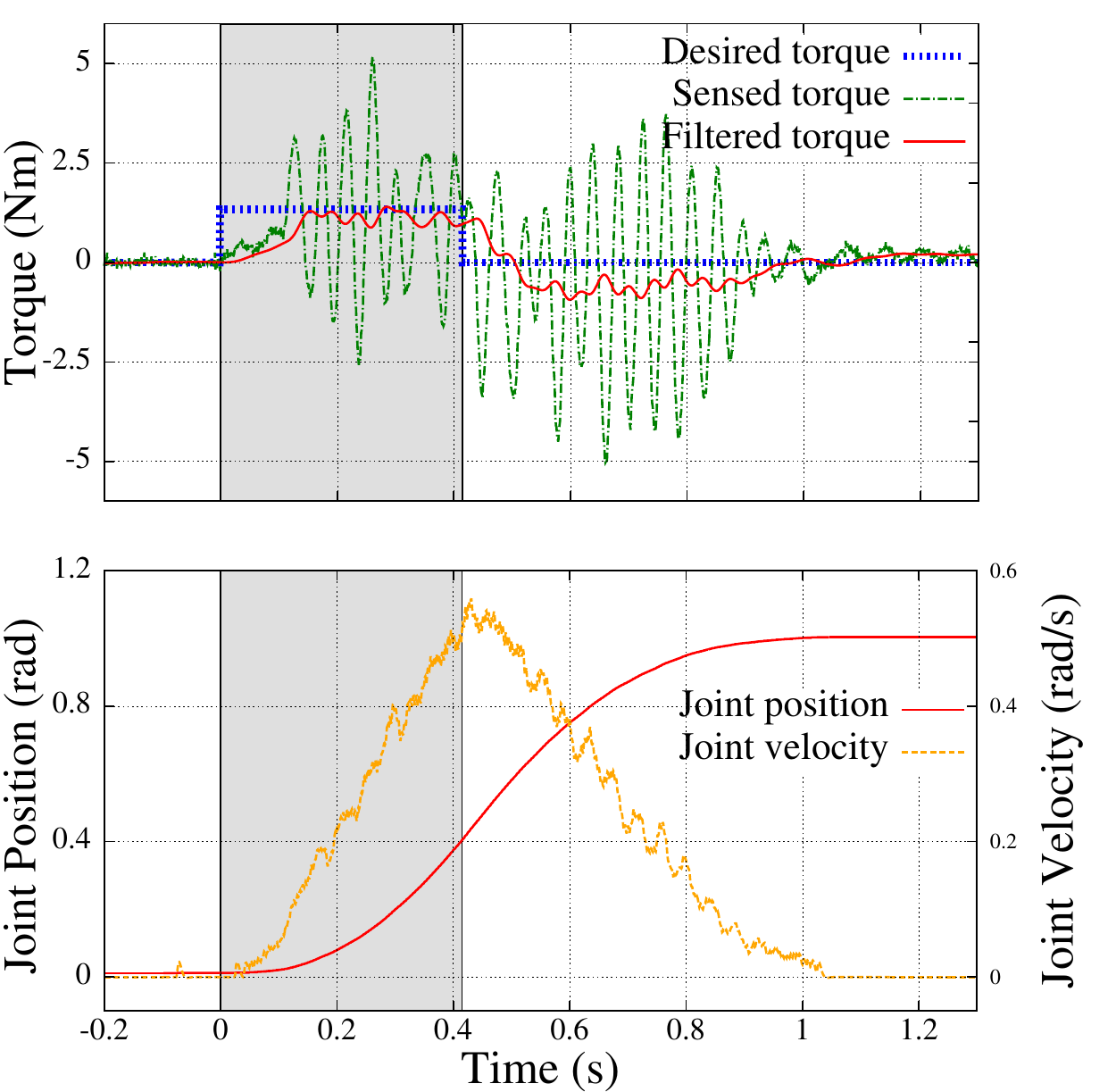}
\caption
[Step Torque Input Response]
{{\bf Step Torque Input Response:}
	A constant desired torque is applied to the controller for $0.4 \rm sec$ (The shaded period). The torque sensor data are filtered by 10Hz cutoff frequency Buttworh low-pass filter to show a trend. }\label{act_torque_01}
\end{figure}
\subsection{Force Control with Step Input in the Testbed}
In this experiment, we applied a constant desired torque to the force controller for $0.4 \rm sec$. 
The graph in Fig. \ref{act_torque_01} shows the result.
The joint angular velocity increases with an approximately constant acceleration during the step torque.   
The joint velocity shows small ripples because the torque sensor data has huge oscillations caused by the Harmonic drive's  torque ripples. Hence, the oscillation only occurs during load movement, and the period is half of the input shaft \cite{lu2012torque}. However, even though the Harmonic drive generates this torque ripples, the torque controller compensates for the effect and makes the system stable.
In the graph, the filtered torque sensor data are filtered by a 10Hz cutoff frequency Butterworth low-pass filter. They show that the controller tried to follow the desired torque. When the desired torque is zero, negative torque sensor data are measured. The negative torque data shows that the friction from the wheel is applied to the drivetrain. That means that the friction is also included in the external force to which the controller should be compliant.

\subsection{Simple Low Impedance Control with Inner Force Controller in the Testbed}
\begin{figure}[p!]
\centering
\includegraphics[width=0.95\linewidth] {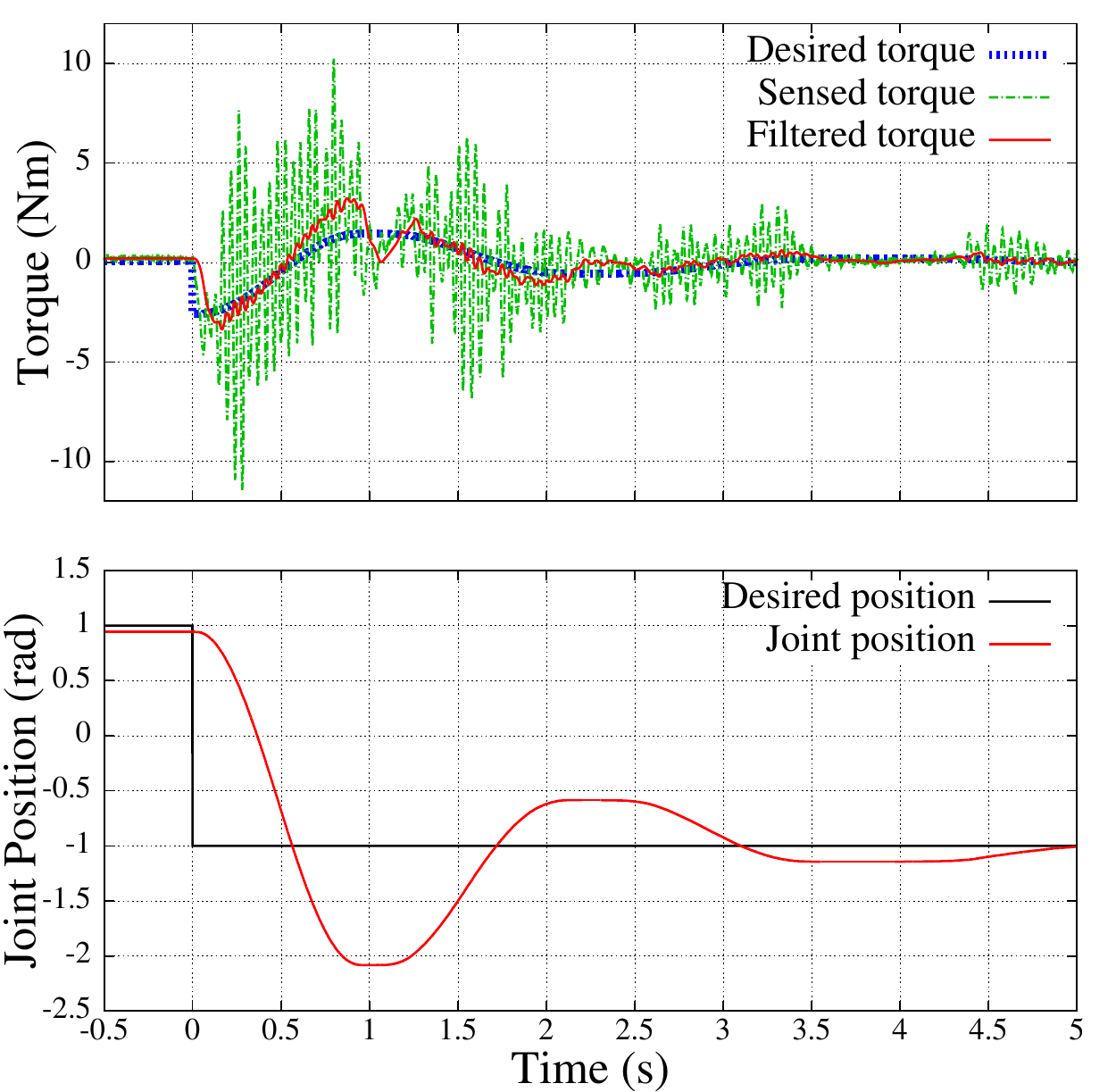}
\caption
[Simple Low Impedance Control]
{{\bf Simple Low Impedance Control}
	shows a low natural frequency of $0.4 \rm Hz$.}\label{act_pos_01}
\end{figure}

In this experiment, we attach an simple low impedance controller to the force controller, and the output of the position controller is fed into the force controller as an input. The system dynamics is simplifed without any damping term.
The controller consists of only a proportional term, which means the outer controller is equivalent to a spring.
The result is shown in Fig. \ref{act_pos_01}.
The joint angle trajectory shows a typical undamped mass-spring-damper dynamics. 
Approximately, we can figure out the equivalent spring stiffness from the frequency of the oscillation, $0.4 \rm Hz$. 
If we assume the frequency is the same as the natural frequency $\sqrt{\frac{k}{m}}$, then the equivalent spring stiffness $k = 4.47 \rm Nm / rad$, which is a low impedance for systems with high friction.

\subsection{External Force Estimation on the Mobile Robot}
\begin{figure}[t]
\centering
\includegraphics[width=0.95\linewidth] {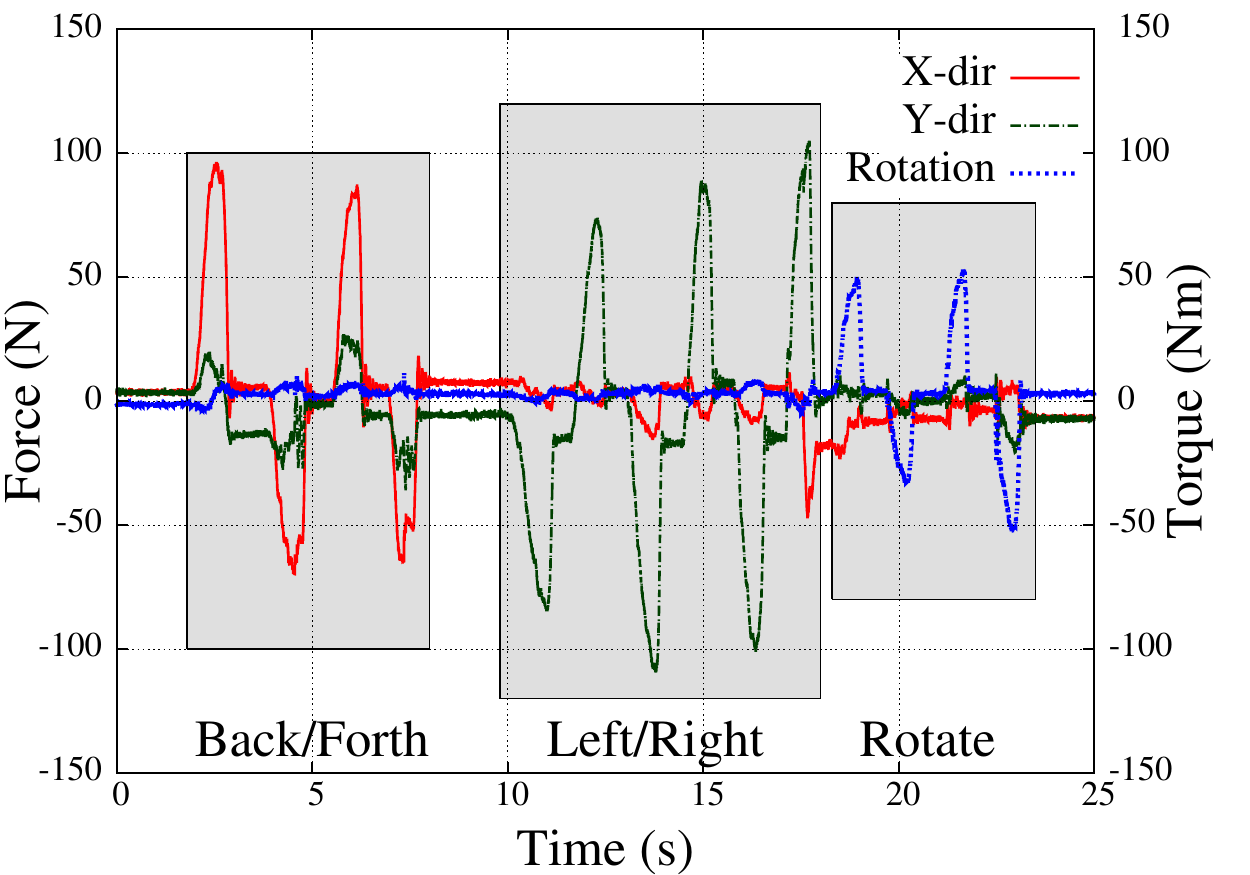}
\caption
[ The Estimated Force/Torque from the Torque Sensors]
{ { \bf The Estimated Force/Torque from the Torque Sensors}
	shows that there are external forces from front, back, left, and right, and external torque rotating the frame.} \label{est_torque}
\end{figure}

Unlike other compliant mobile robots which can sense external forces with their body frame, our mobile robot senses the external force directly from the drivetrain. Therefore, all the external forces in the body can be picked up by the torque sensors (e.g. the external forces normal to the ground cannot be detected). Because the actuation of the robot is holonomic, the external force is estimated easily by the combination of the torques from the torque sensors as follows:
\begin{equation}
\begin{bmatrix}
f_x \\ f_y \\ \tau_z
\end{bmatrix}
=
\frac{1}{r}
\begin{bmatrix}
0 && 
-\cos ( \pi/6) &&
\cos ( \pi/6)  \\
1 && -\sin ( \pi/6) && -\sin ( \pi/6) \\
R && R && R
\end{bmatrix}
\begin{bmatrix}
\tau_0 \\ \tau_1 \\ \tau_2
\end{bmatrix}
\end{equation}
where $\tau_k$ is the measured torque from the torque sensor on k-th drivetrain, and $f_x$, $f_y$, $\tau_z$ are the external force from the front, the external force from the side, and the rotating external force, respectively. $r$ is the radius of each wheel, and $R$ is the distance from the center of the robot to the wheels.
We apply force manually from different directions and measure the torque data while the mobile robot is not operational. The result shown in Fig. \ref{est_torque} proves that the torques sensors located on the drivetrains can detect the external forces precisely.

\subsection{Force Feedback Based Compliance of Trikey}
\begin{figure}[t]
\centering
\includegraphics[width=0.95\linewidth] {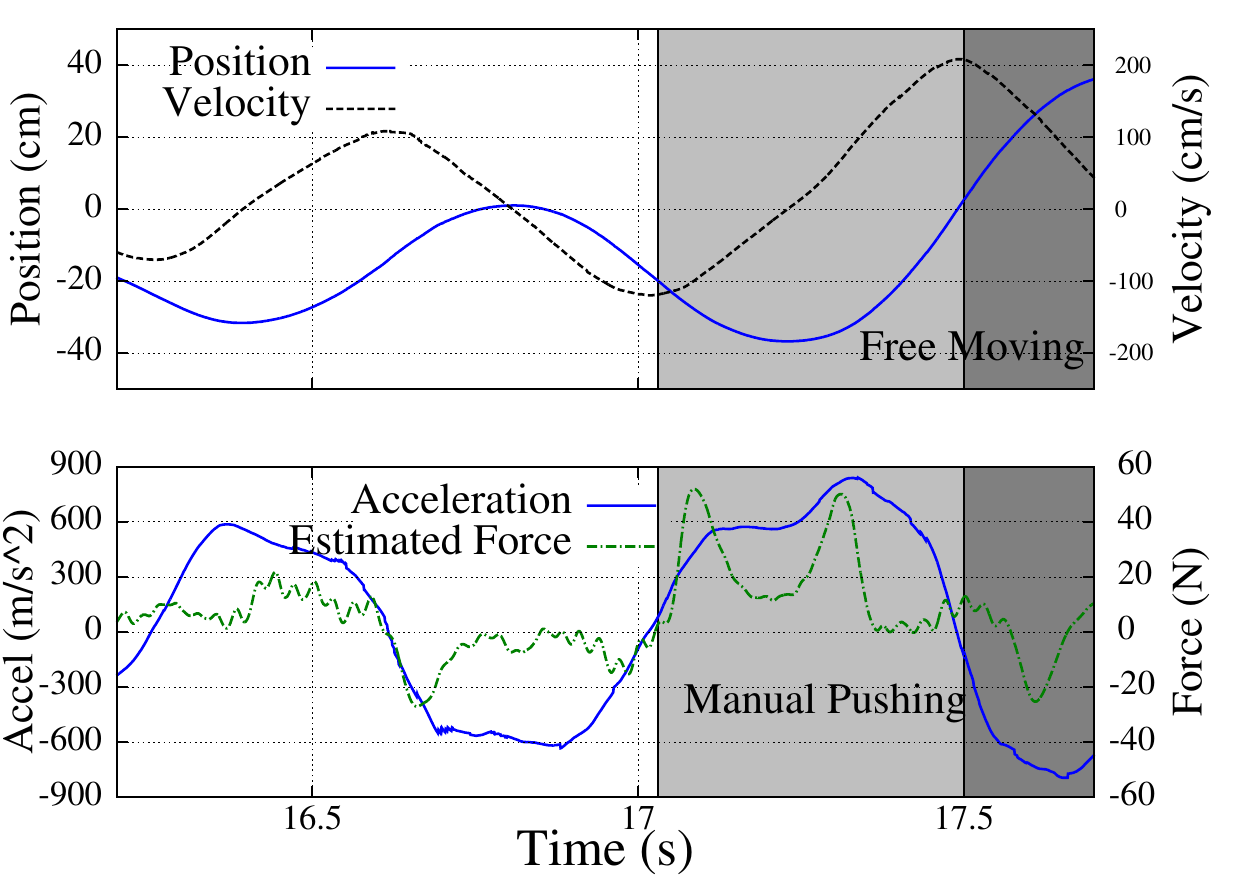}
\caption
[ The Estimated Force and the Odometry during the Compliant Motion]
{ { \bf The Estimated Force and the Odometry during the Compliant Motion:}
	The position of the robot is measured from the motion capture system. The estimated force is estimated from the sensed torque data. The velocity and the acceleration are derived from the position.} \label{est_torque_2}
\end{figure}

We also conduct the zero force control experiment that we did in \ref{zfe} in the mobile robot. We run the actuators in the mobile robot in zero force control mode.
The motion capture system measures the position and the orientation of the robot.
We conduct one directional compliance test. All the actuators run the zero force control previously described and the user applies external forces by pushing and pulling the base.
The result in Fig. \ref{est_torque_2} shows that the acceleration and the estimated force have a similar trend. Also, the mobile base shows compliant movement. However, we did not completely remove the passivity that originates from the robot mass and the friction between the ground and the wheels, so the degree of compliance still needs to be improved.

\begin{figure*}[ht]\centering
\begin{subfigure}[Colliding with compliance]
{
\includegraphics[width=0.95\linewidth] {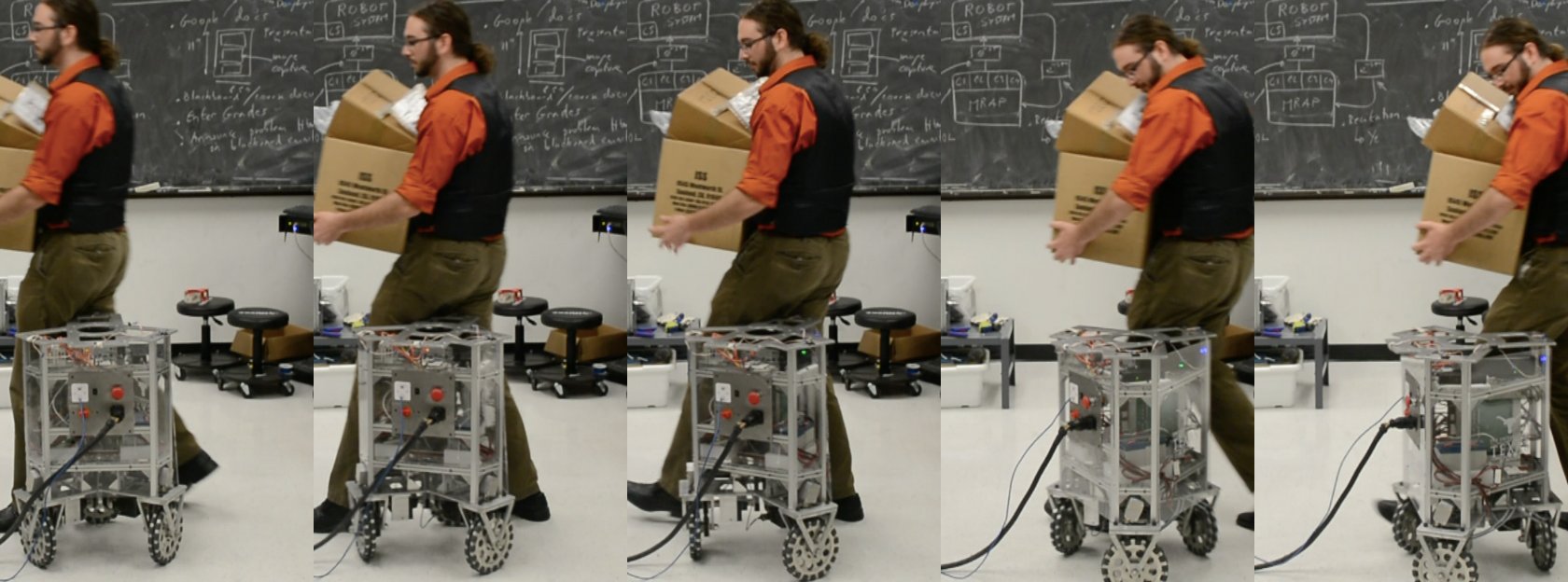}
}
\end{subfigure}
\begin{subfigure}[Colliding without compliance]
{
\includegraphics[width=0.95\linewidth] {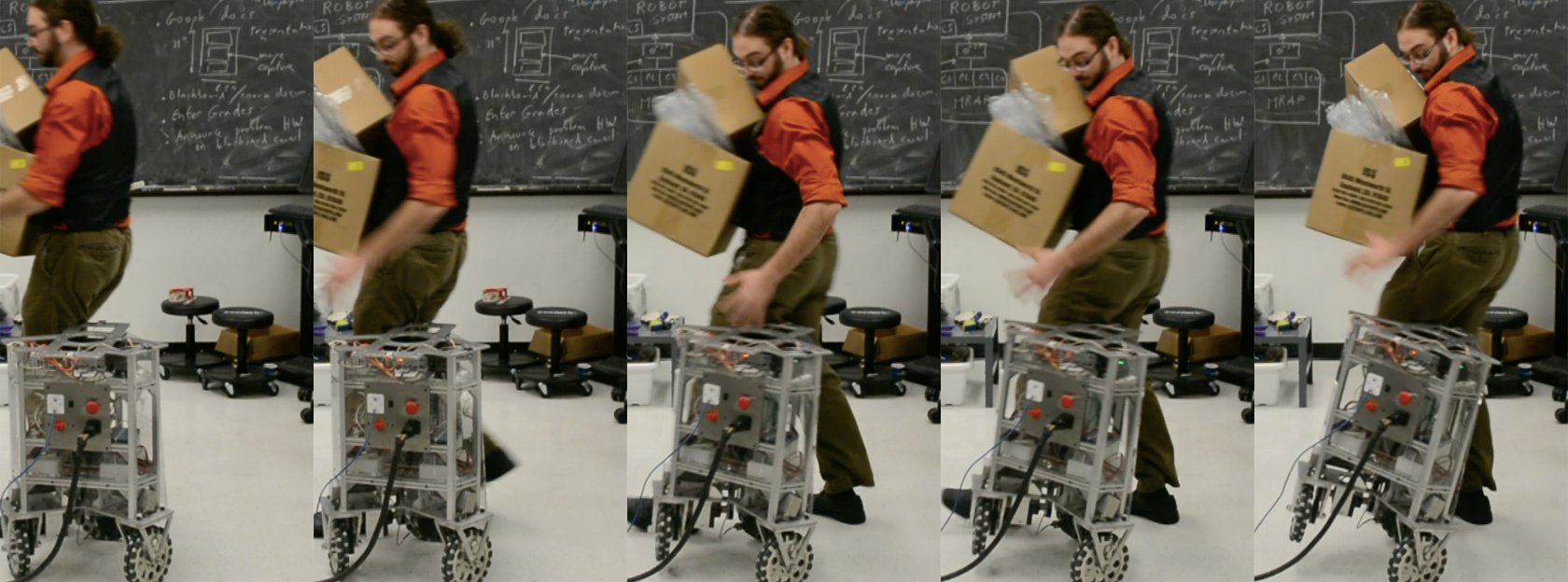}
}
\end{subfigure}
\begin{subfigure}[Passing the robot]
{
\includegraphics[width=0.95\linewidth] {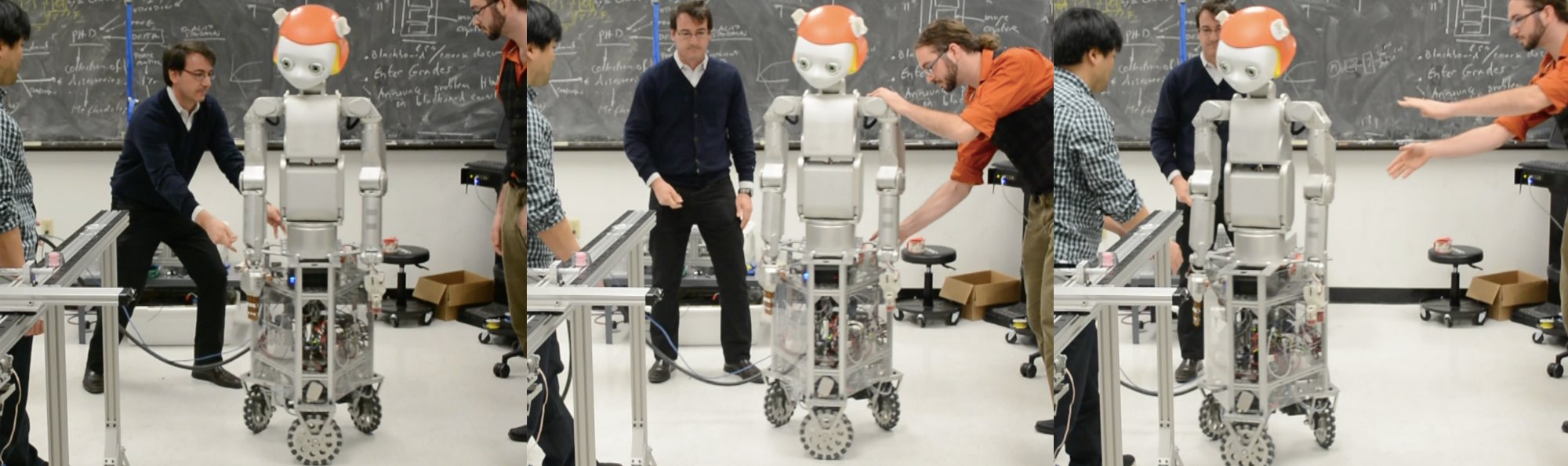}
}
\end{subfigure}
\caption
[ Experiments on Compliant Mobile Robot]
{ {\bf Experiments on Compliant Mobile Robot:}
	a compliant mobile base is implemented by the zero force controller in the actuators.
(a) The mobile robot is compliant during an unexpected collision with a human.
(b) The mobile robot tips over due to a human collision while the force controller is disabled.
(c) The compliance is omni-directional, so the robot can be passed in any direction.
} \label{mobile_base}
\end{figure*}

We also made a trial of involving human-robot collision test. We plan an unexpected collision scenario, conduct it, and observe the collision. Fig. \ref{mobile_base} shows that the mobile robot becomes more stable with the compliance when it collides with human. Also, to show its omni-directinoal compliance, we push the robot in different directions.

\section{Discussion and future work}
We have retrofitted the holonomic Trikey robot with torque sensing to allow a fully omnidirectional compliant behavior, the first of its kind. 
We have implemented a predictive controller on the embedded hardware in order to improve the reaction time and bandwidth of our force controller. 
And we have carried out experiments to demonstrate the safety and compliance of the Trikey sytem, both alone and serving as a base for a humanoid upper body. 
Compliance is one of the most important features necessary for mobile robots to safely cooperate with humans in human environments.

Our work does however leave some unaddressed questions. 
To achieve our ultimate goal, safe mobile robots, we still need to pursuit more 

For instance, while our force controller effectively hides the passivity of the actuator friction, the friction of the robot on the wheel bearings still causes the system to be highly damped. 
Friction modelling should be able to allow less damped behavior in the future. 

Another goal is to improve over compliant zero force control by implementing impedance control. 

Also, since Trikey's actuators are rigid, it relies on high bandwidth control to mitigate the damage to a person in the event of a collision, but is unable to improve the highest frequency impedance without the addition of a lower impedance component between it and the human.  In our experiments the high frequency impedance is low pass filtered by soft tissue acting as a damped spring-mass system, but to make the human robot interaction experience less painful, soft foam will be added as future work. 

Finally, there is an opportunity to improve our force controller by modelling and predicting the torque ripples effect endemic to Harmonic drives.

%% file: chapter-auro.tex
\chapter{Full-Body Collision Detection and Reaction with Omnidirectional Mobile Platforms: A Step Towards Safe Human-Robot Interaction}
\label{chap:auro}

\section{Introduction}
\blfootnote{This chapter has been published in {\it Autonomous Robots} \cite{kim2016}. Travis Llado contributes the design and manufacture of the spring dumper shown in Fig. \ref{setup}.}
As mobile robots progress into service applications, their environments become less controlled and less organized compared to traditional industrial use. In these environments, collisions will be inevitable, requiring a thorough study of the implications of this type of interaction as well as potential solutions for safe operation. With this in mind, we are interested in characterizing the safety and collision capabilities of statically stable mobile bases moving in cluttered environments. The work presented here is the first of which we are aware to address, in depth, the mitigation of the effects of collisions between these types of sizable robots and objects or people.

The majority of work addressing mobility in cluttered environments has centered around the idea of avoiding collisions altogether. However, collisions between robotic manipulators and objects and humans have been investigated before \cite{Haddadin2009,Yamada1996}. Push recovery in humanoid robots allows them to regain balance by stepping in the direction of the push \cite{Pratt2006} or quickly crouching down \cite{Stephens-Thesis:2011}. Inherently unstable robots like ball-bots \cite{Nagarajan2013,Kumagai2008} and Segways \cite{Nguyen2004} have been able to easily recover from pushes and collisions using inertial sensor data. A four-wheel robotic base with azimuth joint torque sensors \cite{Fremy2010} has been able to respond to human push interactions, but only when its wheels are properly aligned with respect to direction of the collision. Also, a non-holonomic base with springs on the caster wheels was recently developed \cite{Kwon2011} and reported to detect pushes from a human, but with very preliminary results and without the ability to detect forces in all directions or detect contacts on the wheels themselves. In this work, we focus on non-stationary robots, as opposed to fixed base manipulators.
In the field of non-stationary robotic systems, such as statically or dynamically balancing mobile bases and legged robots, one of the key deficiencies is the availability of collision reaction methods that can be used across different platforms. Dynamically balancing mobile bases and humanoid robots rely on IMU sensing to detect the direction of a fall and then regain balance along that direction. However, this type of method is limited to robots which naturally tip over at the slightest disturbance.

The main objective of this chapter is to develop general sensing and control methods for quickly reacting to collisions in statically stable mobile bases. 
Specifically, we develop methods that rely on joint level torque sensing instead of inertial measurement sensing to determine the direction and magnitude of the collision forces. If IMUs were used, accelerations would only be sensed accurately once the robot overcomes static friction which, for a sizable robot, could be quite large. Torque sensors, which are mounted next to the wheels, can quickly detect external forces sooner than IMUs and therefore are more suitable for quick collision response. Equally important is the fact that statically stable mobile bases can move in any direction or not at all in response to a collision, whereas dynamically balancing mobile bases and humanoid robots must move in the direction of the collision. This ability makes statically stable mobile bases more flexible when maneuvering in highly constrained environments.

To provide these capabilities, we take the following steps: (1) we develop a floating base model with contact and rolling constraints for an omnidirectional mobile base; (2) we process torque sensor signals using those models and statistical techniques; (3) we estimate roller friction and incorporate it into the constrained dynamics; (4) we implement a controller to quickly escape from the collisions; (5) we present an experimental testbed; and (6) we perform experiments including several calibrated collisions with the testing apparatus, and a proof of concept experiment in which the robot moves through a cluttered environment containing people against whom it must safely collide.

Overall, our contributions are (1) developing the first full-body contact sensing scheme for omnidirectional mobile platforms that includes all of the robot's body and its wheels, (2) being the first to use floating base dynamics with contact constraints to estimate contact forces, and (3) being the first to conduct an extensive experimental study on collisions with human-scale mobile bases.

\begin{figure}[h]\centering
\includegraphics[width=0.6\linewidth, clip=true ] {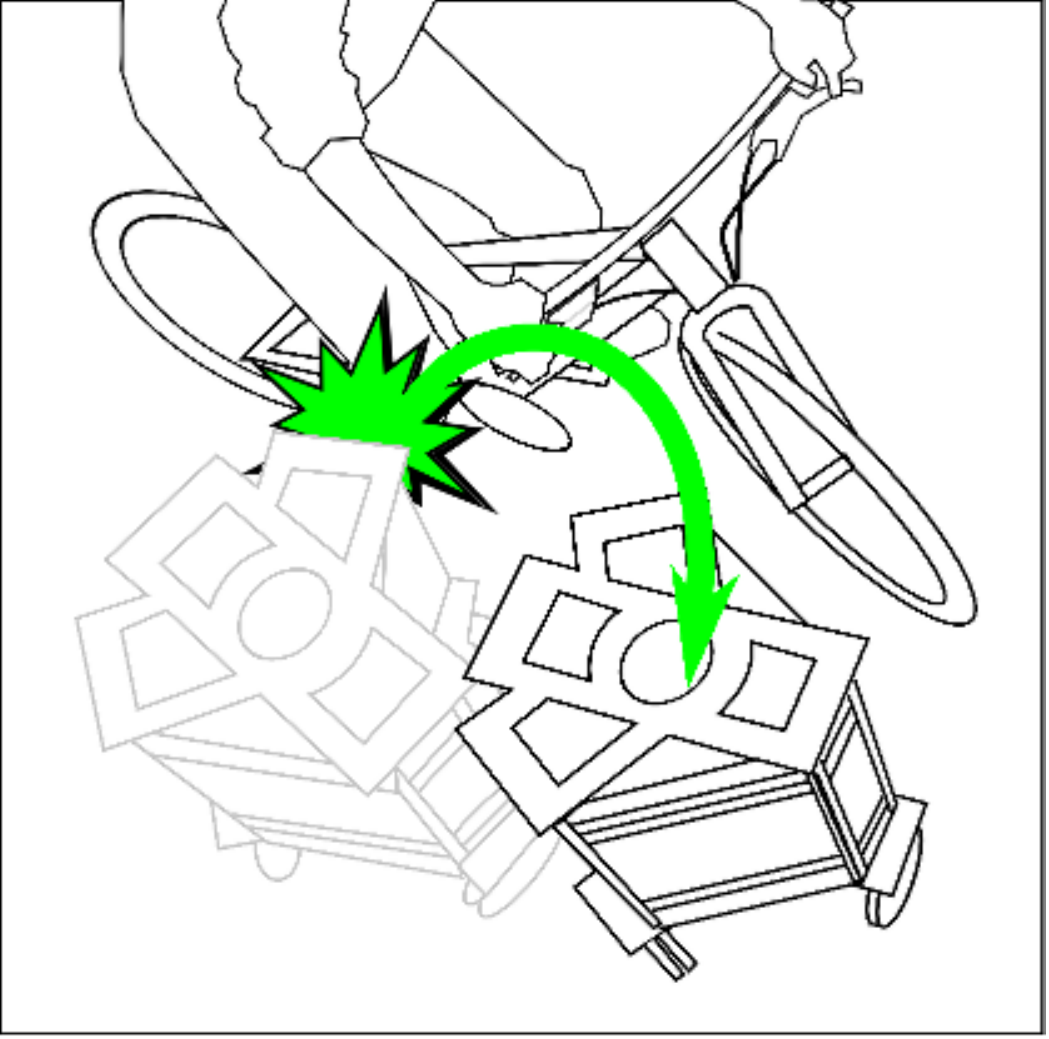}
\caption[Concept]{{\bf Concept:} unexpected collision between a robot and a person on a bicycle, as presented in our supporting video.}
\label{concept}
\end{figure}

\section{Related Work}
\vspace{0.15in}

\subsection{Mobile Platforms with Contact Detection}

To be compliant to external forces, mobile robots have adopted various sensing techniques. One simple way to detect external forces is by comparing actual and desired positions \cite{Kim2013} or velocities \cite{Doisy2012}. This method is easy to implement because it can use the built-in encoders on the robot joints or wheels to detect external forces. However, the ability to detect contacts using this method depends largely on the closed-loop impedance chosen for the control law. 

Another means of detecting external forces is physical force\slash torque sensors such as strain gauges or optoelectronics. This approach has been used in many mobile platform applications such as anticipating user intention with a force-based joystick \cite{Sabatini2002}, developing a handle with force\slash torque sensing capabilities \cite{Spenko2006}, implementing an impedance control law based on force\slash torque sensed on a handle \cite{Chuy2007}, and quantifying user intent and responding with an admittance controller based on a force\slash torque sensor mounted on a stick \cite{Huang2008,Wakita2013}. However, all of these methods rely on detecting forces and torques at a specific location, such as on a handle, or joystick. When the user interacts or collides with other parts of the robot's body, such robots will not be able to respond to the applied forces safety. 

In \cite{Hirata2003}, a force\slash torque sensor measures forces between the mobile robot's body and an external protection cover, providing partial safety, but collisions against the wheels cannot be detected. In \cite{Fremy2010} they introduce a quasi omnidirectional mobile robot that is compliant to external forces by measuring torques on the yaw joints of its caster wheels. This technique can detect collisions on the wheels like ours, but suffers from singularities which limit both the directions in which it can detect force and its freedom of motion. In \cite{Kwon2011} a sensorized spring system is installed on the frame of a mobile base with caster wheels and is used for push interactions. However, the base can respond to forces only in limited directions and is once more insensitive to collisions against the wheels. 

Other sensing properties have been used for contact interactions, notably the tilt measured by an inertial measurement unit on ball-bot robots \cite{Nagarajan2013}. This type of robot, and the associated inertial sensing, have been used effectively to handle contact interactions with people \cite{Kumagai2008}. However, the main drawback of this method is that the robot must move in the direction of the disturbance or it will fall over. In contrast, non-inertial force sensing techniques like ours allow a robot to react in any direction upon collision or force interaction. This ability might be very useful when producing planned movements tailored to the external environment. 

\subsection{Contact Detection via Joint Torque Sensing}

Several existing studies use joint torque sensing to detect contact, like us, but only address serial robotic manipulators. Note that this technique is distinct from the commonly used multi-axis force\slash torque sensor located at the end effector of a manipulator. Many researches have investigated sensing external forces on all parts of a manipulator's body using distributed joint torque sensors~\cite{Wu1985,Luca2006}. 

Like our method, this indirect external force sensing requires estimation that considers dynamic effects such as linkage and motor masses, inertias, momentum, gravitational effects, and friction. Statistical estimation methods~\cite{Fang2011} are used to estimate external forces based on joint torque sensing~\cite{Le2013}. These methods have inspired our research, but we note that we have taken similar approaches for a mobile platform instead of for a robotic manipulator. A mobile platform has different dynamics because it has a non-stationary base and its wheels are in contact with the terrain. Such differences imply different dynamic models and modifications of the estimation methods.

\subsection{Safety Analysis in Robotics}

Pioneering work on safety criteria for physical human-robot interaction are provided in~\cite{Yamada1996}. In particular, curves of maximum tolerable static forces and dynamic impacts on various points of the human body are empirically derived. A method to detect external forces using motor current measurements and joint states is proposed, and a viscoelastic skin is utilized to dampen impacts.

In \cite{Zinn2004} the positive effects on safety of actuators with a series elastic compliance are brought up but linked to lower performance. A double macro-mini actuation approach is proposed to accomplish safe operation while maintaining performance, and the automotive industry's Head Injury Criterion index is used to demonstrate the benefits of this approach in terms of safety.

A comprehensive experimental study on human-robot impact is conducted in \cite{Haddadin2009}. This study suggests that the Head Injury Criterion is not well suited for studying injuries resulting from human-robot interaction. Instead, the authors propose contact forces acting as a proxy to bone fractures as their injury indicator. The low output inertia achievable with their torque control manipulators is shown to be highly conducive to preventing injury during collisions.

Also relevant to our work is the study considering child injury risks conducted in \cite{Fujikawa2013}. Extensive experimental data is obtained from a 200Kg mobile robot moving at speeds of 2Km/h and 6Km/h and colliding against a robot child dummy fixed to a wall. The head injury and neck injury criteria are used to study the consequences of the impacts, and the severity of injury is expressed by the Abbreviated Injury Scale. Those criteria are reinforced with analysis of chest deflection for severity evaluation. In contrast with our work, their mobile platform is uncontrolled and does not have the ability to sense contact. This study is focused purely on impact analysis instead of contact sensing and safe control.

\subsection{Model-Based Control of Omnidirectional Platforms}

A mobile platform colliding or interacting with the environment is not only affected by external forces, but also by static and dynamic effects such as the robot's inertia, its drivetrain and wheel friction, and other mechanical effects. \cite{Zhao2009} considers a simulated system consisting of a 6-DOF omnidirectional mobile robot with caster wheels, and addresses the modeling and control of motion and internal forces in the wheels. \cite{Djebrani2012} derives the dynamic equation including the rolling kinematic constraint for a mobile platform similar to ours, but uses an oversimplified dynamic friction model with respect to the effects of roller friction. Studies that incorporate static friction models include~\cite{Viet2012,BarretoS.2014}, but again these use oversimplified models that ignore omniwheel and roller dynamics. The studies above are mostly theoretical, with few experimental results.

\section{System Characterization}
\label{sec:syschar}

\subsection{Hardware Setup}

To perform experimental studies on human-robot collisions, we have built a series of capable mobile platforms. This study uses the most recent.
We began designing mobile bases to provide omnidirectional rough terrain mobility to humanoid robot upper bodies~\cite{sentis2012}. The newest iteration of our platform, produced in~\cite{Kim2013}, replaced the previous drivetrain with a compact design that minimized backlash by using belts and pulleys. Rotary torque sensors in the drivetrain and harmonic drives on the actuators were incorporated into the base in ~\cite{Kim2014}, enabling accurate force feedback control for impedance behaviors. The electronics in the current system improve over that of ~\cite{Kim2014} in that the once centralized torque sensor signal processing is now divided into each actuator's DSP in order to minimize electrical crosstalk. This study is the first study that uses the torque sensors on the hardware base for full-body model-based estimation of the contact forces.

Rotary torque sensors in the wheel drivetrains produce the unique feature of our base: full-body contact estimation on all parts of its body, including any part of the wheels. An alternative would have been to cover all of a robot's body with a sensitive skin, but this option would have left the wheels uncovered and therefore unable to detect contact. We note that the wheels are often the first part of the base that collides with unexpected objects. Therefore, our solution with three rotary torque sensors in the wheel's drivetrain is the first and only one of which we are aware that can respond to collisions on all parts of the mobile platform. Additionally, the harmonic drives and belt-based drivetrain of the base minimize backlash and therefore achieve more accurate force sensing. 

\begin{figure}\centering
\includegraphics[width=0.9\linewidth, clip=true ] {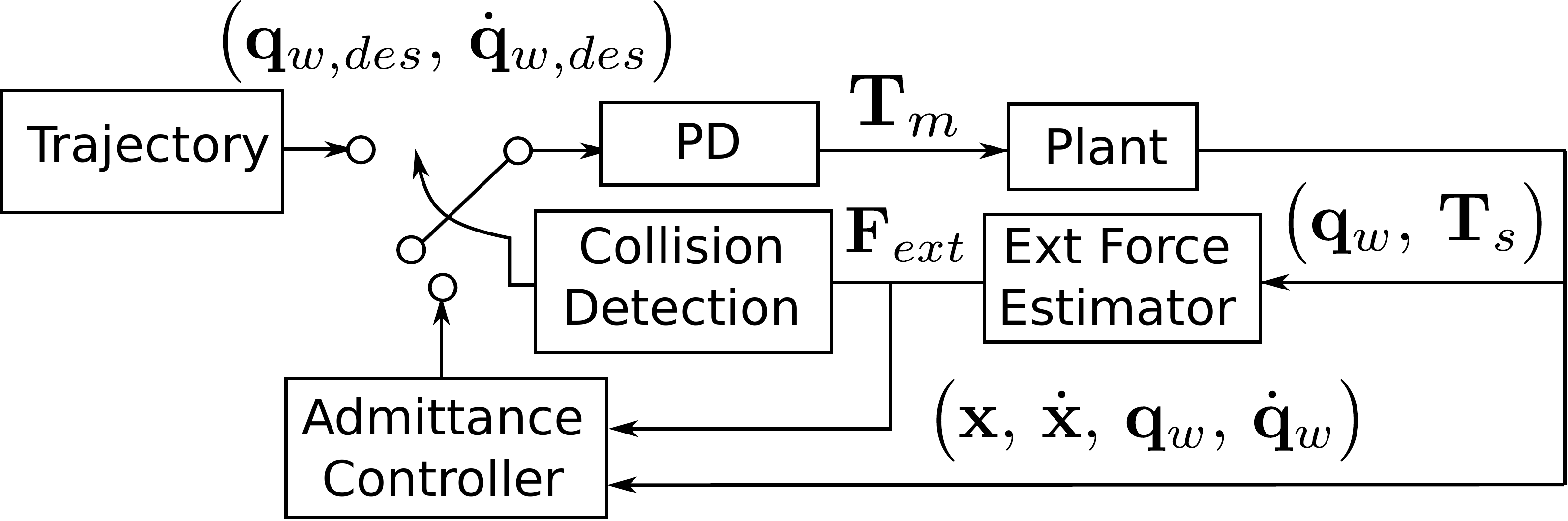}
\caption[Control Diagram]{{\bf Control Diagram} showing how estimated external force, $\mathbf{F}_{ext}$, is fed through the collision detector and ultimately determines the position controller's input. When the robot detects a collision it uses an admittance controller in place of its usual trajectory to escape the contact as fast as it can safely move.}
\label{controller}
\end{figure}

\subsection{Safety Controller Design}

When a mobile base collides with people, two cases can be previously distinguished: In unconstrained collisions a person can be pushed away, whereas in fixed collisions the person is pushed against a wall. In either scenario our robot moves away from the collision as quickly as possible to mitigate injury. 

Fig. \ref{controller} shows our proposed control architecture for detection of and reaction to collisions. Under normal circumstances, the controller tracks a trajectory given by a motion planner or sensor-based algorithm. When an external force breaches our contact threshold, the controller switches on an admittance controller. This admittance controller generates a trajectory that responds to the sensed external force and rapidly leads the robot away from the contact. We tested both an impedance and an admittance controller in this role during the course of our research, but found the admittance controller to be more responsive.

\begin{figure} \centering
\includegraphics[width=\linewidth, clip=true]{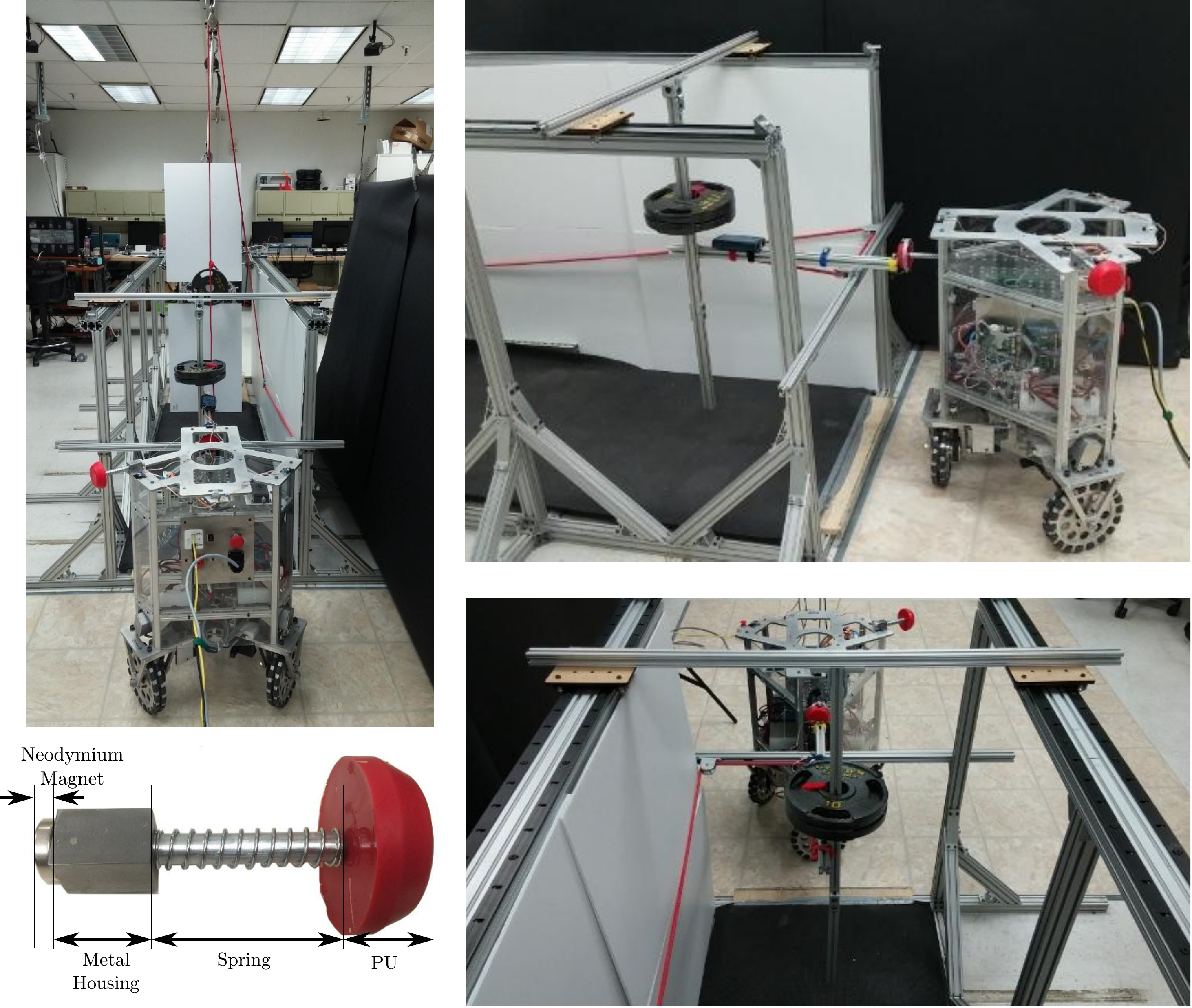}
\caption
[Collision Testing Apparatus]
{{\bf Collision Testing Apparatus} simulates human contact using a 10kg mass on a slider. This one degree of freedom system is accelerated via a second weight hanging from an elaborate pulley system, and can also be used to apply a static force. Motion capture markers attached to the slider and the PU bumper are used to measure their position.}
\label{setup}
\end{figure}

\subsection{Reaction to Collisions}

The admittance controller is designed to provide compliance with respect to the external force. The desired dynamics can be expressed as
\begin{align}\label{eq:simpledyn}
M_{des} \ddot{x} + B_{des} \dot{x} = F_{ext,x}(t),
\end{align}
where $M_{des}$ and $B_{des}$ are the desired mass and damping of a virtual compliant system, and $F_{ext,x}(t)$ is the time dependent force disturbance applied to the system. Assuming the external force is close to a perfect impulse, i.e. a Dirac delta function, the above equation can be solved to produce the desired trajectory,
\begin{equation}\label{eq:traj}
x\left(t\right) = x_0 + \frac{F_{ext,x}}{B_{des}} \left( 1 - e^{-B_{des}/M_{des} \, t }\right),
\end{equation}
where $x_0$ is the position of the system when the collision happens.
An identical admittance controller operates on the $y$ degree of freedom.

Our controller attempts to maintain constant yaw throughout the collision, i.e.
\begin{equation}
\theta(t) = \theta_0.
\end{equation}
Combining the three degrees of freedom, we write the robot's full trajectory as
\begin{equation}
\mathbf x_{des}(t) = \big(x_{des}(t), \;\;\; y_{des}(t), \;\;\; \theta_{des}(t) \big)^T.
\end{equation}
This trajectory is differentiated and then converted into a desired joint space trajectory using the constrained Jacobian, $\mathbf J_{c,w}$ given in Eq. (\ref{eq:jcw}), i.e.
\begin{gather}
\dot{\mathbf q}_{w,des}(t) = \mathbf J_{c,w}(t) \; \dot {\mathbf x}_{des}(t),\\[1mm]
{\mathbf q}_{w,des}(t) = {\mathbf q}_{w,des}(t_0) + \int_{t_0}^t \dot{\mathbf q}_{w,des}(\tau) d\tau,
\end{gather}
and fed to the PD controller of Fig. \ref{controller} to achieve the intended impedance behavior.

\subsection{Collision Testbed}

To assess the safety of our mobile platform, we constructed a calibrated collision testbed. Following the collision test procedure used in the automotive industry \cite{UNECE2011}, we chose a 10kg mass as our leg-form test dummy. The collision dummy is attached to a sliding system which provides a single degree of freedom for impact, and is accelerated by a free falling weight. In Fig. \ref{setup} we illustrate details of the test environment. The absolute positions of the dummy and the mobile base are measured by the Phase Space motion capture system described in \cite{Kim2013}. Four markers on the mobile base measure its position and two markers on the dummy measure its linear motion. 

\subsection{Stiction-Based Bumper}
\label{material}

The\, time \, requirement\,  for our base to detect collision and reverse direction is roughly one hundred milliseconds. Keeping the collision time brief works to reduce injury, but is insufficient to eliminate it\, altogether.\  Though it is impractical to fully pad a robot, some padding can drastically reduce the collision forces due to collision with specific parts of the robot's body. Yet reducing the forces makes the problem of detecting the collision more difficult, and increases the amount of time before the robot acknowledges an impact. We have designed a one DOF springloaded bumper with a relatively long travel to study the design of safe padding for omnidirectional robots. This design features a magnetic lock at peak bumper extension, which works to allow earlier detection of a collision, while simultaneously reducing the overall maximum impact force. Details of the bumper can be found in Fig. \ref{setup}.


\begin{figure}\centering
\includegraphics[width=0.65\linewidth, clip=true ] {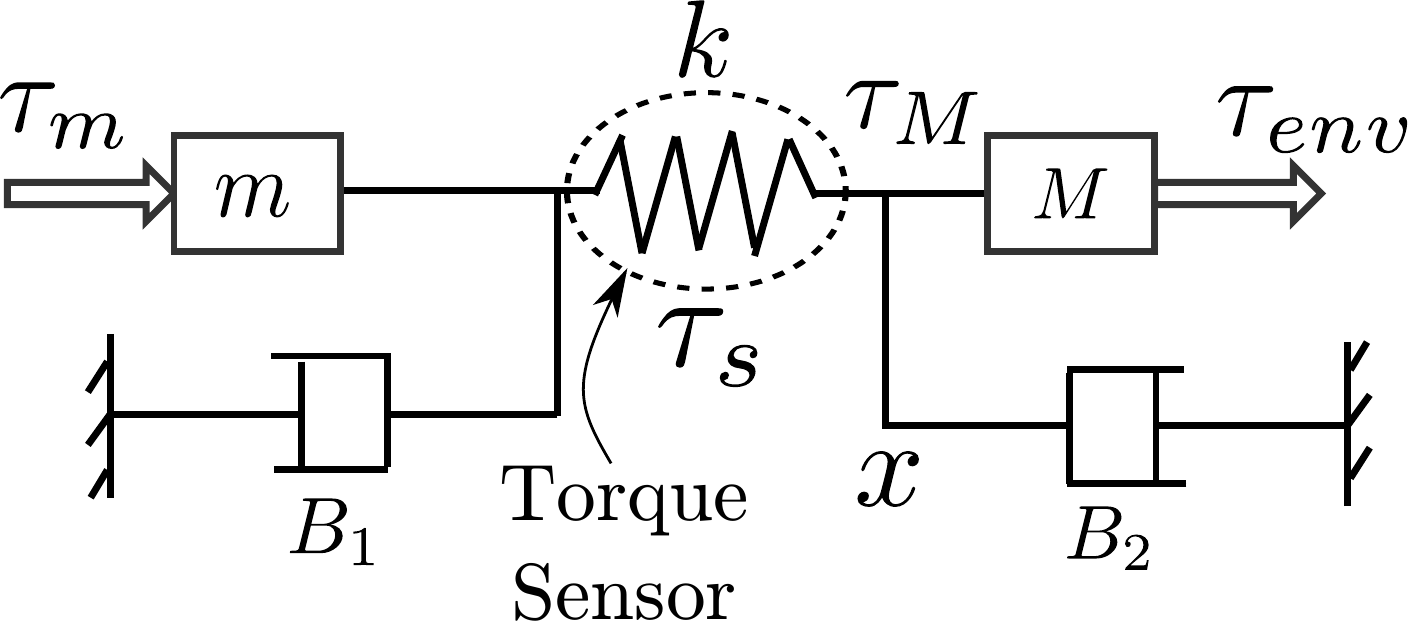}
\caption
[Actuator Model]
{{\bf Actuator Model} including the torque sensor, modeled as a spring. The two masses $m$ and $M$ represent the motor inertia, reflected through the gear system, and the load mass. Motor side friction and load side friction are expressed as the damping terms $B_1$ and $B_2$, respectively.}
\label{model} 
\end{figure}

\section{Full-Body External Force Estimation}
\label{sec:forceestimation}
To estimate external forces based on drivetrain torque sensing, we rely on a model of the actuators, and on the robot's kinematics and dynamics. The constrained kinematic mapping between the base's motion and wheel motion is used to find the base and omniwheel roller velocities based on measured wheel velocities. The actuator model provides a mapping between motion and expected torque sensor values in the absence of external forces. This model is trained empirically to better estimate the friction in the omni-wheel rollers. Ultimately the position, magnitude, and direction of the applied external forces is estimated based on the deviation of the observed wheel torques from those predicted by the force free model, and the kinematics are again invoked to transform this into the Cartesian frame.

To build an intuition of our method for estimating external forces, consider the single actuator system shown in Fig. \ref{model}. In this system, the torque sensor is modeled as a torsional spring, with spring constant $k$, and its displacement is proportional to the torque applied to the sensor. The spring is compressed or extended through the combined action of the motor, the wheel's inertia, and the external environment. Some of the important variables include the motor's torque, $\tau_m$, its rotor's mass, as reflected through mechanical gearing, $m$, the gear friction, $B_1$, the load's mass (i.e. the wheel, or the robot itself in the constrained case), $M$, the friction between the wheel and the external environment, $B_2$. But most importantly, the torque $\tau_{env}$ includes the effect of the wheel traction on the floor and any possible external collision with objects or people,
\begin{equation}
\tau_{env} = \tau_{trac} + \tau_{ext}.
\end{equation}
Assuming that the effect of the wheel traction, $\tau_{trac}$ can be modeled, our goal is to estimate the external forces, $\tau_{ext}$, based on observed sensor torque $\tau_s$:
\begin{equation}\label{extest}
\tau_{ext} = -\tau_{trac} + B_2 + M \ddot x - \tau_s.
\end{equation}
This method can then be applied to the estimation problem of the full base by using the kinematic constraint relationships between the wheels and the ground.

\subsection{Torque Output Dynamics}

To derive wheel and roller kinematics, we consider a planar scenario where the wheel moves omnidirectionally on a flat floor. In~\cite{Kim2013} we developed the following equations relating the contribution of the $i^{th}$ wheel's angular velocity, $\dot q_{w,i}$, and their omniwheel roller's angular velocity, $\dot q_{r,i}$, to the Cartesian velocity of the robot with respect to a fixed inertial frame, $\dot x$ and $\dot y$:
\begin{gather}
\dot{x} = r_r \dot{q}_{r,i} \cos \left( \theta + \phi_i \right) - \left( r_w \dot{q}_{w,i} - R \dot{\theta} \right) \sin \left( \theta + \phi_i \right),\label{const_eq1}\\[-2mm]
\dot{y} = r_r \dot{q}_{r,i} \sin \left( \theta + \phi_i \right) + \left( r_w \dot{q}_{w,i}  - R \dot{\theta} \right) \cos \left( \theta + \phi_i \right).\label{const_eq2}
\end{gather}
Where, $\theta$ is the absolute orientation of the robot's body, $R$ is the distance from the center of the robot's body to the center of the wheel, $r_w$ and $r_r$ are the radii of the wheels and their passive rollers, respectively, and $\phi_i$ is the angle from a reference wheel to the i-th wheel in sequential order, i.e. 0$^\circ$, 120$^\circ$, or 240$^\circ$. The kinematics of $\dot q_{w,i}$ and $\dot q_{r,i}$ are obtained from Eq. (\ref{const_eq1}) 
\begin{align}
&r_w \dot{q}_{w,i}= - \dot{x} \sin\left( \theta + \phi_i \right) + \dot{y} \cos \left( \theta + \phi_i \right) + R \dot{\theta}, \label{const_eq3}\\[1.5mm]
&r_r \dot{q}_{r,i}=\dot{x} \cos \left( \theta + \phi_i \right) + \dot{y} \sin \left( \theta + \phi_i \right). \label{const_eq4}
\end{align} 
Expressing these equations in matrix form,
\begin{gather}\label{eq:dotqw}
\mathbf{\dot q}_w = \mathbf J_{c,w} \, \mathbf{\dot x},\\[1.5mm]\label{eq:dotqr}
\mathbf{\dot q}_r = \mathbf J_{r,w} \, \mathbf{\dot x}
\end{gather}
where
\begin{gather}\label{eq:jcw}
\mathbf{J}_{c,w} \triangleq 
\frac{1}{r_w}
\begin{pmatrix}
-\sin\left( \theta + \phi_0 \right) &&& \cos\left( \theta + \phi_0 \right) &&& R \\
-\sin\left( \theta + \phi_1 \right) &&& \cos\left( \theta + \phi_1 \right) &&& R \\
-\sin\left( \theta + \phi_2 \right) &&& \cos\left( \theta + \phi_2 \right) &&& R
\end{pmatrix} \in \mathbb{R}^{3\times 3}, \\\label{eq:jcr}
\mathbf{J}_{c,r} \triangleq 
\frac{1}{r_r}
\begin{pmatrix}
\cos\left( \theta + \phi_0 \right) &&& \sin\left( \theta + \phi_0 \right) &&& 0 \\
\cos\left( \theta + \phi_1 \right) &&& \sin\left( \theta + \phi_1 \right) &&& 0 \\
\cos\left( \theta + \phi_2 \right) &&& \sin\left( \theta + \phi_2 \right) &&& 0
\end{pmatrix}\in\mathbb{R}^{3\times 3},
\end{gather}
are the Jacobian matrices, $\mathbf{q}_w \triangleq (q_{w,0}, \;q_{w,1}, \; q_{w,2})^T$, $\mathbf{q}_r \triangleq (q_{r,0}, \;q_{r,1}, \; q_{r,2})^T$, and $\mathbf{x} \triangleq (x, \; y, \; \theta)^T$. The system's generalized coordinates combine the wheel and Cartesian states
\begin{equation}
\mathbf{q} \triangleq
\begin{pmatrix}
\mathbf{x}^T & \mathbf{q}_{w}^T & \mathbf{q}_{r}^T 
\end{pmatrix}^{T}.
\end{equation}
Notice that we not only include wheel rotations, $\mathbf q_w$, but also side roller rotations, $\mathbf q_r$. This representation contrasts previous work on modeling that we did in~\cite{Sentis2013}. The main advantage, is that the augmented model will allow us to take into account roller friction which is significant with respect to actuator friction. As such, we will be able to estimate external interaction forces more precisely.

The mappings given in Eqs. (\ref{eq:dotqw}) and (\ref{eq:dotqr}) can be written as the constraint
\begin{equation}
\mathbf{J}_c \ \dot{\mathbf{q}} = 0,
\end{equation}
with 
\begin{equation}
\mathbf{J}_c \triangleq 
\begin{pmatrix}\label{eq:jcw}
\mathbf{J}_{c,w} &&& 
-\mathbf{I} &&&
\mathbf{0} \\
\mathbf{J}_{c,r} &&&
\mathbf{0} &&& 
-\mathbf{I}
\end{pmatrix}\in \mathbb{R}^{6\times 9}.
\end{equation}
Using the above kinematic constraints, one can express the coupled system dynamics in the familiar form
\begin{equation}\label{sys_dyn}
\mathbf{A} \ddot{\mathbf{q}} + \mathbf{B} + \mathbf{J}_c^T \boldsymbol\lambda_c = \mathbf{U}^T \mathbf{T}, 
\end{equation}
where $\mathbf A$ is the mass/inertia generalized tensor, $\mathbf B$ is a vector containing the estimated wheel drivetrain friction and roller to floor friction, and $\boldsymbol \lambda_c$ is the vector of Lagrangian multipliers associated with the traction forces of the wheel, where $\boldsymbol\lambda_{c,w}$ enforces the relationship between Cartesian robot position and wheel angle, and $\boldsymbol\lambda_{c,r}$ enforces the relationship between Cartesian robot position and omniwheel roller angle. In other words 
\begin{equation}
\boldsymbol\lambda_c = \left(\boldsymbol\lambda_{c,w}^T, \; \boldsymbol\lambda_{c,r}^T\right)^T.
\end{equation}
Additionally, $\mathbf U$ is the vector mapping motor torques to generalized forces, and $\mathbf T\in \mathbb{R}^3$ is the vector of output torques on the wheels. As mentioned previously, these are equivalent to the sensed torques,
$
T_s = T.
$
Values for the aforementioned matrices are
\begin{gather}\label{eq:A}
\mathbf{A} =
\begin{pmatrix}
\mathbf{M} & \mathbf{0} & \mathbf{0} \\
\mathbf{0} & I_w \mathbf{I} & \mathbf{0} \\
\mathbf{0} & \mathbf{0} & I_r \mathbf{I}
\end{pmatrix} \in \mathbb{R}^{9\times9}, \;\;\;
\mathbf{M} = 
\begin{pmatrix}
M & 0 & 0 \\
0 & M & 0 \\
0 & 0 & I_b \\
\end{pmatrix},\\[4mm]\label{eq:B}
\mathbf{B} =
\begin{pmatrix}
\mathbf{0} & 
\mathbf{B}_w^T & 
\mathbf{B}_r^T 
\end{pmatrix}^T\in \mathbb{R}^9,\quad
\mathbf{U} =
\begin{pmatrix}
\mathbf{0} & \mathbf{I} & \mathbf{0}
\end{pmatrix}\in \mathbb{R}^{3\times 9},\;
\end{gather}
where $M$, $I_b$, $I_w$, and $I_r$ are the robot's mass, body inertia, wheel inertia, and roller inertia respectively. The damping term, $\mathbf{B}$, consists of the damping at the wheel output (i.e. torque sensor bearings and belt drive), $\mathbf{B}_w$, and the damping from the side rollers, $\mathbf{B}_r$. 
We note that the side rollers do not have bearings and consist of a relatively high friction bushing mechanism. Therefore, the wheel friction is negligible relative to that of the side rollers. Thus we estimate only roller friction in our final controller. Eq.~(\ref{sys_dyn}) can be decomposed into separate equations expressing robot's body, wheel and roller dynamics as
\begin{gather}\label{eq:system}
\begin{cases}
\mathbf{M} \,
\ddot{\mathbf{x}}
+
\begin{pmatrix}
\mathbf{J}_{c,w}^T & \mathbf{J}_{c,r}^T
\end{pmatrix} \boldsymbol\lambda_c 
= \mathbf{0},\\[2mm]
I_w
\ddot{\mathbf{q}}_w
- \boldsymbol\lambda_{c,w}
= \mathbf{T},\\[2mm]%
I_r
\ddot{\mathbf{q}}_r
+ \mathbf{B}_r -\boldsymbol\lambda_{c,r}
= \mathbf{0}.
\end{cases}
\end{gather}
Using the second and third equations above, we can calculate the constraint forces on the wheels and rollers,
\begin{align}
\boldsymbol\lambda_c &= 
\begin{pmatrix}
I_w
\ddot{\mathbf{q}}_w 
- \mathbf{T} \\[2mm]
I_r
\ddot{\mathbf{q}}_r
+ \mathbf{B}_r
\end{pmatrix}.\label{lambda_c}
\end{align}
Substituting this expression into the first equation of the equation system~(\ref{eq:system}) we get
\begin{multline}\label{sys_dyn04}
\mathbf{M}\,
\ddot{\mathbf{x}}
+
\mathbf{J}_{c,w}^T 
\left(
I_w
\ddot{\mathbf{q}_w} 
- \mathbf{T}
\right)
+ \mathbf{J}_{c,r}^T
\left(
I_r
\ddot{\mathbf{q}_r}
+ \mathbf{B}_r
\right)
= \mathbf{0}.
\end{multline}
Solving the above for the output torque, $T$, we get the nominal torque model
\begin{align}\label{f_ext03}
\mathbf{T} &=
\mathbf{J}_{c,w}^{-T}
\Big[
\mathbf{M}\,
\ddot{\mathbf{x}}
+ \mathbf{J}_{c,r}^T
\left(
I_r \mathbf{\ddot{q}}_r +
\mathbf{B}_r
\right)
\Big]
+ I_w \mathbf{\ddot{q}}_w.
\end{align}
This model predicts torque sensor values in the absence of external forces. By comparing the torque sensor data against this estimate, as in Eq. (\ref{extest}), we will be able to infer the external forces. But first we must calibrate the roller friction estimate.

\subsection{Empirical Estimation of Roller Damping}

As we shown in Eqs. (\ref{sys_dyn}) and (\ref{eq:B}), the\, damping\, terms\, associated\, with\, the\, output\, dynamics\, correspond\, to\, wheel output damping, $\mathbf B_w$ and roller damping, $\mathbf B_r$. Wheel output damping consists of the friction sources between the torque sensor and the wheel, which correspond to sensor bearings and the belt connecting the sensor to the wheel. Notice that gear friction is not included, as the torque sensor is located after the gears. When we lift the robot of the ground and rotate the wheels, the mean value of the torque sensor signal is close to zero, meaning that the drivetrain output friction is negligible compared to roller friction. On the other hand, roller friction is relatively large as the rollers do not have bearings and therefore endure high friction when rotating in their shaft. In the next lines we will explain our procedure to estimate roller damping based on torque sensor data.

\begin{figure}[p!] \centering
\includegraphics[width=\linewidth, clip=true]{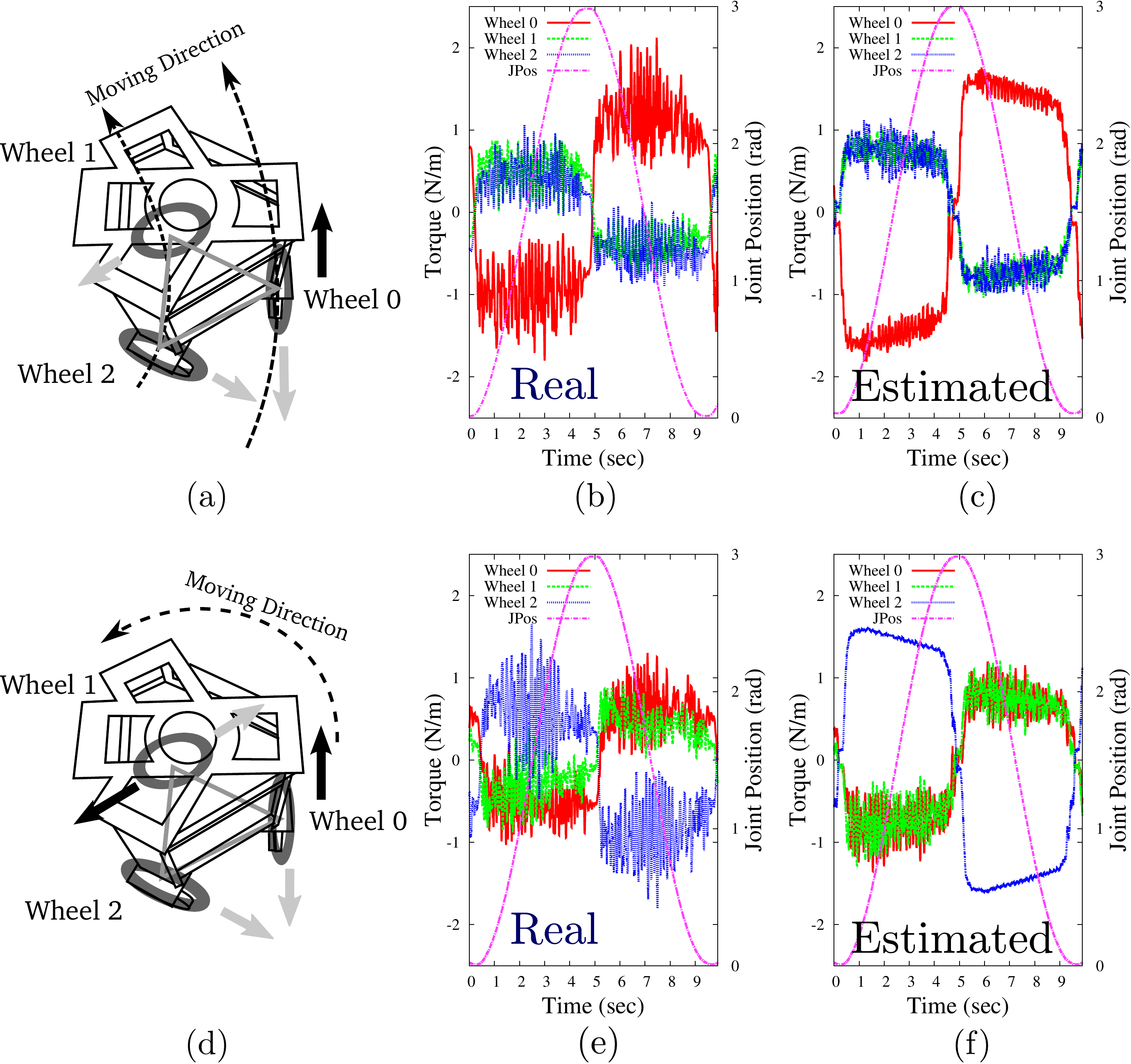}
\caption
[ Torque Signals from Simple Motions]
{{\bf Torque Signals from Simple Motions}
	\label{roller_friction} are used to calibrate the roller friction model. No external forces are applied to the robot in this test. The JPos lines represent the motion along the two simple arc trajectories. Torque signals from the calibrated model are shown to the right of the graphs representing the actual data on which they were trained. Subfigures (a-c) represent the rotation of the robot about a virtual pivot outside the base of support, while (d-f) show a pivot centered on Wheel~2. Gray arrows in figures (a) and (d) represent the torque sensed at the wheels, while the black arrows represent wheel motion. By comparing (b) against (c) and (e) against (f), we can conclude that the expected roller friction torque model at least partially captures the gross shape of the data.
}
\label{roller_friction}
\end{figure}

Fig.~\ref{roller_friction} demonstrates the two experiments under which the roller friction model was calibrated. In these tests, joint position controllers for each wheel, simple high gain servos, push the robot through a nominal path, and the resulting torque sensor values are measured in the absence of any external force. In Subfig. \ref{roller_friction} (a) we show an experiment in which wheel 0 moves sinusoidally with time while the other two wheels remain fixed, resulting in an arc motion of the entire robot. In Subfig. \ref{roller_friction} (b) we plot the sinusoidal joint trajectory of wheel 0 and the torque sensor readings from the three wheels. The torque signals on all wheels show an approximately square wave shifting phase according to the direction of wheel's 0 motion. Because of this pattern, we assume that most of the friction is due to Coulomb effects instead of dynamic friction effects. We approximate this Coulomb friction in our model using a $tanh$ softening of the signum function, i.e.
\begin{align}\label{eq:br}
B_{r,i} = B_r \tanh \left( \alpha \, \dot{q}_{r,i}  \right),
\end{align}
where the magnitude $B_r$ and scaling factor $\alpha$ are tunable parameters that we adjust based on the empirical data. To do the tuning, we implemented Eq. (\ref{f_ext03}) in a software simulation and compared its output to the experimental data. In that equation, the accelerations of the wheels, the robot's body and the side rollers must be known. We calculate them using the wheel trajectories, 
$q_{w,0} = 3/2 - 3/2 {\rm cos}(2 \pi \omega t)$, $q_{w,1} = q_{w,2} = 0$, which can be easily differentiated twice to obtain $\ddot{\mathbf q}_w$. To obtain the robot's body acceleration we use the inverse of Eq. (\ref{eq:dotqw}) and take the second derivative, yielding 
\begin{equation}\label{eqddotx}
\ddot{\mathbf{x}} = \mathbf J_{c,w}^{-1} \ddot{\mathbf q}_{w} + \dot{\mathbf J}_{c,w}^{-1} \, \dot{\mathbf q}_{w}.
\end{equation}
Eq. \ref{eq:dotqr} then provides the roller accelerations 
\begin{equation}
\ddot{\mathbf q}_r = \mathbf J_{r,w} \ddot{\mathbf x} + \dot{\mathbf J}_{r,w} \dot{\mathbf x}.
\end{equation} 
Plugging these values into the simulation of Eq. (\ref{f_ext03}), with $\mathbf B_r$ given by the model of Eq. (\ref{eq:br}) we searched over $B_r$ and $\alpha$ until the simulation matched the real data. In Fig. \ref{roller_friction} (c) we show the result of the simulation using Eq. (\ref{f_ext03}) which can be compared to the real data of Fig. \ref{roller_friction} (b). Our final model parameters were $B_r = 0.2 Nm$ and $\alpha = 0.4$. 

To further validate the procedure we conducted a second estimation process, shown in the Figs. \ref{roller_friction} (d), (e) and (f) in which two wheels of the mobile base track a sinusoidal trajectory while one of them remains at a fixed joint position. As we can see, the simulated torques with the estimated roller friction model of Fig. \ref{roller_friction} (f) has a good correspondence to the actual data of Fig. \ref{roller_friction} (e).

\subsection{Model-Based Force Estimation}
Following the simplified estimation of external torques from Eq. (\ref{extest}), we modified Eq. (\ref{sys_dyn}) to account for external forces, yielding
\begin{align}\label{ext_dyn}
\mathbf{A} \ddot{\mathbf{q}} + \mathbf{B} + \mathbf{J}_c^T \boldsymbol\lambda_c + \mathbf{J}_{ext}^T \mathbf{F}_{ext} = \mathbf{U}^T \mathbf{T}. 
\end{align}
where $\mathbf{J}_{ext}$ is the Jacobian corresponding to the location of the external forces, and $F_{ext}$ is an external wrench containing a Cartesian force and a torque, i.e.
\begin{align}\label{f_ext}
\mathbf{F}_{ext} \triangleq
\begin{pmatrix}
F_{ext,x} & F_{ext,y} & \tau_{ext}
\end{pmatrix}^T.  
\end{align}
The differential kinematics of the point on the exterior of the body at which the external force is applied can be expressed in terms of the robot's differential coordinates as
\begin{equation}\label{eq:dotxext}
\dot{\mathbf{x}}_{ext} = 
\dot{\mathbf{x}} + \dot{\theta}\, \mathbf{i}_z \times {\mathbf{d}}
= \mathbf{J}_{ext,b} \dot{\mathbf{x}}
\end{equation}
where 
$\mathbf{x}_{ext} \triangleq \begin{pmatrix} x_{ext} & y_{ext} & \theta_{ext}\end{pmatrix}^T$,  $\dot \theta$ is the angular velocity of the base, $i_z$ is the unit vector in the vertical, $z$, direction, $\times$ is the cross product, and $\mathbf{d}$ is a vector describing the distance from the center of the robot to the collision point. Developing the above equations, we can define
\begin{equation}\label{eq:jext}
\mathbf J_{ext,b} \triangleq
\begin{pmatrix}
1 &&& 0 &&& y - y_{ext} \\
0 &&& 1 &&& x_{ext} - x \\
0 &&& 0 &&& 1 
\end{pmatrix}\in\mathbb{R}^{3\times 3}.
\end{equation}
Extending Eq. (\ref{eq:dotxext}) with respect to the full generalized coordinates yields
\begin{equation}
\dot{\mathbf{x}}_{ext} = 
\mathbf{J}_{ext} \, \dot{\mathbf{q}}, \quad {\rm with} \quad
\mathbf{J}_{ext} \triangleq
\begin{pmatrix}
\mathbf{J}_{ext,b}
&
\mathbf{0}_{3\times 6}
\end{pmatrix}.
\end{equation}
Using the above expression for $J_{ext}$ in the extended dynamics of Eq. (\ref{ext_dyn}), and neglecting the effect of the wheel and roller inertias, $I_w\approx 0$, and $I_r \approx 0$ with respect to the robot's mass, and the effect of the wheel friction, $\mathbf B_w \approx 0$ with respect to the roller friction, we get a similar system of equations than that shown in Eqs. (\ref{eq:system}), i.e.
\begin{gather}
\begin{cases}
\mathbf{M} \,
\ddot{\mathbf{x}} + \Big( \mathbf{J}_{c,w}^T\;\; \mathbf{J}_{c,r}^T \Big)
\boldsymbol\lambda_c + \mathbf{J}_{ext,b}^T F_{ext} = \mathbf{0},\\[2mm]
- \boldsymbol\lambda_{c,w} = \mathbf{T},\\[2mm]
\mathbf{B}_r -\boldsymbol\lambda_{c,r}
= \mathbf{0}.
\end{cases}
\end{gather}
Substituting $\boldsymbol{\lambda}_c \triangleq (\mathbf \lambda_{c,w}, \mathbf \lambda_{c,r})$ on the first equation above by the values of $\lambda_{c,w}$ and $\lambda_{c,r}$ obtained from the second and third equations we get
\begin{equation}\label{eq:mddotxfext}
\mathbf{M} \,
\ddot{\mathbf{x}} -\mathbf{J}_{c,w}^T \, T + \mathbf{J}_{c,r}^T \mathbf B_r + \mathbf{J}_{ext,b}^T F_{ext} = \mathbf{0}.
\end{equation}
In the absence of external forces, we can solve for the torques
\begin{equation}
\mathbf T \big\rvert_{\mathbf F_{ext}=0} = \mathbf{J}_{c,w}^{-T} \Big[\mathbf M \ddot{\mathbf{x}} + \mathbf{J}_{c,r}^T \mathbf B_r \Big].
\end{equation}
\begin{figure}\centering
\includegraphics[width=90mm, clip=true ] {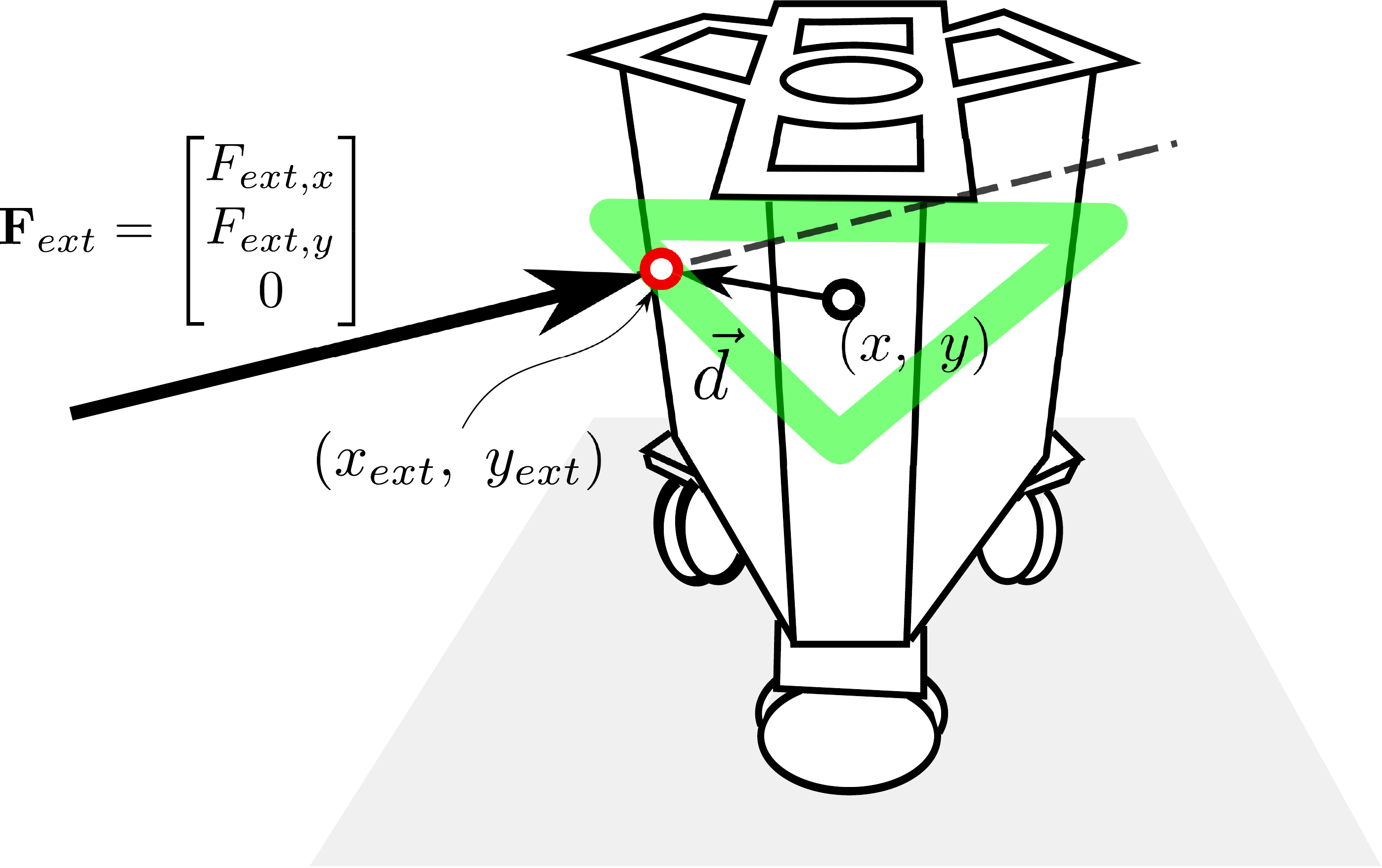}
\caption
[External Force Estimation]
{{\bf External Force Estimation} is predicated on the assumption that the external force is a purely translational push applied to the robot's surface, as approximated by a triangular prism. The green triangle is the approximated robot body shape in a horizontal plane, and the perceived contact point, a red circle, occurs at the first of two intersections between this triangle and the line of zero external moment.}
\label{ext_force}
\end{figure}
The important point of the mapping above is that it can be numerically solved using the model and the acceleration estimate of Eq. (\ref{eqddotx}). On the other hand, when the robot collides with the environment, the torque sensors read values according to the dynamics of Eq. (\ref{eq:mddotxfext}). Assuming the output torque is equal to the value given by the torque sensors, i.e. $\mathbf T_s = \mathbf T$, we can use the previous two equations to solve for the external forces
\begin{equation}
\left(\mathbf T \big\rvert_{\mathbf F_{ext}=0} - \mathbf T_s\right) = \mathbf{J}_{c,w}^{-T} \mathbf{J}_{ext,b}^T F_{ext},
\end{equation}
which can be written in the alternative form
\begin{equation}\label{eq:alter}
\mathbf{J}_{ext,b}^T F_{ext} = \mathbf{J}_{c,w}^T \left(\mathbf T \big\rvert_{\mathbf F_{ext}=0} - \mathbf T_s\right).
\end{equation}
We now make the following simplifying assumptions:
\begin{itemize}
\item The external wrench has no net torque.
\item The external wrench is applied at a point on the triangular prism approximation of the body
\item The external wrench is always of a pushing nature
\end{itemize}
With those premises and the expression of Eq. (\ref{eq:jext}), the above equation becomes
\begin{align}\label{f_ext04}
\Big[
F_{ext,x}, \;\;
F_{ext,y}, \;\;
\left(x_{ext} - x \right) F_{ext,y} -\left(y_{ext} - y \right) F_{ext,x}
\Big]^T
= \mathbf{J}_{c,w}^T \left(\mathbf T \big\rvert_{\mathbf F_{ext}=0} - \mathbf T_s\right).
\end{align}
This equation has four unknowns, $\{F_{ext,x}, F_{ext,y}, x_{ext},$ $y_{ext} \}$ but only three entries. It is attempting to simultaneously solve the external force and its location. Let us focus on the third entry of the above equation. The third row can be written in the form
\begin{equation}\label{eq:line}
\left(x_{ext} - x \right) F_{ext,y}-\left(y_{ext} - y \right) F_{ext,x}
=
I_b\, \ddot{\theta} - \frac{R}{r_w} \sum_{i=0}^{2} \tau_{s,i}.
\end{equation}
This derivation comes from first comparing Eqs. (\ref{eq:mddotxfext}) and (\ref{eq:alter}), which lead to
\begin{equation}
\mathbf{J}_{c,w}^T \left(\mathbf T \big\rvert_{\mathbf F_{ext}=0} - \mathbf T_s\right) = \mathbf{M} \,
\ddot{\mathbf{x}} -\mathbf{J}_{c,w}^T \, T + \mathbf{J}_{c,r}^T \mathbf B_r,
\end{equation}
and then deriving the third row of the right hand side of the above equation, yielding
\begin{equation}
\mathbf{J}_{c,w}^T \left(\mathbf T \big\rvert_{\mathbf F_{ext}=0} - \mathbf T_s\right)\bigg\rvert_{row \; 3}
= I_b \ddot \theta - \frac{R}{r_w} \sum_{i=0}^{2} \tau_{s,i}.
\end{equation}
The above results are obtained from the third rows of the transpose of Eqs. (\ref{eq:jcw}) and (\ref{eq:jcr}), i.e.
\begin{gather}
J_{c,w}^T\big\rvert_{row \; 3} = \frac{1}{r_w} 
\begin{pmatrix}
R&&R&&R
\end{pmatrix},\\
J_{c,r}^T\big\rvert_{row \; 3} =
\begin{pmatrix}
0&&0&&0
\end{pmatrix}.
\end{gather}
Because Eq. (\ref{eq:line}) corresponds to a geometric line, the location of the contact point can be solved using solely Eq. (\ref{eq:line}) and our previously stated assumptions. The line is parallel to the direction of the external force, $\mathbf{F}_{ext}$, and can be used to find the distance from the center of the robot to the intersection of the line with the robot's body. The shape of our mobile base can be approximated as a triangular prism, and its planar section is a triangle, which is convex. Thus, there are only two points on its body where the line meets the premises. Therefore, we solve for the location where the external force is applied using those geometric constraints as shown in Fig. \ref{ext_force}.

Once we find the location of the contact point, we now solve for the external force using the first and second row of Eq. (\ref{f_ext04}).

\begin{figure}[p!]
\centering
\includegraphics[width=\textwidth]{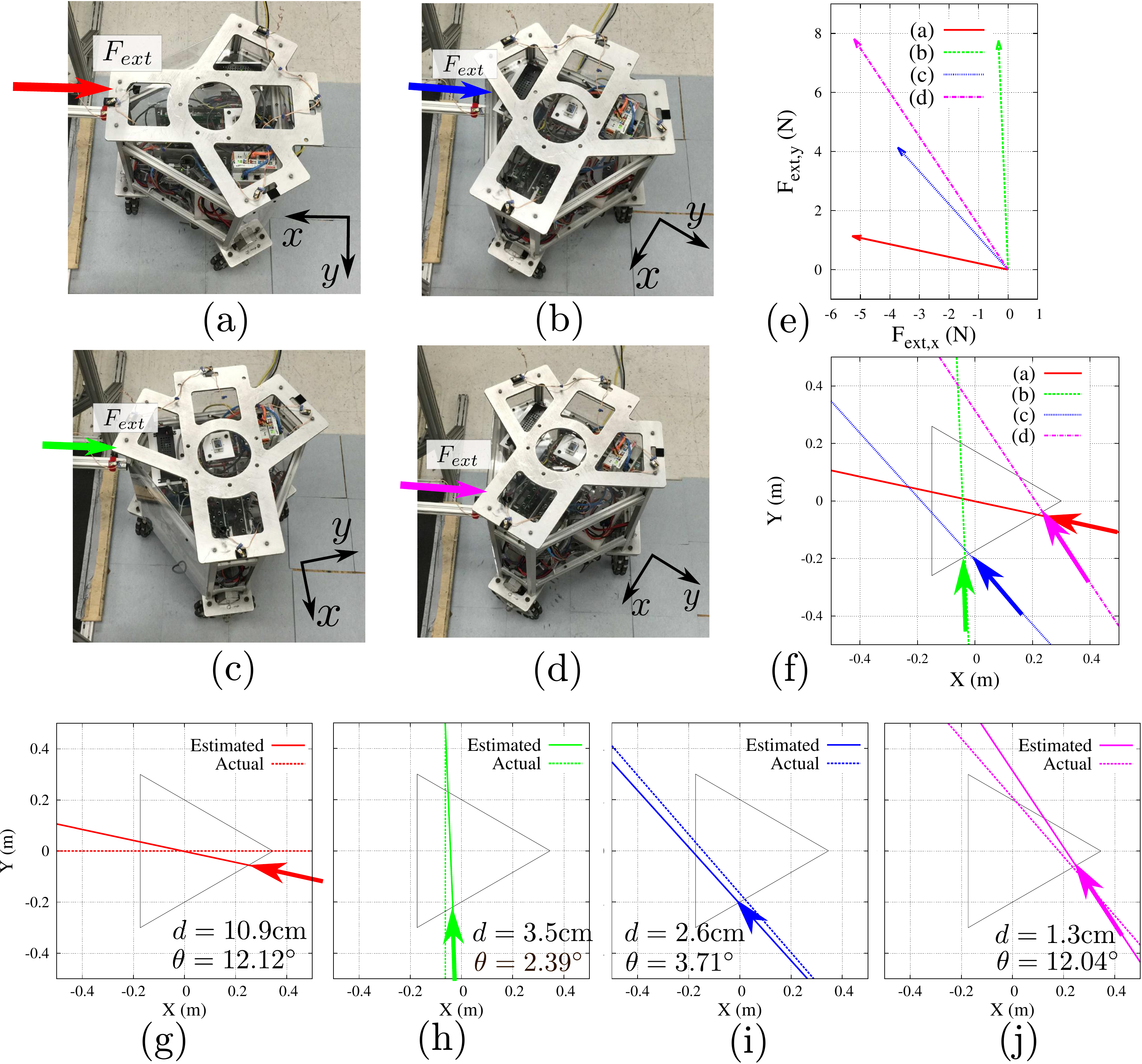}
\caption
[ Estimating Static External Forces]  
{{\bf Estimating Static External Forces} 
	using only the torque sensors results in an accurate estimate of their location, angle, and magnitude. In Subfigures (a-d), the bar on the left side of the image confers an external push of roughly 9 Newtons onto the robot, above which is overlaid a triangle and a dot. This overlay is meant to reveal the robot's internal coordinate system, for clarity. The estimated forces from all four robot positions are shown directionally in Subfigure (f), where they are represented in the coordinate system of the robot. Subfigure (e) illustrates the magnitude of these forces, in the same coordinate system, emanating from the origin.}
\label{detection}
\end{figure}

\section{Experimental Results and Assessment}


Throughout the previous sections we have established the following infrastructure: (1) full-body collision detection capabilities using constrained models and including wheel and side roller dynamics; (2) estimation of roller damping which is dominant in the behavior of the output robot dynamics; (3) fast collision response capabilities by achieving desired impedances through an admittance controller; (4) an experimental infrastructure including, a mobile base with torque sensors on the wheel drivetrains, a calibrated collision dummy, and a motion capture system.

The goal of this section is multi-objective: (1) to characterize the performance of our infrastructure in terms of accuracy of force detection and the impact location, (2) to measure the amount of time that takes our robot to detect collisions, (3) to measure the amount of time it takes our robot to respond to collisions once they have been detected, (4) to poke the robot in various places to proof that we can detect collisions in all parts of the robot including its wheels, and (5) to give an idea of what are the implications of our methodology for providing safety in human-scale mobile bases.

To do so, we conduct five calibrated experiments where we measure performance using a combination of the wheel torque sensor data, the wheel odometry and the motion capture data on the robot and the collision dummy. Additionally, we conduct a proof of concept experiment on safety, where the robot roams freely around people in all sorts of postures and collides with them safely.

\subsection{Detection of External Force and Contact Location}
In this experiment we evaluate our method's ability to detect the point of contact on the robot's body, the direction of the external force, and the magnitude of the external force. In particular we will use only the wheel drive-train torque sensors to identify those quantities without any use of external sensor mechanisms. In other words, the robot does not utilize motion capture data or wheel odometry to detect those quantities. 

\begin{figure}[p!]\centering
\includegraphics[width=\linewidth, clip=true]{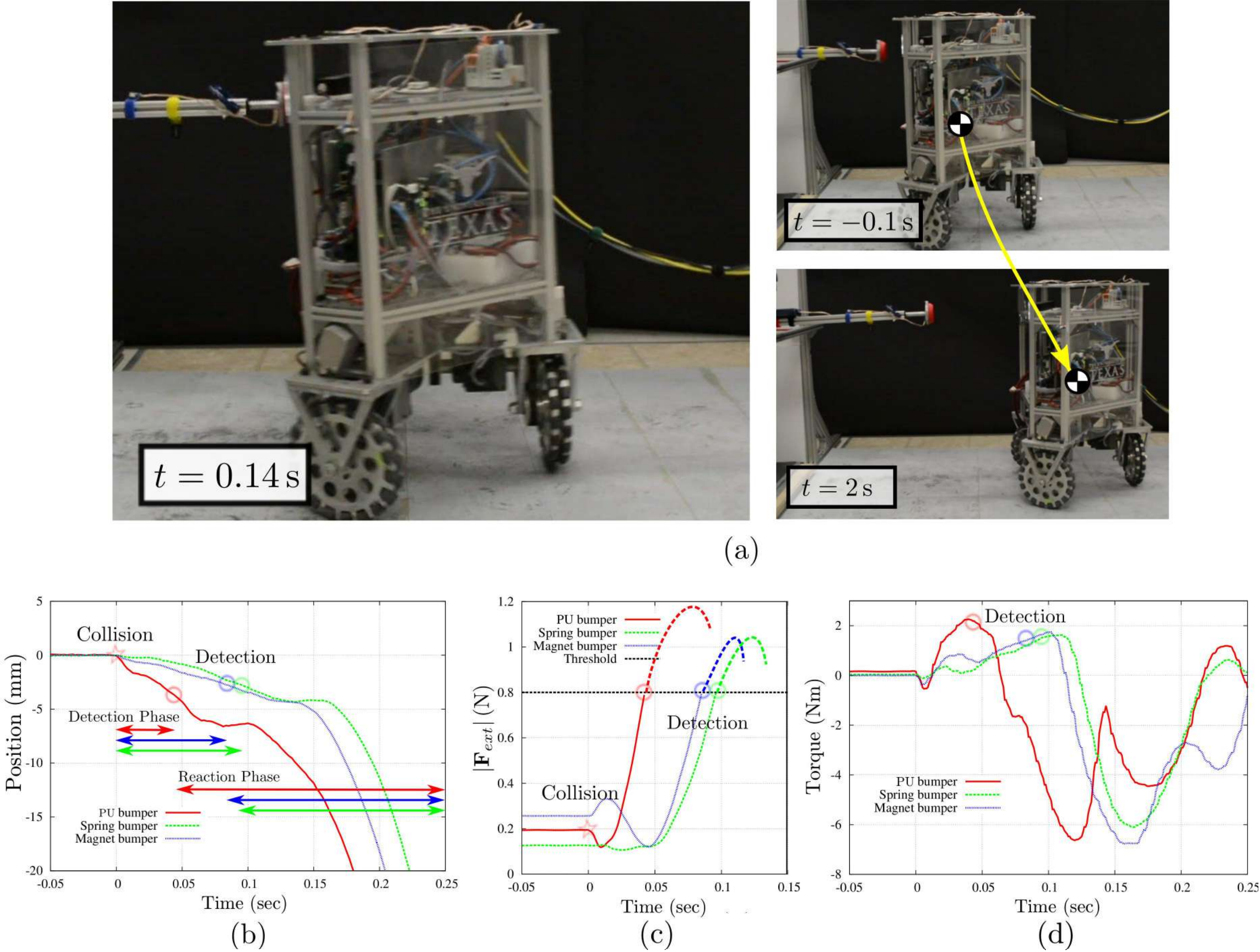}
\caption[ Upper Body Collision Testing] 
{{\bf Upper Body Collision Testing} 
	illustrates the robot's collision avoidance behavior with respect to the three different bumper designs when the impact occurs above its center of mass. The magnet bumper impact is shown at three representative frames in (a), with $t=0.14 \rm s$ representing the peak of force and spring deflection. Subfigure (b) shows the position evolution of the robot from the instant of contact, and highlights the instant when the robot's software registered the impact in each of the three trials. After the initial impact but before the robot recognized the impact, that is, during the detection phase, the force of impact pushed Trikey backwards. In (c), the measured torque sensor value on the Wheel 0 for each trial is plotted against the same time range. Note that the initial dip in torque is due the propensity of an upper body impact to tip the robot over, rolling the wheels forward. By virtue of being a more direct transfer of energy, the PU bumper is detected first, causes more initial motion, and results in a higher peak torque than the other experiments The faster detection is due to the larger torque, since external force measurement is based on a moving average filter of the torque sensor signals.}
\label{high_static_collision}
\end{figure}

To conduct these tests, we use the infrastructure depicted in Fig. \ref{setup}. The horizontally sliding dummy is connected to a pulley system that runs to an overhead system with a vertical weight of $1$Kg. As a result a constant force of $10$N is applied to the slider. In Fig. \ref{detection} we show images of the experimental setup where the slider is placed in contact with the base before conducting the estimation process. The robot's wheels are  powered off, and because of the high friction of the harmonic drives, the forces applied by the dummy are not enough to push the robot away. 

Subfigs. \ref{detection} (a)-(d) show the procedure that we conduct. We first place the robot in different directions and orientations with respect to the dummy. Using only the torque sensor data, we proceed to use the force estimation techniques described in Sec. \ref{sec:forceestimation} to identify the point of contact, the direction of the force, and its magnitude. We repeat the same experiment for 4 different scenarios applying the same amount of force. Without loss of generality, all the external forces are applied to the same side of the robot as the robot is symmetrical.

Fig. \ref{detection} (e) and (f) shows the results of the estimation process. Subfig. \ref{detection} (e) shows that the magnitude and direction of the estimated forces and Subfig. \ref{detection} (f) shows the contact point and the force direction with respect to the base geometry and orientation. The magnitude of the forces estimated ranges from $5.5$N to $10$N. Those values are $(0\%-45\%)$ smaller than the $10$N of force applied by the contact dummy. We believe that the reason is due to stiction of the overall mechanical structures standing between the contact point, the wheels in contact with the ground, and the pulley system connecting the wheel to the torque sensors. The maximum error in detecting the direction of the forces is $3.3\%$ with respect to the full circle, or equivalently $12$deg over $360$deg with a mean value of $\pm 2\%$. Finally, the maximum error in detecting the point of contact is $11$cm with a mean value of $4.5$cm, or equivalently, $18\%$ of error with a mean value of $7.5\%$ with respect to the $61$cm of length of the robot's side walls. 

Overall we accomplish maximum errors of $45\%$ for the magnitude, $3.3\%$ for the direction and $18\%$ for the location of the external forces. The good accuracy of the location and direction of the estimated force can be leveraged to respond safely to impacts by moving away from the colliding bodies with precision. The medium accuracy of the estimated force's magnitude is probably due to the mechanical structure and not due to the estimation strategy. Nonetheless, it is sufficient for the controller to execute the admittance control model. However, if we wish to achieve the target impedance with high precision, the external force's magnitude will have to be estimated with higher accuracy. In that case improved designs of the mobile robot that minimize stiction should be sought.

\subsection{Collisions with Motionless Robot}
In this experiment we evaluate our method's ability to not only detect collisions but to quickly react in a manner that is perceived as safe. Moreover, the tests discussed here will analyze collisions with the mobile base standing motionless close to the collision dummy. Responding safely to collisions when the robot is still is one of the hardest case scenarios that a robot may encounter. In such case, the safe response of the robot solely depends on its ability to estimate the external forces with accuracy. In contrast, when a robot collides while in motion its controller knows the trajectory where it came from. As such a simple safe response would be to reverse direction towards that trajectory. 

Once more we use the infrastructure of Fig. \ref{setup}. However, this time around we connect the pulley system to a vertical weight of $4.54$Kg producing a constant horizontal force of $44.54$N on the contact dummy. The contact dummy is also now connected to a sliding weight of $9.08$Kg which constitutes the effective mass that collides with the mobile base. The sliding dummy is released at a certain distance to the robot and when it collides with the robot it has reached a velocity of $0.5m/s$. The robot is initially at rest and when it detects contact it moves away from the collision in the direction of the collision.

\begin{figure}[p!]\centering
\includegraphics[width=\linewidth, clip=true ] {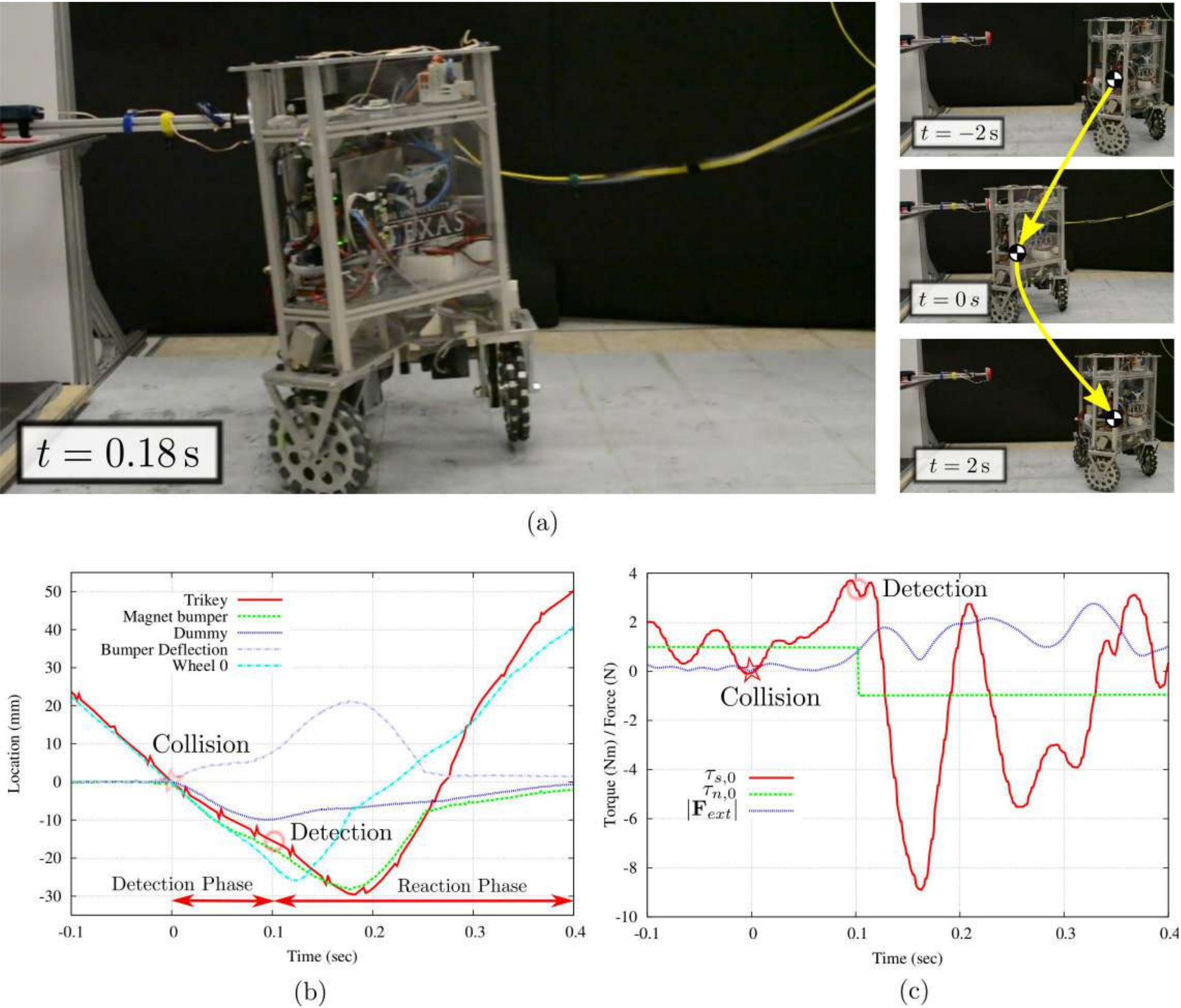} 
\caption
[ Collision Against a Static Obstacle]
{{\bf Collision Against a Static Obstacle} 
	tests Trikey's ability to reverse direction when moving at full speed, after an impact with the magnet bumper. Various stills in (a), including the $t=0.18 \rm s$ frame with maximum spring deflection, illustrate the experimental procedure. Subfigure (b) plots various reference positions including the position of Trikey itself, the position of the bumper, the position of the slider, the spring deflection, and the angle of Wheel 0 (times a scaling factor), with all positions normalized to zero at the instant of collision.
Subfigure (c) plots the torque sensor from Wheel 0, the expected Wheel 0 torque sensor value, and the estimated external force. This external force exceeds the predefined collision threshold when $t= 105\rm ms$, corresponding to the Detection timestamps in both (b) and (c).}
\label{high_moving}
\end{figure}

In Fig. \ref{high_static_collision} we show the procedure that we conduct. We first place the robot next to the collision dummy with the dummy separated from the robot. Once more, we only use torque sensor data to estimate the direction, location and magnitude of the collision and respond to it. We implement the force estimation procedure of Sec. \ref{sec:forceestimation} and the admittance controller of Sec. \ref{sec:syschar}. The desired impedance that we implement for the controller is $M_{des} = 2kg$ and $B_{des} = 1.6N/m^2$. The motivation for these values is first to maximize the reaction speed by setting a low target mass. However, if we make $M_{des}$ too small, the robot accelerates too quickly in reaction to the collision and it tips over. Therefore, we decrease it to just over the limit where it tips over. In order to select $B_{des}$ we follow the subsequent procedure. Using Eq. (\ref{eq:traj}), the position achieved by the controller on a particular direction, e.g. $x$, after impact at time $\infty$ is
\begin{equation}\label{eq:xdes}
x_{des}(t\rightarrow \infty) = x_0 + \frac{F_{ext,x}}{B_{des}}.
\end{equation}
Based on this equation, we design $B_{des}$ such that the robot moves away by $0.5$m upon collision, i.e. 
\begin{equation}
x_{des}(t\rightarrow \infty) = x_0 + 0.5.
\end{equation}
Taken into account that we use a threshold of $|\mathbf F_{ext}| = 0.8N$ to initiate the admittance controller (see Fig. \ref{controller}), solving Eq. (\ref{eq:xdes}) for these values we get $B_{des} = 0.8/0.5 = 1.6 N/m^2$.

We conducted the collision experiments using three different materials on the collision dummy: the default thin polyurethane plastic (PU bumper), a thin polyurethane foam with a spring (Spring bumper), and the same thin polyurethane foam with the spring and a magnetic latch as described in Subsection \ref{material} (Magnetic bumper).

\begin{figure}[t]
\centering
\includegraphics[width=\textwidth] {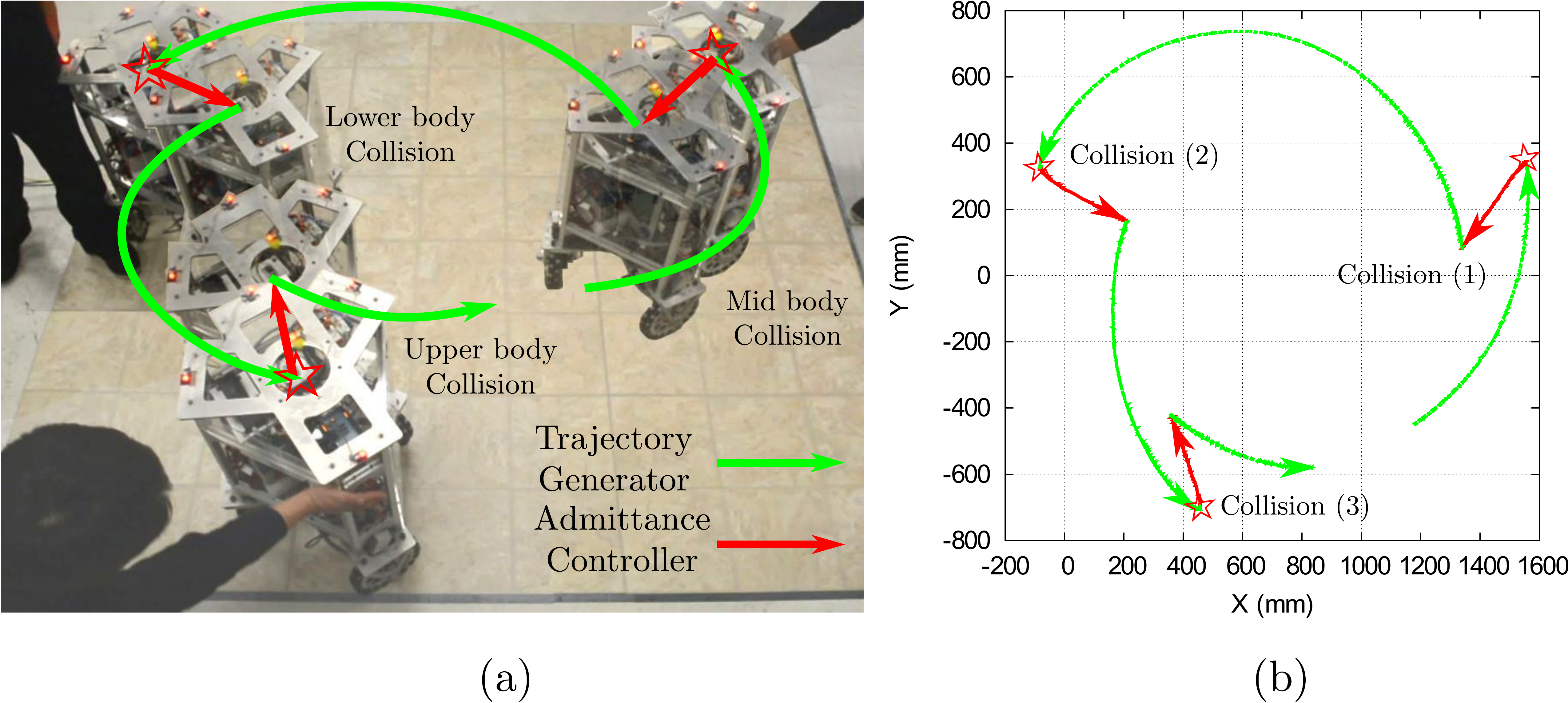} 
\caption
[Omni-directional Motion with Unplanned Collisions] 
{{\bf Omni-directional Motion with Unplanned Collisions} 
	demonstrates Trikey's full motion capability as it moves about a $1.5 \rm m$ diameter circle at $0.16 \rm m/s$. A composite image of several frames, (a) shows the motion, the escapes, and the human obstacles in the experiment. The trajectory captured by the motion capture system is shown in (b).}
\label{omni_collision}
\end{figure}

Additionally, to compare performance, we conduct collision tests both in the upper and lower parts of the mobile base. As shown in Fig. \ref{high_static_collision} (c), the collision was detected in $45$ms (PU bumper), $95$ms (Spring bumper), and $85$ms (Magnet bumper). 

In Subfig. \ref{high_static_collision} (c) we observe that that the estimated external force in the PU bumper case reaches the reaction threshold at $t=0.05$s. As a result, the admittance controller kicks in causing the robot to move quickly away. As shown in Subfig. \ref{high_static_collision} (b), after detecting the contact, the robot's change in position seems to hit a plateau for about $50$ms. The reason is due to the robot accelerating quickly and lifting the front wheel (see Subfig. \ref{high_static_collision} (a) for that effect). After that plateau, the robot quickly moves away from the collision.

Let us focus on Subfig. \ref{high_static_collision} (d). Positive wheel torques result from the impact forces on the robot and negative torques result from the robot moving away from the impact. As we can see, using the spring and magnet bumpers reduces the impact torques by about $20\%$ with respect to the peak value of the PU bumper. Additionally, if we focus on Subfig. \ref{high_static_collision} (b) we can see during the collision time, $t\in(0,t_{detection})$, the robot's trajectory associated with the response to the PU bumper accelerates much more quickly than that for the spring or magnetic bumpers. It is this combination of lower peak force and lower acceleration that will make the use of the spring or magnetic bumpers safer.



\subsection{Collisions with Moving Robot}
The setup for this experiment is similar to the one before. However, the robot now moves towards the resting contact dummy and produces a collision to which it needs to respond. This experiment tests the reaction time and peak torques of the moving robot upon collision.

The collision dummy is initially at rest with a total sliding mass of $13.62$Kg. The robot moves towards the dummy and hits it with a velocity of $0.22 \rm m/s$. This time around, we only conduct the experiment with the magnetic bumper. The same estimation and control methods used in the previous section are applied.

Similarly to the tests before that contain the spring or magnetic bumper, it takes $105$ms to detect the collision threshold. Fig. \ref{high_moving} (c) shows the measured torque from the torque sensor in Wheel 0 and the magnitude of the estimated external force. 

Overall, the reaction time is similar to the motionless experiment before and the peak torque values are about twice the values we had obtained with the spring or magnetic bumpers. This increase in value is due to the robot having an initial velocity which causes a higher force collision due to the robot's heavy weight.

\subsection{Additional Experiments}

In Fig. \ref{omni_collision} an experiment involving our mobile base executing circular arc trajectories while being pushed away is presented. The trajectory of the mobile base is recorded using the motion capture system. As we can see, collisions are promptly detected resulting in the robot reacting to them in the opposite direction of the colliding force.

\begin{figure}[p!]\centering
\includegraphics[width=\textwidth] {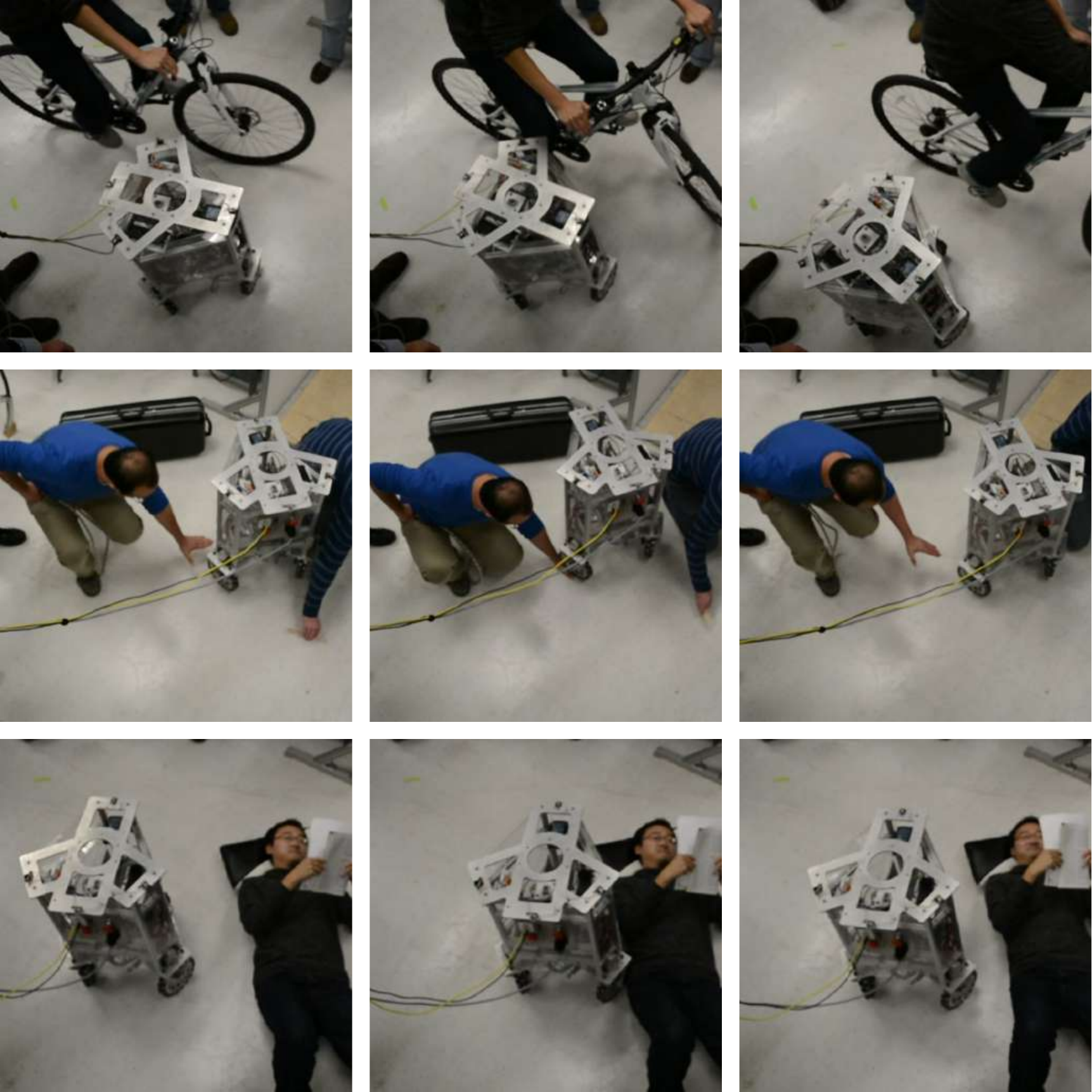}
\caption
[ Collisions in Human Environments] 
{{\bf Collisions in Human Environments} 
	points towards our long term vision for mobile robots. Here Trikey collides with humans in various scenarios, and reacts to the collisions safely.}
\label{collisions}
\end{figure}

Finally, as a proof of concept, we conducted an experiment where we let the mobile base move around performing circular arc trajectories while people provide it with simulated accidental collisions. Fig. \ref{collisions} shows the robot's reaction to collision with a bicycle, a hand placed on the floor in the wheel path, and a person lying down.

\section{Conclusions}

Mobile robots will not be truly useful until they are very safe in cluttered environments. We have presented a methodology for these types of robots to quickly react and achieve low impedance behaviors upon discovering an unexpected collision. It is the first study to accomplish full-body collision detection on all parts of a mobile platform.

Our estimation method has been shown to estimate the contact location of the collisions with $18\%$ error, direction of the contact forces with $3.3\%$ error, and magnitude of those forces with $45\%$ error. The lower accuracy of the magnitude is due to mechanical limitations of the structure of the base and the connection of the wheel to the torque sensors. Those could be improved by having a stiffer structure and improved connections from the wheel to the torque sensor.

Empirically estimating roller dynamics has been key to enhancing the external force sensing accuracy. We have chosen to use only model based estimation and have achieved good precision but feel we could benefit in the future from statistical methods such as \cite{Fang2011}. Our detection and reaction to collisions rely solely on the on-board torque sensor data. They do not rely on wheel odometry or external global pose estimation. As such they can attain a very fast reaction time.

As shown in the experiment of Fig. \ref{omni_collision}, the admittance controller developed in Sec. \ref{sec:syschar} has been effective at providing the desired impedances. In particular, it decreased the peak contact torques without tipping over the base. At the same time, the desired closed loop damping prevents the robot from moveing too far away from the collision source. These parameters can be tuned depending on the application.


In the future we would like to conduct experiments with test dummies that are clamped against a wall. Such scenario is one of the most dangerous ones. We would also like to apply safety criteria that compare the static and dynamic forces of our base to the maximum tolerable curves obtained from previous empirical studies. Additionally, we would like to study collisions of the base at moderately high speeds. Because bases are heavy and have limited braking ability, their reaction capabilities are similar to those of cars. As such we would like to apply injury indicators from the automotive industry to explore these types of collisions.

%% file: chapter-lidar.tex
\chapter{Human Body Part Multicontact Recognition and Detection Methodology}


\section{Introduction}
\blfootnote{This chapter has been published in {\it 2017 IEEE International Conference on Robotics and Automation} \cite{kim2017}.}
In this chapter, we study physical Human Robot Interaction (pHRI) in personal robotic platforms using human contact detection and contact gesture queues. Previously, to interact with robots, various methods have been devised such as those relying on dedicated input devices, body language visual recognition, voice recognition, and sensorized skins or touch devices \cite{Yohanan2012} among others. In our study, we focus on fusing visual recognition with contact sensing on mobile platforms, which allows close contact with people. Existing work on fusing visual and contact recognition \cite{Magrini2015} has primarily focused on robotic manipulators and on using external structured light 3D sensing. However, this type of method does not directly apply to omnidirectional mobile platforms like our studies. First mobile platforms require different dynamical models that incorporate traction and roller constraints as well as models of roller friction. Second, mobile platforms need to carry the 3D sensor on board, and therefore time-of-flight 3D sensing is more suitable for close range detection. Third previous methods have focused on detecting contacts and the corresponding forces. In addition to these capabilities, we focus on inferring which human body parts are in contact with the robot. Ultimately, these capabilities endow richer pHRI behaviors. Finally, previous work on multicontact detection has focused on detection only, but not on multicontact gesture communications as explored here. As such, the aim of this work is to enable mobile ground platforms to physically interact with humans by means of multicontact gestures.

Ultimately, if robots are to be used as personal companions for boosting our comfort and productivity, we believe they will benefit by these type of close contact capabilities. The goal is to increase the communication modalities with robots to become more intuitive by exploiting contact interactions.

One type of intuitive method for HRI is based on using body language. A popular method for gesture recognition is using depth image data made from structure light such as the Kinect$^{\rm TM}$ sensor. One of the drawback is that this type of sensor is limited to indoor environments because the infrared light (IR) is vulnerable to sunlight. An alternative to structure light sensing is a laser scanner based on time-of-flight measurements. In our case we use this optical sensing modality because it allows to operate at close ranges and also in outdoor environments.
To leverage gesture recognition to HRI in robot companion applications, we focus both on the detection phase but also on the identification of the human body parts approaching or in contact with the robotic platform. Such capability allows to closely analyze the nature of the intended behaviors.

Additionally touching has became pervasive for handheld devices. We feel that our technology will allow to interact with robots in similar ways, enabling complex touch behaviors. In our case, we feel that the possibilities are enormous as many body parts can be used for communication and the entire robot structure can be touched.

One approach to detect contact is to use a vision system recognizing contact with nearby object \cite{Ebert2002}. Also, tactile sensors on robot skins are popular to detect a contact. In such case, tactile sensors are attached to an outer skin of the robot. The problem is that all the exterior of the robot needs to be covered by the sensorized skin to guarantee whole-body contact detection. Another limitation is that touch skins cannot recognize human body parts.
Lastly, joint torques on robotic manipulators have been used to infer contact information \cite{Luca2006}. Usually, sensing contact forces has enabled some level of safety during accidental collisions \cite{Haddadin2009}. 
Previously, we have investigated whole-body contact detection on mobile platforms to provide safety \cite{Kim2015}, but not for human intention recognition. 
Since a contact force is just one part of touch-based gestures, we cannot fully estimate the human intention solely based on forces. 
To push the boundaries of physical HRI, we suggest a multi-contact gesture recognition method with human body-part awareness. Our contact-based gesture recognition method allows  differentiation between intentional gestures and unintentional human activity. The reason is that the robot can recognize what contacts are being made and match them against the human body parts in contact. This allows to be precise on the detection of intention. Compared to other input devices, a physical contact interaction can be more intuitive for a certain class of communication queues. It could ultimately relief operators from getting distracted during their interaction with personal mobile platforms.

In our approach, a human behavior is estimated by a depth image generated by a rotary laser scanner. The estimated gesture is not immediately recognized as an intentional command. When an external force is detected by the mobile platform's torque sensors, contact gestures can be recognized from the behavior information of the human. As a response, the robot can trigger a behavior that services the estimated intention.
\begin{figure}\centering
\includegraphics[width=0.7\linewidth, clip=false ] {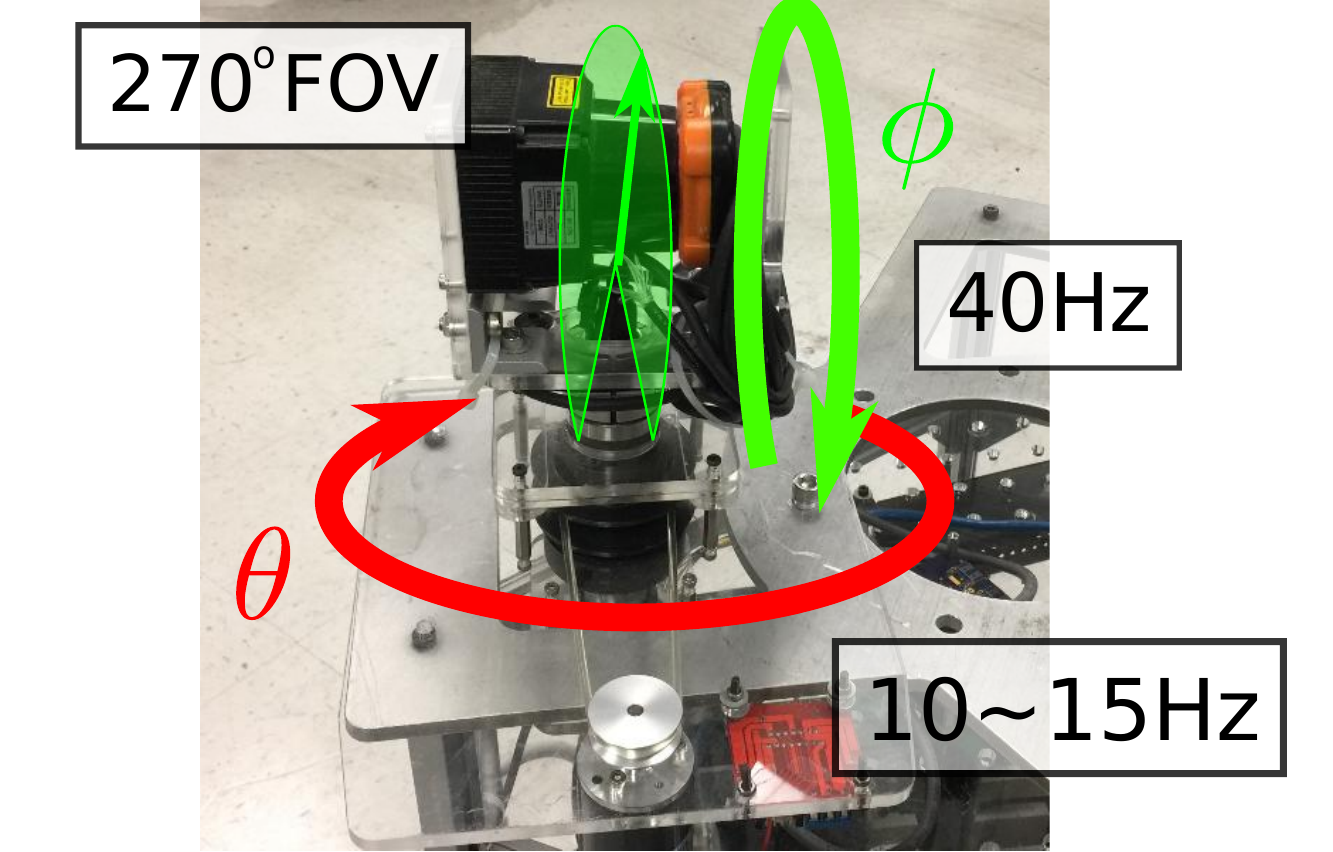}
\caption[3D scanner]{{\bf 3D scanner} made from a 2D LIDAR, Hokuyo UTM-30LX, can scan all around the robot. While the 2D lidar scans a vertical plane (green plane), the rotating gimbal rotates the plane with $10\sim15$Hz speed.}
\label{fig:lidar}
\end{figure}

\section{Related Works}
To detect humans and objects in unstructured environments, exteroceptive sensors such as cameras and range finders are often used. One of them is the Kinect sensor \cite{Chen2013}. The original application of this structured light sensor was as a gaming input device, so the sensor did not require $360^\circ$ scanning capabilities and had a relatively narrow field of view. Additionally, the Kinect and similar sensors have a relatively large minimum focal length making them unable to detect proximity at close range.
Recently, \cite{Magrini2015} detected multicontact on a robotic manipulator using a Kinect sensor. The sensor was located outside of the manipulator to allow it to work at the prescribed operating distances. This placement constraint highlights that Kinect type of sensors cannot be used to detect close range contacts or proximity in mobile platform { compared to the proposed system which has with the minimum distance of $10 \rm cm$}. 
Video cameras can also be used to detect humans around robots, but their narrow field of view limits their effectiveness.
Additionally, video cameras have very noisy and low resolution depth sensing making them less suitable to detect proximity behaviors.
A third method consists on using a laser range finder. A large number of research on human detection with these types of devices focus on finding and tracking pedestrians with 2D scanning \cite{Panangadan2010} which is used in autonomous cars to avoid collisions, for instance.
To scan a 3D environment with a range finder, multiple laser rays are shot and multiple planes are scanned simultaneously by a multilayer 3D LIDAR \cite{Wachs2010} or multiple 2D LIDARs \cite{Zhao2005}.
On the other hand, some researchers reconstruct a 3D environment with a single LIDAR by rotating it over its axis \cite{Panangadan2010}. 
2D LIDARs on rotary mounts are embedded in quadrotors \cite{Droeschel2014}, on mobile robots \cite{Ohno2009}, or on handheld poles \cite{Matsumoto2011}.
There are several types of 3D scanning methods addressing the rotating axes of the LIDAR and the rotary mount \cite{Wulf2003}.
In \cite{Panangadan2010}, the authors track human behavior via HMM's base on a planar LIDAR scanning.
In \cite{Anderson2014}, the authors suggested combining an interlaced 3D scanner with a 2D planar LIDAR.
This scanner is also known as Lissajous scanner, and is used for multiresolution microscopes \cite{Tuma2012}.
In \cite{Ebert2002}, the authors attempted to find a contact based on video sequences, but their estimated contacts did not include magnitude and direction of the applied forces.

\section{Hardware}\label{sec:hw}
\subsection{Omni-directional Mobile Robot}
Trikey shown in Figs. \ref{fig:exp02}, \ref{fig:poc} is a holonomic omnidirectional mobile robot which has torque sensors on its drivetrain \cite{Kim2015}. 
The external forces applied by users are detected by the torque sensors via model based whole-body sensing algorithms. We can estimate the location, magnitude and direction of external forces and collisions.
\begin{figure}[t]\centering
\includegraphics[width=0.8\linewidth, clip=false ] {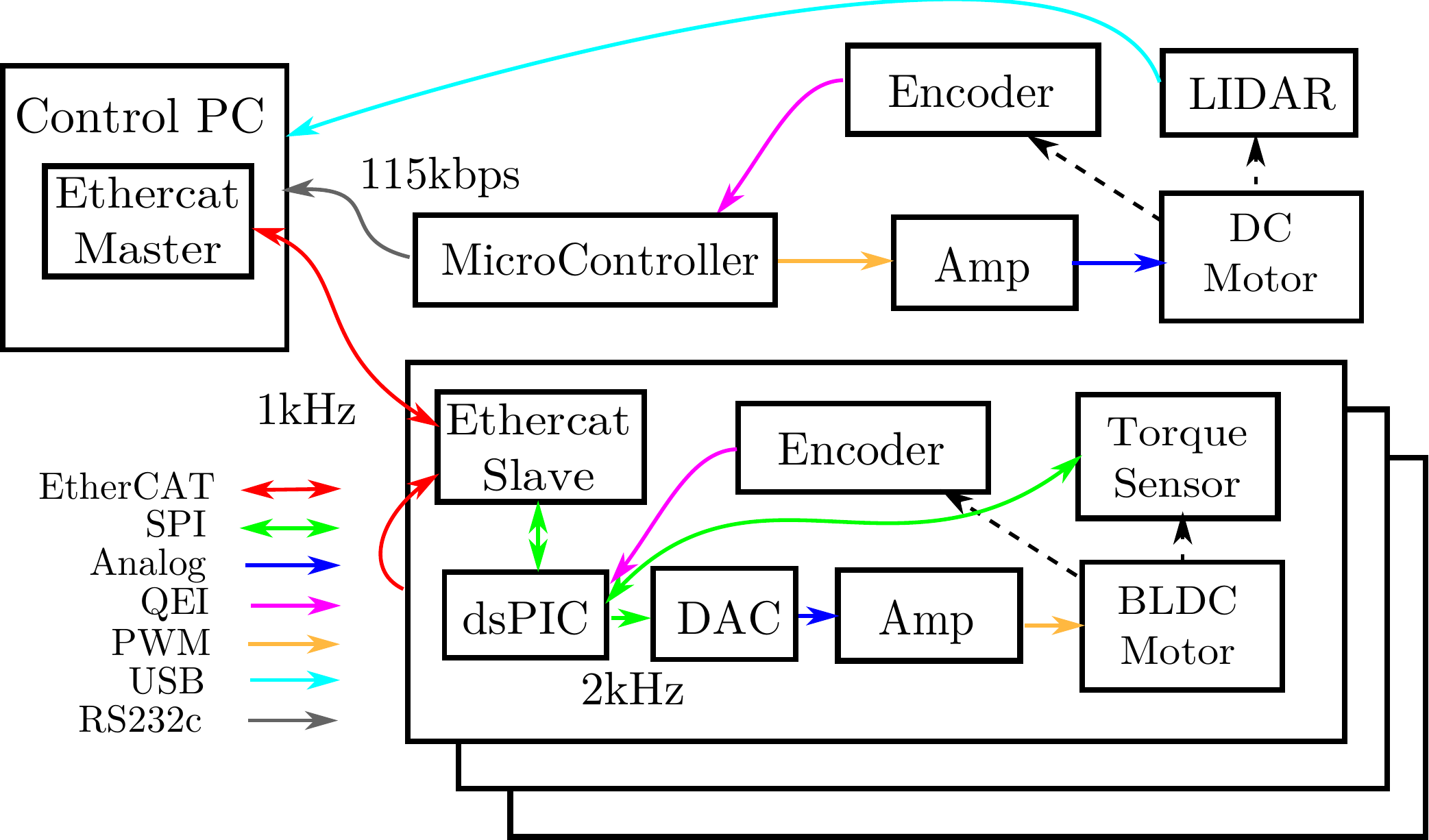}
\caption
[ The electrical system of the mobile platform] 
{{\bf The electrical system of the mobile platform} 
	consists of actuation parts and the 3D scanner. The microcontroller, which is Raspberry PI 2, for the 3D scanner controls the rotational speed of the gimbal and measures the orientation of the laser ray. It generates the orientation from the synchronization signal from the 2D LIDAR and the encoder on the gimbal.}
\label{fig:elec}
\end{figure}

\subsection{Interlaced Scanning}
We implement a 3D scanner by employing a 2D LIDAR (Hokuyo UTM-30LX) on top of a gimbal as shown in Figs. \ref{fig:lidar}, \ref{fig:elec}. The LIDAR triggers a synchronization signal to the GPIO port of the microcontroller. Whenever the laser in the sensor rotates once with 40Hz and the LIDAR generates a sequence of distance, we interpolate the timestamps of the signals, and generate a sequence of a tuple which consists of a timestamp, angle, and distance. 
Also, we attach an 2500-CPS optical encoder from US Digital Inc. (E6-2500-1000-IE-S-H-D-B) to the gimbal, and the QEI signals from the encoder are also fed into the GPIO ports. To deal with the signals from the LIDAR and the encoder, we execute a sequential program on the microcontroller, and its loop period is 300kHz. By merging the signals, we generate the orientation of the laser ray and deliver it to the control PC through RS232c serial communication with 115kbps baudrate.
The gimbal actuator is powered by a 12V DC motor which is controlled by L298N DC motor driver. The rotational speed is controlled by PWM signals and kept between 10 and 15Hz.
There are two rotational axes in the 3D scanner. $\phi$ is the angle of the laser ray in the 2D LIDAR, and $\theta$ is that of the gimbal rotating the 2D LIDAR as shown in Fig \ref{fig:lidar}.
When the laser ray shot by the 2D LIDAR collides with an object, the collision point with distance $d$ can be expressed in Cartesian coordinate as follows.
\begin{align}
\mathbf{p} &= 
\begin{bmatrix}
d \, \cos \theta \, \cos \phi  &
d \, \sin \theta \, \cos \phi  &
d \, \sin \phi
\end{bmatrix}^T
\end{align}
The scanned 3D position, $\mathbf{p}$, is specified by three variables $d$, $\phi$, and $\theta$, which represent polar coordinates. 
There are two different types of scanning methods: progressive (raster) scanning and interlaced (Lissajous) scanning \cite{Tuma2012,Anderson2014}. Progressive scanning methods are implemented in most of the 3D scanners made from 2D LIDARs. 
Typically, the rotation of the gimbal is much slower than the 2D scanning, and the points captured during the previous rotation are replaced with the new points.
Therefore, the points in the constructed point cloud from the progressive scan have high correlation between position and time because the scanning is conducted sequentially from one scan line to another scan line.
However, in the case of the interlaced scanning, the scanning speed of the gimbal is comparable to the LIDAR scanning which means the scanning method is more suitable for tracking objects moving fast because the interlaced scanning is more responsive than the progressive scanning, and the points clouds are generated from the points captured during several rotations of the gimbal.
Considering the point cloud generated from the progressive scanning, the temporal sequence of the points is usually ignored meaning that we get a snapshot taken at a given time.
On the other hand, the points from interlaced scanners have a low correlation between position and time because adjacent points can belong to different scan lines.
Therefore, each point has its own generation time which is not related to the position, so we need to deal with not only position of the point but also its timestamp.
\begin{figure}\centering
\includegraphics[width=\linewidth, clip=false ] {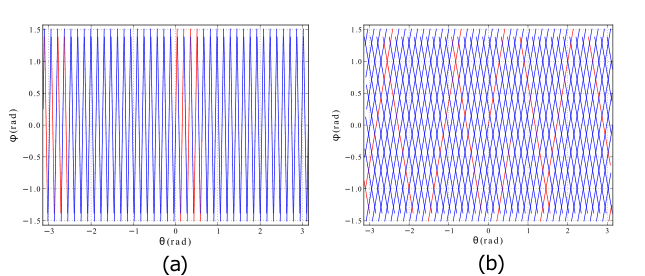}
\caption
[ $\phi$-$\theta$ coordinates of (a) progressive scanning and (b) interlaced scanning for 1 second] 
{{\bf $\phi$-$\theta$ coordinates of (a) progressive scanning and (b) interlaced scanning for 1 second} 
	show that both scanning methods have similar spatial resolutions, but the 0.1 second scan lines (red) show that the interlaced scan lines are scattered more which means the scan lines cover more area. }
\end{figure}

\section{Contact Gesture Recognition}
An external force estimation method has been derived with certain limitations as described in \cite{Kim2015}. 
With the help of a 3D scanner, we can also identify the location of the contact which generates an external force and, relax some of the constraints of the estimation.
In addition to the estimated contact forces, the point clouds generated by the 3D scanner include the information about the object making the contact.
We assume that the object is a human body, and identify which parts of the human body make contacts with the mobile platform. All the estimated contact information is used to generate a contact gesture.
\begin{figure}[t]\centering
\includegraphics[width=0.9\linewidth, clip=false ] {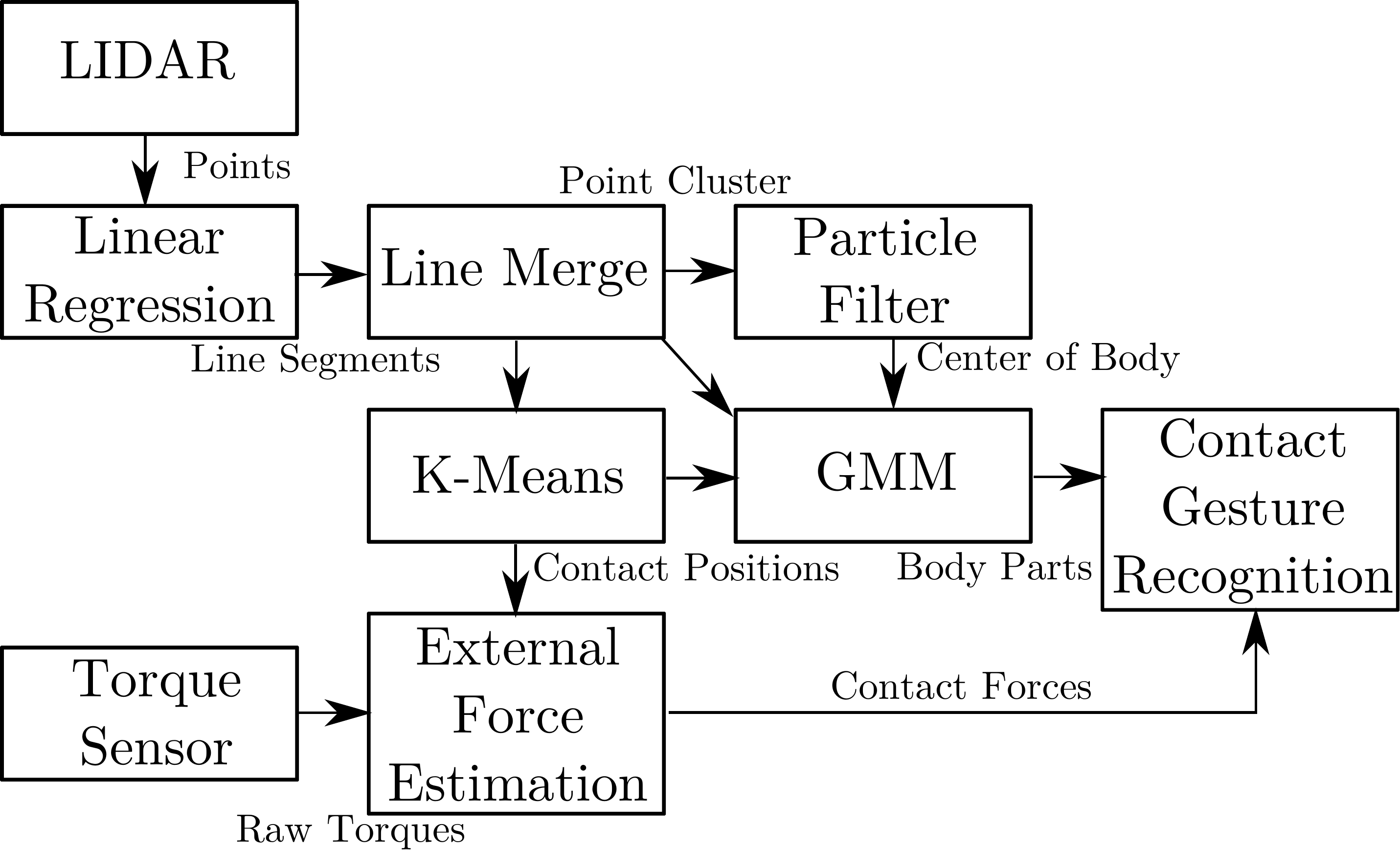}
\caption
[ Contact Gesture Estimation]
{{\bf Contact Gesture Estimation} 
	is implemented using the 3D scanner and the torque sensors. The estimated contact gesture consists of the number of contacts, the location and the force vector of each contact, and the human body part that makes the contact.
\label{fig:est}}
\end{figure}

\subsection{Point Cloud Registration}
\label{sec:pcloud}
\begin{figure}[t]\centering
\includegraphics[width=0.9\linewidth, clip=false ] {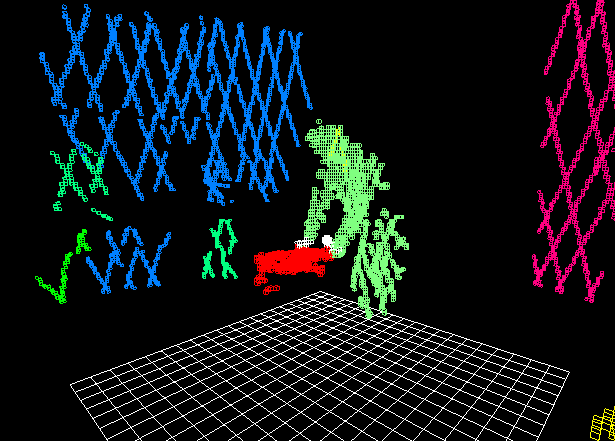}
\caption
[ An occupancy map evaluated by Octree] 
{{\bf An occupancy map evaluated by Octree} 
	includes the mobile platform and nearby objects. The mobile platform (red voxels) is identified from the predefined shape.}
\label{fig:octree}
\end{figure}

When the LIDAR sensor and the encoder of the gimbal system generate a polar coordinate of the shooting laser, it occupies one voxel in an octree \cite{meagher1980} with the resolution of $2 \rm cm \times 2 \rm cm \times 2 \rm cm$.
We register the shape of the top plate of the mobile platform as a triangular plane, and voxels outside of the plane are considered to be separate objects.
So far, all the voxels in the space are separated into two groups: platform voxels and object voxels. To identify whether there is an object making a contact with the platform, we measure the distance from each object voxel and the triangular plane, and if the distance is less than a given threshold, we identify the voxel as just next to the platform. 
Fig. \ref{fig:octree} shows the platform voxels and the object voxels which are making contacts with the platform. 
The object voxels are grouped together, and the mean position of the voxels in each group corresponds to the location of the contact which is used in the subsequent section.

\subsection{Point Clustering and Object Tracking}
\label{sec:object}
After the points are generated from the 2D LIDAR scanning and added to the octree, the points are classified into several point clusters. 
To reduce the computation time for the clustering, we form a line segment from the LIDAR data using the incremental least square linear regression algorithm, and each line segment is classified into the nearest point cluster. 
Then, we can deal with each point cluster as an object. Some objects can be human bodies and others are environments. 
In this study, we assume that any objects close to the robot are human operators who want to interact physically with the robot.

To estimate the contact gesture, we start by tracking human body which results in a set of time trajectories of the human operator.
To incorporate both the hitting laser and the missing laser, we use a particle filter for the object tracking, and each particle represents the candidate center point of the object.
We assume that the shape of the tracking object is an ellipsoid which can be expressed with the covariance matrix of the point cluster distribution, $\mathbf{\Sigma}$, as follows.
\begin{align}
(\mathbf{x}-\mathbf{x}_c)^T \mathbf{\Sigma}^{-1} (\mathbf{x}-\mathbf{x}_c)  = 1
\end{align}
where $\mathbf{x}_c$ is the center of the object.
If the j-th laser ray from the 3D scanner hits a point, $z_j$, then we can test whether the point is located in the object whose center is the i-th particle through the weighted distance, $d_{ij}$ between them as follows.
\begin{align}
d_{ij} = (\mathbf x_i-\mathbf z_j)^T \mathbf{\Sigma}^{-1} (\mathbf x_i-\mathbf z_j) 
\end{align}
where $\mathbf{x}_i$ is the postion of the i-th particle.
If $d_{ij}$ is smaller than or equal to $1$, that means the point belongs to the object.
Therefore, we can generate a conditional probability of the observation where the j-th laser ray hits on $\mathbf z_j$ if the center of the object is located on the i-th particle, $\mathbf x_i$ as follows.
\begin{align}
P \left( \mathbf z_j | \mathbf x_i \right) = 
\left\{ 
\begin{matrix}
\epsilon\ : \ d_{ij} \le 1\\
0\ : \ d_{ij} > 1\\
\end{matrix}
\right.
\end{align}
Then, the conditional probability that the center of the object is located on the i-th particle $\mathbf{x}_i$ given the observation, $\mathbf{z}_j$ can be derived from the Bayesian rule as follows.
\begin{gather}
P\left(\mathbf{x}_i | \mathbf{z}_j \right) = \frac{P(\mathbf{x}_i) \, P \left(\mathbf{z}_j | \mathbf{x}_i \right)}{P(\mathbf{z}_j)}
\end{gather}
where 
\begin{gather}
P\left(\mathbf{z}_j\right) = \sum_{\mathbf{x}_i, d_{ij} \le 1} \epsilon P\left(\mathbf{x}_i\right) \label{eq:bay}
\end{gather}

When the laser misses the object, the missed laser ray can also be used to confirm that the object does not exist on the ray.
The laser ray can be represented as a matrix equation $\mathbf{A}\,\mathbf{z} = \mathbf{0}$ because the laser comes from the origin. Then, the conditional probability of the missed laser ray given the object position, $\mathbf x_i$ can be expressed as follows.
\begin{align}
P \left( \mathbf{A}\,\mathbf z = \mathbf{0} | \mathbf x_i \right) = 
\left\{ 
\begin{matrix}
\eta\ : \min_{\mathbf{z}_j \in \mathbf{z}} d_{ij} > 1\\
0\ : \min_{\mathbf{z}_j \in \mathbf{z}} d_{ij} \le 1
\end{matrix}
\right. 
\end{align}
The conditional probability of the i-th particle with respect to the given missed laser ray events, $P\left(\mathbf{x}_i | \mathbf{A}\,\mathbf z= \mathbf 0 \right)$ also can be derived by the Bayesian rule similar to Eq. (\ref{eq:bay}).

\subsection{Contact Position and Human Posture Estimation}
\label{sec:contact}
From the previous section, the robot can generate the point cluster of a human body and identify its center position. 
As described in Sec. \ref{sec:pcloud}, when the point cloud is registered in the occupancy map, we check how close each point is to the predefined mobile platform shape.
Therefore, we can build a set of all the points close enough to the robot by selecting points whose distance to the platform is smaller than a given threshold, and name it as a contact point set. 
The points in the set are clustered by k-means algorithm \cite{macqueen1967}, then the number of contacts the human body makes can be identified by the number of the clusters.
The initial states of the clusters are established from the center of the points in the set, and all the points are classified into one of the clusters after the algorithm converges. 
Cluster with no points are removed, and the remaining clusters are considered as contact positions.
In this study, $k\,=\,3$ which means we assume that the contact points can be both hands and the body of the human operator.
By identifying contact positions, it can be assumed that the external forces are applied on the contact positions. 
The locations of the contacts are fed into the external force estimator as described in Sec. \ref{sec:multi}.

The object point cloud, the center of the cloud, and the contact positions to the mobile platform are estimated from the 3D scanner.
The body posture of the human operator can also be estimated from the point cloud, then we can identify how the human operator makes a contact with the mobile platform which we call a contact gesture.
To determine body posture from the point cloud, we express a human body as a Gaussian mixture model (GMM) with expectation-maximization (EM) algorithm. Each Gaussian distribution of GMM represents a human body part, and we can identify the contact gesture from the locations of the Gaussian distributions.
In the study, we use 4 mixture components which represent left and right arms, a upper body, and a lower body if there are two contact locations. If there is only one contact location, we use 3 mixture components.
The initial distributions of the mixture components are initialized with the center of the point cluster and the contact locations, and the EM algorithm iterates until the change of distributions is below a given threshold.
Even though the human body posture is oversimplified as only 3 or 4 mixture components, the GMM representation has enough information to identify which part of human body makes a contact with the mobile platform.
With the help of estimation methods described in this section, we can track a human operator around the mobile platform, and when the operator makes a contact, we can identify which parts of the operator make contacts with which parts of the mobile platform. The whole estimation process is depicted in Fig. \ref{fig:est}.

\subsection{Multicontact Force Estimation}
\label{sec:multi}
In the previous section, we have figured out where the contacts happen on the mobile platform, which are described as 3D coordinates on the robot's frame. 
Assume that there are $N$ objects making contacts with the top of the platform, and the location of the i-th contact location is $\left(x_i, y_i\right)$ which is in the local frame of the mobile platform, and the contact force is $\mathbf{F}_{i} = \left( F_{i,x}, F_{i,y}\right)$ which includes no torque. Then, the sum of all the forces satisfies the following condition with respect to the net force and torque on its center, $\mathbf{F}_{ext} =
\begin{pmatrix} F_{ext,x} & F_{ext,y} & \tau_{ext}\end{pmatrix}^T$.  
which can be expressed as the following matrix form.
\begin{align}
\mathbf{F}_{ext} =
\mathbf{H}_N \, \mathbf{F}_N 
\end{align}
where 
\begin{align}
\mathbf{H}_N &=
\begin{pmatrix}
\begin{matrix}
\mathbf{I}_{2\times2} \\
\begin{matrix}
-y_1 & x_1 &
\end{matrix}
\end{matrix}
&
\cdots 
&	
\begin{matrix}
\mathbf{I}_{2\times2} \\
\begin{matrix}
-y_N & x_N
\end{matrix}
\end{matrix}
\end{pmatrix} \in \mathbb{R}^{ 3 \times 2N }\\
\mathbf{F}_N &=
\begin{pmatrix}
F_{1,x} & F_{1,y} & \cdots & F_{N,x} & F_{N,y}
\end{pmatrix}^T
\end{align}
From \cite{Kim2015}, the external net force on the center of the mobile platform, $\mathbf{F}_{ext}$ can be derived from the joint torques as follows.
\begin{align}\label{eq:fext}
\mathbf F_{ext} &= \mathbf{J}_{c,w}^T \left(\mathbf \Gamma \big\rvert_{\mathbf F_{ext}=0} - \mathbf \Gamma_s\right)
\end{align}
The size of $\mathbf{H}_N$ is determined by the number of contacts, and the estimation can be either overdetermined or underdetermined.
In either case, we can estimate the contact forces, $\widetilde{\mathbf{F}}_N$ as follows.
\begin{align}
\widetilde{\mathbf{F}}_N = \mathbf{H}_N^T \left(\mathbf{H}_N \mathbf{H}_N^T \right)^{+} \, \mathbf{F}_{ext}\label{eq:norm}
\end{align}
where $\left(\cdot\right)^+$ is a pseudoinverse. The solution is equivalent to a minimum norm estimation if it is underdetermined and a least mean square error estimation if overdetermined.

When there are more than one contact point, the estimation is underdetermined because 
the joint torque sensors cannot sense all the contact forces such as squeezing and stretching, which means the estimated contact forces, $\widetilde{\mathbf{F}}_N$ are different from the actual contact forces, $\mathbf{F}_N$. 
However, the multicontact may happen simultaneously, so the estimated first contact information can be exploited during a multicontact estimation.
For the first contact, we can determine the unit vector of the force, $u_{1} = \begin{pmatrix} u_{1,x} & u_{1,y} \end{pmatrix}^T$ and apply it to the minimum norm estimation process as follows.
\begin{gather}
\mathbf{H^\prime} = 
\begin{pmatrix}
\mathbf{H}_N \\ 
\begin{matrix}	
u_{1,y} && -u_{1,x} && 0 && \cdots & 0
\end{matrix}
\end{pmatrix} \in \mathbb{R}^{(3+1) \times 2N} \nonumber\\
\mathbf{F}^\prime_{ext} = 
\begin{pmatrix}
\mathbf{F}_{ext} \\ 0
\end{pmatrix}
\label{eq:Hprime}
\end{gather}
Using Eq. (\ref{eq:Hprime}), contact forces are estimated with the first contact information as follows.
\begin{gather}
\widetilde{\mathbf{F}}^{\prime}_N = \mathbf{H}^{\prime T} \left(\mathbf{H}^{\prime} \mathbf{H}^{\prime T} \right)^{+} \, \mathbf{F}^{\prime}_{ext} \label{eq:norm2}
\end{gather}

\subsection{Reaction to Human Intention}
\label{sec:reaction}
Using a new algorithm shown in Fig. \ref{fig:est}, we identify all contacts on the mobile platform and corresponding human body parts in contact.
Subsequently, the mobile platform can respond according to the estimated contact gestures.
Inspired by touch-based APIs in mobile devices, we define a multi-touch event including location of a contact, contact force vector, and human body part in contact.
When the mobile platform detects $n$ contacts, each contact is labeled with the corresponding human body part, $i \in \mathbf{P} \triangleq \left\{ left\_hand,\ right\_hand,\ body \right\}$, and touch event, $t_{e,i}$ belonging to a set of multiple touches, $\mathbf{T}_e$. Each touch event includes a location vector ($\mathbf{l}_{e,i}$) and the force vector corresponding to the touch ($\mathbf{f}_{e,i}$).
To react to the contact gestures, a command set, $\mathbf{C}$ is defined. Each command, $c \in \mathbf{C}$ includes a set of triggering touch information, $\mathbf{T}_c$. A triggering touch made by the human body part, $i$, is denoted as $t_{c,i} \in \mathbf{T}_{c}$, and it consists of the location vector ($\mathbf{l}_{c,i}$) and the touch force vector ($\mathbf{f}_{c,i}$).
Given the estimated touches, we can find the desired command, $c$, from the command set from the following equation.
\begin{align}
\argmin_{c \in \mathbf{C}} \sum_{i \in \mathbf{P}} 
\mu_i \left( \mathbf{T}_e,\ \mathbf{T}_c \right)
\left( 
	w_l \left| \mathbf{l}_{e,i} - \mathbf{l}_{c,i} \right| 
	+w_f \left| \mathbf{f}_{e,i} - \mathbf{f}_{c,i} \right| 
\right) \label{eq:cmd}
\end{align}
where $w_l$ and $w_f$ are weights for distance and torque, respectively, and 
$\mu_i$ is a function of sets which has the following property
\begin{align}
\mu_i \left( \mathbf{T}_1, \mathbf{T}_2 \right) = 
\left\{ 
\begin{matrix}
	1,  & \exists\ t_{1,i} \in \mathbf{T}_1,\ \rm and\ \exists\ t_{2,i} \in \mathbf{T}_2 \\
	0,  & t_{1,i} \not \in \mathbf{T}_1,\ \rm and\ t_{2,i} \not \in \mathbf{T}_2 \\
	\infty, & \rm otherwise
\end{matrix}
\right.
\end{align}
\section{Experiments}
\input{experiments.tex}

\section{Conclusion}
In this study, we have devised a methodology for identifying contact gestures between humans and omnidirectional mobile platforms.  
To estimate contact gestures, we combine data from a 3D scanner which is constructed using a rotating 2D LIDAR and rotary torque sensors on the platform's drivetrains. We use this infrastructure to determine which human body parts make contact with mobile platforms and how much forces they apply to it.
To achieve responsive and omnidirectional contact detection using the 3D scanner, we choose an interlaced scanning procedure, where its meshlike scan map enables instantaneous contact detection. 
Even though it is hard to reconstruct a sophisticated 3D environment with the coarse scanning sensor, our method can obtain enough contact information for effective physical HRI. 
{ We assume that all the nearby objects are human operators, but this limitation can be relaxed by adopting pervasive object classification methods.}
The contact information includes the location of a contact and the human body parts making contacts, which are determined via clustering methods.
Also, by fusing these contact information with the rotary torque sensory data, we can estimate the contact force on each contact location. The estimation is underdetermined, so we apply a minimum norm estimation and prior contact information. Finally, we demonstrate the possibility of using contact gestures as a physical HRI tool through various proof of concept experiments. In those experiment, the mobile platform identifies the predefined contact command queues and response according to the commanded gestures. In the future, we will focus on achieving higher accuracy on detection and faster responsiveness to human gestures. 
{The complexity of our detection algorithm is proportional to the number of nearby human operators, so it can be extended without a great effort.
}
We will also focus on developing a more meaningful language for contact based communications for effective pHRI.

%% file: experiments.tex

In this section, we conduct experiments with the mobile platform, and prove that the proposed algorithm guarantee the effective retrieval of the contact gestures.
The experiments consist of
1) contact force estimation experiment in which contact positions are identified by the 3D scanner, and we estimate the contact force on each position; 
2) contact gesture recognition experiment in which the posture of a human operator is estimated, and we figure out which parts of the operator make contacts with the mobile platform; and 
3) proof of concept experiment in which we show examples of how the proposed contact gesture recognition can be used as a physical HRI tool.
\subsection{Contact Force Estimation}
\begin{figure}\centering
\includegraphics[width=0.8\linewidth, clip=false ] {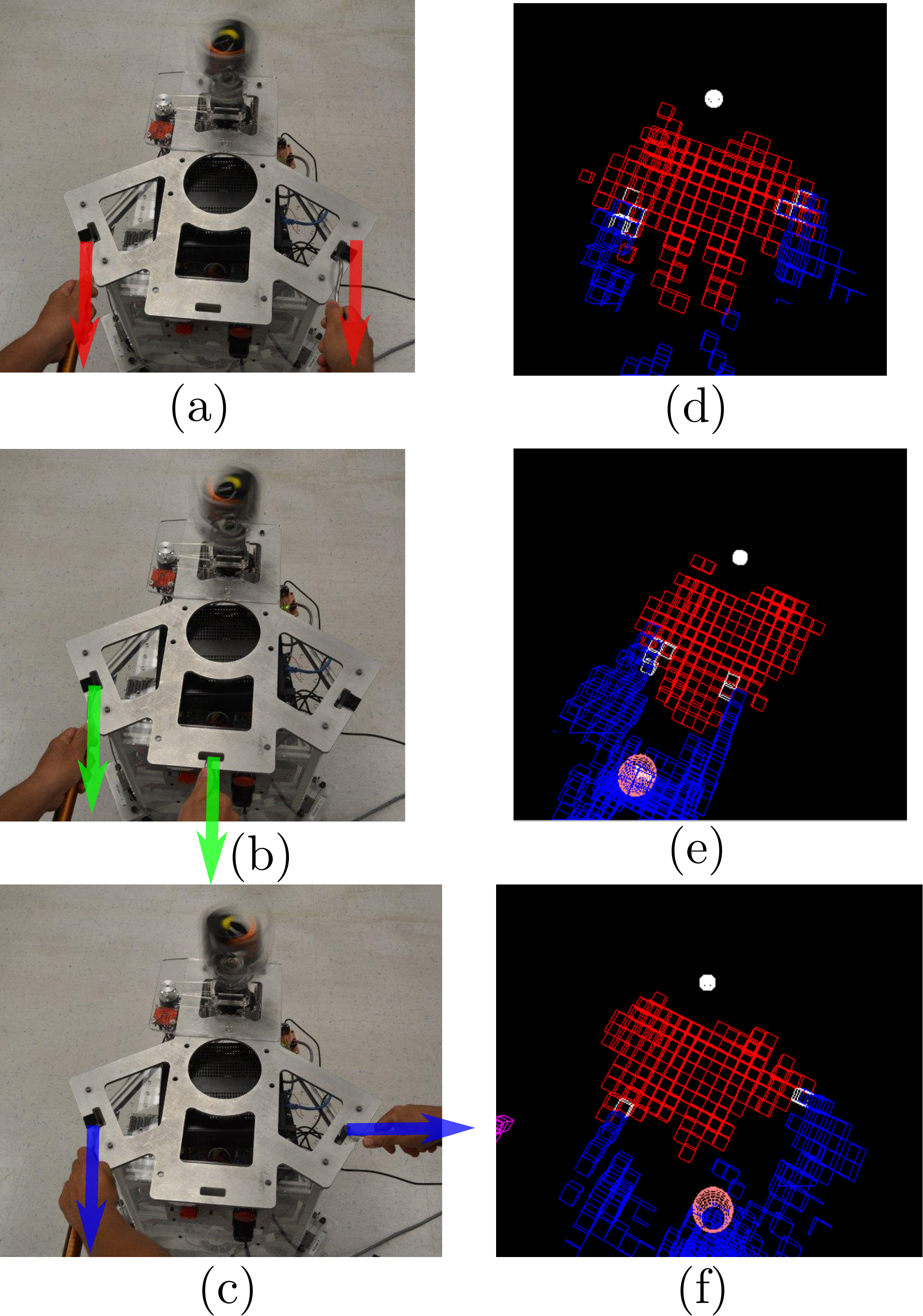}
\caption
[ Multicontact Force Estimation] 
{{\bf Multicontact Force Estimation} 
	is conducted on the top of the mobile platform. Both hands of the human operator make contacts and apply some forces in (a)$\sim$(c). In (d)$\sim$(f), the corresponding occupancy maps are shown. The red voxels are the mobile platform, the blue voxels are the human operator, and the white voxels are the voxels of the human operator which are making contacts.}\label{fig:exp01}
\end{figure}
In this experiment, we determine contact forces when a human operator makes contacts with the mobile platform. 
To generate the calibrated contact forces, the human operator applies $10\rm N$ of contact forces by pulling the mobile platform with spring scales as shown in Figs. \ref{fig:exp01}-(a)$\sim$(c). 
With the 3D scanner, the contact positions are observed, and the contact forces are estimated from the contact positions and the net force generated from the rotary torque sensors.

Figs. \ref{fig:exp01}-(a)$\sim$(c) show the configurations of the contact positions. In the case of Figs. \ref{fig:exp01}-(a) and (c), the contact positions are identical, and there is no difference in the 3D scanner data. However, the operator applies different contact forces, and the contact gesture recognition method is able to resolve for the different forces.
In Figs. \ref{fig:exp01}-(a) and (b), the contact forces are applied in the same direction, while in Fig. \ref{fig:exp01}-(c)the contact forces are perpendicular.
In Figs. \ref{fig:exp01}-(d)$\sim$(f), the occupancy maps are generated from the point clouds of the configurations of Figs. \ref{fig:exp01}-(a)$\sim$(c), respectively. 
The red voxels are the mobile platform, and the blue voxels are the human operator. 
The white voxels belong to the human operator, and are close enough to be considered as making a contact with the mobile platform.
The pink sphere is the estimated center of the human operator from the particle filter in Sec. \ref{sec:object}. As described before, the contacts in Figs. \ref{fig:exp01}-(a) and (c) are identical in position.

\begin{figure}[p!]\centering
\includegraphics[width=\linewidth, clip=false ] {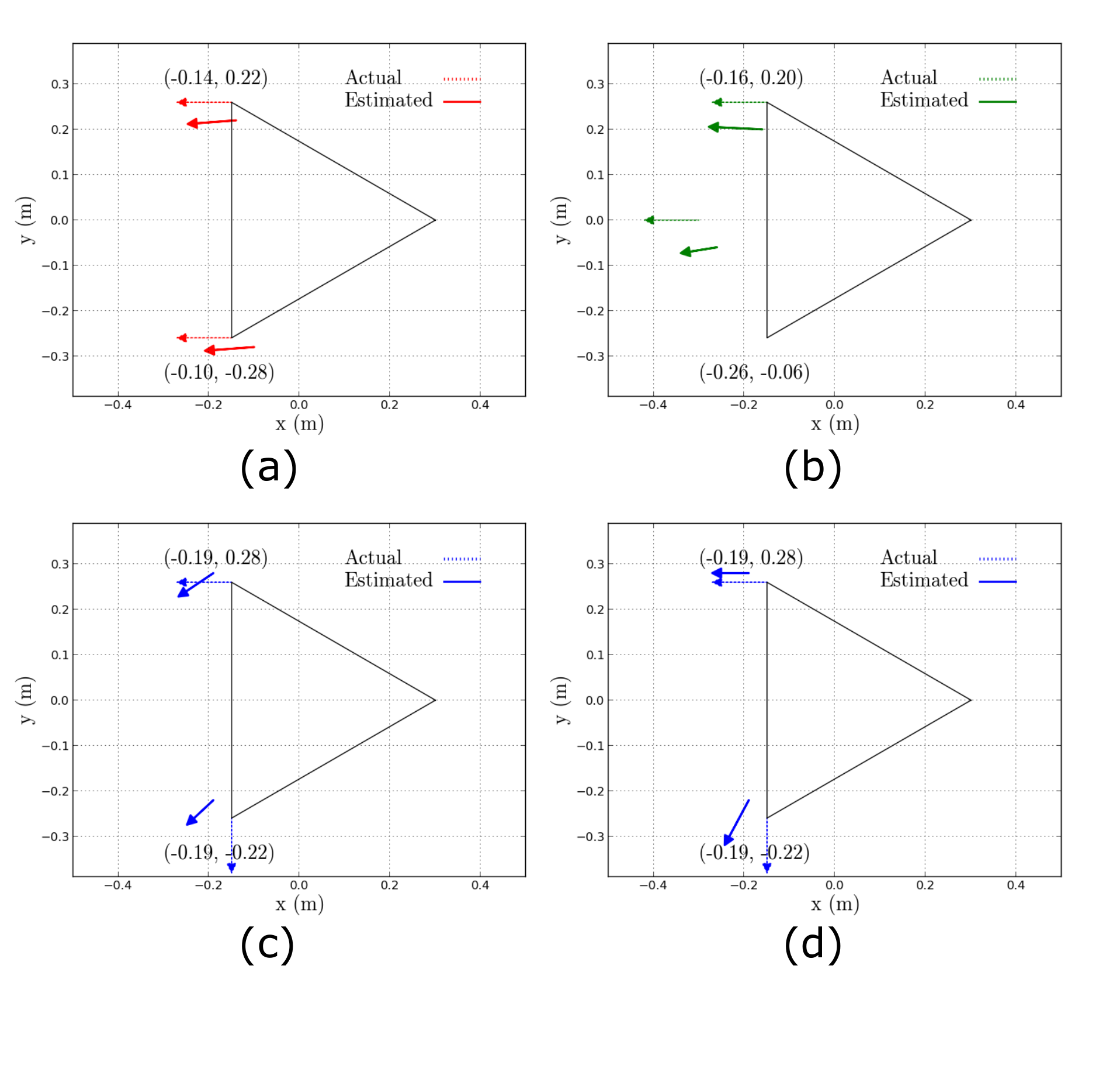}
\caption
[ Estimated Multicontact Positions and Forces] 
{{\bf Estimated Multicontact Positions and Forces} 
	are shown in the graphs. The forces in Fig (c) and (d) are estimated from the same data, but (d) uses a prior information of the left hand contact (upper one).}\label{fig:exp01_02}
\end{figure}

Figs. \ref{fig:exp01}-(d)$\sim$(f) show both the estimated contact positions and the contact forces. The contact positions are calculated from the average of contact voxels in the occupancy maps, and the contact forces are estimated from the net force and the contact positions with Eq. (\ref{eq:norm}).
Because the applied contact forces in Figs. \ref{fig:exp01_02}-(a) and (c) are in the same direction, the estimated fores are identical from the minimum norm estimation process.
On the other hand, in Fig. \ref{fig:exp01_02}-(c), the forces are not in the same direction, and therefore some components cancel each other out, resulting in the estimated forces being different from the actual ones.
To estimate contacts more precisely, the human operator makes contacts sequentially, and the information from the first contact is used as a prior. The left hand makes a contact earlier than the right hand, the direction of the first contact is added to $\mathbf{H}^\prime$ in Eq. (\ref{eq:Hprime}), and the contact forces are estimated with Eq. (\ref{eq:norm2}).
The estimation with the sequential contacts in Fig. \ref{fig:exp01_02}-(d) shows that the estimated forces are close to the actual forces.

\subsection{Contact Gesture Recognition}
\begin{figure}[t]\centering
\includegraphics[width=\linewidth, clip=false ] {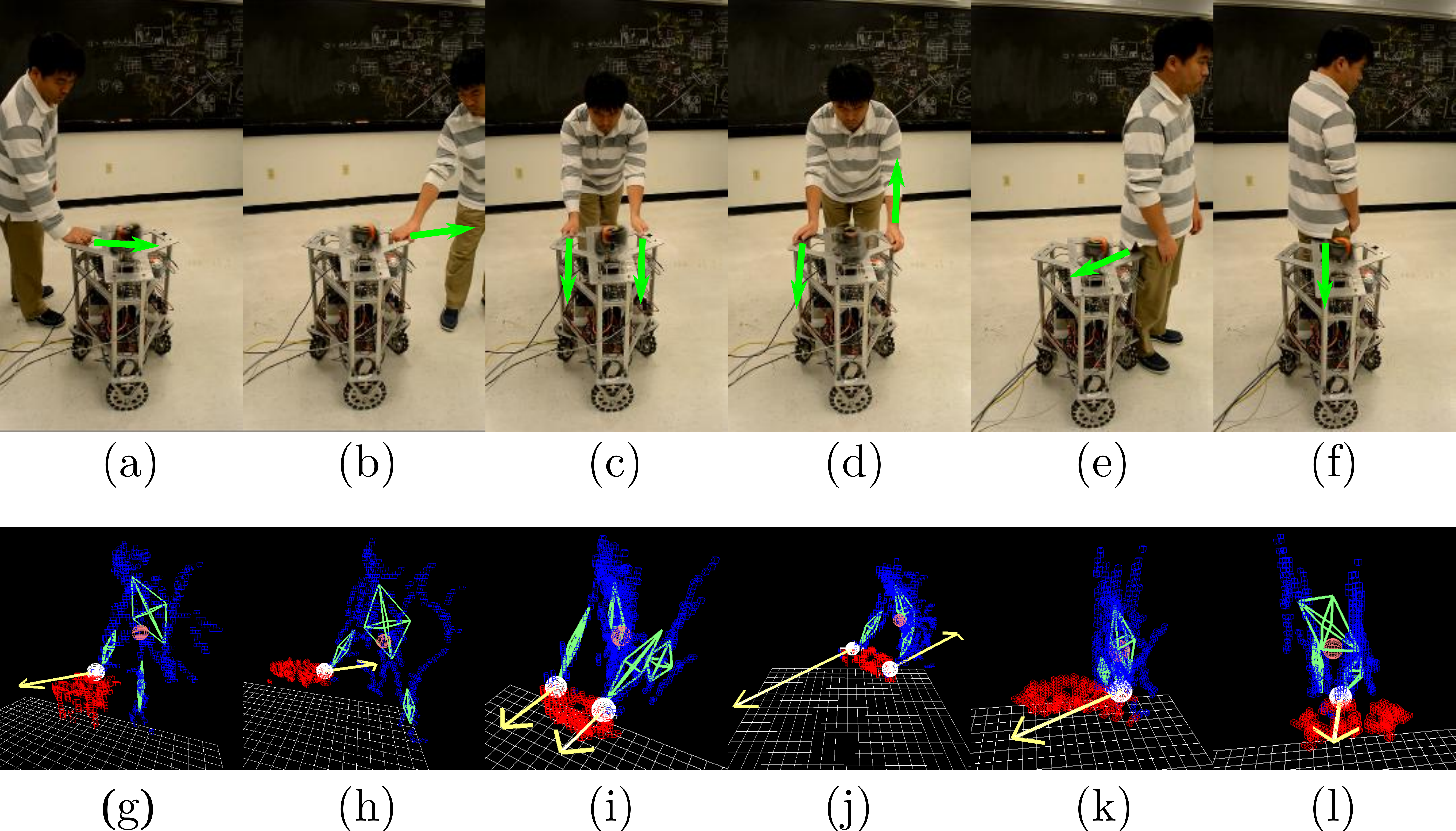}
\caption
[ Contact Gesture Recognition] 
{{\bf Contact Gesture Recognition} 
	for multiple contact situations are shown in Fig (a)$\sim$(f). Their corresponding occupancy maps are shown in Fig (g)$\sim$(l), respectively. The white spheres show the contact locations, the green octahedrons represent the identified human body parts, and the yellow arrows are the estimated contact forces.}\label{fig:exp02}
\end{figure}

\begin{table}[] \centering
\caption
[Command Set]
{\bf Command Set}
\vspace{0.15in}
\label{tbl:cmd}
\begin{tabular} {|c|c|c|c|c|}
\hline
Name & Part & Location (m)& Force (N) & Action\\[1.2ex]
\hline
\hline
Collide & body & (0,0) & (5,0) & Move away quickly\\[1.2ex]
\hline
Push & right & (0,-0.3) & (5,0) & Go straight\\[1.2ex]
\hline
Pull & right & (0,-0.3) & (-5,0) & Go straight and \\
		 &&&& come back 
\\[1.2ex] 
\hline
Rotate & left & (0,-0.3) & (5,0) & Rotate\\[1.2ex]
\cline{2-4}
& right & (0,0.3) & (-5,0) &
	\\[1.2ex]
\hline
\end{tabular}
\end{table}
In addition to the previous experiment, we estimate the body posture of the human operator making contacts with the mobile platform. In this experiment, the following body parts of the human operator make contact: 1) a single hand, 2) both hands, and 3) a thigh. 
We simplify the human body and express it as three or four parts: one or two arms, an upper body, and a lower body.
The number of arms is determined by the number of contacts which is identified by the k-means method. 
Also, while making contacts, the human operator applies forces by pulling or pushing.
Fig. \ref{fig:exp02} shows the experimental results.

In Figs. \ref{fig:exp02}-(a)$\sim$(f), the human operator's posture is rendered, and the estimated contact gestures are shown in Figs. \ref{fig:exp02}-(g)$\sim$(l).
In those figures, the white spheres are the estimated contact positions identified from the point clouds, and the green octahedron shows the Gaussian distribution of the estimated human body parts. Each vertex of the octahedron is 1-$\sigma$ boundary of the distribution. The yellow arrows are the estimated contact forces. 

In Figs. \ref{fig:exp02}-(a) and (b), single pulling and pushing forces are applied to the mobile platform, respectively, and The green arrows show the applied forces.
As shown in Figs. \ref{fig:exp02}-(g) and (h), 
a single contact point is found, and the three Gaussian distributions are identified as the human body parts.
The arm clusters converge to the real arm distributions, and the contact is determined to be at the end of the arms. Thus, we can determined what hands make contacts with the mobile platform. Also, the contact forces estimated from the external force estimator in Fig. \ref{fig:est} have the correct directions with respect to the applied forces.

In the case of the multicontact experiments shown in Figs. \ref{fig:exp02}-(c) and (d), a pushing and twisting forces are applied to the mobile platform, respectively. 
The k-means algorithm identifies that there are two contact points, and therefore four Gaussian distributions are used for all contact situations. Both arm clusters converge towards the actual visualized arms, allowing to determine that both hands make contacts with the mobile platform. Figs. \ref{fig:exp02}-(i) and (j) show the estimated forces with the contact gestures, and the directions of the contact forces are identical to the actual forces.  

Contact with the lower human body is tested in Fig. \ref{fig:exp02}-(e) and (f). Typically, this kind of contact means a collision that needs to be avoided. In the experiment, the human operator leans toward the mobile platform, and a pushing force is accidentally applied to it. As shown in Fig. \ref{fig:exp02}-(k) and (l), all three distributions are located on the lower human body area. 
By comparing the covariance of the Gaussian distributions to those of other contact events, we can distinguish between lower body contacts and hand contacts.

\subsection{pHRI through Contact Gesture}
\begin{figure*}\centering
\includegraphics[width=\linewidth, clip=false ] {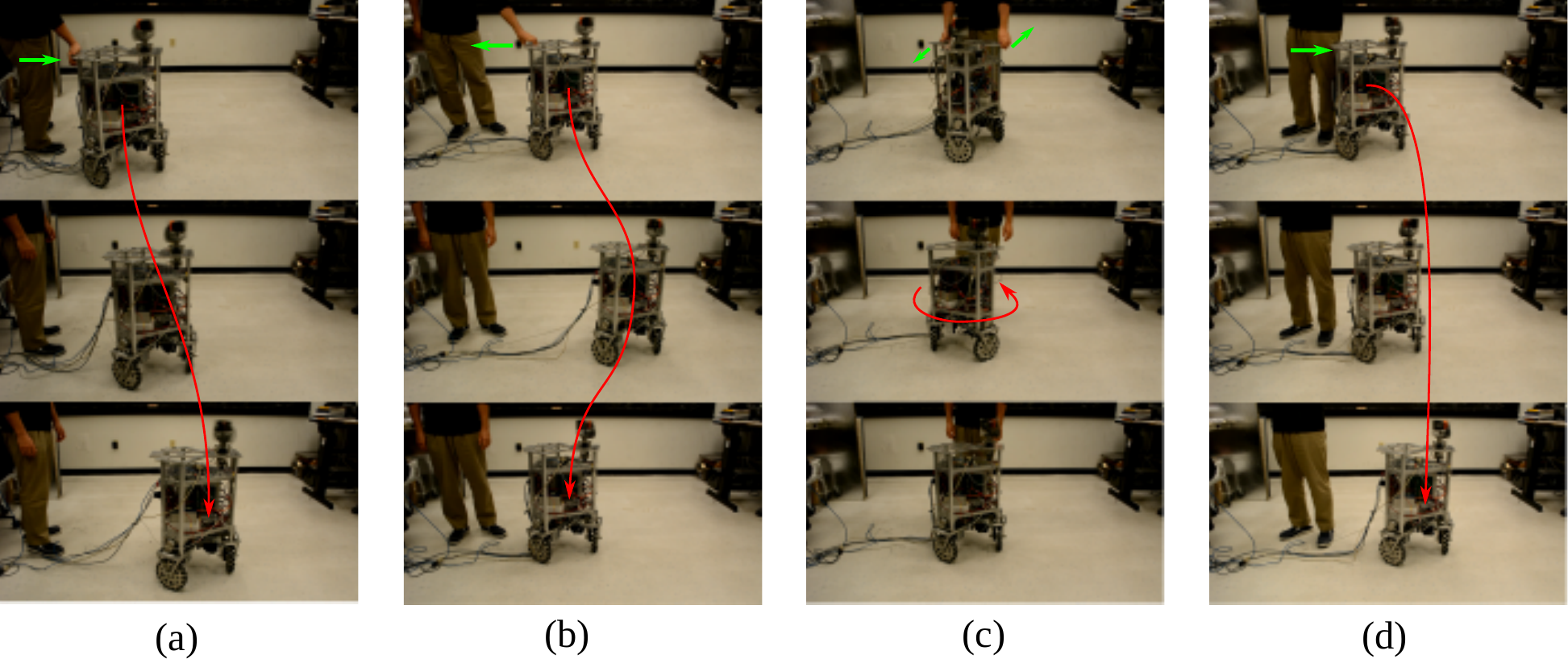}
\caption
[ pHRI Experiments using Contact Gesture Recognition] 
{{\bf pHRI Experiments using Contact Gesture Recognition} 
	demonstrate four use cases. Each use case starts from the top row. When the mobile platform detects contact forces, it operates following the predefined reaction table shown in Table \ref{tbl:cmd}.}\label{fig:poc}
\end{figure*}

In the last experiment, we prove the concept of HRI with the proposed contact-based gesture recognition.
For the experiments, we define four commands in Table \ref{tbl:cmd}.
If the magnitude of the estimated contact force is greater than a triggering threshold of $5\rm N$, the mobile platform determines one command in the set which has the smallest test value from Eq. (\ref{eq:cmd}), and executes the predefined action.
Fig \ref{fig:poc} shows the operations of the mobile platform commanded through the recognized contact gesture. Even though the difference between the gestures in Fig. \ref{fig:poc}-(a) and (b) are insignificant, the commands are different by the estimated forces. 
Also, by identifying the human body part in contact, we can differentiate an intentional action from an undesired collision as shown in Fig. \ref{fig:poc}-(a) and (d).

%% file: chapter-tro.tex
\ifdefined\JOURNAL
\else
\chapter{Intelligent Collision Intervention to Increase Peoples Safety}
\fi

\section{Introduction}
Is it okay for robots to stop objects or other robots that could collide with people? Although ground robots and autonomous cars have operated for a while in human populated environments, it is unusual to see them intervening over collisions that might happen externally to their own bodies. Therefore, we believe it is unique a study focusing on intervention processes to stop a likely event from happening, in our case a collision between external objects and nearby people.

With today's advancements on autonomous systems and drones, we are prompted to study such a problem, that of measuring the probability of an accident about to happen followed by a decision process for a robot to stop the likely event if physically possible. A posit here is that a robot might benefit from three levels of safety: i) collision avoidance, ii) in the case that collision avoidance fails, then collision detection and fast reaction, iii) in the case that a collision between an external object and a person is likely to happen, then intervene to stop the collision if ethically and physically possible. In this chapter we explore case iii). And we have first studied this case from an engineering perspective without entering into ethical, moral, medical, psychological or anthropological questions. These other questions are of course very important, but we wanted first to understand what technologies could be devised and employed to intervene over external events using human-centered robots. For this study we use an upper body humanoid robot called Dreamer, consisting of an articulated torso and two articulated arms. For sensing we use a Kinect structured light sensor mounted outside of the robot. The point cloud sensor is able to simultaneously ``see'' the objects in the scene, the humanoid upper body robot, and people nearby. 

In order to make this study possible, the humanoid upper body robot needs two endow two capabilities: some sort of estimation of the risk of a collision of nearby people, and an intervention strategy that is likely to stop the object. We further give a twist to this study by considering that the robot might be engaged into a task prior to considering intervening. In that case, the robot must face the question if it can still stop the accident from occurring without stopping the task at hand. This question results in a study on constrained motion planning to engineer possible approaches. Further details are broadly discussed next.

First, we use a point cloud sensor and a prediction methods to estimate the risk that an object might collide with a person. Observing the current state of the surrounding objects to predict the future movements, the probability of collision of two objects or an object and a person is computed and the robot is controlled to prevent that collision from happening given a practical threshold risk value. Second, after it is decided that a collision prevention should be attempted, it is necessary to compute intervention paths between the current robot parts and the estimated object path. For this purpose, we use a sampling-based motion planning algorithm and an analysis of the reachable robot part volumes to determine the posture and path of the robot to intervene with the object's trajectory. We complement this computational capabilities with various experiments of the humanoid upper body stopping objects using its elbow, shoulder, torso, and end-effector in scenes where a human is likely to be striked by an object or by another robot. In light of this discussion, the main contribution of this chapter is on devising engineering methods to practically stop collisions between external objects and humans on behalf of reducing the risk of injury, and when possible achieve such capability without stopping the previous task that the intervening robot was performing.

\begin{figure}[t]\centering
\includegraphics[width=\linewidth] {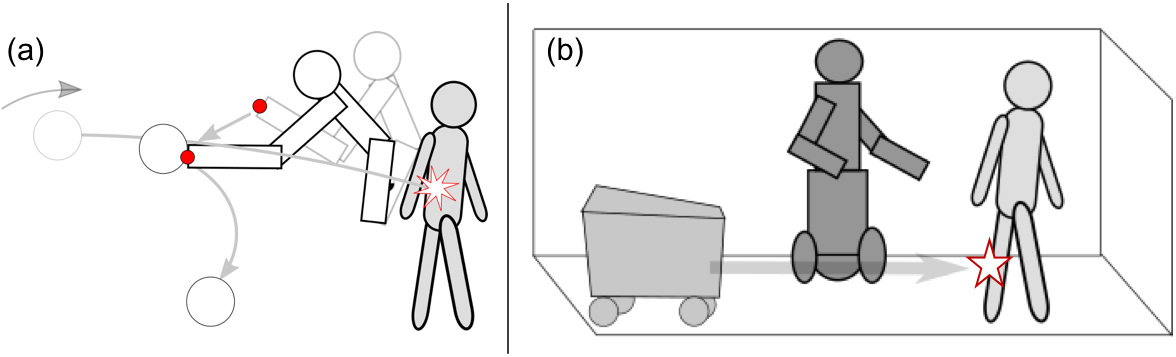}
\caption
[ Two types of intervention scenes] 
{{\bf Two types of intervention scenes.} 
	Scene (a) depicts the action of a robot when a ball is likely to hit a human nearby. The robot acts to stop the ball. Scene (b) depicts a case where a moving cart is likely to hit a human unaware of the danger. A robot nearby could intervene to stop the cart and possibly mitigate injury to the human.} \label{fig:intervention_scenes} 
\end{figure}




\section{Risk analysis}
\label{sec:colman}
We model here a cluttered environment with a robot agent around multiple objects, and their possible collisions with other objects or people. In this environment, there is one robotic agent and $\rm N_o$ objects. The agent observes the movement of objects around its reachable space using exteroceptive sensors, considering those objects as stochastic processes independent of the robot agent. As such, each object is expected to change acceleration and direction based on a stochastic hypothesis, and we assign to it a ``risk level'' which determines its collision probability with respect to other moving objects. As objects move in the environment, their collision probability could increase or decrease. In addition, the relative velocity during a collision between external objects and people can be evaluated via an impact dynamic model. The timing of the robot agent to stop a likely collision can be controlled within the robot's physical and computational limits.

A $\rm N_d\in \left\{2,3\right\}$-dimensional environment is considered and labeled as $\mathcal{S}$, the set of Cartesian coordinates of all objects in the environment. The robot agent in the environment occupies a region, $\mathcal{A} \subset \mathcal{S}$, and an i-th object occupies another region, $\mathcal{O}^i \subset \mathcal{S}$. The region $\mathcal{O}^i$ is a function of time, which we discretize as $t = m\, \Delta t$, yielding time indexed regions $\mathcal{O}_m^i$ where $\Delta t$ is the sampling time of the observer and controller. For our study, we assume that all objects are round (2D) or spherical (3D) bodies. The position and radius of the i-th object are labeled as $p_{o,m}^i$ and $r_o^i$, respectively. Given the previous descriptions, object states and their collision probabilities will be evaluated. Those predictions are based on the current states and observations at time $t_0$. For notation clarity, the time indexes $m$ or $k$ appear on symbols as a subscript, and probability density functions, probabilities, and predicates are denoted as 
$\mathbf{P}$, 
$\mathbf{p}$, and
$\mathbf{q}$, respectively.

%

\subsection{Object dynamics}
\label{sec:obj_dyn}
As mentioned before, the objects are circular rigid bodies, so the configuration of each object can be denoted by its center position. We assume that the linear acceleration of each object is produced by an unknown source which is a zero-mean normal distribution, and the dynamics of the objects need to be reconstructed and exploited for collision prediction. The state of the i-th object at the m-th time index is defined as follows.
\begin{align}
P_m^i \triangleq
\begin{pmatrix}
p_{o,m}^{i} \\ 
\dot{p}_{o,m}^{i} 
\end{pmatrix}
\end{align}
Also, each object has the following dynamic equations which includes unknown disturbances $w^i$ as follows.
\begin{align}
P_m^i&= 
\underbrace{
\begin{pmatrix}
I & \Delta t I \\
0 & I
\end{pmatrix}
}_{A_p}
P_{m-1}^i
+
w^i \label{eq:st}\\
w^i &=
\begin{pmatrix}
{\xi^d}  \\ \alpha \Delta t
\end{pmatrix} \in \mathbb{R}^{2\rm N_d} \\
&\sim \mathcal{N} \bigg( 
\begin{pmatrix}
0 \\ 0
\end{pmatrix},
\underbrace{
\begin{pmatrix}
\Sigma_d^i & 0 \\ 0 & \Sigma_a^i \Delta t
\end{pmatrix}
}_{\Sigma_w^i}
\bigg)
\end{align}
where $\xi^d$ is a velocity disturbance and $\alpha$ is an unknown acceleration input. Both of these disturbances are assumed to be random variables with zero-mean normal distributions in the bases of the Cartesian coordinate, and their covariances are $\Sigma_d^i$ and $\Sigma_a^i$, respectively. We also assume that the position sensor value of the i-th object at time index $m$, $y_m^i$, has a normal distribution noise, $\xi_s^i$ with covariance $\Sigma_s^i$ as follows.
\begin{align}
y_m^i &= 
\underbrace{\begin{bmatrix}
I & 0 \end{bmatrix}}_{C_p}
P_m^i + \xi_s^i \\
\xi_s^i &\sim \mathcal{N} \left( 0, \Sigma_s^i \right)
\end{align}
Then, the state of the i-th object and its covariance matrix, $\Phi_m^i$ can be estimated via Kalman filter as follows.
\begin{align}
\hat{P}_m^i &= \hat{P}_{m | m-1}^i + K_m^i \left( y_m^i - C_p \hat{P}_{m | m-1} \right) \\
\Phi_m^i &= \left( I - K_m^i C_p \right) \Phi_{m|m-1}^i
\end{align}
where
\begin{align}
\hat{P}_{m | m-1}^i &= A_p \hat{P}_{m-1} \\
\Phi_{m|m-1}^i &= A_p \Phi_{m-1} A_p^T + \Sigma_w^i \\
S_m^i &= C_p \Phi_{m|m-1}^i C_p^T + \Sigma_s^i \\
K_m^i &= \Phi_{m|m-1}^i C_p^T {S_m^i}^{-1} 
\end{align}

Also, we define a new random variable to estimate the future states of the objects, $X_k^i$, and its initial value can be defined from the result of the Kalman filter as follows. 
\begin{align}
X_0^i & \sim \mathcal{N} \bigg( 
\hat{P}_m^i
,~ 
\Phi_m^i
\bigg) 
\end{align}
Given the initial value and the state transition equation, Eq. (\ref{eq:st}), the random variable $X_k^i$ which is the position of the i-th object at time index k can be expressed using the following normal distribution.
\begin{align}
\begin{split}
X_k^i &= \begin{pmatrix} x_k^i \\ \dot{x}_k^i \end{pmatrix}  \\
&\sim
\mathcal{N} \Bigg( 
\underbrace{
\begin{pmatrix}
\overline{x}_{k}^i \\
\dot{\overline{x}}_{k}^i
\end{pmatrix}
}_{\overline{X}_k^i}
,~
\underbrace{
\begin{pmatrix}
\Sigma_{x,k} & \Sigma_{xv,k}^i \\ \Sigma_{vx,k}^i &  \Sigma_{v,k}^i 
\end{pmatrix}
}_{
\Sigma_{X,k}^i
}
\Bigg) 
\\
&=
\mathcal{N} \left( 
A_p \overline{X}_{k-1}^i
,~
A_p 
\Sigma_{X,k-1}^i
A_p^{T}
+
\Sigma_w^i 
\right) 
\label{eq:X}
\end{split}
\end{align}
where $\overline{X}_k^i$ is the expected object state. In the state prediction, the time index $k$ of the current time is set to be zero. In Sec. \ref{sec:simple}, the random variables are approximated to normal distributions for computational efficiency.

\begin{figure}\centering
\includegraphics[width=0.8\linewidth, clip=false ] {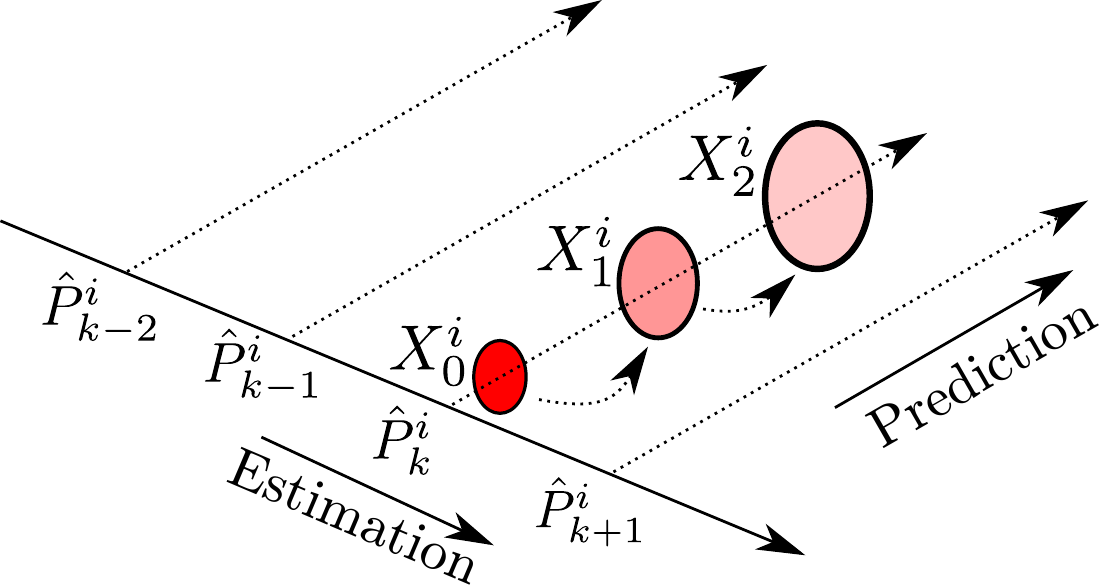}
\caption
[ State estimation and prediction]
{{\bf State estimation and prediction} . Whenever a new object observation comes in from the external sensors, the state is estimated, and its future state is predicted.}\label{fig:prediction}
\end{figure}

We also define a conditional random variable, $X_f^{ij} \triangleq {\begin{pmatrix} x_f^{ij\,T} & \dot{x}_f^{ij\, T} \end{pmatrix}}^T$, which represents the position of the i-th object under the condition that it has not collided with the j-th object. The probability distribution of the random variable is the complement of the collision probability between i-th and j-th objects, so $X_f$ is not a normal distribution. $X_f$ also propagates in time similarly to $X$ in Eq. (\ref{eq:X}) such that we can define another random variable, $\hat{X}_f$ which is a one-step propagated random variable given $X_f$. Even though $X_f$ and $\hat{X}_f$ are not normal distributions, the probabilistic properties of $\hat{X_f}$ can be derived from those of $X_f$ as follows.
\begin{align}
\begin{split}
{\rm E} \left( \hat{X}_{f,k} \right) &= A_p {\rm E} \left( X_{f,k-1} \right) \\
{\rm V} \left( \hat{X}_{f,k} \right) &= A_p {\rm V} \left( X_{f,k-1} \right) A_p^T + \Sigma_w \label{eq:Xhat}
\end{split}
\end{align}
%
The estimation and prediction of object states are illustrated in Fig. \ref{fig:prediction}.

\subsection{Instantaneous collision probability and its complement} 
\label{sec:ic}
Based on the object dynamics of Eq. (\ref{eq:st}), we can predict the future locations of the objects and also determine whether a collision will take place. 
By observing the current states of objects, we can anticipate their future states as random variables, and estimate the possibility of their collisions as stochastic processes as shown in Eq. (\ref{eq:X}).

For convenience, we define a probability function, $\mathbf{P}_o^i: \mathbb{R}^{\rm N_d} \rightarrow \mathbb{R} $ being the probability that the center of the i-th object is located at a given point, $p^i$, as follows.
\begin{align}
\mathbf{P}_o^i\left(p_o\right) \triangleq \mathbf{P}( x^i = p^i )
\end{align}
where $x^i$ is the center position of the i-th object.
Also, because we assume that all the objects are circles or spheres in $\mathbb{R}^{\rm N_d}$, a predicate whether two objects collide with each other is defined as follows.
\begin{align}
\mathbf{q}_{ic}(i,j) &\triangleq \left\| x^i - x^j \right\| \le r^i + r^j  
\end{align}

Before computing the cumulative collision probability over time, we need to consider the probability of collision at a given time instance. The probability that the i-th object located at $p^i$ collides with the j-th object can be derived from the probability density function of the i-th object, $\mathbf{P}_o^i$ and that of the j-th object, $\mathbf{P}_o^j$, and it can be denoted as $P_{ic}$ and expressed as follows. 
\begin{align}
\begin{split}
&{P}_{ic} ^{ij}\left( p^i, \mathbf{P}_o^i, \mathbf{P}_o^j \right) \\
\triangleq& 
\mathbf{P}_o^i\left(p^i\right)\, 
\mathbf{P}\left( \mathbf{q}_{ic} (i,j) \Big| x^i = p^i, \, x^j \sim \mathbf{P}_o^j \right) \\
=& 
\mathbf{P}_o^i\left(p^i\right)\, 
\int_{\mathcal{S}} f_c^{ij}(p^i, p^j) 
	\mathbf{P}_o^j\left(p^j\right) dp^j \label{eq:pic} 
\end{split}
\end{align}
where 
$f_c^{ij}$ is a function that checks collision events between the i-th and j-th objects defined as follows.
\begin{align}
&f_c^{ij}(p^i, p^j) \triangleq \left\{
\begin{matrix}
1 & \left\| p^i - p^i \right\| \le r^i + r^j \\
0 & \rm otherwise
\end{matrix}
\right.
\end{align}
The integration of the probability density function around $\mathcal{S}$ corresponds to the probability that the two objects collide with each other. This collision probability is based on the probability density functions at a given time instance, so we call it the {\it instantaneous collision probability}, $\mathbf{p}_{ic}^{ij}$. 
\begin{align}
\mathbf{p}_{ic}^{ij} \triangleq \int_{\mathcal{S}} {P}_{ic} ^{ij}\left( p^i, \mathbf{P}_o^i, \mathbf{P}_o^j \right) dp^i \label{eq:pic2}
\end{align}
	
We have derived the probability that given two objects collide with each other. Therefore, we can also derive the probability that they do not collide with each other, and the conditional probability density function of the collision-free object position. The collision probability density function in Eq. (\ref{eq:pic}) is the probability that the i-th object located at $p^i$ collides with the j-th object. The corresponding probabilistic density function describing that the i-th object at $p^i$ is free from colliding with the j-th object is denoted as $P_{of}^{ij}$ and expressed as follows.
\begin{align}
\mathbf{P}_{of}^{ij} \left( p^i, \mathbf{P}_o^i,\, \mathbf{P}_o^j \right)
\triangleq
\frac{
\mathbf{P}_o^i \left( p^i \right) - {P}_{ic}^{ij} \left( p^i,\, \mathbf{P}_o^i,\, \mathbf{P}_o^j \right)\label{eq:pof}
}
{ 1 - \mathbf{p}_{ic}^{ij} }
\end{align}
where $\mathbf{P}_o$ is the probability density function of the collision-free object at the given time. Therefore, Eq. (\ref{eq:pof}) can be indexed at time $k$ as follows.
\begin{align}
\mathbf{P}_{of,k}^{ij} \left( p^i \right)
\triangleq
\frac{
\hat{\mathbf{P}}_{of,k}^{ij} \left( p^i \right) - {P}_{ic}^{ij} \left( p^i,\, \hat{\mathbf{P}}_{of,k}^{ij},\, \hat{\mathbf{P}}_{of,k}^{j} \right)\label{eq:pofk}
}
{ 1 - \mathbf{p}_{ic}^{ij} }
\end{align}
where $\hat{\mathbf{P}}_{of,k}$ is the collision-free probability density function at time $k$ which is predicted from that at time $k-1$. From the definition of $X_f$ and $\hat{X}_f$ in Sec. \ref{sec:ic}, $\mathbf{P}_{of}$ and $\hat{\mathbf{P}}_{of}$ correspond to their probabilistic density functions.  
\begin{align}
\begin{split}
X_f &\sim \mathbf{P}_{of} \\
\hat{X}_f &\sim \hat{\mathbf{P}}_{of}
\end{split}
\end{align}
If $\mathbf{P}_{of,k-1}$ is a normal distribution, $\hat{\mathbf{P}}_{of,k}$ can be estimated from Eq. (\ref{eq:X}). Though $\mathbf{P}_{of}$ is not a normal distribution, we approximated to be a normal distribution in Sec. \ref{sec:simple}.
\begin{figure}[t]\centering
\includegraphics[width=0.9\linewidth, clip=false ] {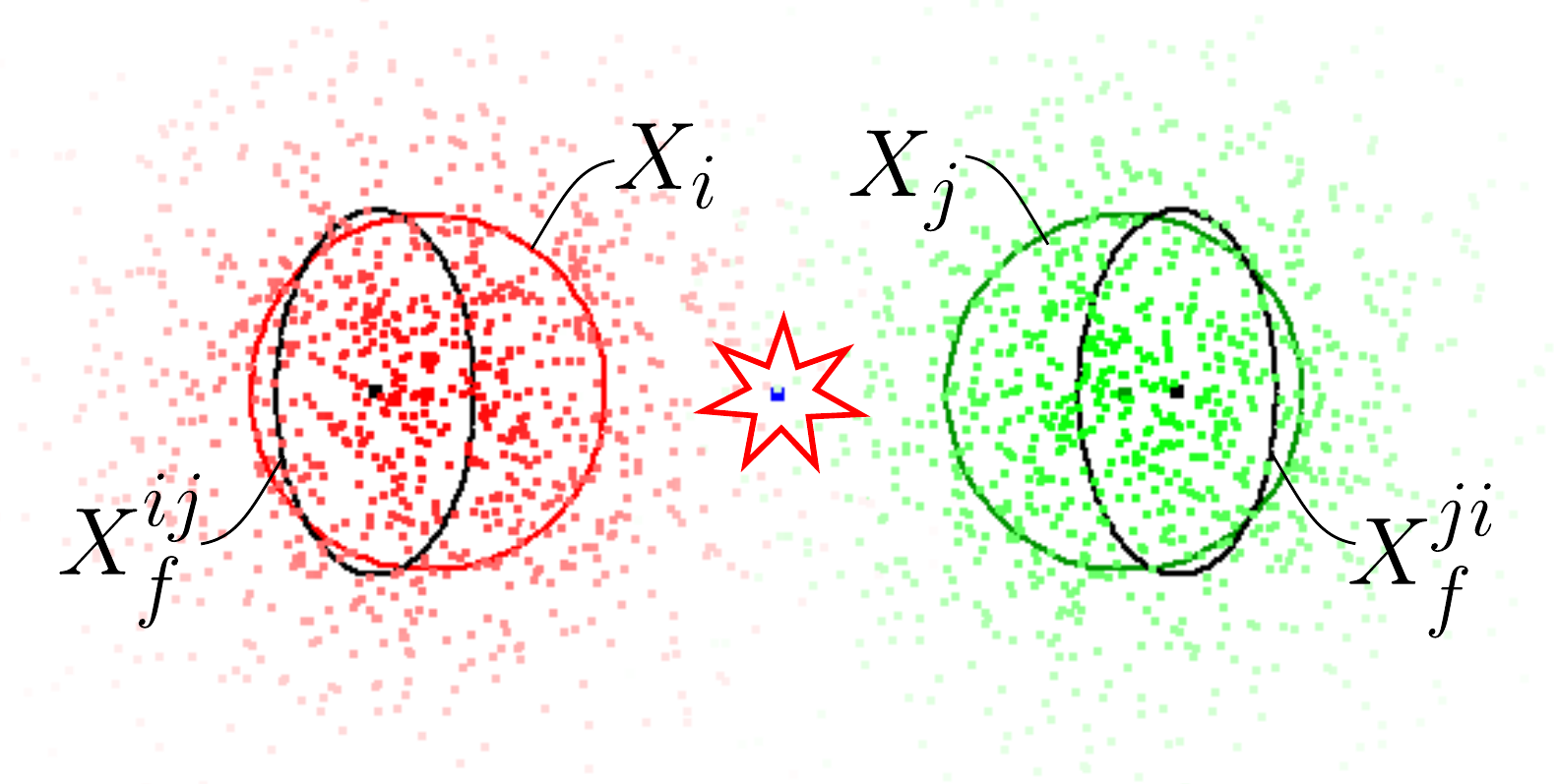}
\caption
[ Probability density function of the object positions random variables, $X_i$ and $X_j$ and those of the collision-free random variables, $X_f^{ij}$ and $X_f^{ji}$]
{{\bf Probability density function of the object positions random variables, $X_i$ and $X_j$ and those of the collision-free random variables, $X_f^{ij}$ and $X_f^{ji}$.}
	The collision-free random variables are the complement of the collision probability.
} \label{fig:pdf} 
\end{figure}
\begin{figure}[t]\centering
\ifdefined\JOURNAL
\includegraphics[width=\linewidth, clip=false ] {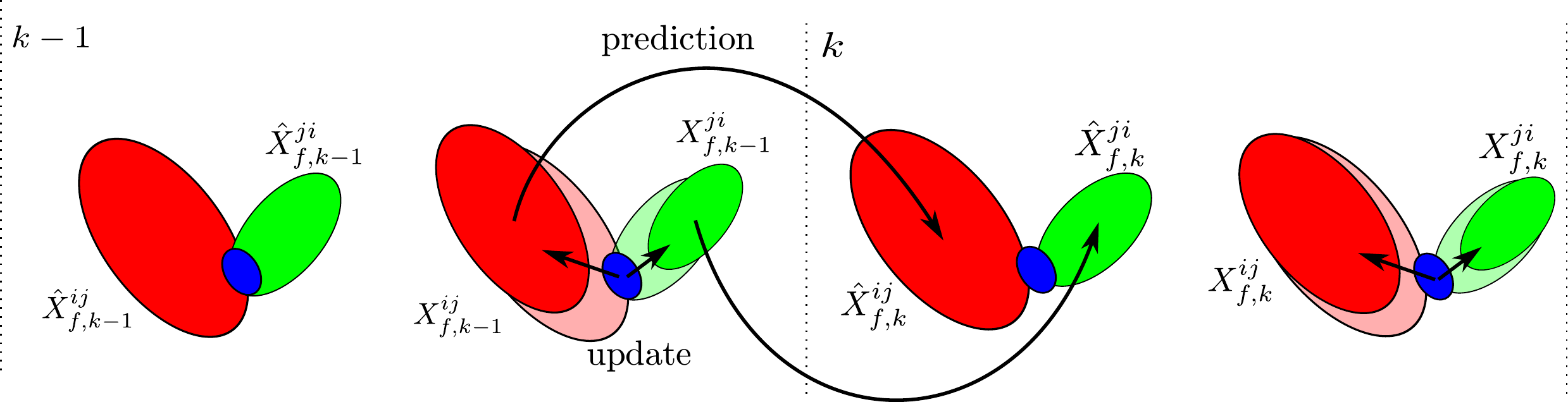}
\else
\includegraphics[width=1.0\linewidth, clip=false ] {x_free.pdf}
\fi
\caption
[ Prediction and update of the collision-free random variable, $X_f$]
{{\bf Prediction and update of the collision-free random variable, $X_f$}
} \label{fig:x_free} 
\end{figure}



\subsection{Cumulative collision probability}
The collision probability of Eq. (\ref{eq:pic2}) estimates only the instantaneous probability at a given time, so we cannot use that probability to determine how likely a collision happens at that time because a collision might already have happened and the objects involved in it keep colliding. Thus, we need to recursively accumulate the collision probability and exploit the conditional probability that collisions do not happen before the considered time. The predicate that describes this situation can be defined as follows.
\begin{align}
\mathbf{q}_{ac,k}(i,j) &\triangleq \bigvee_{n \le k} 
\left\| x_{n}^i - x_{n}^j \right\| \le r^i + r^j 
\end{align}
The corresponding {\it accumulated collision probability} is expressed and computed as follows.
\begin{align}
\mathbf{p}_{ac,k}^{ij} 
=& ~\mathbf{p} \left( \mathbf{q}_{ac,k} (i,j) \right) \nonumber \\
=&		~\mathbf{p}_{ac,k-1}^{ij} + 
\label{eq:pac}
\\
&	~	\big( 1 - \mathbf{P}_{ac,k-1} ^{ij}\big) 
\mathbf{p}\left(  
		\mathbf{q}_{ic,k}\left(i,j\right) \bigg| \neg \mathbf{q}_{ac,k-1}\left( i,j\right) \right) 
\nonumber
\\
\mathbf{P}_{ac,0}^{ij} =& \mathbf{P}\left(\mathbf{q}_{ic,0}\left(i,j\right)\right) = \mathbf{P}_{ic,0}\left(i,j\right) \label{eq:pac0}
\end{align}
The conditional probability on the last term of Eq. (\ref{eq:pac}) is equivalent to the integral of the collision probability of Eq. (\ref{eq:pic}), yielding the following.
\begin{align}
\mathbf{P}_{ac,k}^{ij} 
=&		~\mathbf{P}_{ac,k-1}^{ij} + \nonumber\\
&	~	\big( 1 - \mathbf{P}_{ac,k-1} ^{ij}\big) 
\int_{\mathcal{S}}
{P}_{ic}^{ij} \left( p^i, \mathbf{P}_{of}^{ij}, \mathbf{P}_{of}^{ji} \right) dp^i
\label{eq:pac2}
\end{align}

The cumulative collision probability, $\mathbf{P}_{ac}$, increases over time and can be computed recursively starting at time zero. However, the derivation of the cumulative collision probability, $P_{ic}$, based on non-normal distributions is computationally expensive, prompting us to use an approximation.
In the next section, the probability will be simplified for real-time computation.

Given the cumulative collision probability and a probability threshold, $\eta$, we can derive the minimum time index, $k_c$, at which the probability of a collision between any two objects from taking place exceeds the threshold as follows.
\begin{align}
&k_c = \min k \nonumber \\
&\text{such that} \nonumber\\
&\bigvee_{i,\,j \le \rm N_o} 
\bigg( \left( i \neq j \right) \wedge \Big(\mathbf{P}_{ac,k}\left(i,j\right)  \ge \eta \Big) \bigg) \label{eq:tc2}
\end{align}
$k_c$ embodies the likelihood that at least one object pair in the environment will collide with each other at time $k_c\cdot \Delta t$ with a probability $\eta$.




\subsection{Implementation considerations}
\label{sec:simple}
To predict an imminent collision time index, $k_c$, we would ideally need to compute the instantaneous collision probability of objects using non-Gaussian distributions according to Eq. (\ref{eq:pic2}). This computation would take a large amount of time to finally predict the collision-free object distributions of Eq. (\ref{eq:pof}). Sampling-based probability derivations have been proposed for computational efficiency in \cite{lambert2008}, however the computations using that approach remains expensive for real-time usage. Therefore we compromise on the accuracy of the estimation using other simplifications.

First, we simplify the instantaneous collision probability of Eq. (\ref{eq:pic2}), which consists of an integration over the considered environment and the shapes of the objects on it. By assuming the distributions of the objects to be normal, the computation becomes greatly efficient. The integration operation then becomes a single normal distribution, which allows the use of pervasive efficient numerical solutions for the integration of multivariate normal distributions.
The detailed approximation and computation of $\mathbf{p}_{ic}$ are explained in Appendix \ref{sec:a_pc}. If the object distributions are assumed to be normal, the collision probability of the two objects can be simplified as follows.
\begin{align}
\mathbf{p}_{ic}^{ij} &= \int_{\mathcal{B}} \mathbf{P}_{conv} \left( x \right) dx
\end{align}
where $\int_{\mathcal{B}} \left(\cdot\right)dp$ denotes the integration along the Minkowski sum of the i-th object and j-th object, and $\mathbf{P}_{conv}$ is a normal distribution with the following probability.
\begin{align}
\mathbf{P}_{conv} &\sim \mathcal{N} \left( 
		\mu_i - \mu_j,\ 
		\Sigma_i + \Sigma_j
\right)
\end{align}
The approximate integration of multivariate normal distributions around elliptic shapes is studied in \cite{sheil1977}.
Therefore, the instantaneous collision probability can be computed effectively, such that the conditional probability on the last term of Eq. (\ref{eq:pac2}) can be simplified. 

Second, we revisit the conditional probability density function $\mathbf{P}_{of}$ of Eq. (\ref{eq:pof}), and approximate it as a normal distribution. Initially, it has a normal distribution because it comes from the Kalman filter of Eq. (\ref{eq:X}).
\begin{align}
\mathbf{P}_{of,0}^{ij} = \mathcal{N} \left( \overline{x}^i, \Sigma_{x,0}^i\right)
\end{align}
Because $\mathbf{P}_{of,0}^{ij}$ is a normal distribution, the probability at the next iteration, $\hat{\mathbf{P}}_{of,1}^{ij}$ can be estimated as follows.
\begin{align}
\hat{X}_{f,1} \sim \mathcal{N} \left( A_p X_0, A_p \Sigma_{X,0}^{i} A_p^T + \Sigma_w^i \right) \label{eq:hatxf}
\end{align}
If we assume that all the subsequent probabilities are also normal distributions as shown in Appendix \ref{sec:xf}, any $\hat{X}_{f,k}$ can be predicted from its previous estimate as follows.
\begin{align}
\hat{X}_{f,k}^{ij} \sim \mathcal{N} \left( A_p X_{k-1}^{ij}, A_p \Sigma_{X_f,k-1}^{ij} A_p^T + \Sigma_w^i \right) \label{eq:hatxf}
\end{align}
\begin{figure}[t]\centering
\ifdefined\JOURNAL
\includegraphics[width=\linewidth, clip=false ] {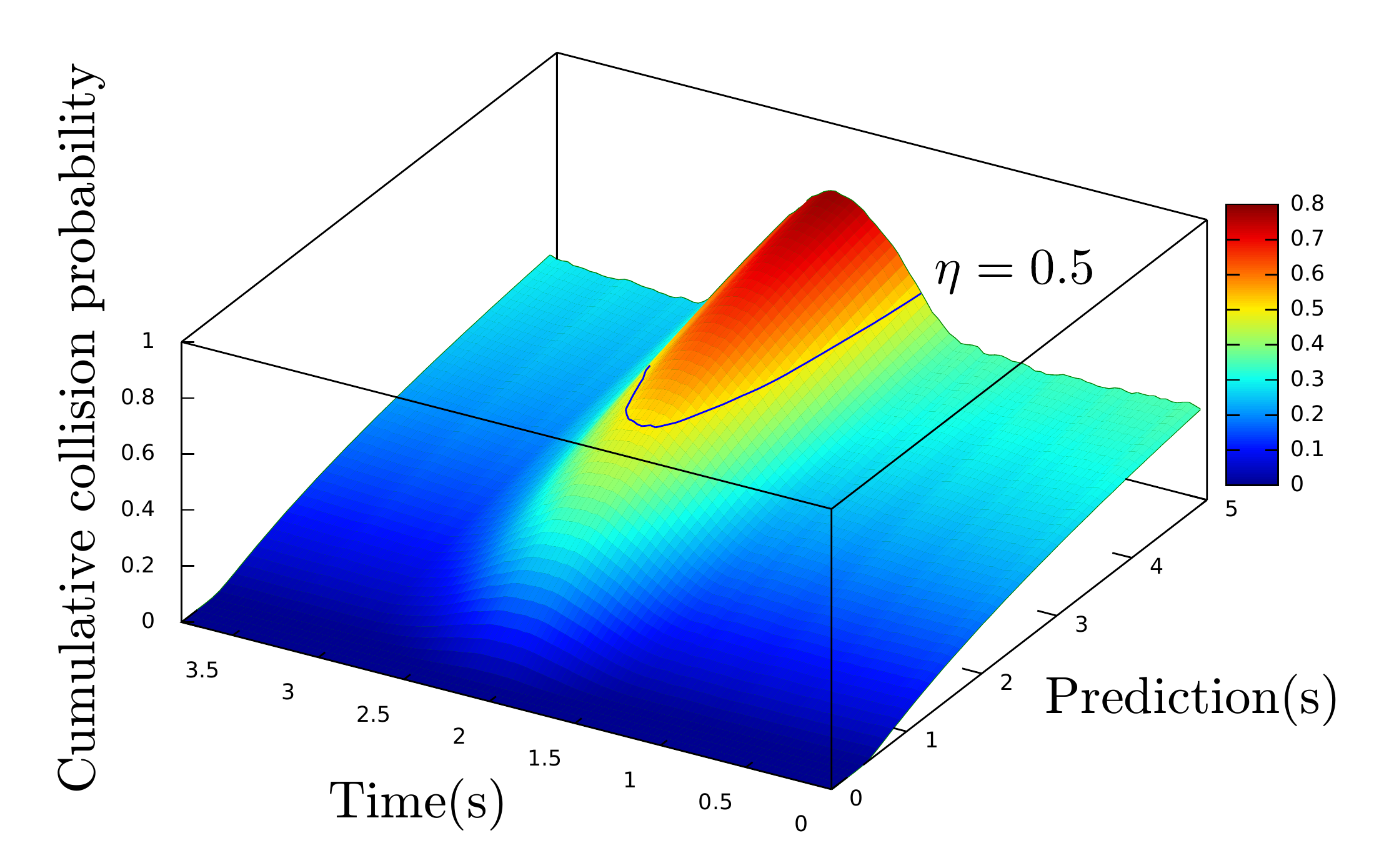}
\else
\includegraphics[width=0.7\linewidth, clip=false ] {cum_col.pdf}
\fi
\caption
[ Cumulative collision probabilities over time] 
{{\bf Cumulative collision probabilities over time} 
	shows the estimated cumulative collision probabilities when an object is getting closer to another one (0-2sec) and receding from it (2-4sec). Whenever a new observation is produced by an external sensor, the agent computes the cumulative probabilities.
} \label{fig:cum_col} 
\end{figure}
\begin{figure}[t]\centering
\ifdefined\JOURNAL
\includegraphics[width=0.8\linewidth, clip=false ] {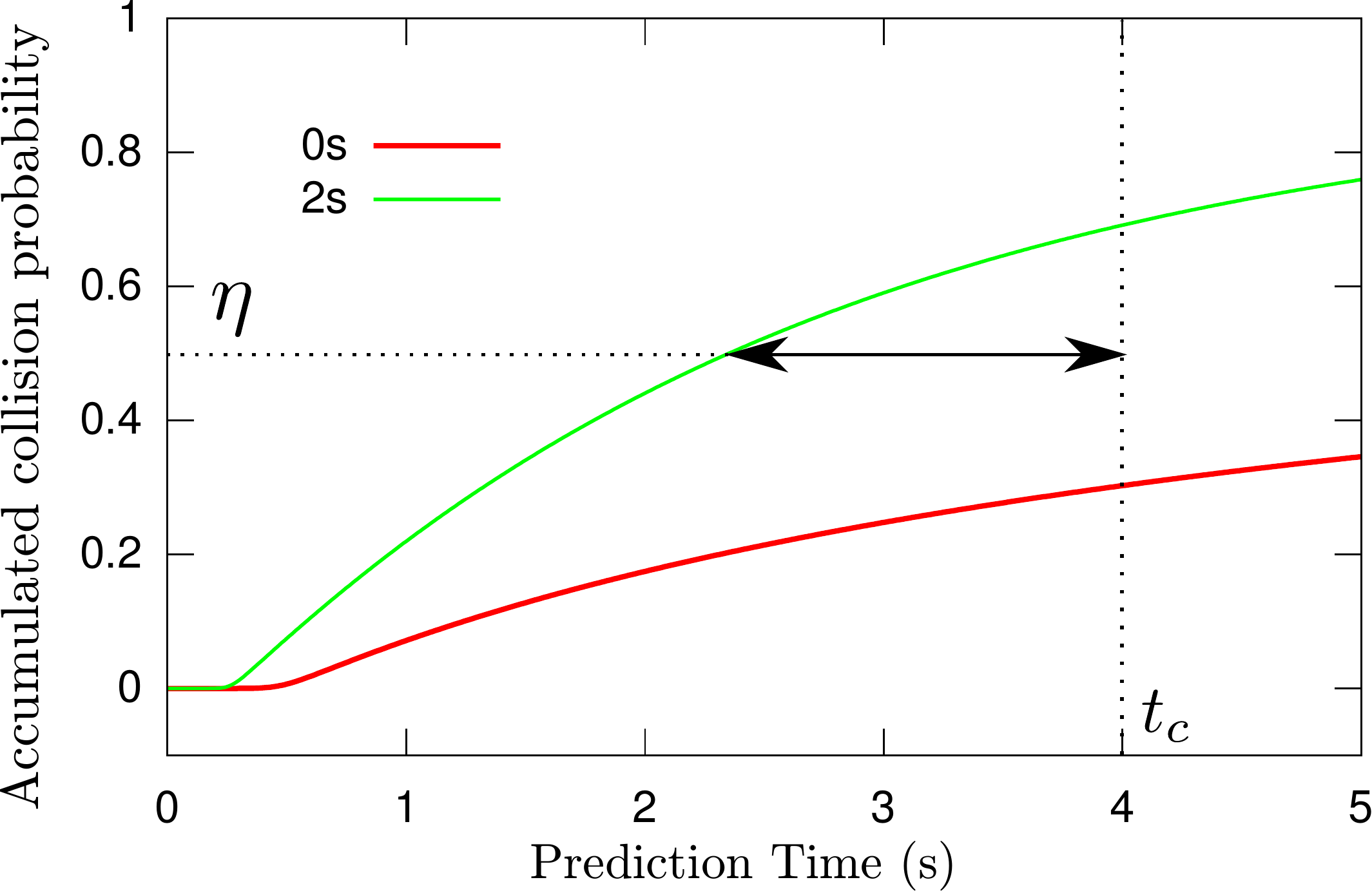}
\else
\includegraphics[width=0.7\linewidth, clip=false ] {acc_col.pdf}
\fi
\caption
[ Cumulative collision probabilities]
{{\bf Cumulative collision probabilities}
	are extracted from Fig. (\ref{fig:cum_col}) at 0 and 2 second. The sample at 0 seconds shows the estimated collision time when the collision probability exceeds the collision threshold $\eta$. If the time is earlier than the threshold, $t_c$ , then the agent intervenes by attempting to stop the collision. In the case of the sample at 0 seconds, the cumulative probability does not exceed the threshold over the prediction horizon. As such, the agent will not intervene to stop the collision until somewhere around the 2 seconds instant.
} \label{fig:acc_col} 
\end{figure}
\section{Task planning}
\label{sec:agent}
Given the environment and the prediction process described in Section~\ref{sec:colman}, we want our robot agent to keep conducting its primary manipulation task as long as possible but at the same time intervene to stop collisions between two external objects when needed. The objective of the agent is 1) to prevent collisions between external objects by moving its body to intercept likely colliding objects, and 
2) to keep performing its operational task such as using its end-effector around a desired process region. Can we fulfill 1) and 2) simultaneously? Under what circumstances do we need to stop 2) to give priority to 1)? To prevent collisions from taking place, the robot agent continuously and quickly senses and anticipates objects' future trajectories as previously discussed. The anticipated collision probabilities will eventually converge to 1, given a sufficiently large prediction horizon, so it is always expected that a collision will happen. However, depending on the states of the objects, the prediction time of the collision event largely varies. If the prediction time is far away from the current instant of time, the collision is ignored, but if not, the agent decides to intervene intercepting the colliding objects using a convenient part of its body. 

\begin{figure}\centering
\ifdefined\JOURNAL
\includegraphics[width=0.8\linewidth, clip=false ] {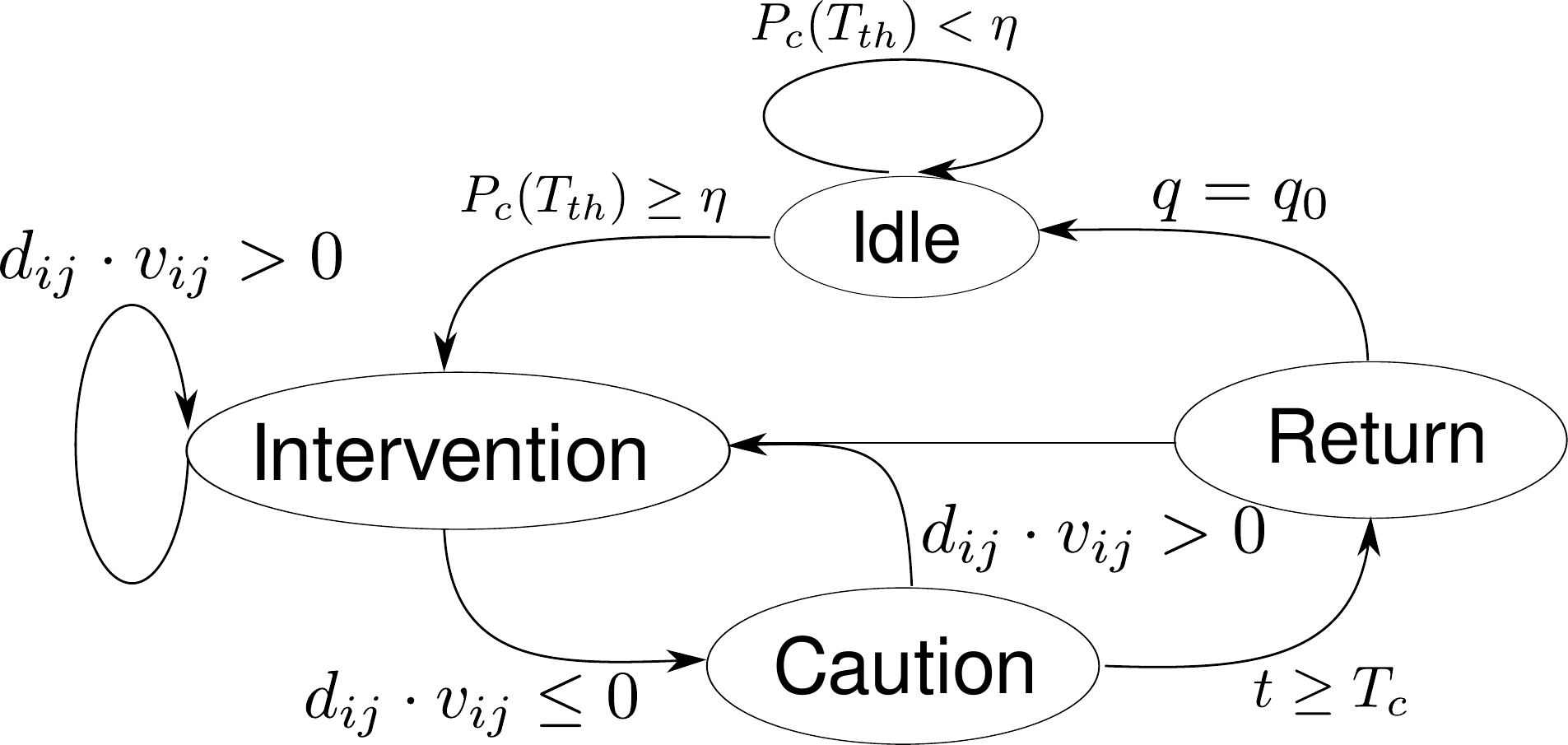}
\else
\includegraphics[width=0.7\linewidth, clip=false ] {col_fsm.pdf}
\fi
\caption
[ State transition diagram]
{{\bf State transition diagram}
} \label{fig:col_fsm} 
\end{figure}

To deal with the anticipated collisions, the robot agent uses four operating modes: idle, intervention, caution, and return. In the idle mode, the agent keeps predicting collisions and continuously planning a motion to intervene the most likely collision, but does not start a collision intervention behavior. In the intervention mode, the agent uses the most convenient body part to intercept the likely colliding objects with the hope to deviate one of them from the set trajectory. In the caution mode, the agent expects that anticipated collisions are unlikely but have enough risk to approach threatening objects with its body parts without inflicting contact.
Finally, in the return mode, the agent estimates that there are no collision threats and returns to its normal operation. The modes and their transitions are described in the state diagram of Fig. \ref{fig:col_fsm}


\subsection{Idle mode}
The estimated cumulative probability of a collision at every instant over a prediction horizon asymptotically increases as suggested by Eq. (\ref{eq:tc2}). However, the time when the probability exceeds a given threshold depends on the initial state and the statistics of the random variables. If a collision will happen in the distant future, the agent does not need to be concerned. Otherwise, the agent, according to premises, will intervene and attempt to stop the collision. Therefore, a probability threshold and a time threshold to differentiate the distant and the near futures need to be set to reason about interventions. In Fig. \ref{fig:acc_col}, the probability threshold, $\eta$ and the time threshold, $T_{th}$ are characterized. If the collision probability of any two object at $T_{th}$ exceeds the probability threshold, $\eta$, then the agent intervenes by switching its mode from idle mode to intervention mode.
Therefore, if any pair of objects are expected to collide with each other within a given time period, the mode of the agent changes as follows.
\begin{align}
\Bigg( \bigvee_{i,\,j \le \rm N_o,\ i \neq j} 
\Big(P_{ac,T_{th}/\Delta t}^{ij} \ge \eta \Big) \bigg) \label{eq:trans1}
\nonumber \\ 
\rightarrow
\left( 
mode \gets \rm INTERVENTION	
\right)
\end{align}
The pair of objects for which the collision probability exceeds the threshold at $T_{th}$ 
form a new set, $\mathcal{P}_c$, corresponding to the set of imminent collisions as follows. 
\begin{align}
\mathcal{P}_c = \left\{  \left(i,j\right) \bigg| P_{ac,T_{th}/\Delta t}^{ij} \ge \eta \right\} \label{eq:pc_set}
\end{align}

\subsection{Intervention mode}
\begin{figure}\centering
\ifdefined\JOURNAL
\includegraphics[width=0.6\linewidth, clip=false ] {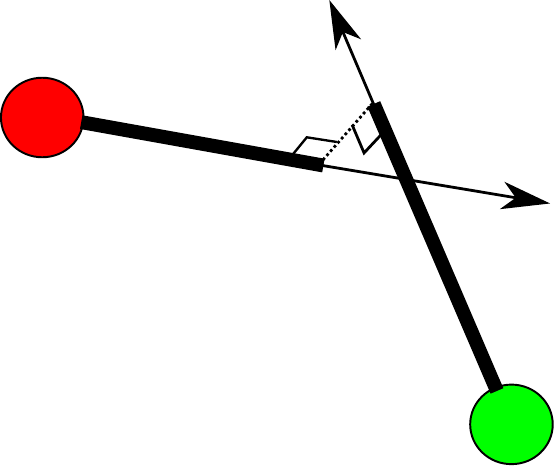}
\else
\includegraphics[width=0.4\linewidth, clip=false ] {col_pair.pdf}
\fi
\caption
[ A pair of objects are getting close to each other]
{{\bf A pair of objects are getting close to each other.}
	Their moving directions are represented as rays from their current position (lines with arrows), and line segments for the current positions to the closest points can also be determined (bold lines).
} \label{fig:col_pair} 
\end{figure}

In the intervention mode, the agent attempts to intercept the trajectory of the most imminent collision event. The most imminent collision is chosen from the set $\mathcal{P}_c$ of Eq. (\ref{eq:pc_set}) by considering the positions and velocities of the objects. Each anticipated collision consists of two objects, and the time when these two objects will make a contact or come the closest to each other can be derived from their initial positions and velocities. Let the position and velocity of the i-th object be $p_i$ and $v_i$, respectively, then the time when the distance between them is minimum can be derived as follows.
\begin{align}
\frac{d}{dt} \Big\| \left(p_i + v_i\, t\right) - \left(p_j + v_j\, t \right) \Big\| = 0 \\
\left( p_i - p_j \right) \cdot \left( v_i - v_j \right) < 0
\end{align}
The time at which the i-th and j-th object approach the closest or collide is defined as $t_c^{ij}$ and derived analytically as follows.
\begin{align}
t_c^{ij} =  
- \frac{\left( p_i - p_j \right) \cdot \left( p_i - p_j \right)} 
{\left( p_i - p_j \right) \cdot \left( v_i - v_j \right)} 
\label{eq:xxx}
\end{align}
Then the pair of objects that will be involved in the most imminent collision can be derived as follows.
\begin{align}
( i_c, j_c ) = \underset{
	\left( i,j\right)}
	{\rm argmin} \ 
	t_c^{ij}, \left(i,j\right) \in \mathcal{P}_c 
\end{align}
Once the pair of the most imminent collision is identified, the positions and velocities of the objects are used to determine the robot agent motion to intercept them. From that pair of positions and the velocities, the time when they come the closest to each other has been derived in Eq. (\ref{eq:xxx}), and the trajectories of the objects from the current positions to the closest positions can be expressed as line segments as shown in Fig.~\ref{fig:col_pair}. 
Subsequently, a motion planner described in the next section generates a plan to intercept the line segments, and the planned trajectory is executed.

Since the most imminent collision is identified with our method and the robot agent mode switches to the intervention mode, the agent keeps updating the expected trajectories of the objects and 
determines whether to continue being in the intervention mode. The decision criteria to stay in that mode considers collision probability and the inner product of the relative position and velocity of the two objects. If the probability is still higher than the threshold or the inner product is negative \-- meaning that the distance between the objects is decreasing \-- then the agent remains in the intervention mode. Otherwise, the mode switches to caution.
\begin{align}
\Bigg( \bigwedge_{i,\,j \le \rm N_o, i \neq j } 
\Big(P_{ac,T_{th}/\Delta t}^{ij} < \eta \Big)  
\wedge \Big( d_{ij} \cdot v_{ij} \le 0 \Big)
\Bigg) \nonumber \\ 
\rightarrow
\left( 
mode \gets \rm CAUTION	
\right)
\label{eq:trans2}
\end{align}

Once the mode switches to intervention and the most imminent collision is determined, the object pair expected to collide with by the agent will not change until the collision becomes unlikely due to the robot agent intercepting the objects. The agent will finish intervening that collision even though a more imminent collision is newly identified.

\subsection{Caution mode}
If the collision probability between two objects is lower than the given threshold and the two objects are estimated to move away from each other, we assume that the two external objects are unlikely to collide with each other. If this happens during the idle mode, the agent can keep ignoring the collision. However, if it happens during the intervention mode, the robot agent needs to change its policy because the intervention process for which it is involved with will suddenly be labeled as less likely to happen.

Instead of forgetting the intervention task it started, the agent keeps being concerned about it by switching to the caution mode. In the caution mode, the agent neither intervenes nor returns to the primary goal task, but instead stays in the caution mode for a while. During the caution mode, if all of a sudden a collision trajectory increases its likelihood to be of concern, the agent switches to the intervention mode.
\subsection{Return mode}
When the likelihood of an imminent collision is low before the threshold time, $T_{ca}$, based on Eq. (\ref{eq:trans2}) the agent doesn't anticipate a collision to be likely, and the agent goes back to its normal operation. To achieve the default posture used in normal operation, the agent switches its mode to return. In the return mode, it generates a new motion plan to return to the original pose. 
The transition condition for the return mode is shown below.
\begin{align}
\left( 
mode = \rm CAUTION	
\right) \wedge
\Bigg( \bigwedge_{i,\,j \le \rm N_o, i \neq j } 
\Big(P_{ac,T_{th}/\Delta t}^{ij} < \eta \Big)  
\Bigg) \nonumber \\ 
\wedge ( t > T_C) 
\rightarrow
\left( 
mode \gets \rm RETURN	
\right)
\label{eq:trans3}
\end{align}

\section{Motion planning}
The agent intervenes over anticipated collisions by launching its body parts towards the predicted collision position and intercepting the object pair as shown in Sec. \ref{sec:agent}.
To be general, there are many types of robot types, but in this study we consider a manipulator with an articulated body with $N_j$ joints and $N_l$ links, being redundant for performing end-effector tasks. The robot will attempt to use its most convenient body part to intervene and if possible exploit its torso and arm redundancy to stop the collision without stopping the process task of the end effector. Any of the robot's body parts other than the end-effector will be explored for collision intervention unless there is no other choice than using the end-effector itself. To realize such a behavior, we exploit the concept of reachable volumes where we consider the reachable space of all body parts, and search in the configuration space for the posture in which both the primary end-effector task and the intervention task are satisfied. Then, a motion planner generates a configuration trajectory from the current configuration to the desired one. We assume that the manipulator conducts a primary goal task with its end-effector, so the reachable volumes and the motion planner are constrained to satisfy the primary end-effector task. Even though the agent has redundant degrees of freedom to explore the null space of the end-effector task space, reaching the collision target may not be possible. In that case, the agent will be prompted to give up on the end-effector task and intervene the collision without task constraints. As such, we will both consider constrained and unconstrained reachable volumes for the motion plans.

Because the intervention motion planning process needs to be very quick, both planning and execution should be achievable at the desired intervention time. As such, the motion planning process is divided into an off-line and an on-line process similar to two-phase motion planning discussed in \cite{kavraki1996}. During the off-line process, the reachable volumes of every joint and linkage of the robot are generated, and the configurations that correspond to the volumes are interconnected as tree structures according to the probabilistic roadmap (PRM) method \-- see \cite{kavraki1996}. In the roadmap generation, dynamic properties of the configurations are also generated.

\subsection{Computing reachable volumes}
\begin{figure}\centering
\ifdefined\JOURNAL
\includegraphics[width=0.8\linewidth, clip=false ] {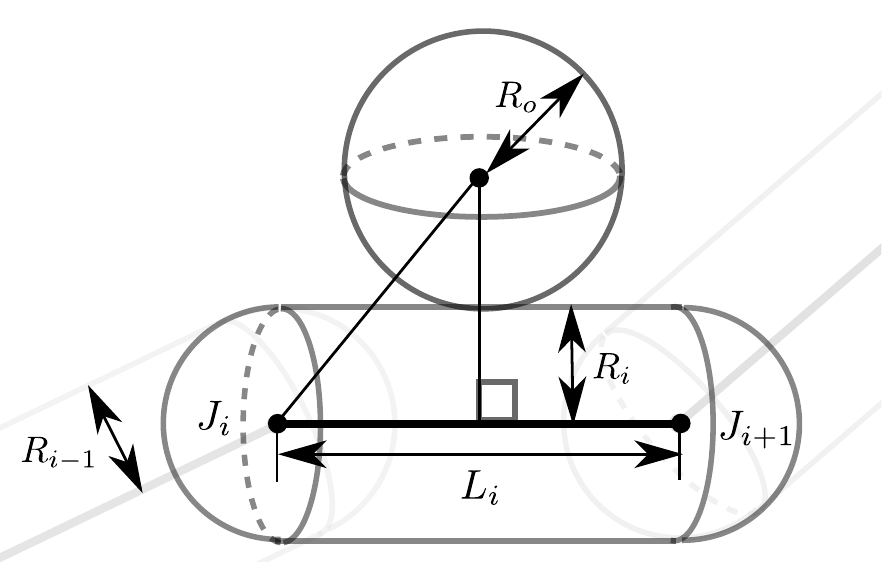}
\else
\includegraphics[width=0.6\linewidth, clip=false ] {int_link.pdf}
\fi
\caption
[ A configuration in which a $i$-th link makes a contact with an external object of radius $R_o$] 
{{\bf A configuration in which a $i$-th link makes a contact with an external object of radius $R_o$} 
} \label{fig:int_l} 
\end{figure}

To be able to use any of the robot's body parts for collision intervention, the position of each  part needs to be evaluated during the motion planning process. We approximate each body part as two hemispheres connected with a cylinder as shown in Fig. \ref{fig:int_l}.

\begin{figure}\centering
\includegraphics[width=\linewidth, clip=false ] {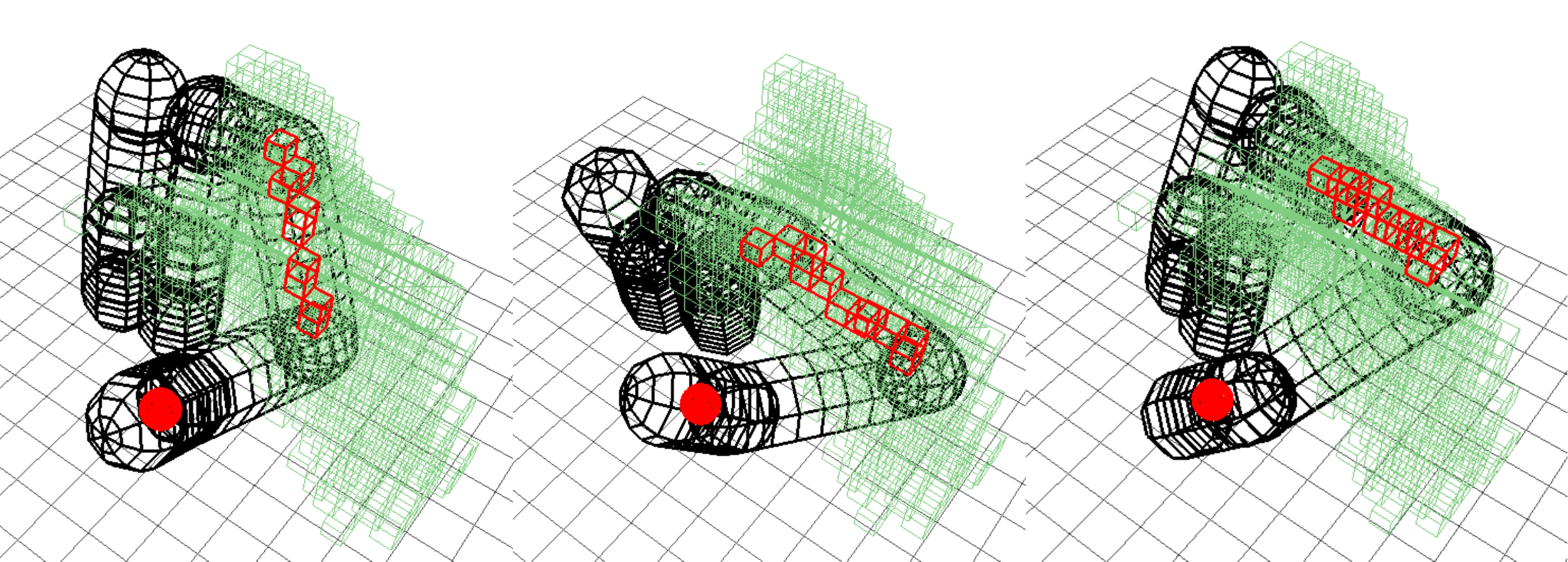}
\caption
[ The reachable volume]
{{\bf The reachable volume}
	corresponding to the robot's upper arm while fulfilling the end-effector task is generated by an occupancy map of all the possible postures.
} \label{fig:occupancy_l} 
\end{figure}

To enable real-time performance while exploring the aforementioned combinatorics, we generate reachable volumes for each body linkage a priori as described in \cite{mcmahon2014}. In contrast, our reachable volumes are generated by sampling feasible configurations rather than analytically computing the Minkowski sum of the reachable volumes. This approach allows us to deal with more  complex shapes and joint limits. We first model each link as a line segment and generate reachable volumes as shown in Fig. \ref{fig:occupancy_l}.
Adjacent Cartesian coordinates of the link positions are clustered, and an 3D cell array structure is generated similarly to \cite{csucan2012}. Each cell represents a set of adjacent points in 3D space and consists of the configurations at which the body link positions are located in the cell. We use an octree \cite{meagher1980} to efficiently represent the reachable volumes. There are $N_l$ octrees for the constrained space and another $N_l$ octrees for the unconstrained space. The computation time for building octrees is $\mathcal{O} ( \log N )$, which is considered as not too expensive. The algorithm for building octrees is shown in Algorithm 1.
\begin{algorithm}[t]
\caption*{Algorithm 1. Building octree}
\begin{algorithmic}
\Function {\rm BuildOctree}{ $Link$, $Octree$ }
\State $p_1 \gets Link.from$
\State $p_2 \gets Link.to$
\State $r \gets Link.radius$
\If { $Octree.depth$ = \rm MAX\_DEPTH }
\State {$Octree.links \gets Link$}
\State \Return
\EndIf
\If { $Octree.depth = \emptyset$ }
\State {\rm CreateChildren($Octree$) }
\EndIf
\For {$child$ in $Octree.children$}
\State { $d \gets $ {\rm getDistToLineSeg}($child.center,\, p_1,\,p_2)$}
\If { $d \le  child.width + r$ }
\State { \rm BuildOctree} ( $Link$, $child$ )
\EndIf
\EndFor
\State \Return
\EndFunction
\end{algorithmic}
\end{algorithm}

The sampled configuration fed into the Octree algorithm is first generated by the PRM method {\cite{kavraki1996}. As we mentioned, constrained configurations are those which keep the end-effector on at a desired task location while unconstrained configurations do not impose any requirement. To consider both constrained and unconstrained options, we generate two different roadmaps. The algorithm for roadmap generation is shown in Algorithm 2. In that algorithm, the variable $Link$ consists of four tuples, $from$, $to$, $r$, $q$ corresponding to the position of one end of the link, the position of the other end, the radius of the link, and the configuration to which the link belongs to, respectively. Algorithm 2 is similar to Algorithm 1, but in the former, the sampled configuration is used to generate the reachable volumes. Also, in the case of constrained motion planning, we exploit the null-space projection of the primary goal task to fulfill the primary end-effector task. The detailed explanation of the constrained planner is described in Sec. \ref{sec:wbmp}. The constrained and unconstrained reachable volumes of our robot's left upper arm are shown in Fig. \ref{fig:constrained}.
\begin{figure}[t]\centering
\includegraphics[width=\linewidth, clip=false ] {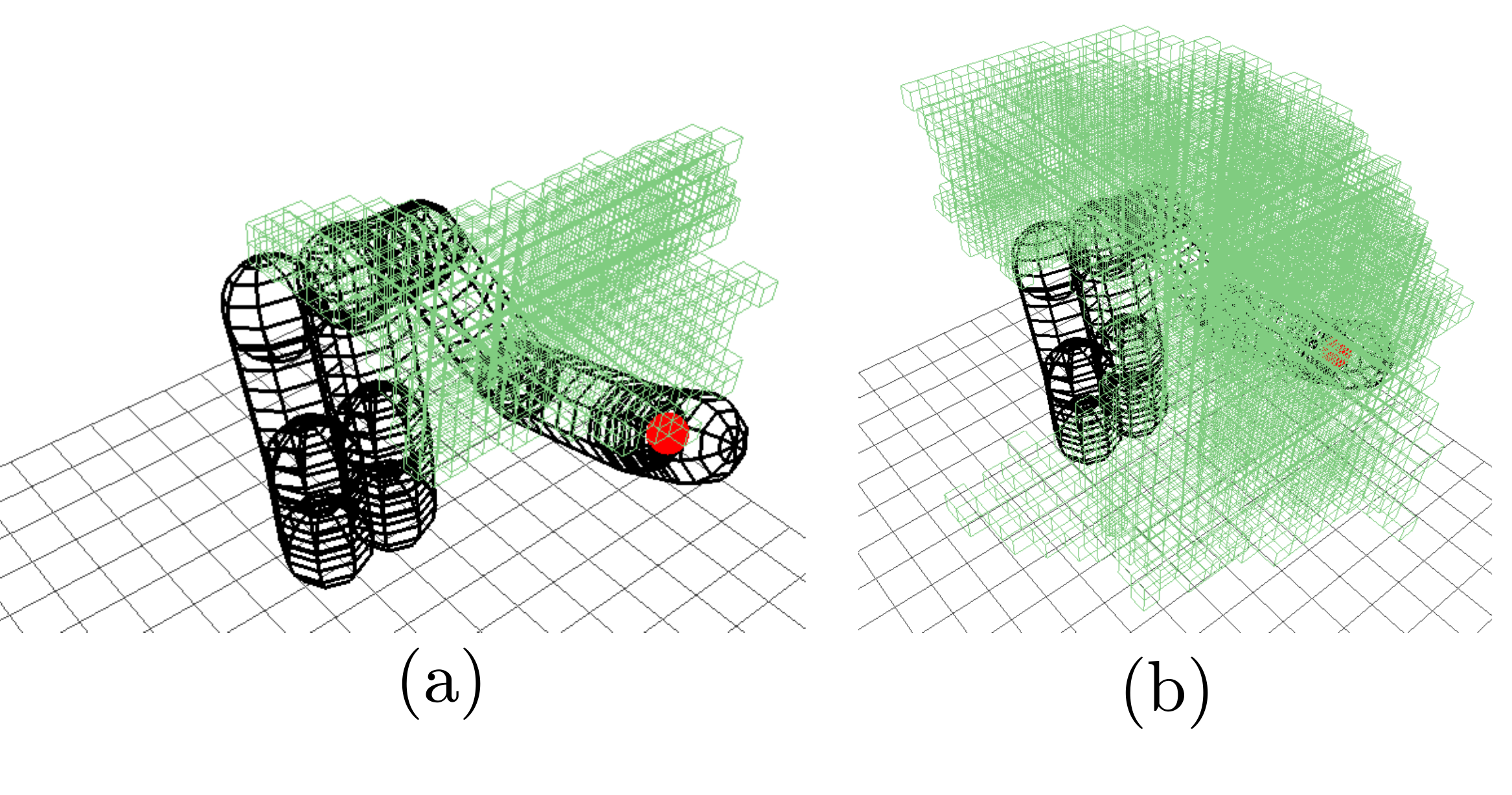}
\caption
[ Constrained and unconstrained reachable volumes of the robot's upper arm]
{{\bf Constrained and unconstrained reachable volumes of the robot's upper arm}. The location of the end-effector for computing constrained reachable volumes is assumed to be stationary. The unconstrained case does not impose end-effector task location requirements. The red ball shown at the end-effector illustrates its goal position in the constrained scenario.
} \label{fig:constrained} 
\end{figure}
Whenever a new sampled configuration is considered, the sample is registered as a new node in a search tree for sampling-based motion planning. In addition, the Cartesian coordinates of each link of the sampled configuration is also registered in a corresponding 3-dimensional cell of the reachable volume. The roadmap also provides paths to connect all configurations, so the motion plan from the current state to the desired intervention state can be effectively derived by the roadmap.
\begin{algorithm}[t]
\caption*{Algorithm 2. Learning phase}
\begin{algorithmic}
\Function {\rm Learn}{}
\State $N$ $\gets$ $\emptyset$,  
$E$ $\gets$ $\emptyset$,
$O$ $\gets$ $\emptyset$
\While{ 1 }
\State $q_{new}$ $\gets$ RandomConf()
\State $q_{near}$ $\gets$ NearestNeighbor($q_{new}$, $N$)
\State $dq$ $\gets$ Project($q_{near}$, $q_{new} - q_{near}$) 
\State $\Delta q$ $\gets$ $dq / \left|dq \right| \times \bf STEP$
\State $q_{prev}$ $\gets$ $q_{near}$
\State $q$ $\gets$ $q_{prev}$ + $\Delta q$
\While{ IsFeasible($q$) }
\State $N$ $\gets$ $N \cup q$
\State $E$ $\gets$ $E \cup \left( q,\, q_{prev}\right)$
\For {$l_i$ in {Links($q$)} }
\State {BuildOctree($l_i$, $O_i$)}
\EndFor 
\State $q_{prev}$ $\gets$ $q$
\State $q$ $\gets$ $q$ + $\Delta q$
\EndWhile
\EndWhile
\EndFunction
\end{algorithmic}
\end{algorithm}


\subsection{Searching for intervening body parts}
During the collision intervention process described in Sec. \ref{sec:agent}, the robot needs to determine which body part can intercept the predicted collision path. Because the reachable volumes of all the links are computed in advance, all the configurations for intervention can be determined by searching the reachable volumes. Given an object position, $p_o$, its velocity, $v_o$, and its radius, $r_o$, we can define the set of configurations, $C_{int,j}$ at which the j-th link intersects the trajectory as follows.
\begin{align}
\begin{split}
C_{int,j} = \big\{& l_k.q \in C \ \bigg|\ l_k \in O_j.links,\ \\ 
&Distance(l_k.from,\, l_k.to,\, p_o,\, v_o \cdot t_d ) \le r_o + r \big\} \label{eq:c_intj}
\end{split}
\end{align}
where the function, $Distance$ returns the minimum distance between two line segments. The four arguments for that function are the pairs defining the line segments. The configurations of the previous set have at least one link that intersects the object trajectories. If the set has at least one element, the manipulator can intervene to stop the collision using the corresponding body link. Our robot consists of $N_l$ body links, and there are also constrained and unconstrained motion plans. Therefore, given a collision trajectory, $2 N_l$ sets of intersecting configurations need to be considered.

The decision policy for the intervening body link is shown in Fig. \ref{fig:decision}.
\begin{figure*}\centering
\includegraphics[width=\linewidth, clip=false ] {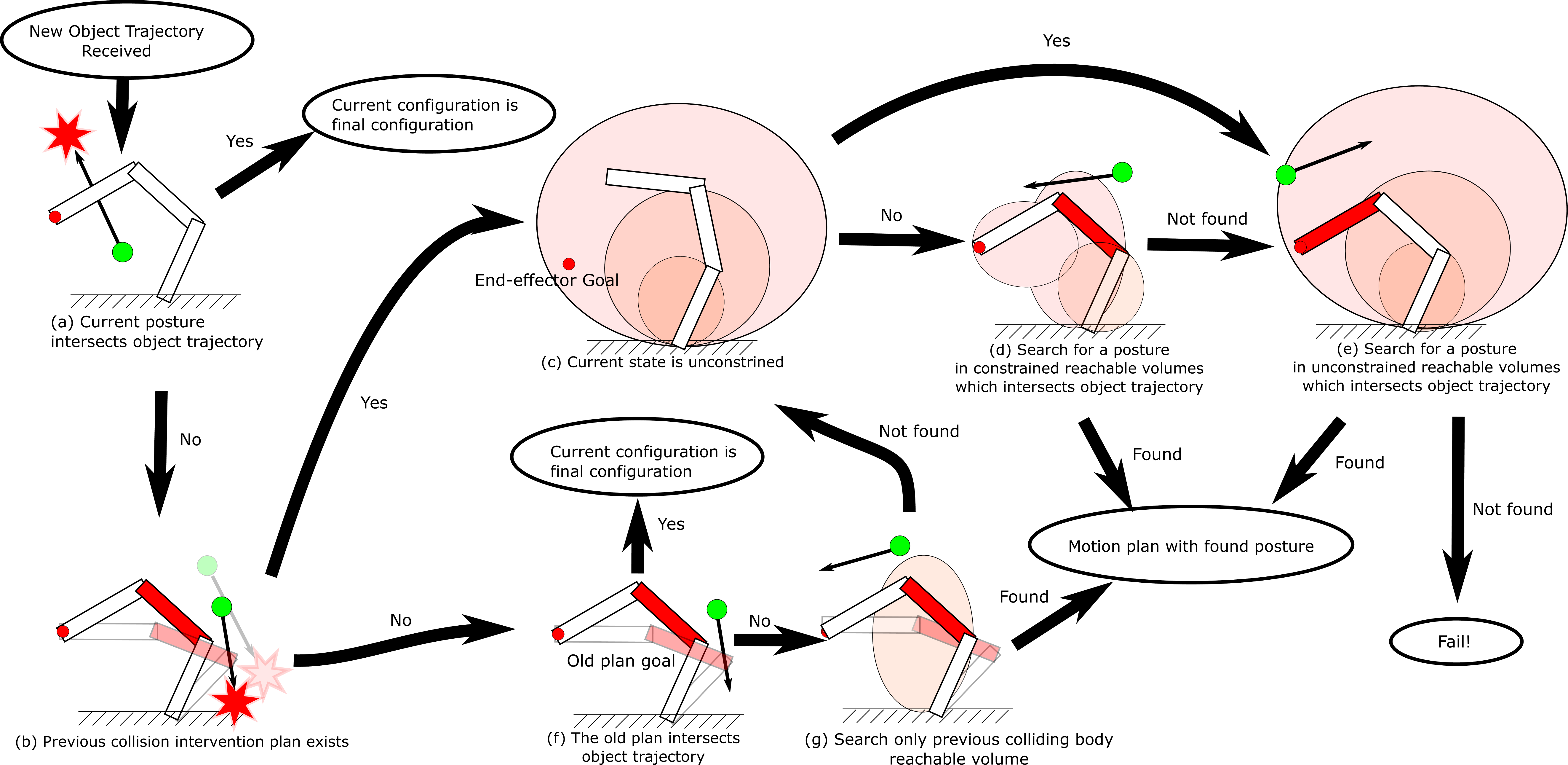}
\caption
[Decision policy of the intervening body link]
{{\bf Decision policy of the intervening body link}. When there is an existing motion plan, the robot agent will reuse it. If not, a new motion plan is generated.
} \label{fig:decision} 
\end{figure*}
Whenever the state of the objects to be stopped changes, the agent determines new intervening body parts. If the current robot configuration intersects the trajectory of the object (Fig. \ref{fig:decision} (a)), then it should remain at that configuration. 
If not, the decision system checks whether the robot is operating in the intervention mode (Fig. \ref{fig:decision} (b)) such that it can determine whether to reuse the previous motion plan. If the agent is already intervening in a collision path, the final destination of the previous motion plan is evaluated to see whether it intersects with the collision trajectory (Fig. \ref{fig:decision} (f)). If it intersects the trajectory, we reuse the previous motion plan. In the case when the final destination does not intersect the trajectory, we replan the intervention task. To replan the task seamlessly, we attempt to reuse the chosen intervening body link from the previous motion plan (Fig. \ref{fig:decision} (g)). If the reachable volume of the link intersects the collision trajectory, we build a new motion plan to reach one of the configurations in the volume. If not, we replan the motion plan without any consideration with respect to the previous motion plan.

If the current state violates the end-effector task (Fig. \ref{fig:decision} (c)), we search the unconstrained reachable volumes.
If the current state fulfills the end-effector task, we search the constrained reachable volumes. If the robot agent does not find a configuration which intersects the collision trajectory within the constrained reachable volumes, it searches the unconstrained volumes. If the search fails, the intervention fails because the agent cannot reach the collision trajectory.

\subsection{Constrained motion planning}
\label{sec:wbmp}
In this section, we deal with the constrained manifold imposed by the primary end-effector goal task such that the robot can intervene using whole-body operation space control (WBOSC). Thanks to the task hierarchy features of WBOSC, we can generate a motion plan compliant with higher priority tasks. The task hierarchy of WBOSC is provided via null space projections of higher priority tasks. The basis of the constraint manifold of higher priority tasks is given by WBOSC operators. An introduction to WBOSC and detailed descriptions and derivations are provided in Appendix \ref{sec:TH}.

The rank of $\Lambda_2^*$ in Appendix \ref{sec:TH}, which is equivalent to that of $\Lambda_2^{*\,+}$, is the same as the dimension of the task space coordinates (e.g. the end-effector position). Thus, by the definition of $\Lambda_2^*$, the rank of $J_2^* N_1^* \Phi^* N_1^{*\,T} J_2^{*\,T}$ emerges from the following model.
\begin{align}
\begin{split}
\Lambda_2^* &= J_2^* N_1^* \Phi^* N_1^{*\,T} J_2^{*\,T} \\
			&= J_2 \overline{UN_c} N_1^* UN_c A^{-1} \left(UN_c\right)^T N_1^{*\,T} J_2^{*\,T}\\
&= J_2 N_1 A^{-1} N_1^T J_2^T \label{eq:l2}
\end{split}
\end{align}
where $N_1$ is the null space projection of the higher priority task into the lower priority task, and is defined as follows.
\begin{align}
	N_1 &= I - A^{-1} N_c^T J_1^T \left(J_1 A^{-1} N_c^T J_1^T\right)^+ J_1 N_c \label{eq:n1}
\end{align}
Because $A$ is a positive definite matrix, we only check the rank of $J_2 N_1$ to identify the dimension of the lower priority task. The rank of $N_1$ is determined from Eq. (\ref{eq:n1}). Taking the singular value decomposition of $A^{-1/2}N_c^T J_1^T$, Eq. (\ref{eq:n1}) can be expressed as follows.
\begin{align}
\begin{split}
N_1 &= I - A^{-1/2} \mathcal{U} \Sigma \mathcal{V}^T \left( \mathcal{V} \Sigma^T \Sigma \mathcal{V}^T \right)^+ \mathcal{V} \Sigma \mathcal{U}^T A^{1/2} \\
&= A^{-1/2} \left( I - \mathcal{U} \Sigma \cancel{\mathcal{V}^T \mathcal{V}} \Sigma^+ \Sigma^{+\,T} \cancel{\mathcal{V}^T \mathcal{V}} \Sigma \mathcal{U}^T \right) A^{1/2} \\
&= A^{-1/2} \left( I - \mathcal{U} \Sigma \Sigma^+ \Sigma^{+\,T} \Sigma \mathcal{U}^T \right) A^{1/2} \label{eq:n1_2}
\end{split}
\end{align}
If the rank of the diagonal matrix, $\Sigma \in \mathbb{R}^{r \times n}$ is $r < n$,  then $\Sigma \Sigma^+ \Sigma^{+\,T} \Sigma = \begin{pmatrix} I_{r \times r} & 0 \\ 0 & 0 \end{pmatrix}$ and Eq. (\ref{eq:n1_2}) becomes the following.
\begin{align}
\begin{split}
N_1 &= A^{-1/2} \left( I - \mathcal{U} \begin{pmatrix} I_{ r \times r} & 0 \\ 0 & 0 \end{pmatrix} \mathcal{U}^T  \right) A^{1/2} \\
&= A^{-1/2} \mathcal{U} \begin{pmatrix} 0 & 0 \\ 0 & I_{ (n-r) \times (n-r)} \end{pmatrix} \mathcal{U}^T  A^{1/2}
\end{split}
\end{align}
Thus, the rank of $N_1$ is characterized as the dimension of the robot's generalized coordinates minus the rank of $J_1 N_c$.

The basis of the constrained manifold can be derived by substituting Eq. (\ref{eq:l2}) into Eq. (\ref{eq:x2}) as follows.
\begin{align}
\ddot{x}_2 &= \Lambda_2^{*\,+} \Lambda_2^* \ddot{x}_{2,des} + \tau_1^{\prime \prime} \label{eq:x2u}\\
\intertext{ Because $\Lambda_2^*$ is a symmetric matrix, we can take its singular value decomposition and express it as $\mathcal{U} \Sigma \mathcal{U}^T$. Then, $\ddot{x}_2$ can be derived as follows}
\begin{split}
\ddot{x}_2&= \mathcal{U} \Sigma^+ \cancel{\mathcal{U}^T \mathcal{U}} \Sigma \mathcal{U}^T \ddot{x}_{2\,des} + \tau_1^{\prime \prime}\\
&= \mathcal{U} \begin{pmatrix} I_{r_2 \times r_2} & 0 \\ 0 & 0 \end{pmatrix} \mathcal{U}^T \ddot{x}_{2\,des} + \tau_1^{\prime \prime} \label{eq:x2u2}
\end{split}
\end{align}
where $r_2$ is the rank of $\Lambda_2^\ast$. In this equation, $\mathcal{U}$ consists of singular vectors, $\begin{pmatrix} u_1 & u_2 & \cdots & u_n \end{pmatrix}$ and $\mathcal{U}^T \ddot{x}_{2\,des}$ can be defined as an arbitrary vector, $z = \begin{pmatrix} z_1 & z_2 & \cdots & z_n \end{pmatrix}^T $. Then, $\ddot{x}_2$ can be expressed as the weighted sum of the independent basis as follows
\begin{align}
\ddot{x}_2 &= z_1 u_1 + z_2 u_2 + \cdots + z_{r_2} u_{r_2} + \tau_1^{\prime \prime} \label{eq:sv}
\end{align}
Therefore, the singular vectors of the task space mass matrix, $\Lambda_2^*$ are the basis of the constraint manifold considering the null space of higher priority tasks.

From the previous section, the basis of the lower priority intervention task has been derived. With this basis, we can generate the $Project$ function of Algorithm 1. The arguments of $Project$ are the current configuration and the displacement of the joint positions. To make the displacement compliant with the end-effector task constraint, it needs to be projected into the null space of the end-effector based on Eq. (\ref{eq:x2u}). $\Lambda_2^\ast$ is the dynamic term which is a function of $q$, such that the projected displacement of Algorithm 1 can be derived as follows.
\begin{align}
\ddot{x}_2 = \Lambda_2^{\ast\, +}(q) \Lambda_2^\ast(q) \left( x_{disp} - \tau^\prime \right) + \tau^\prime
\end{align}




\section{Experiments}
\subsection{Experimental setup}
\begin{figure}[t]\centering
\ifdefined\JOURNAL
\includegraphics[width=.7\linewidth, clip=false ] {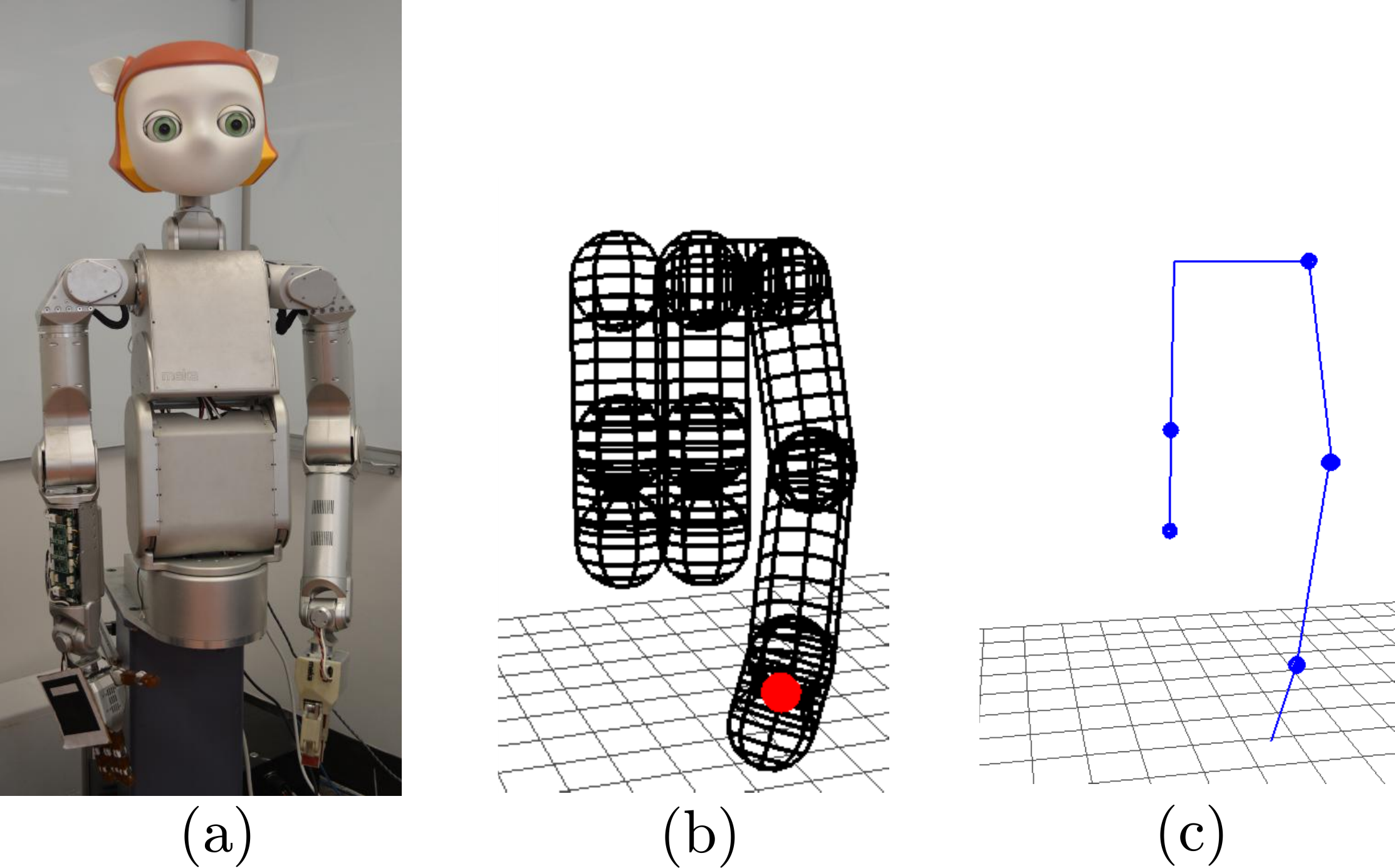}
\else
\includegraphics[width=\linewidth, clip=false ] {dreamer.pdf}
\fi
\caption
[Dreamer]
{{\bf Dreamer}, an upper-body humanoid robot used for the experiment is shown in (a). Dreamer consists of two 7-DoF arms a 3-DoF torso, the latter including two bending coupled joints. Each robot arm consists of 7 revolute joints, which we characterize by 3 capsules as shown in (b). In addition, each torso link is approximated by two capsules. The simplified stick figure of the robot (only considering its left arm) is shown in (c) and determines the robot's kinematic posture. 
} \label{fig:dreamer} 
\end{figure}
To validate the proposed methodology, we conduct collision intervention experiments with an upper-body humanoid robot, Dreamer, shown in Fig. \ref{fig:dreamer}. Detailed information about the robot is provided in \cite{sentis2012}.

To sense its environment, the robot uses a Kinect sensor to detect external objects. Objects in the robot's environment are identified via a color profile and markers. The color profile of a tennis ball is considered in advance, and markers are identified using the Aruco library \cite{munoz2012}. Therefore, whenever the Kinect sensor receives a new RGB image frame with a corresponding depth image, the robot can track a green tennis ball and various registered markers such as the one shown in Fig. \ref{fig:markers}, while the locations and the velocities of the objects are estimated via Kalman filters as shown in Eq. (\ref{eq:st}). The parameters for the Kalman filter are shown in Table \ref{tab:kalman}. Given those states, the collision probability between two objects can be estimated as shown in Sec. \ref{sec:colman}.
The whole experimental setup with Dreamer and the Kinect sensor is shown in Fig. \ref{fig:setup}.
\begin{figure}[t]\centering
\ifdefined\JOURNAL
\includegraphics[width=0.8\linewidth, clip=false ] {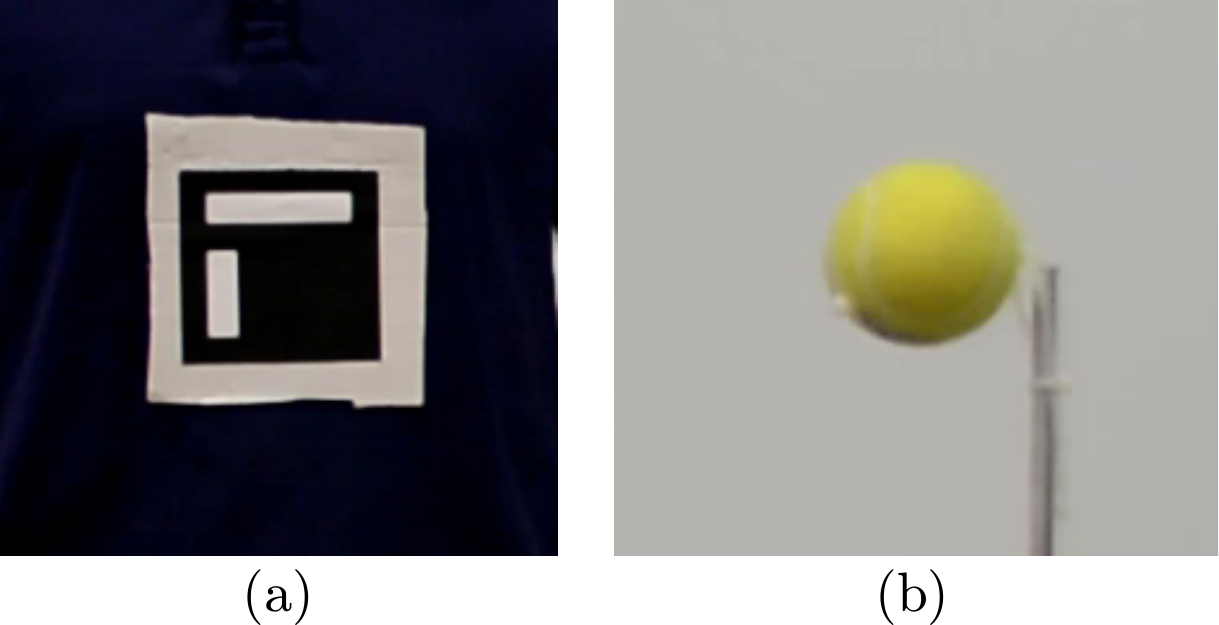}
\else
\includegraphics[width=0.7\linewidth, clip=false ] {markers.pdf}
\fi
\caption
[ Objects to be detected]
{{\bf Objects to be detected}. The robot tracks the marker in (a) and the green tennis ball in (b) using Aruco and the OpenCV library. During tracking, their positions and velocities are estimated via Kalman filter. Kinect sensing updates the input video sequence every 33msec, so the Kalman filter also updates the states every 33msec.
} \label{fig:markers} 
\end{figure}
\begin{figure*}[t]\centering
\ifdefined\JOURNAL
\includegraphics[width=0.9\linewidth, clip=false ] {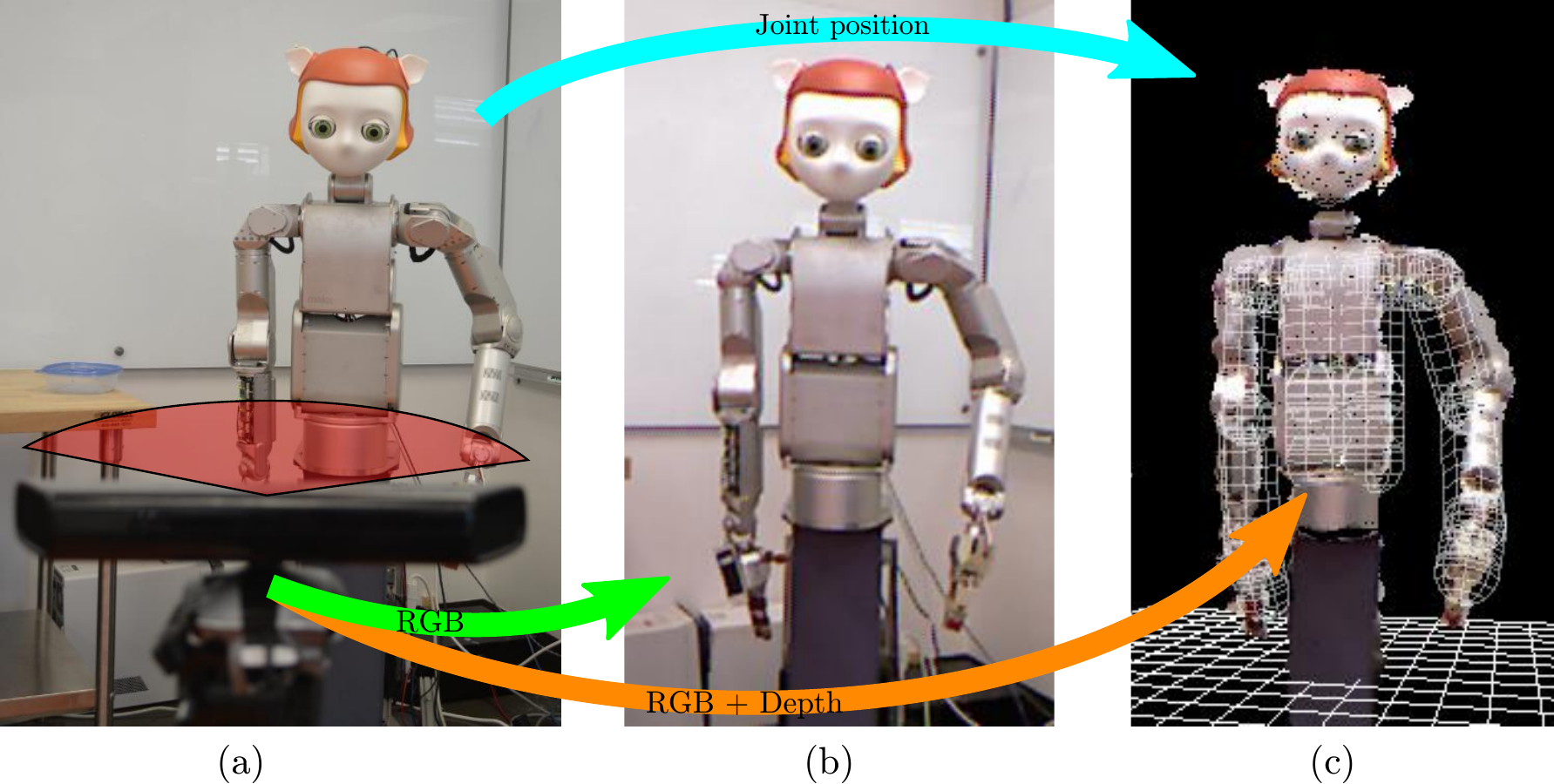}
\else
\includegraphics[width=\linewidth, clip=false ] {setup.pdf}
\fi
\caption
[The experimental setup]
{{\bf The experimental setup}
	consists of an external Kinect camera and the Dreamer robot (a). The transformation between the robot coordinates and the Kinect coordinates is calculated beforehand, so all the RGB points from Kinect in (b) can be represented in the robot's coordinates in (c). The robot's body mesh can also be inferred via joint positions as shown in (c).
} \label{fig:setup} 
\end{figure*}
\begin{table}
\caption
[The Kalman filter parameters for object tracking]
{\rm The Kalman filter parameters for object tracking}\label{tab:kalman}
\begin{center}
\begin{tabular}{|c|l|c|c|}
\hline
Name & Definition & Value & Metric \\
\hline
\hline
$\Delta t$ & Sampling time & 33 & \rm msec \\
\hline
$\Sigma_d$ & Covariance of velocity disturbance & diag(0.01, 0.01, 0.01) & $\rm \left( m / s \right)^2 $ \\
\hline
$\Sigma_\alpha$ & Covariance of acceleration & diag(1.5, 1.5, 1.5) & \rm $\rm \left( m / s^2 \right)^2$ \\
\hline
$\Sigma_s$ & Position sensor noise & diag(0.01, 0.01, 0.01) & $\rm m^2 $ \\
\hline
\end{tabular}
\end{center}
\end{table}

Dreamer is given a primary goal task in which the end-effector has a goal position, and the collision intervention task has a constraint to operate the robot under the null space of the primary task when possible. The end-effector stationary goal is shown as a red ball in Fig. \ref{fig:constrained}-(a) and Fig. \ref{fig:dreamer}-(b). Before collision intervention takes place, the robot generates the constrained and unconstrained probabilistic roadmaps using 10,000 samples, and the reachable volumes of the arm and shoulder links in Fig. \ref{fig:dreamer}-(b) corresponding to the roadmaps are also generated. The volume of the whole environment is $2 \rm m \times 2 \rm m \times 2 \rm m = 8 \rm m^3$, and the space is segmented via the Octree algorithm with 6 levels of depth. Therefore, we consider 8 reachable volumes corresponding to the constrained/unconstrained roadmaps and the four links shown in Fig. \ref{fig:dreamer}(c). The dimension of each cube in the Octrees is $3 \rm cm \times 3 \rm cm \times 3 \rm cm$. The generated reachable volumes of the constrained roadmap for the 4 links are shown in Fig. \ref{fig:occu}. If the object trajectory overlaps any of these constrained volumes, the robot can simultaneously stop the object movement while keeping the end-effector on its goal position. The search for this process is conducted based on the decision policy shown in Fig. \ref{fig:decision}.
\begin{figure}[t]\centering
\includegraphics[width=0.8\linewidth] {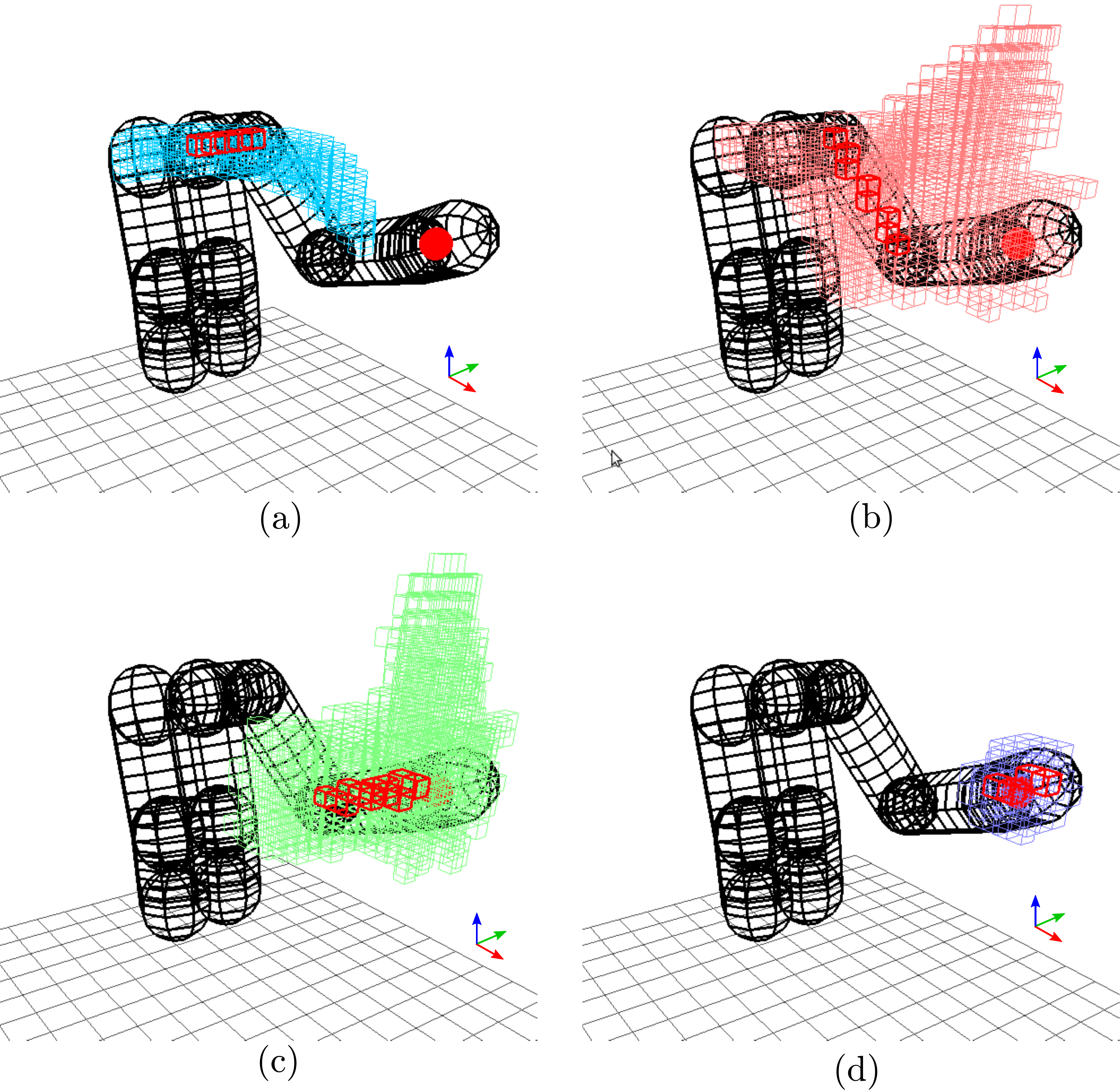}
\caption
[ The reachable volumes of the links corresponding to the constrained roadmap]
{{\bf The reachable volumes of the links corresponding to the constrained roadmap}
	are generated for the occupancies of the shoulder, the upper arm, the lower arm, and the end-effector. 
} \label{fig:occu} 
\end{figure}

\subsection{Constrained collision intervention}
\begin{figure}[t]\centering
\includegraphics[width=0.8\linewidth] {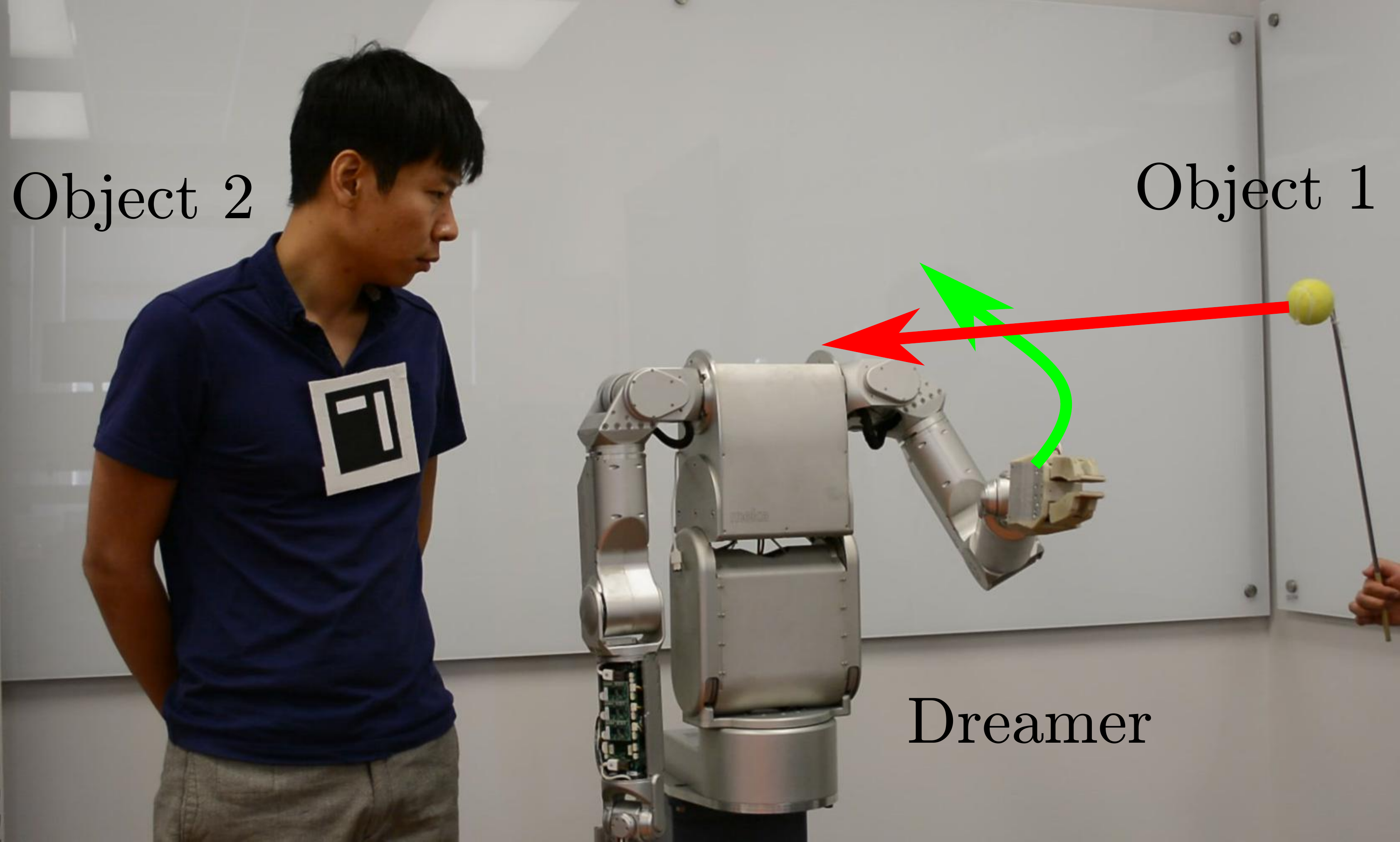}
\caption
[ Concept image of the experiments]
{{\bf Concept image of the experiments}.  The object 1 approaches the object 2, and the robot collides with the object 1 if it will collides with the object2. We assume that the object 2 is a human operator who should be protected from the object 1, which is a ball.
} \label{fig:concept} 
\end{figure}
In this experiment, the object with the marker is assumed to be a human operator working next to the robot and the ball represents a moving object around the workspace and harmful to the operator. Therefore, the objective of the robot is to 1) observe both objects, 2) determine whether these objects will collide with each other, and 3) stop external collisions between the ball and the ``human operator'' by using one of its body parts. The illustration of this scene is shown in Fig. \ref{fig:concept}. For these experiments, object 1 is the ball, which approaches object 2, which is the human operator. The movements of the three objects is observed, and the probability that they collide with each other is predicted based on Eq. (\ref{eq:pac}). To determine whether there will be a collision or not, the threshold described in Fig. \ref{fig:acc_col} needs to be defined. In this experiment, we define $\eta = 0.5$ and $t_c = 4$, which means there will be a collision if the cumulative collision probability will be 0.5 or above and within a 5 seconds prediction horizon. If the collision probability exceeds the threshold, the robot searches its reachable volumes for postures overlapping the future trajectory of the external objects as shown in Fig. \ref{fig:int_l}. According to the search policy shown in Fig. \ref{fig:decision}, the robot searches the constrained reachable volumes first. If it fails to find a feasible intervening body, it searches the unconstrained reachable volumes. If the robot finds an overlapping posture, it will generate a motion plan from the current posture to the intervening posture found using PRM.
\begin{figure}[p]\centering
\begin{minipage}[t]{0.45\linewidth}
\includegraphics[width=\linewidth, clip=false ] {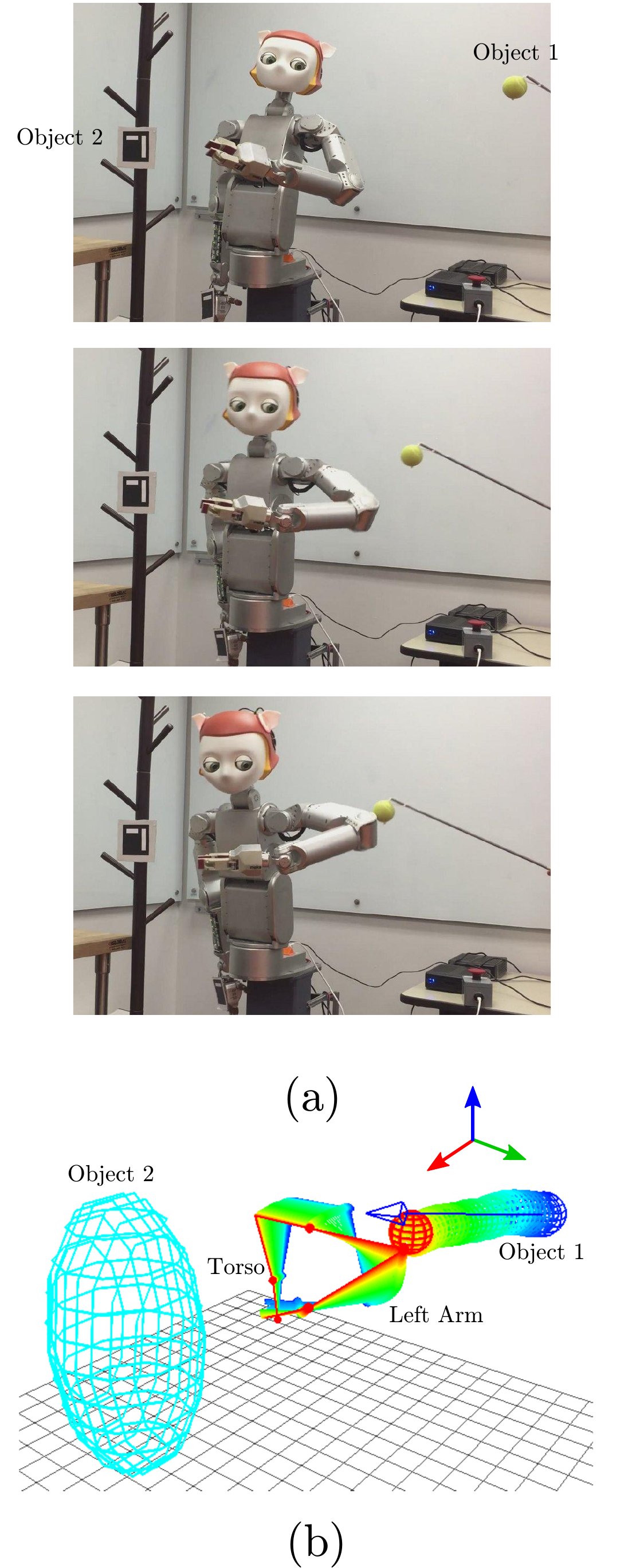}
\caption
[Collision intervention using the robot's upper arm]
{{\bf Collision intervention using the robot's upper arm}. The robot generates a motion plan which makes the upper arm collide with object 1 (a). Subfig (b) shows the posture and ball trajectories.
} \label{fig:col1} 
\end{minipage}
\begin{minipage}[t]{0.45\linewidth}
\includegraphics[width=\linewidth, clip=false ] {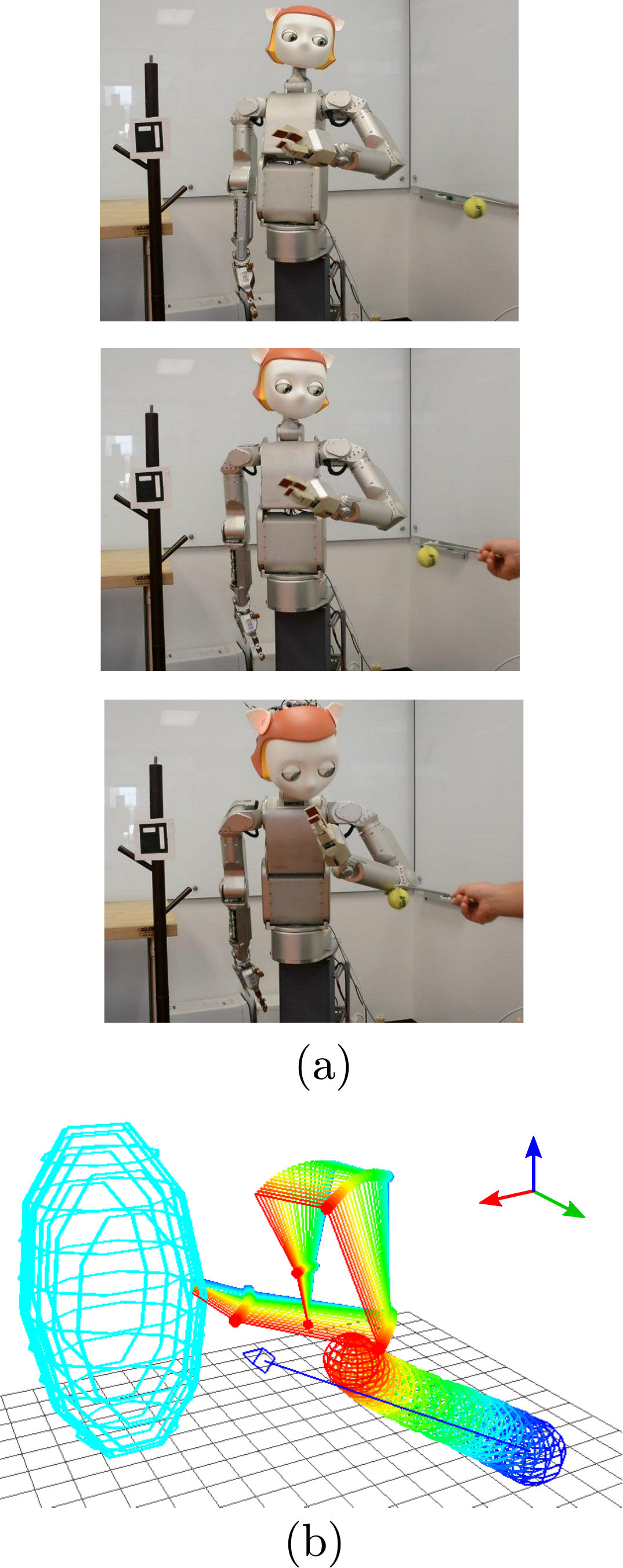}
\caption
[Collision intervention with the robot's lower arm]
{{\bf Collision intervention with the robot's lower arm}. The lower arm of the robot collides with object 1 (a). The blue arrow corresponds to the anticipated object motion.
} \label{fig:col2} 
\end{minipage}
\end{figure}

Three separate collision intervention processes are shown in Figs. \ref{fig:col1}, \ref{fig:col2}, and \ref{fig:col3}. In all of these images the first set of images depicts the robot determining that objects 1 and 2 will collide against each other, and switching its state to the intervention mode. In the intervention situation shown in Fig. \ref{fig:col1}, the robot intervenes using its upper arm, while in Fig. \ref{fig:col2} it determines that the best body part is the lower arm. In addition, Fig. \ref{fig:col3} shows an intervention process using the robot's shoulder. Reachable volume data analysis is further shown in Fig. \ref{fig:occu1} for the first two experiments. 

\begin{figure}[h]\centering
\ifdefined\JOURNAL
\includegraphics[width=0.9\linewidth, clip=false ] {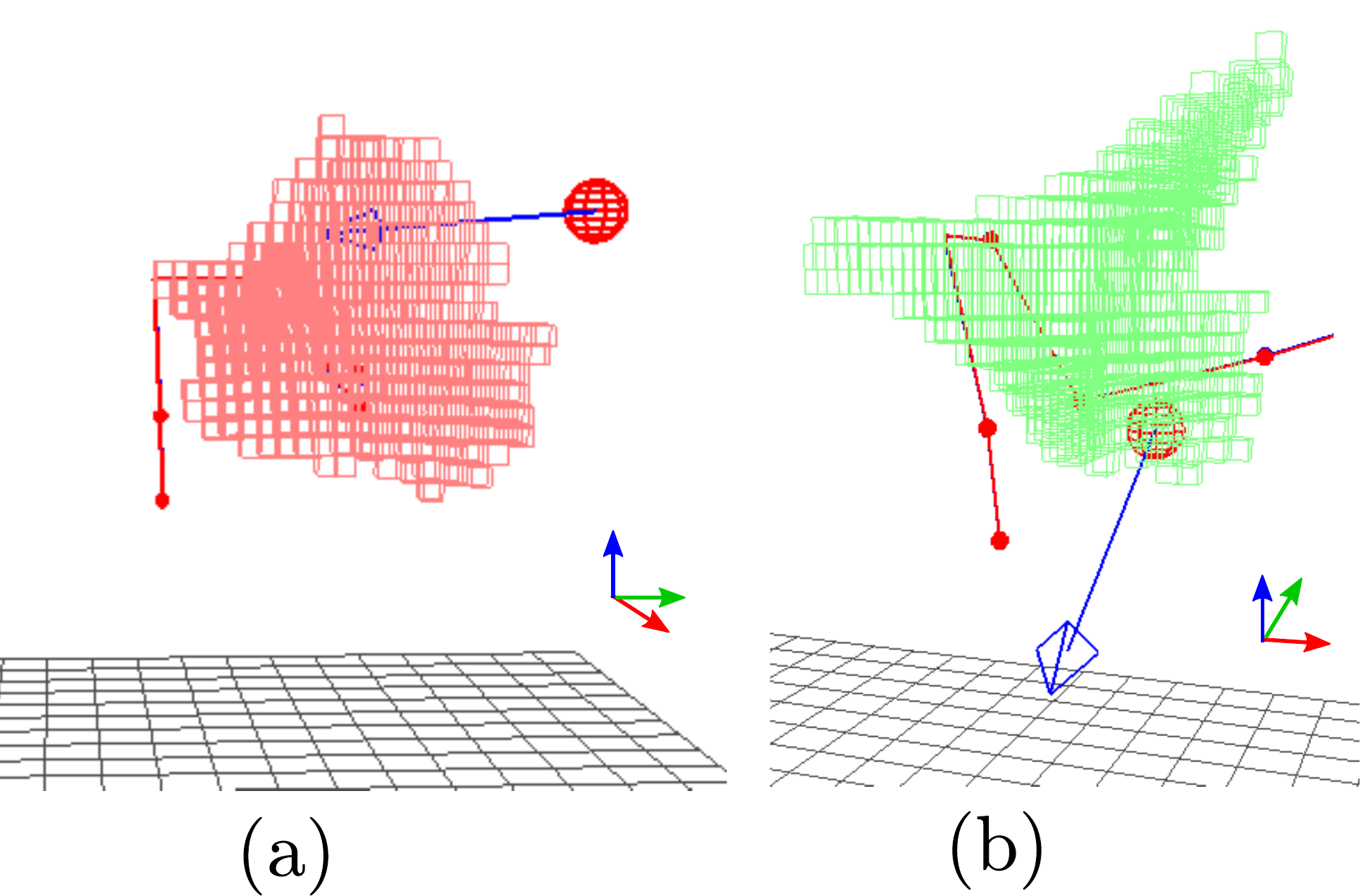}
\else
\includegraphics[width=0.7\linewidth, clip=false ] {occu1.pdf}
\fi
\caption
[ Reachable volume during collision intervention]
{{\bf Reachable volume during collision intervention}. When object 1 approaches object 2 in Fig. \ref{fig:col1}, the reachable volume of the upper arm overlaps the ball trajectory shown as a blue arrow in (a). In contrast, the estimated trajectory of the ball shown in Fig. \ref{fig:col2} overlaps with that of the lower arm as shown in (b).} \label{fig:occu1} 
\end{figure}

\subsection{Unconstrained collision intervention}
\begin{figure}[p]\centering
\begin{minipage}[t]{0.42\linewidth}
\includegraphics[width=\linewidth, clip=false ] {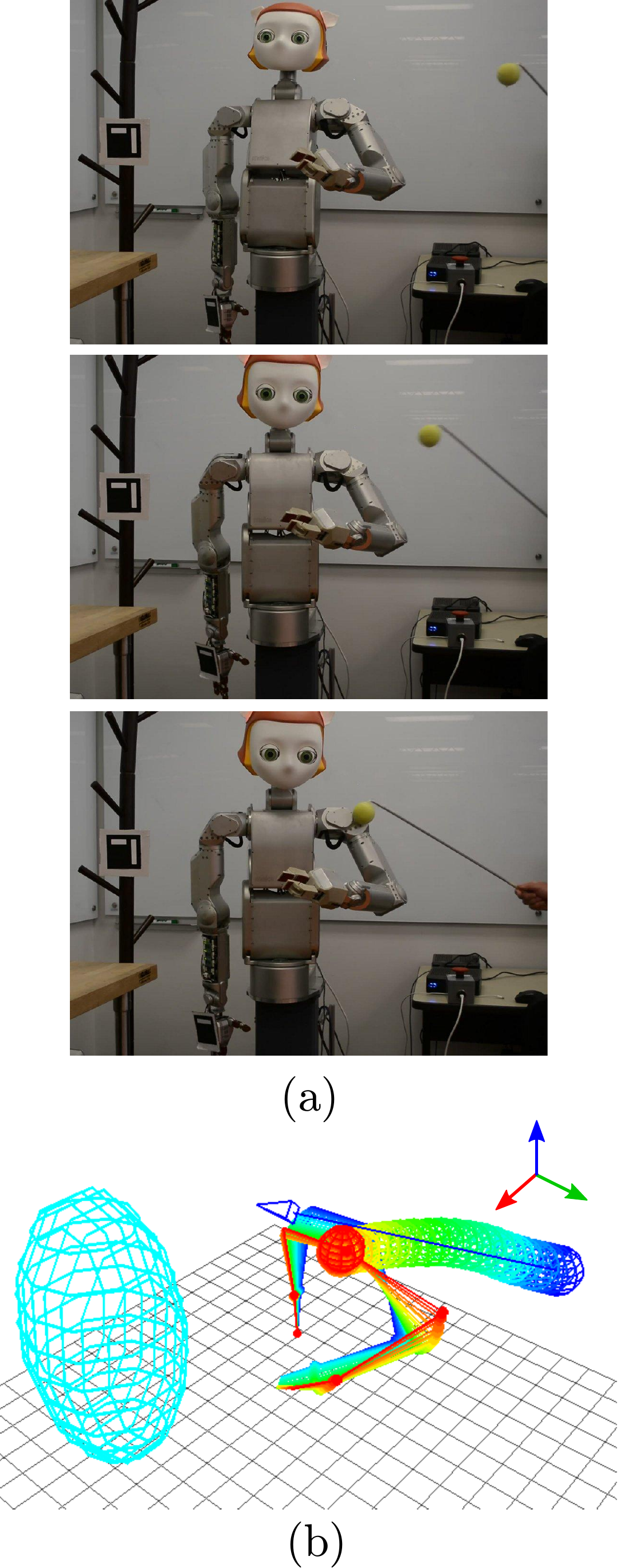}
\caption
[ Collision intervention using the robot's shoulder]
{{\bf Collision intervention using the robot's shoulder}. The reachable volume of the robot's shoulder overlaps with the estimated ball trajectory and as a result the collision intervention is chosen to occur using the shoulder.
} \label{fig:col3} 
\end{minipage}
\begin{minipage}[t]{0.42\linewidth}
\includegraphics[width=\linewidth, clip=false ] {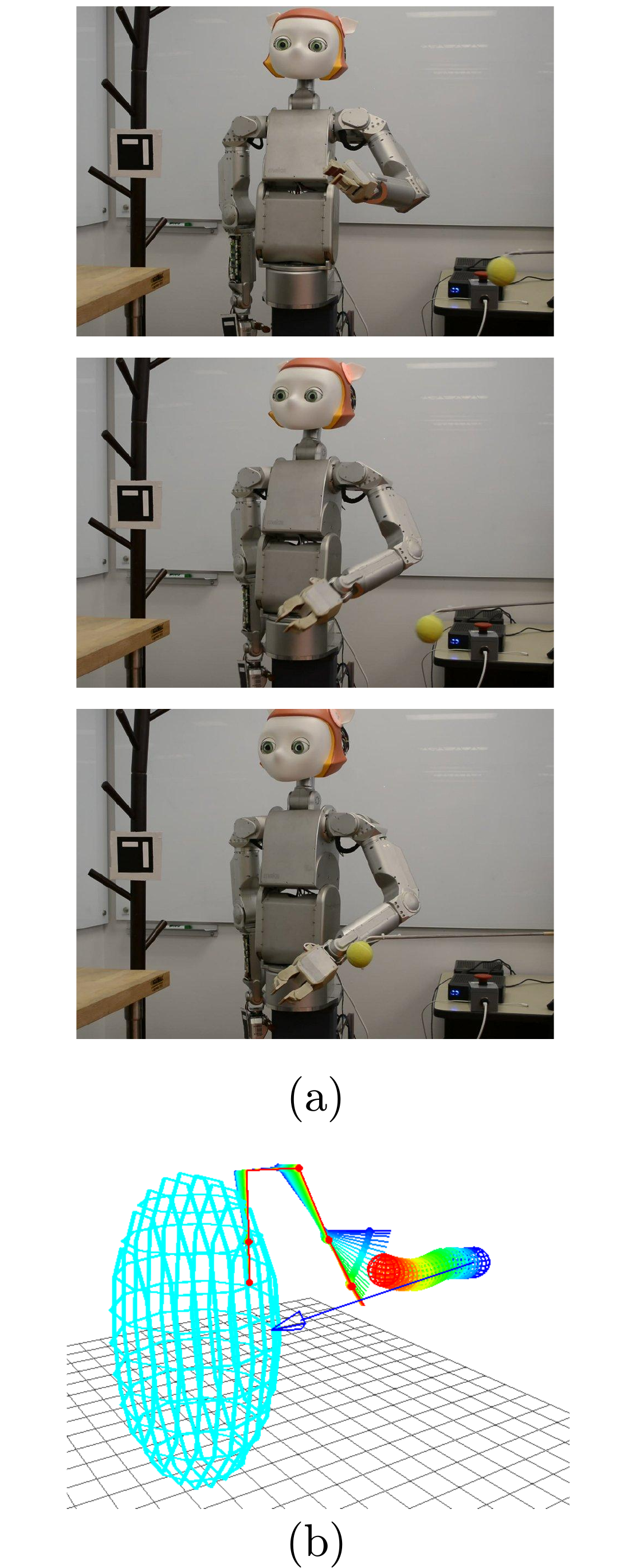}
\caption
[Unconstrained collision intervention using the robot's lower arm]
{{\bf Unconstrained collision intervention using the robot's lower arm}. Here, there is no constrained reachable volume that can be used for collision intervention. As a result the robot intervenes using the unconstrained lower arm.
} \label{fig:col4} 
\end{minipage}
\end{figure}
\begin{figure}[h]\centering
\ifdefined\JOURNAL
\includegraphics[width=0.9\linewidth, clip=false ] {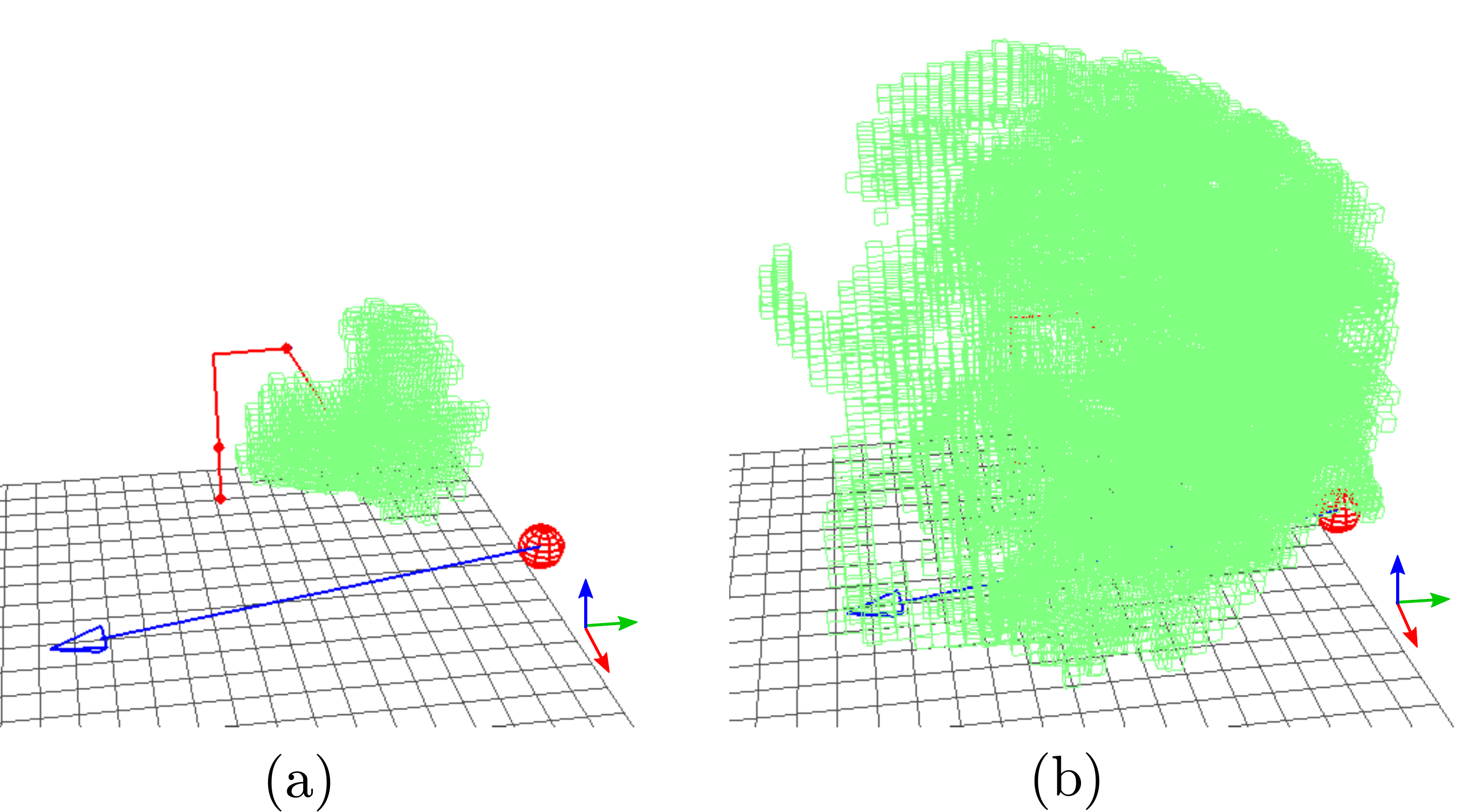}
\else
\includegraphics[width=0.8\linewidth, clip=false ] {occu2.pdf}
\fi
\caption
[Constrained and unconstrained reachable volumes]
{{\bf Constrained and unconstrained reachable volumes} are shown in subfigures (a) and (b), respectively. Those volumes correspond to the reasoning process of the experiment shown in Fig.~\ref{fig:col4}. The estimated object trajectory of Fig. \ref{fig:col4} is shown as blue arrows. Because some cubes of the unconstrained reachable volume shown in (b) overlap the trajectory, the unconstrained lower arm is chosen to be used for intervention. } \label{fig:occu2} 
\end{figure}
If the anticipated object's trajectory does not overlap with any of the constrained reachable volumes, the robot cannot intervene while satisfying the primary end-effector goal task.
In the experiment of Fig.~\ref{fig:col4}, the robot is allowed to violate the goal task. Fig. \ref{fig:occu2} shows the constrained and unconstrained reachable volumes when corresponding to that experiment, which justify the decision to violate the primary goal task. 
In the first image sequence of Fig. \ref{fig:col4}, object 1 approaches object 2, and when collision probability exceeds the designed threshold, then the robot decides to intervene.

\subsection{Proof of concept experiments of intervention between two robots}
Fig. \ref{fig:trikey_dreamer} shows experiments depicting intervention procedures between an upper-body humanoid robot Dreamer and a mobile robot, Trikey to avoid injury to a human. While the ground mobile robot approaches the standing human operator, the humanoid robot estimates the probability of an external collision. If it exceeds a designed threshold, the humanoid intervenes the collision by blocking the mobile robot. In addition, the ground mobile robot incorporates a control algorithm that changes direction once it senses a contact as described in Chapter \ref{chap:auro}. As a result the mobile robot will move away from the humanoid robot upon contact and thus preventing injuring the human. Of course, it is undesirable that mobile robot's threaten with these kind of potential accidents as they should be designed to avoid obstacles. However, our studies are based on the hypothesis that every now and then collision avoidance will fail to work and other means of safety could potentially be beneficial like the ones considered here.
\begin{figure}[p]\centering
\includegraphics[width=0.7\linewidth] {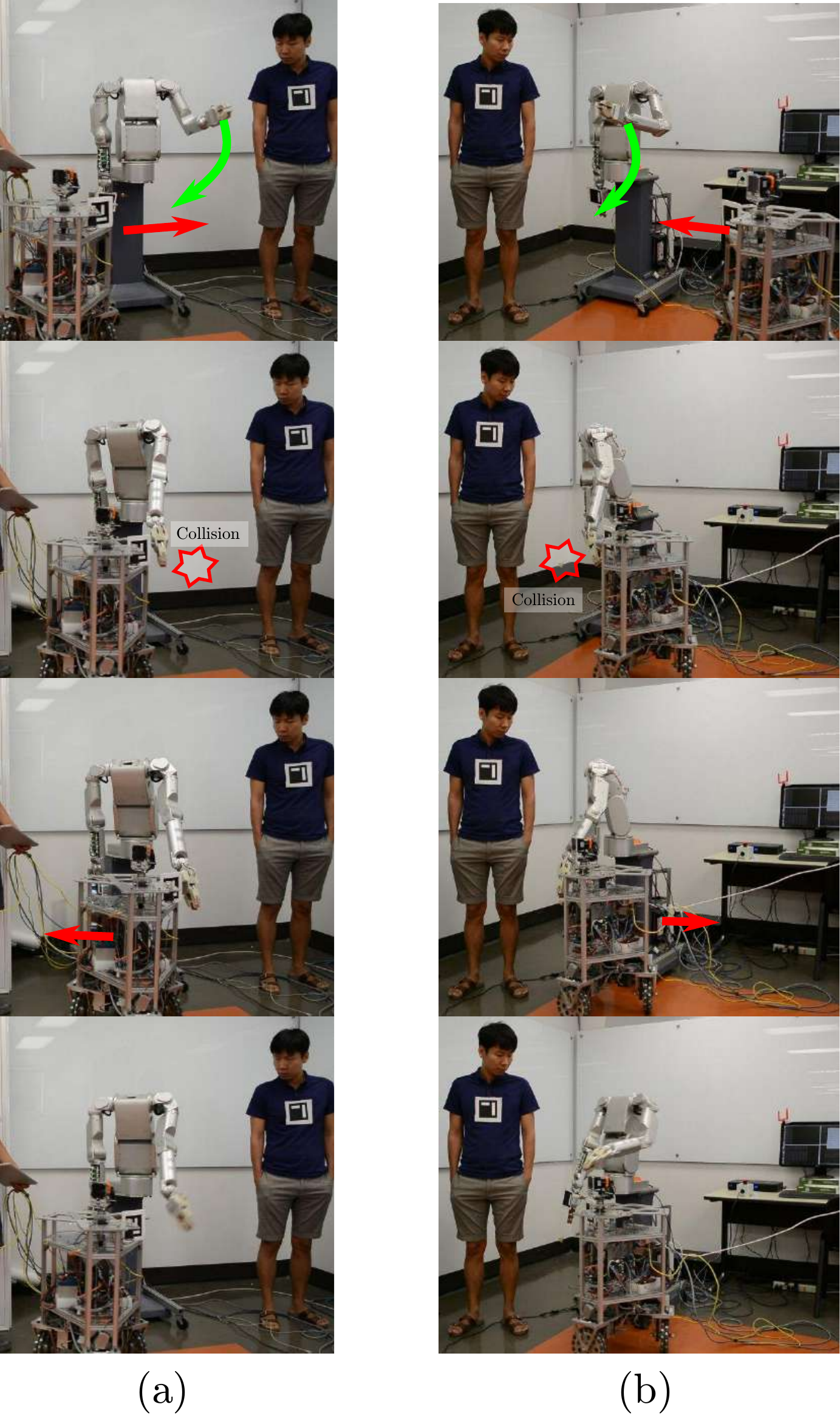}
\caption
[ Proof of concept experiment between two robots]
{{\bf Proof of concept experiment between two robots}. The ground mobile robot previously shown in Chapter \ref{chap:auro}, first approaches a human operator. The humanoid robot decises to intervenes to stop the likely collision between the human operator and the mobile robot to prevent injury to the human.
} \label{fig:trikey_dreamer} 
\end{figure}

\section{Conclusion}
In this chapter, we extend our focus on collision management to prevent collisions between objects and people or between robots and people. To explore the collision intervention process, first, we estimate collision probabilities and compared with a designed probability threshold. Second, we generate a motion plan to intervene the likely collision with using a multi-link robot. To intervene, we assume that any body part of the robot is allowed to stop the collision. We also assume that the robot has a primary goal task using its end-effector, and utilize the redundant degrees of freedom of the robot to stop collisions when possible. We build constrained reachable volumes for all body parts and connect the corresponding postures with PRM planning, and check whether the reachable volumes overlap with the object trajectory. If any of the reachable volumes overlaps, a motion plan is generated to achieve intervention. If not, the robot searches for overlapping posture in the unconstrained reachable volume. This framework is implemented in real robotic systems to stop a multitude of collisions. Depending on the ball trajectories, different body parts are automatically chosen to prevent accidents. Also, when overlapping postures cannot be found using constrained reachable volumes, the unconstrained reachable volumes are exploited. As future extensions, fast visual sensors and time-optimal controllers need be adopted to reduce the response times. Also, the task planner needs to be enhanced to consider not only the collision probability but also the severity of collisions, and the importance of each object in the environment.

%% file: chapter-conclusion.tex
\chapter{Conclusion and future work}
In this dissertation, I have focused on various key studies in which collisions between robots and humans cannot be avoided. I have investigated i) Collision detection and motion control with walls in tight and inclined spaces, ii) Fast collision sensing and reactions against delicate objects including objects on the floor and people, iii) Preventing collisions between moving objects or people via intelligent intervention strategies, and iv) Human robot interaction with mobile platforms via sensor fusion and a simple contact language.

Some of the methods I have used include, i) Embedded system enhancement on human-centered robots to improve physical safety, ii) Time delayed robust force controllers for compliance, iii) Very fast contact sensing via floating-based multi-body dynamics and system identification, iv) Very fast reactions via admittance controllers, v) Probabilistic inference of collisions between multiple objects, vi) Safety-oriented sampling-based whole-body motion control, and vii) Logic planning for intervention to mitigate danger in human-populated environments.

The main contributions include, i) mobile bases sensing and adapting to dynamic collisions in clutered and rough terrains, ii) devising force feebdack controllers for omnidirectional mobile bases, iii) developing the first full-body contact sensing scheme for omnidirectional mobile platforms that includes all of the robot's body and its wheels, iv) conducting an extensive experimental study on collisions with human-scale mobile bases, v) enabling mobile platforms to physically interact with people via fusing visual and multi-contact recognition while inferring human body parts, and vi) having humanoid robots using any part of their bodies to stop objects from colliding against humans while performing other tasks. Let's discuss some technical accomplishments.


First, we devise external force estimators. In Chapter 1, the external force is estimated from the position error of the mobile robot against a wall. Therefore collision detection in tight and inclined spaces is investigated. In Chapter 2, torque sensors are incorporated to the drivetrain of the mobile robot to contact torques applied to the wheels of the robot. In Chapter 3, Cartesian external forces are accurately estimated via floating-base multi-body dynamics and system identification of rotor friction. Since the torque sensors are connected to the coupled drivetrain external forces can be estimated by using the whole-body equation of motion of the mobile robot. Using this force estimation, fast collision detection and reaction against delicate objects including objects on the floor and people becomes possible. In addition, with realistic assumptions, the position, direction, and magnitude of the applied external force can be effectively estimated. In Chapter 4, we investigate the capabilities of sensor fusion of 3D LIDAR and torque sensor information to reason about multi-contacts and human body parts in contact with a mobile base. This capability is exploited as contact based language for physical human-robot interaction. 

Second, due to our ability to estimate external force on mobile platforms, we investigate fast collision/contact reactive controllers. In Chapter 1, the estimated external force via position error is converted into a holonomic constraint and fed to the model as a null space projection to modify the control trajectory. Such strategy is an effective way to achieve safety without the need to perform complex optimization calculations.
In Chapter 2, a zero-force controller is devised for the mobile base to achieve compliant contact reactions. Using the zero-force controller, the mechanical passivity of the robot's actuators can be reduced resulting in faster contact dynamics. However, by removing the passivity, the system becomes more sensitive to noise and time delays which limit how the robot's responds to collisions. To improve upon this problem, stability is increased by employing disturbance observers. Although we achieve compliant interactions of mobile humanoid robots, force-feedback-based impedance controllers are not enough to achieve the fast reactions needed to mitigate collision impacts. 
A solution is explored in Chapter 3 where we study very fast collision detection and reaction of mobile robots using the drivetrain torque sensors. The estimated external force is fed to an admittance controller that effectively transforms the base into a different object with much lighter feel to contact forces. The result is a very rapid response to contact which in turn mitigates danger during collisions with the platform's body or it wheels. Finally, in Chapter 5, we perform an in-depth study on a robots that are faced with the dilemma of stopping an object from hitting a human. We investigate probabilistic inference of collisions between multiple objects and logic planning for intervention to mitigate danger in human-populated environments. We build an object tracker to observe objects around the robot, and compute collision probabilities of objects based on their current states. When the probability of collision between these objects exceeds a given threshold, the collision is physically prevented by the robot, and it does so without interrupting other tasks when possible. Through this investigation, we highlight some observations that could increase safety of human-centered robots. The first observation is when an object collides with a robot, it is practical for the robot to react moving away as quickly as possible and possibly in the opposite direction. The second observations is that if collisions between objects and people or other external objects are likely to happen, it might be beneficial for robots to stop the collisions. Finally, we observe that if a robot is reacting by moving away and there is a person or external object on the escape direction, the robot should move away in a complex pattern that avoids or mitigates future collisions.

\section{Future work}
We have conducted multiple studies about the physical interaction between robots and their environments, but there are many areas we would like to explore in the future. The collision reaction method devised in Chapter 3 is insufficient to endow safety in all collision cases. It takes 80ms for collision detection to happen and about the same time to respond to those collisions. This is not enough to minimize the damage caused by the collisions. To improve upon this, more research on the role of the structural and mechanical parts of the robot and its drivetrain should be investigated. In order to minimize the noise of the torque sensors, the robot structure should be designed so as not to be too sensitive to vibrations. In the case of the drivetrain sensing, the noise model of its components such as the harmonic drives should be investigated and applied to the external force estimator to eliminate effectssuch as torque ripples. It is also necessary to improve the time response of the actuator drivetrain including the motors, gears and belts, for faster reaction speeds. In this study, we have proposed to exploit torque sensors in the drivetrain fo the mobile platform together with a 3D LIDAR for external force estimation. The the type and number of sensors could be extended such as accelerometers, stereo cameras, and tactile sensors. Each sensor has its own advantages and disadvantages, but it is possible to implement better estimator by fusing the multiplicity of sensors. In the case of using the articulated humanoid robot of Chapter 5, the collision intervention performance can be improved by speeding up dynamic response of upper body and its manipulators. Time-optimal motion planning could be introduced for the high degree-of-freedom system. Lastly, we have investigated collisions only from the viewpoint of the robot. When we consider collisions between humans and robots, ethical and moral questions arise. It is an open question whether a robot should intervene if an object is on the path to hit a human. In addition, we need to study how much contact forces should be applied for humans not to feel uncomfortable.

%% file: chapter-appendix-lower.tex
\chapter{Whole-body operational space control}
\label{sec:TH}
The constraint dynamic equation describing free-floating dynamics has the following expression \cite{sentis2007}.
\begin{align}
A \ddot{q} + N_c^T bg = \left(U N_c\right)^T \tau
\end{align}
where $A$, $q$, $N_c$, $bg$, $U$, and $\tau$ are the mass matrix, the robot's generalized coordinates, the constraint null space projection, the coriolis-centrifugal-gravity term, the selection matrix of actuated joints, and the torque/force input to the system, respectively.
According to the WBOSC framework, the torque input for the highest priority task is computed as follows
\begin{multline}
A \ddot{q} + N_c^T bg =
\left(U N_c\right)^T \left( J_1^{*T}\Lambda_1^*\ddot{x}_{1,des} + N_1^{*\,T} \tau_2 \right)
+ \left(U N_c\right)^T \left( \overline{UN_c}^T bg\right)
\end{multline}
where $J_1^*$, $\Lambda_1^*$, $N_1^*$, $\overline{(\cdot)}$, and $\tau_2$ are the Jacobian matrix of the highest priority task, the highest priority task space mass matrix, the null space of the task, the dynamically consistent pseudoinverse of $(\cdot)$, and the torque input of lower priority tasks, respectively. If the system is fully controllable under the hard constraints, i.e. $\overline{UN_c} UN_c = N_c$, then $bg$ cancels out as follows
\begin{equation}
A \ddot{q}  = \left(U N_c\right)^T \left( J_1^{*T}\Lambda_1^*\ddot{x}_{1,des} + N_1^{*\,T} \tau_2 \right) \label{eq:dyn2}
\end{equation}
Lower priority tasks fulfill higher priority task constraints. In Eq. (\ref{eq:dyn2}), the lower priority task input $\tau_2$ is projected into the null space of the higher priority task by left-multiplying by $N_1^*$. $N_1^*$ is defined as follows
\begin{align}
N_1^* &= I - \overline{J_1^*} J_1^*
\end{align}

By left-multiplying Eq. (\ref{eq:dyn2}) by $J_2 A^{-1}$, we obtain the dynamic equation of the lower priority task as follows
\begin{align}
	\ddot{x}_2 + \dot{J}_2 \dot{q}  &= J_2 A^{-1}\left(U N_c\right)^T \left( \tau_1 + N_1^{*\,T} J_2^{*\,T}\Lambda_2^* \ddot{x}_{2,des} \right) \nonumber \\
&= \tau_1^\prime + J_2^* \Phi^* N_1^{*\,T} J_2^{*\,T} \Lambda_2^* \ddot{x}_{2,des}\nonumber\\
&= \Lambda_2^{*\,+} \Lambda_2^* \ddot{x}_{2,des} + \tau_1^\prime \label{eq:x2}
\end{align}
If the lower priority task space mass matrix, $\Lambda_2^*$ is full rank, the task is fully controllable and $\ddot{x}_2$ is equal to the desired task acceleration, $\ddot{x}_{2,des}$. Otherwise, the task acceleration is a combination of the basis vectors corresponding to the non-zero eigenvalues of $\Lambda_2^*$.

%% file: chapter-appendix1.tex
\chapter{Optimization approach for whole-body operational space control}

\section{Optimization in Operational Space Control}
A simple multibody dynamic system and its task space are described as follows.
\begin{align}
\mathbf{A} \ddot{\mathbf{q}} + \mathbf{bg} = \boldsymbol{\tau} \\
\mathbf{J} \dot{\mathbf{q}} = \dot{\mathbf{x}}
\end{align}
Then, we can generate a dynamic equation in terms of a desired task space acceleration.
\begin{align}
\mathbf{J} \ddot{\mathbf{q}} = \ddot{\mathbf{x}} - \dot{\mathbf{J}} \dot{\mathbf{q}}\\
\mathbf{J} \ddot{\mathbf{q}} +
\mathbf{J} \mathbf{A}^{-1} 
\mathbf{bg} = 
\mathbf{J} \mathbf{A}^{-1} 
\boldsymbol{\tau} \\
\mathbf{J} \mathbf{A}^{-1} 
\boldsymbol{\tau} =
\ddot{\mathbf{x}} - \dot{\mathbf{J}} \dot{\mathbf{q}} +
\mathbf{J} \mathbf{A}^{-1} 
\mathbf{bg} 
\end{align}
We can derive the torque inputs with the following performance index which minimizes torques.
\begin{align}
\min_{\boldsymbol{\tau}}\, \left\| \boldsymbol{\tau} \right\| _{\mathbf{A}^{-1}}^2 = 
\min_{\boldsymbol{\tau}} \boldsymbol{\tau}^T \mathbf{A}^{-1} \boldsymbol{\tau}
\end{align}
subject to
\begin{align}
\mathbf{J} \mathbf{A}^{-1} 
\boldsymbol{\tau} &=
\ddot{\mathbf{x}}_{des} - \dot{\mathbf{J}} \dot{\mathbf{q}} +
\mathbf{J} \mathbf{A}^{-1} 
\mathbf{bg} 
\end{align}
From the minimum norm solution for the given performance index, the following torques are obtained.
\begin{align}
\boldsymbol{\tau} = 
\mathbf{J}^T \left( \mathbf{J} \mathbf{A}^{-1} \mathbf{J}^T \right)^{-1} \left[
\ddot{\mathbf{x}}_{des} - \dot{\mathbf{J}} \dot{\mathbf{q}} +
\mathbf{J} \mathbf{A}^{-1} 
\mathbf{bg} \right]
\end{align}

\section{Optimization in Whole-Body Control}
A whole-body control consists of two optimization processes. One process minimizes the weighted torques on all of the joints with respect to a given task, and the other process minimizes the error of the constrained torques. Whole-body control includes unactuated joints which are usually used to represent a free-floating body or biarticular transmissions.

\subsection{Minimizing All Joint Torques}
The constrained dynamics resulting from a hard constraint, $\mathbf{J}_c\dot{\mathbf{q}} = \mathbf{0}$ result in the following equation.
\begin{align}
\mathbf{A} \ddot{\mathbf{q}} + \mathbf{N}_c^T \mathbf{bg} = \left(\mathbf{U\,N}_c\right)^T \boldsymbol{\tau} \label{eq:const}
\end{align}
where 
\begin{align}
\mathbf{N}_c \triangleq \mathbf{I} - \overline{\mathbf{J}}_c\mathbf{J}_c
\end{align}
We can define a virtual input $\boldsymbol{\tau}^\prime$ which applies torque to all of the joints including the unactuated ones. Then, the dynamic equation becomes
\begin{align}
\mathbf{A} \ddot{\mathbf{q}} + \mathbf{N}_c^T \mathbf{bg} = \mathbf{N}_c^T \mathbf{\tau}^\prime 
\end{align}.
The task space can be expressed using the task Jacobian, $\mathbf{J}$, which has the following property.
\begin{align}
\mathbf{J} \dot{\mathbf{q}} = \dot{\mathbf{x}}
\end{align}
Then, we can express a desired task space acceleration, $\ddot{\mathbf{x}}_{des}$ as follows.
\begin{align}
\ddot{\mathbf{x}}_{des} = \mathbf{J} \ddot{\mathbf{q}}_{des} - \dot{\mathbf{J}} \dot{\mathbf{q}}
\end{align}
The constrained dynamics can incorporated the desired task space acceleration as follows.
\begin{align}
\ddot{\mathbf{x}}_{des} - \dot{\mathbf{J}} \dot{\mathbf{q}} + \mathbf{J\,A}^{-1}\mathbf{N}_c^{T} \mathbf{bg} = 
\mathbf{J\,A}^{-1}\mathbf{N}_c^{T} \boldsymbol{\tau}^\prime
\end{align}

If the coefficient of the input $\boldsymbol{\tau}^\prime$ has full row rank, we can satisfy the desired task space acceleration. To specify the input, we can derive a minimum norm solution with the following performance index.
\begin{align}
\min_{\boldsymbol{\tau}^\prime}\, \left\| \boldsymbol{\tau}^\prime\right\|_{\mathbf{A}^{-1}}^2 = 
\min_{\boldsymbol{\tau}^\prime} \boldsymbol{\tau}^{\prime\,T} \mathbf{A}^{-1} \boldsymbol{\tau}^\prime
\end{align}
Then, the input minimizing the performance index can be obtained as follows.
\begin{align}
\boldsymbol{\tau}^\prime &= 
\mathbf{N}_c^T \mathbf{J}^T \left( \mathbf{J} \mathbf{N}_c \mathbf{A}^{-1} \mathbf{N}_c^T \mathbf{J}^T\right)^{-1}
\left( \ddot{\mathbf{x}}_{des} + \mathbf{J} \mathbf{A}^{-1} \mathbf{N}_c^T \mathbf{bg}  
- \dot{\mathbf{J}} \dot{\mathbf{q}} \right) \\
&=
\mathbf{N}_c^T \mathbf{J}^T \left( \mathbf{J} \mathbf{N}_c \mathbf{A}^{-1} \mathbf{N}_c^T \mathbf{J}^T\right)^{-1}
\ddot{\mathbf{x}}_{cmd}\label{taup}
\end{align}
where
\begin{align}
\ddot{\mathbf{x}}_{cmd}
\triangleq
\ddot{\mathbf{x}}_{des} + \mathbf{J} \mathbf{A}^{-1} \mathbf{N}_c^T \mathbf{bg}  
- \dot{\mathbf{J}} \dot{\mathbf{q}} 
\end{align}

\subsection{Minimizing the Underactuated Torque Error}
So far, we have derived the minimum torque values for a desired task space acceleration that complies with the hard constraints. This torque input includes the unactuated joints, which are coupled with the actuated joints via the hard constraints. Thus, we need to determine feasible constrained joint torques as close as possible to the optimized input $\tau^\prime$. This can be done via the following optimization process.
\begin{align}
\min_{\tau} J = \min_{\tau}\, \left\| \mathbf{N}_c^T \tau^\prime -  \left(\mathbf{U\, N}_c\right)^T \tau \right\|_{\mathbf{A}^{-1}}^2 
\end{align}
Then, the first derivative of the performance index with respect to the input, $\tau$ can be derived as follows, needing to be zero to make the performance index minimal.
\begin{align}
\frac{\partial J}{\partial \tau} = 2 \tau^T \mathbf{U\, N}_c \mathbf{A}^{-1} \left( \mathbf{U\, N}_c \right)^T - 2 \tau^{\prime\, T} \mathbf{N}_c \mathbf{A}^{-1} \left( \mathbf{U\, N}_c \right)^T = 0
\end{align}
If the coefficient of $\tau$, $\mathbf{U\, N}_c \mathbf{A}^{-1} \left( \mathbf{U\, N}_c \right)^T$ has full rank, the actuation torque can be derived as follows.
\begin{align}
\tau &= \left\{\mathbf{U\, N}_c \mathbf{A}^{-1} \left( \mathbf{U\, N}_c \right)^T\right\}^{-1}
\mathbf{U\, N}_c \mathbf{A}^{-1} \mathbf{N}_c^T \tau^{\prime} \\
 &= \left\{\mathbf{U\, N}_c \mathbf{A}^{-1} \left( \mathbf{U\, N}_c \right)^T\right\}^{-1}
\mathbf{U\, N}_c \mathbf{A}^{-1} \tau^{\prime}\label{tau1}
\end{align}
By substituting $\tau^\prime$ of Eq. (\ref{taup}) into Eq. (\ref{tau1}), we can derive the input to the system as follows.
\begin{align}
\tau &= \left\{\mathbf{U\, N}_c \mathbf{A}^{-1} \left( \mathbf{U\, N}_c \right)^T\right\}^{-1}
\mathbf{U}
\underbrace{\mathbf{N}_c \mathbf{A}^{-1} \mathbf{N}_c^T}_
{\mathbf{N}_c \mathbf{A}^{-1} }
\mathbf{J}^T \left( \mathbf{J} \mathbf{N}_c \mathbf{A}^{-1} \mathbf{N}_c^T \mathbf{J}^T\right)^{-1} 
\ddot{\mathbf{x}}_{cmd}
	\\
 &= \underbrace{\left\{\mathbf{U\, N}_c \mathbf{A}^{-1} \left( \mathbf{U\, N}_c \right)^T\right\}^{-1}
\mathbf{U\, N}_c \mathbf{A}^{-1}
 }_{\overline{\mathbf{U\,N}_c}^T}
\mathbf{J}^T 
\underbrace{\left( \mathbf{J} \mathbf{N}_c \mathbf{A}^{-1} \mathbf{N}_c^T \mathbf{J}^T\right)^{-1}}_
{\mathbf{\Lambda}^*}
\ddot{\mathbf{x}}_{cmd} \\
 &= 
\underbrace{\overline{\mathbf{U\, N}_c}^T\,
\mathbf{J}^T}_{\mathbf{J}^{*\,T}} 
\mathbf{\Lambda}^*
\ddot{\mathbf{x}}_{cmd} \\
&= \mathbf{J}^{*\,T} \mathbf{\Lambda}^* 
\ddot{\mathbf{x}}_{cmd} 
\label{tau2}
\end{align}
In this derivation, we assume that (1) $\mathbf{J\,N}_c$ has full row rank, which means the task can be achieved regardless of the hard constraints $\mathbf{N}_c$; and (2) 
$\mathbf{U\, N}_c \mathbf{A}^{-1} \left( \mathbf{U\, N}_c \right)^T$ has full rank, which means the system is fullyu controllable within the constrained dynamics.


\subsection{Alternative Method for Minimizing Underactuated Torque Error}
In the previous section, the torques minimizing the error between the virtual and the real actuators is obtaine under some conditions that may not be satisfied. In this section, the torque is derived using an SVD decomposition, whith does not require $\mathbf{U\, N}_c \mathbf{A}^{-1} \left( \mathbf{U\, N}_c \right)^T$ to be full rank. The performance index which minimizes the error is explored again.
\begin{align}
\min_{\boldsymbol{\tau}} J 
\end{align}
where 
\begin{align}
J &= 
\left\| \mathbf{N}_c^T \boldsymbol{\tau}^\prime -  \left(\mathbf{U\, N}_c\right)^T \boldsymbol{\tau} \right\|_{\mathbf{A}^{-1}}^2  \\
&= 
\left( \mathbf{N}_c^T \boldsymbol{\tau}^\prime -  \left(\mathbf{U\, N}_c\right)^T \boldsymbol{\tau} \right)^T \mathbf{A}^{-1} 
\left( \mathbf{N}_c^T \boldsymbol{\tau}^\prime -  \left(\mathbf{U\, N}_c\right)^T \boldsymbol{\tau} \right)^T \\
&= 
\left\| \mathbf{A}^{-1/2} \mathbf{N}_c^T \boldsymbol{\tau}^\prime -  
\mathbf{A}^{-1/2} \left(\mathbf{U\, N}_c\right)^T \boldsymbol{\tau} \right\|^2 
\end{align}
By taking the singular value decomposition on $\mathbf{A}^{-1/2} \left(\mathbf{U\, N}_c\right)^T$ as $\mathbf{U \Sigma V}^T$ and defining a new vector, $\mathbf{b} \triangleq 
\mathbf{A}^{-1/2} \mathbf{N}_c^T \boldsymbol{\tau}^\prime$ and $\mathbf{t} \triangleq \mathbf{V}^T\boldsymbol{\tau}$, the performance index can be simplified as follows.  
\begin{align}
J &= \left\| \mathbf{U\Sigma\,t}- \mathbf{b} \right\|^2 \\
&= \left\| \mathbf{\Sigma \, t} - \mathbf{U}^T \mathbf{b} \right\|^2\\ 
&= \sum_{i=1}^{n} \left( \sigma_i t_i - \mathbf{u}_i^T \mathbf{b} \right)^2 
\end{align}
where $\sigma_i$, $t_i$, and $\mathbf{u}_i$ are the $\left(i,i\right)$-th element of $\mathbf{\Sigma}$, i-th element of $\mathbf{t}$, and i-th singular vector of $\mathbf{U}$, respectively. If the rank of $\mathbf{U \Sigma V}^T$ is $r$, $\sigma_{r+1}\, \cdots, \sigma_n$ are equal to zero, then the above summation can be written as follows. 
\begin{equation}
J = \sum_{i=1}^{r} \left( \sigma_i t_i - \mathbf{u}_i^T \mathbf{b} \right)^2 +
 \sum_{i=r+1}^{n} \left( \mathbf{u}_i^T \mathbf{b} \right)^2 
\end{equation}
To minimize $J$, the following condition should hold.
\begin{align}
t_i &= \mathbf{v}_i^T \boldsymbol{\tau} \\
&= 
\left\{
\begin{matrix}
\frac{1}{\sigma_i}{\mathbf{u}_i^T\,\mathbf{b}} \, & i \le r \\
	0\, & i > r
\end{matrix}
\right.\\
\mathbf{t} &= \mathbf{V}^T \boldsymbol{\tau} \\
	&= 
	\begin{pmatrix}
	1/\sigma_1 & 0 & \cdots & 0\\
		0 & 1/\sigma_2 & \cdots & 0\\
		\vdots & \ddots & \vdots & 0 \\
		0 & 0 & \cdots &  1/\sigma_n\ \rm or \ 0
	\end{pmatrix}
	\begin{pmatrix}
	\mathbf{u}_1^T \mathbf{b} \\
	\mathbf{u}_2^T \mathbf{b} \\
	\cdots\\
	\mathbf{u}_n^T \mathbf{b}
	\end{pmatrix} \\
&= \mathbf{\Sigma}^{+} \mathbf{U}^T\mathbf{b}
\end{align}
Therefore, we can obtain the torque input which minimizes the given performance index as follows.
\begin{align}
\boldsymbol{\tau} &= \mathbf{V} \mathbf{\Sigma}^{+} \mathbf{U}^T \mathbf{b} \\
&= \left(\mathbf{U\Sigma V}^T\right)^+\mathbf{b}\\
&= \left(\mathbf{A}^{-1/2} \left(\mathbf{U\, N}_c\right)^T \right)^+ 
\mathbf{A}^{-1/2} \mathbf{N}_c^T \boldsymbol{\tau}^\prime
\end{align}
Also, $\boldsymbol{\tau}$ is also equivalent to the following expression.
\begin{align}
\boldsymbol{\tau} &= \mathbf{V} \mathbf{\Sigma}^{+} \mathbf{U}^T \mathbf{b} \\
&= \mathbf{V} \mathbf{\Sigma}^{+\, 2} \mathbf{V}^T \mathbf{V} \mathbf{\Sigma} \mathbf{U}^T\mathbf{b} \\
&= \left(\mathbf{V} \mathbf{\Sigma}^2 \mathbf{V}^T\right)^+ \mathbf{V} \mathbf{\Sigma} \mathbf{U}^T\mathbf{b} \\
&= \left(\mathbf{V} 
\mathbf{\Sigma}
\mathbf{U}^T
\mathbf{U}
\mathbf{\Sigma}
\mathbf{V}^T\right)^+ \mathbf{V} \mathbf{\Sigma} \mathbf{U}^T \mathbf{b} \\
&= \underbrace{\left(\mathbf{U\, N}_c \mathbf{A}^{-1} \left( \mathbf{U\, N}_c \right)^T\right)^+
\mathbf{U\, N}_c 
\mathbf{A}^{-1/2}
\mathbf{A}^{-1/2} \mathbf{N}_c^T}_{\overline{\mathbf{UN}_c}}
\tau^{\prime} \\
&=
\overline{\mathbf{UN}_c}
\tau^{\prime} 
\end{align}
This optimization method needs no conditions, so $\mathbf{U\, N}_c \mathbf{A}^{-1} \left( \mathbf{U\, N}_c \right)^T$ can be singular.

\subsection{Substituting the Optimal Solution}
When the optimal solution is substituted into the original constrained dynamics of Eq. (\ref{eq:const}), the equation simplifies as follows.
\begin{align}
\mathbf{A} \ddot{\mathbf{q}} + \mathbf{N}_c^T \mathbf{bg} &= 
\left(\mathbf{U\,N}_c\right)^T 
\mathbf{J}^{*\,T} \mathbf{\Lambda}^* 
\ddot{\mathbf{x}}_{cmd}  \\
&= \left(\mathbf{U\,N}_c\right)^T 
\underbrace{\overline{\mathbf{U\, N}_c}^T\,
\mathbf{J}^T}_{\mathbf{J}^{*\,T}} 
\mathbf{\Lambda}^*
\ddot{\mathbf{x}}_{cmd}  \\
&= 
\left(\overline{\mathbf{U\, N}_c}\,
\mathbf{U\,N}_c\right)^T 
\mathbf{J}^T
\mathbf{\Lambda}^*
\ddot{\mathbf{x}}_{cmd}
\end{align}
If the rank of $\mathbf{N_c}$ and $\mathbf{UN_c}$ are equal to each other and smaller than or equal to that of $\mathbf U$, the right-hand-side of the above equation simplifies as follows.
\begin{equation} 
= 
\mathbf{N_c}^T
\mathbf{J}^T
\mathbf{\Lambda}^*
\ddot{\mathbf{x}}_{cmd}  
\end{equation}

To investigate the acceleration in the operational space, we can left-multiply $\mathbf{J A}^{-1}$ to replace $\ddot{\mathbf{q}}$ with $\ddot{\mathbf{x}}$ as follows.
\begin{align}
\underbrace{\mathbf{J A}^{-1}
\mathbf{A} \ddot{\mathbf{q}}}
_{
\ddot{\mathbf{x}} - \dot{\mathbf{J}} \dot{\mathbf{q}}
} + 
\mathbf{J A}^{-1}
\mathbf{N}_c^T \mathbf{bg} 
&= 
\mathbf{J} 
\underbrace{
\mathbf{A}^{-1}
\mathbf{N_c}^T}
_
{
\mathbf{N_c}
\mathbf{A}^{-1}
\mathbf{N_c}^T
}
\mathbf{J}^T
\mathbf{\Lambda}^*
\ddot{\mathbf{x}}_{cmd} 
\end{align}
If $\mathbf{J N}_c$ is full rank, then $\mathbf{\Lambda}^{*\, +} \mathbf{\Lambda}^*$ cancells out, and the dynamic equation further simplifies as follows.
\begin{align}
\ddot{\mathbf{x}} - \dot{\mathbf{J}} \dot{\mathbf{q}}
+ 
\mathbf{J A}^{-1}
\mathbf{N}_c^T \mathbf{bg} 
&=
\underbrace{
\mathbf{J} 
\mathbf{N_c}
\mathbf{A}^{-1}
\mathbf{N_c}^T
\mathbf{J}^T
}_{\cancel{\mathbf{\Lambda}^{*\, +}}}
\cancel{\mathbf{\Lambda}^*}
\ddot{\mathbf{x}}_{cmd}  \\
\ddot{\mathbf{x}} - \cancel{\dot{\mathbf{J}} \dot{\mathbf{q}}}
+ 
\cancel{\mathbf{J A}^{-1}
\mathbf{N}_c^T \mathbf{bg}} 
&=
\ddot{\mathbf{x}}_{des} + \cancel{\mathbf{J} \mathbf{A}^{-1} \mathbf{N}_c^T \mathbf{bg}}- \cancel{\dot{\mathbf{J}} \dot{\mathbf{q}}}
\end{align}
Then, we can directly control the task space acceleration accounting for hard constraints and dynamic other disturbances.

%% file: chapter-appendix2.tex
\chapter{Proof of $\overline{UN_c} \, UN_c = N_c$ }
\section{Problem Description}
The following equality will be proved.
\begin{align}
\overline{UN_c} \, UN_c &= N_c 
\end{align}
To satisfy the equation above, the rank of $N_c$ and $UN_c$ need to be equal, while being smaller or equal to the rank of $U$.
\begin{align}
Rank(N_c) = Rank(UN_c) \le Rank(U)
\end{align}
By the definition of $\overline{UN_c}$, the first equation can be expanded to be.
\begin{align}
A^{-1}\left(UN_c\right)^{T} \left( UN_c A^{-1} \left(UN_c\right)^T \right)^+ UN_c &= N_c
\end{align}
where $(\cdot)^+$ is the Moore-Penrose pseudoinverse. Because $A$ is full-rank, $A^{-1}$ can right-mutiplied the above equation.
\begin{align}
A^{-1}\left(UN_c\right)^{T} \left( UN_c A^{-1} \left(UN_c\right)^T \right)^+ UN_c A^{-1} &= N_c A^{-1} \\
N_c A^{-1} U^{T} \left( U N_c A^{-1} U^T \right)^+ UN_c A^{-1} &= N_c A^{-1} 
\end{align}
By defining a new matrix, $X \triangleq N_c A^{-1} = A^{-1} N_c^T \in \mathbb{R}^{n\times n}$ and 
$S \triangleq \begin{pmatrix} I_{m \times m} & 0_{m \times (n-m)} \end{pmatrix} \in \mathbb{R}^{m \times n}$,
the above equation takes the more general form.
\begin{align}
X S^{T} \left( S X S^T \right)^+ S X &= X
\end{align}
with the following conditions.
\begin{align}
&Rank\left(N_c\right) = Rank\left(N_c A^{-1}\right) \nonumber\\
=& Rank\left(X\right) \nonumber\\
=&Rank\left(UN_c\right) = Rank\left(UN_c A^{-1/2}\right) = Rank\left(UN_c A^{-1/2} \left(UN_c A^{-1/2}\right)^T\right) \nonumber\\
=& Rank\left( X S X^T\right) \nonumber\\
\le& Rank\left(S\right)
\end{align}
We can define $S$ as a rectanglar matrix with 1s on its diagonal without loss of generality.

\section{General Symmetric Matrix Proof}
The following equation will be proved.
\begin{align}
 &X S^{T} \left( S X S^T \right)^+ S X  = X
\label{eq:lhs-one}
\end{align}
where $X$ and $S$ are defined as matrices with block matrices as follows.
\begin{align}
X &= X^T = 
\begin{pmatrix}
X_{11} & X_{12} \\
X_{12}^T & X_{22}
\end{pmatrix} \\
S &= \begin{pmatrix} I_{m \times m} & 0_{m \times (n-m)} \end{pmatrix} \in \mathbb{R}^{m \times n}
\end{align}
where $X_{11} \in \mathbb{R}^{m \times m}$, $X_{12} \in \mathbb{R}^{m \times (n-m)}$, 
$X_{22} \in \mathbb{R}^{(n-m) \times (n-m)}$, and the ranks of the matrices are conditioned as follows. 
\begin{align}
		Rank(X) = Rank(SXS^T) = Rank(X_{11}) = r \le Rank(U) = m
\end{align}
\begin{figure}\centering
\includegraphics[width=0.7\linewidth, clip=false ] {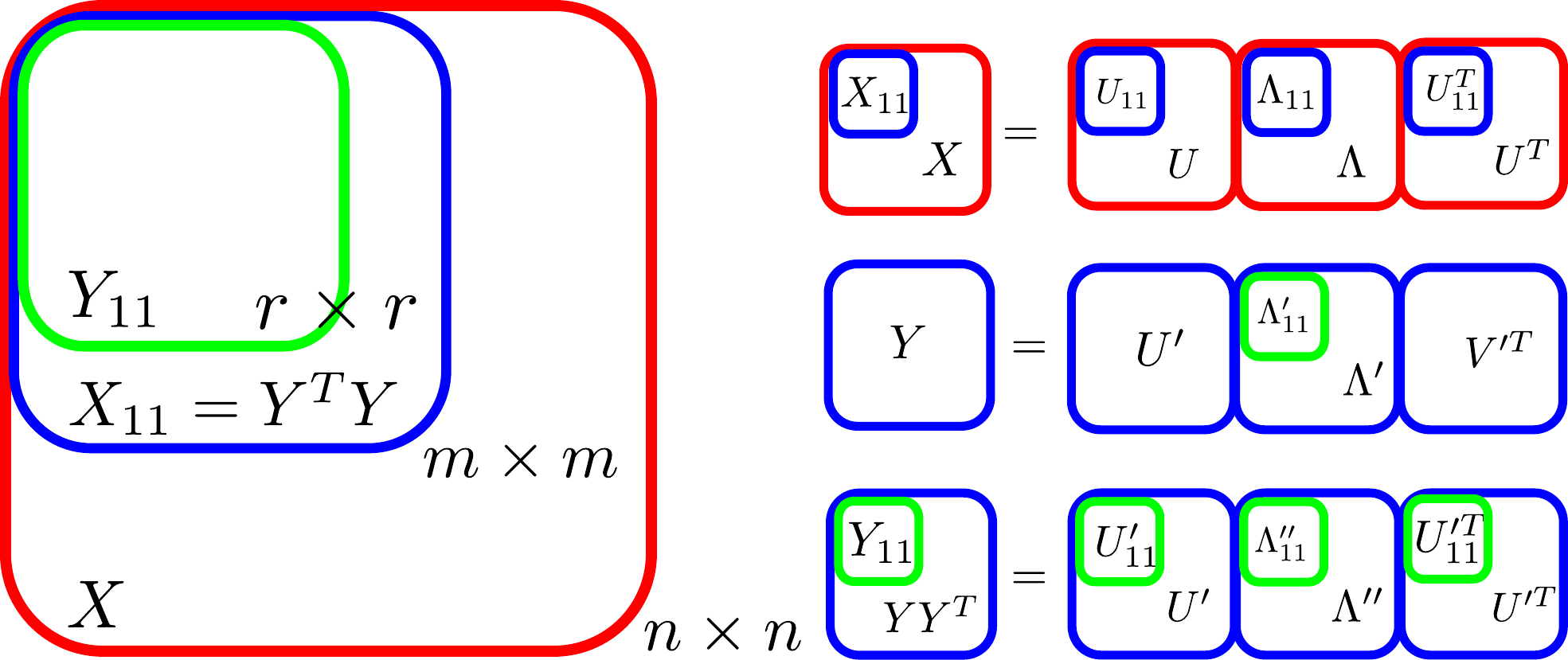}\label{fig:mtx}
\caption{{\bf Visualization of the matrices }}
\end{figure}
Then, the left hand side of the eq. (\ref{eq:lhs-one}) can be expressed as follows.
\begin{align}
 &X S^{T} \left( S X S^T \right)^+ S X  \nonumber\\
=&
\begin{pmatrix}
X_{11} \\ X_{12}^T
\end{pmatrix}
\left( X_{11} \right)^{+} 
\begin{pmatrix}
X_{11} & X_{12}
\end{pmatrix} \nonumber\\
=&
\begin{pmatrix}
X_{11} X_{11}^{+} X_{11} & 
X_{11} X_{11}^{+} X_{12} \\
X_{12}^T X_{11}^{+} X_{11} &
X_{12}^T X_{11}^{+} X_{12}
\end{pmatrix} \nonumber\\
=&
\begin{pmatrix}
X_{11} &
X_{11} X_{11}^{+} X_{12} \\
\left(X_{11} X_{11}^{+} X_{12}\right)^T &
X_{12}^T X_{11}^{+} X_{12}
\end{pmatrix} \label{eq:stat}
\end{align}

To prove eq. (\ref{eq:lhs-one}), the following equations should be proved.
\begin{align}
X_{11} X_{11}^{+} X_{12}  &= X_{12} \\
X_{12}^T X_{11}^{+} X_{12} &= X_{22}
\end{align}
To analyze these terms, $X$ needs to be broken down into its singular value decomposition (SVD) as follows.
\begin{align}
X 
&= 
U \Lambda U^T \nonumber\\
&=
\begin{pmatrix}
U_{11} & U_{12} \\
U_{21} & U_{22}
\end{pmatrix}
\begin{pmatrix}
\Lambda_{11} & 0 \\
0 & 0
\end{pmatrix}
\begin{pmatrix}
U_{11} & U_{12} \\
U_{21} & U_{22}
\end{pmatrix}^T \nonumber\\
&=
\begin{pmatrix}
U_{11}\Lambda_{11} U_{11}^T & 
U_{11}\Lambda_{11} U_{21}^T \\
U_{21}\Lambda_{11} U_{11}^T & 
U_{21}\Lambda_{11} U_{21}^T
\end{pmatrix} \nonumber\\
&= 
\begin{pmatrix}
X_{11} & X_{12} \\
X_{12}^T & X_{22}
\end{pmatrix} 
\end{align}
where $U_{11} \in \mathbb{R}^{m \times m}$, 
	  $U_{12} \in \mathbb{R}^{m \times (n-m)}$, 
	  $U_{21} \in \mathbb{R}^{(n-m) \times m}$, 
	  $U_{22} \in \mathbb{R}^{(n-m) \times (n-m)}$, and $\Lambda_{11} \in \mathbb{R}^{m \times m}$.
Because the rank of $X$ is less than or equal to $m$, all the non-zero singular values can be contained in the $m \times m$ matrix, and $\Lambda_{11}$ may not be full rank.

We need to analyze the term, $X_{11}^+$ of Eq. (\ref{eq:stat}). Let us define a new matrix, $Y$ as follows.
\begin{align}
Y &\triangleq \Lambda_{11} ^{1/2} U_{11}^T \in \mathbb{R}^{m \times m}\\
\intertext{which is a half of $X_{11}$ as follows.}
X_{11} &= U_{11} \Lambda_{11} U_{11}^T  = Y^{T} Y 
\end{align}
We can also decompos $Y$ into its SVD, and $X_{11}$ and $X_{11}^+$ can be expressed with the singular values and the singular vectors as follows.
\begin{align}
Y &= U^\prime \Lambda^\prime V^{\prime\, T} \\
X_{11} &= Y^T Y = V^\prime \Lambda^\prime U^{\prime\, T}
 U^\prime \Lambda^\prime V^{\prime\, T} \nonumber\\
&= V^\prime \Lambda^{\prime 2} V^{\prime\, T} \\
X_{11}^+ 
&= V^\prime \Lambda^{\prime + 2} V^{\prime\, T} 
\end{align}
where $U^\prime, \ V^\prime,\ \Lambda^\prime \in \mathbb{R}^{m \times m}$, and $\Lambda^{\prime\,+}$ is formed by replacing all the non-zero elements of $\Lambda^\prime$ with their multiplicative inverses. Also, the following rank conditions hold.
\begin{align}
Rank(X_{11}) = Rank(Y^T Y) = Rank(Y) = Rank(Y Y^T) = r
\end{align}
By substituting $Y$, $X_{11} X_{11}^{+} X_{12}$ can be expressed as follows.
\begin{align}
&X_{11} X_{11}^{+} X_{12}  \nonumber\\
=&
U_{11}\Lambda U_{11}^T 
\left(U_{11}\Lambda U_{11}^T \right)^{+}
U_{11}\Lambda U_{21}^T  \nonumber\\
=&
U_{11}\Lambda^{1/2} Y 
\left(Y^T Y \right)^+ 
Y^T \Lambda^{1/2} U_{21}^T  \nonumber\\
=& U_{11}\Lambda ^{1/2}
U^\prime \Lambda^\prime \cancel{V^{\prime\, T}} 
\left(\cancel{V^\prime} \Lambda^{\prime + 2} \cancel{V^{\prime\, T}}  \right) 
\cancel{V^\prime} \Lambda^\prime U^{\prime\, T} 
\Lambda ^{1/2}U_{21}^T
\end{align}
Because the rank of $\Lambda^\prime$ is $r$, 
$\Lambda^\prime \Lambda^{\prime + 2} \Lambda^\prime$ embeds an identity matrix with the rank of $r$ such that the above equation becomes
\begin{align}
=& U_{11}\Lambda ^{1/2}
\underbrace{
U^\prime }_
{
\begin{pmatrix}
U_{11}^\prime & 0 \\ 0 & U_{22}^\prime
\end{pmatrix}
}
\underbrace{
\Lambda^\prime 
\Lambda^{\prime + 2} 
\Lambda^\prime
}_
{
\begin{pmatrix}
I_{r\times r} & 0 \\ 0 & 0
\end{pmatrix}
}
\underbrace{
U^{\prime\,T} }
_{
\begin{pmatrix}
U_{11}^{\prime\,T} & 0 \\ 0 & U_{22}^{\prime\,T}
\end{pmatrix}
}
\Lambda ^{1/2}U_{21}^T  \nonumber\\
=& U_{11}
\begin{pmatrix}
\Lambda_{11}^{1/2} & 0 \\ 0 & 0
\end{pmatrix}
\begin{pmatrix}
U_{11}^\prime U_{11}^{\prime\,T} & 0 \\ 0 & 0
\end{pmatrix}
\begin{pmatrix}
\Lambda_{11}^{1/2} & 0 \\ 0 & 0
\end{pmatrix}
U_{21}^T  
\end{align}
$U^\prime$ is an orthogonal matrix, but it is not guaranteed that $U_{11}^{\prime}$ which is an upper left corner block matrix is also an orthogonal matrix. Therefore, it is necessary to prove that $U_{11}^\prime$ is an orthogonal matrix. $U^\prime$ has left singular vectors, which are also the singular vectors of $Y Y^T$. $Y Y^T$ is a matrix where all the non-zero components are located in the $r \times r$ left upper block as follows.
\begin{align}
YY^T &= \left(
\Lambda ^{1/2} U_{11}^T \right)
\left(
U_{11} 
\Lambda ^{1/2} 
\right) \nonumber\\
&= 
\begin{pmatrix}
\Lambda_{11}^{1/2} & 0 \\ 0 & 0
\end{pmatrix}
U_{11}^T
U_{11} 
\begin{pmatrix}
\Lambda_{11}^{1/2} & 0 \\ 0 & 0
\end{pmatrix} \nonumber\\
&=
\begin{pmatrix}
Y_{11} & 0 \\ 0 & 0
\end{pmatrix} \label{eq:yyt1}
\end{align}
where $\Lambda_{11}, \ Y_{11} \in \mathbb{R}^{r \times r}$, and $\Lambda_{11}$ is full rank ($Rank(\Lambda_{11}) = r$). Because the rank of $Y$ is $r$, the ranks of $YY^T$ and $Y_{11}$ are $r$, which means $Y_{11}$ is full rank. The singular values of $Y Y^T$ can also be derived, and the singular vectors, $U^\prime$ are equal to the left singular vectors of $Y$.
\begin{align}
YY^T &=
U^\prime \Lambda^{\prime\prime} U^{\prime\, T} \nonumber\\
&= 
\begin{pmatrix}
U_{11}^\prime & U_{12}^\prime \\
U_{21}^\prime & U_{22}^\prime
\end{pmatrix}
\begin{pmatrix}
\Lambda_{11}^{\prime\prime}& 0 \\
0 & 0
\end{pmatrix}
\begin{pmatrix}
U_{11}^\prime & U_{12}^\prime \\
U_{21}^\prime & U_{22}^\prime
\end{pmatrix}^T  \nonumber\\
&=
\begin{pmatrix}
U_{11}^\prime\Lambda_{11}^{\prime\prime} U_{11}^{\prime\,T} & 
U_{11}^\prime\Lambda_{11}^{\prime\prime} U_{21}^{\prime\,T} \\
U_{21}^\prime\Lambda_{11}^{\prime\prime} U_{11}^{\prime\,T} & 
U_{21}^\prime\Lambda_{11}^{\prime\prime} U_{21}^{\prime\,T}
\end{pmatrix}\label{eq:yyt2}
\end{align}
where $U_{11}^\prime \in \mathbb{R}^{r \times r}$, $U_{12}^\prime \in \mathbb{R}^{r \times (m-r)}$, $U_{22}^\prime \in \mathbb{R}^{(m-r) \times (m-r)}$, and $\Lambda_{11}^{\prime \prime} \in \mathbb{R}^{r \times r}$.
Because the rank of $UX$ is the same to that of $X$, $\Lambda_{11}^{\prime \prime}$ is also full rank. From Eqs. (\ref{eq:yyt1}) and (\ref{eq:yyt2}), the following condition can be derived.
\begin{align}
U_{11}^\prime\Lambda_{11}^{\prime\prime} U_{21}^{\prime\,T} &= 0 \label{eq:u21}
\end{align}
$Y_{11} = U_{11}^\prime\Lambda_{11}^{\prime\prime} U_{11}^{\prime\,T}$, so both $U_{11}^\prime$ and $\Lambda_{11}^{\prime \prime}$ are full rank. Therefore, to satisfy Eq. (\ref{eq:u21}), $U_{21}^\prime$ is zero.

By the definition of the singular vectors of the SVD, $U^\prime$ is an orthogonal matrix which satisfies the following condition.
\begin{align}
U^\prime U^{\prime\,T} =&
\begin{pmatrix}
U_{11}^\prime & U_{12}^\prime \\
0 & U_{22}^\prime
\end{pmatrix}
\begin{pmatrix}
U_{11}^\prime & U_{12}^\prime \\
0 & U_{22}^\prime
\end{pmatrix}^T \nonumber\\
=&
\begin{pmatrix}
U_{11}^\prime U_{11}^{\prime\,T} +
U_{12}^\prime U_{12}^{\prime\,T} \label{eq:u11p}
& 
U_{12}^\prime U_{22}^{\prime\,T} \\
U_{22}^\prime U_{12}^{\prime\,T} & 
U_{22}^\prime U_{22}^{\prime\,T}
\end{pmatrix}
\nonumber\\
=&
\begin{pmatrix}
I & 0 \\ 0 & I
\end{pmatrix}
\end{align}
From that condition, the following equations can be obtained.
\begin{align}
U_{12}^\prime U_{22}^{\prime\,T} = 0\label{eq:u12p}\\
U_{22}^\prime U_{22}^{\prime\,T} = I
\end{align}
Because $U_{22}^\prime$ is a full rank, $U_{12}^\prime$ should be zero to satisfy Eq. (\ref{eq:u12p}). Then, by substituting $U_{12}^{\prime} = 0$  into Eq. (\ref{eq:u11p}), the following equation can be derived.
\begin{align}
U_{11}^\prime U_{11}^{\prime\,T} &= I
\end{align}

Also, $\Lambda_{11}$ and $U_{11}^\prime$ have the same rank and $U_{11}^\prime U_{11}^{\prime\,T} = I$, so $\Lambda^{1/2}$ terms can be combined as follows.
\begin{align}
&X_{11} X_{11}^{+} X_{12}  \nonumber\\
=& U_{11}
\begin{pmatrix}
\Lambda_{11}^{1/2} & 0 \\ 0 & 0
\end{pmatrix}
\begin{pmatrix}
U_{11}^\prime U_{11}^{\prime\,T} & 0 \\ 0 & 0
\end{pmatrix}
\begin{pmatrix}
\Lambda_{11}^{1/2} & 0 \\ 0 & 0
\end{pmatrix}
U_{21}^T   \nonumber\\
=& U_{11}
\begin{pmatrix}
\Lambda_{11} & 0 \\ 0 & 0
\end{pmatrix}
U_{21}^T  \nonumber\\
=& U_{11} \Lambda U_{21}^T  \nonumber\\
=& X_{12}
\end{align}
In the same manner, it can be shown that $X_{12}^T X_{11}^{+} X_{12}$ is equal to $X_{22}$ as follows.
\begin{align}
&X_{12}^T X_{11}^{+} X_{12}  \nonumber\\
=&
U_{21}\Lambda U_{11}^T 
\left(V^\prime \Lambda^{\prime + 2} V^{\prime\, T}  \right) 
U_{11}\Lambda U_{21}^T  \nonumber\\
=& U_{21}\Lambda ^{1/2}
U^\prime \Lambda^\prime \cancel{V^{\prime\, T}} 
\left(\cancel{V^\prime} \Lambda^{\prime + 2} \cancel{V^{\prime\, T}}  \right) 
\cancel{V^\prime} \Lambda^\prime U^{\prime\, T} 
\Lambda ^{1/2}U_{21}^T  \nonumber\\
=& U_{21}\Lambda ^{1/2}
U^\prime \Lambda^\prime 
\Lambda^{\prime + 2} 
\Lambda^\prime U^{\prime\, T} 
\Lambda ^{1/2}U_{21}^T  \nonumber\\
=& U_{21}\Lambda U_{21}^T  \nonumber\\
=& X_{22}
\end{align}
Finally, by substituting the previous results into Eq. (\ref{eq:stat}), we can prove equation eq. \ref({eq:lhs-one}) as follows.
\begin{align}
 &X S^{T} \left( S X S^T \right)^+ S X  \nonumber\\
=& 
\begin{pmatrix}
X_{11} & X_{12} \\
X_{12}^T & X_{22}
\end{pmatrix} \nonumber\\
=&
X
\end{align}

%% file: chapter-appendix3.tex
\chapter{Commutation of task null-space projections}
\section{Prioritized task hierarchy}
In highly redundant robots, the degrees of freedom for the operational tasks are smaller than the degrees of freedom of the entire robotic system. Therefore, the remaining redundancy can be used for lower priority tasks. In the whole-body control framework, lower priotity tasks operate in the null space of higher priority tasks. For example, if there are two tasks, and task 1 has higher priority than task 2, we express this as follows.
\begin{align}
	A \ddot{q} + N_c^T bg = \left(UN_c \right) \left( \tau_1 + N_{prec(2)}^{T} \tau_2 \right)
\end{align}
where $N_{prec(2)}$ is a null space projection matrix for the lower priority task. When the equation is projected to the higher priority task space by left-multiplying by $J_1 A^{-1}$, the task space dynamics are as follows.
\begin{align}
	\ddot{x}_1 - \dot{J}_1 \dot{q} + J_1 A^{-1} N_c^T bg = J_1 A^{-1} \left(UN_c\right)^T \left( \tau_1 + N_{prec(2)}^{T} \tau_2 \right)
\end{align}
To remove the effect of the lower priority task on the higher priority, the $\tau_2$ term should fulfill the following null projection.
\begin{align}
J_1 A^{-1} \left(UN_c\right)^T N_{prec(2)}^{T} = 0
\end{align}

If there are $n_t$ control tasks to be peformed by the system, the lower j-th priority task $\tau_j$ should be zero within the higher priority task as follows.
\begin{align}
J_1 A^{-1} \left(UN_c\right)^T N_{1|prec(1)}^{T} = 0 \nonumber\\
J_2 A^{-1} \left(UN_c\right)^T N_{2|prec(2)}^{T} = 0 \nonumber\\
\cdots\nonumber\\
J_{j-1} A^{-1} \left(UN_c\right)^T N_{j|prec(j)}^{T} = 0 \label{eq:zero}
\end{align}
$N_{prec(j)}$ is a product of $N_{i|prec(i)}$ which corresponds to the null space projection of the i-th task as follows.
\begin{align}
N_{prec(j)} = \prod_{i=1}^{j-1} N_{i|prec(i)}
\end{align}
If $N_{i|prec(i)}$ is defined as follows, the condition of Eq. (\ref{eq:zero}) can be satisfied.
\begin{align}
	N_{i|prec(i)} = 
	I - \overline{J}_{i|prec(i)}^*
	J_{i|prec(i)}^*
\end{align}
where $J_{i|prec(i)}^* \triangleq J_i^* N_{prec(i-1)}$, which is the i-th task Jacobian that is projected into the null space of all the higher priority tasks.

\section{Null space projection commutation}
Assume that the first ($i-1$) task null space projections commute ($N_{i|prec(i)},\ i=1,\cdots,i-1$). Then, we check the commutation of the i-th null space projection and the k-th null space projection, where k is smaller than i as follows.
\begin{align}
N_{k|prec(k)}\, N_{i|prec(i)} &=&
	\left(I - \overline{J}_{k|prec(k)}^*
	J_{k|prec(k)}^* \right)
	\left(I - \overline{J}_{i|prec(i)}^*
	J_{i|prec(i)}^*\right)\nonumber \\
&=&
	I - \overline{J}_{k|prec(k)}^*
	J_{k|prec(k)}^* 
	- \overline{J}_{i|prec(i)}^*
	J_{i|prec(i)}^*\nonumber \\
&&	+ \overline{J}_{k|prec(k)}^*
	J_{k|prec(k)}^* 
	\overline{J}_{i|prec(i)}^*
	J_{i|prec(i)}^*
\end{align}
To investigate the commutation, we need to analyze the last term. Especially, we analyze the two matrices, $J_{k|prec(k)}^* \overline{J}_{i|prec(i)}^*$ as follows.
\begin{align}
	J_{k|prec(k)}^* 
	\overline{J}_{i|prec(i)}^*
 &= 
	J_{k|prec(k)}^* 
 \Phi^* 
 N_{prec(i)}^T J_i^{*\, T} \nonumber\\
 &=
J_{k|prec(k)}^* 
 \Phi^* 
 \prod_{l=1}^{i-1}
 N_{l|prec(l)}^T J_i^{*\, T}
 \Lambda_{2|prec(2)}^*
\end{align}
Because $\Phi^*$ and any $N_{l|prec(l)}$ commute, the following equations hold.
\begin{align}
	J_{k|prec(k)}^* 
	\overline{J}_{i|prec(i)}^*
	&=
 J_{k|prec(k)}^*
 \prod_{l=1}^{i-1}
 N_{l|prec(l)} 
 \Phi^* 
 J_i^{*\, T}
 \Lambda_{2|prec(2)}^*
\end{align}
Based on that assumption, all the task null space projections, $N_{l|prec(l)}$'s where $l < i$ commute with each other, so the term, $N_{k|prec(k)}$ in the second product can be moved next to the first matrix $J_k^*$ as follows.
\begin{align}
	J_{k|prec(k)}^* 
	\overline{J}_{i|prec(i)}^*
	&=
 J_{k|prec(k)}^* N_{k|prec(k)} 
 \prod_{l=1,\, l \ne k }^{i-1}
 N_{l|prec(l)} 
 \Phi^* 
 J_i^{*\, T}\label{eq:jjbar}
 \Lambda_{2|prec(2)}^*
\end{align}
By the definition of pseudo inverse, Eq. (\ref{eq:jjbar}) becomes zero as follows.
\begin{align}
J_{k|prec(k)}^* 
\overline{J}_{i|prec(i)}^*
&=
J_{k|prec(k)}^* N_{k|prec(k)} \cdots \nonumber\\
&=
J_{k|prec(k)}^* \left( I - \overline{J}_{k|prec(k)}^* J_{k|prec(k)}^*\right) \cdots \nonumber\\
&= 0
\end{align}
Therefore, the following equation holds.
\begin{align}
N_{k|prec(k)}\, N_{i|prec(i)} &=
	I - \overline{J}_{k|prec(k)}^*
	J_{k|prec(k)}^* 
	- \overline{J}_{i|prec(i)}^*
	J_{i|prec(i)}^* 
\end{align}
In the same manner, $N_{i|prec(i)}\, N_{k|prec(k)}$ also can be derived as follows.
\begin{align}
N_{i|prec(i)}\, N_{k|prec(k)} &=
	I 
	- \overline{J}_{i|prec(i)}^*
	J_{i|prec(i)}^* 
	- \overline{J}_{k|prec(k)}^*
	J_{k|prec(k)}^* 
\end{align}
In sum, if all the null space projections of the control tasks for which the priorities are from the first to (i-1)-th levels ($N_{i|prec(i)},\ i=1,\cdots,i-1$), then the i-th null space projection and the k-th null space projection where $k < i$ commute. Also, we can easily check the commutation of the first two tasks' null space projections as follows.
\begin{align}
N_{1|prec(1)}\, N_{2|prec(2)} &=&
	\left(I - \overline{J}_{1|prec(1)}^*
	J_{1|prec(1)}^* \right)
	\left(I - \overline{J}_{2|prec(2)}^*
	J_{i|prec(2)}^*\right) \nonumber\\
&=&
	I - \overline{J}_{1|prec(1)}^*
	J_{1|prec(1)}^* 
	- \overline{J}_{2|prec(2)}^*
	J_{2|prec(2)}^* \nonumber\\
&&	+ \overline{J}_{1|prec(1)}^*
	J_{1|prec(1)}^* 
	\overline{J}_{2|prec(2)}^*
	J_{2|prec(2)}^* 
\end{align}
Again, the last term should be analyzed to check the commutation property.
\begin{align}
&
 \overline{J}_{1|prec(1)}^*
	J_{1|prec(1)}^* 
	\Phi^* N_{1|prec(1)}^T J_2^{*\, T}
	J_{2|prec(2)}^* \nonumber\\
=&
 \overline{J}_{1|prec(1)}^*
 \underbrace{
	J_{1|prec(1)}^* 
	N_{1|prec(1)}}_{0} \Phi^* J_2^{*\, T}
	J_{2|prec(2)}^* 
\end{align}
Therefore, the last term of $N_{1|prec(1)}\, N_{2|prec(2)}$ disappears as follows.
\begin{align}
N_{1|prec(1)}\, N_{2|prec(2)} 
=
	I - \overline{J}_{1|prec(1)}^*
	J_{1|prec(1)}^* 
	- \overline{J}_{2|prec(2)}^*
	J_{2|prec(2)}^* 
\end{align}
In the same manner, $N_{2|prec(2)}\, N_{1|prec(1)}$ can be expressed as follows.
\begin{align}
N_{2|prec(2)}\, N_{1|prec(1)} 
=
	I 
	- \overline{J}_{2|prec(2)}^*
	J_{2|prec(2)}^* 
	- \overline{J}_{1|prec(1)}^*
	J_{1|prec(1)}^* ,
\end{align}
which means $N_{1|prec(1)}$ and $N_{2|prec(2)}$ commute with each other. Therefore, by mathematical induction, any $N_{i|prec(i)}$'s commutes.

%% file: chapter-appendix5.tex
\chapter{Collision probability between two objects}
\label{sec:pb}
\subsection{Probability of collision}
\label{sec:a_pc}
For easy understanding, one-dimensional collision of two objects with labels $i$ and $j$ is considered here. It is assumed that there are two one-dimensional objects corresponding to line segments with length $l_i$ and $l_j$, and their center positions are distributed with the probability density functions, $\mathbf{P}_o^i$ and $\mathbf{P}_o^j$. Then, the probability of collision between them is as follows.
\begin{align}
\begin{split}
\mathbf{p}_{ic}^{ij} =
&\int_{-\infty}^{\infty}
\int_{-\infty}^{\infty}
f_{ij}(x_i,\, x_j) \mathbf{P}_o^i (x_i) \, \mathbf{P}_o^j(x_j) dx_j\, dx_i
\\
=&\int_{-\infty}^{\infty}
\mathbf{P}_o^i (x_i) 
\left(\int_{x_i-w}^{x_i+w}
	 \mathbf{P}_o^j(x_j) dx_j\, \right)dx_i\\
=&\int_{-\infty}^{\infty}
\mathbf{P}_o^i (x_i) 
\left(\int_{-w}^{+w}
	 \mathbf{P}_o^j(x_i+x) dx\, \right)dx_i \label{eq:a_pc}
\end{split}
\end{align}
where $x \triangleq x_j - x_i$ is the relative position of $x_j$ with respect to $x_i$, and $f_{ij}$ is the predicate function which returns $1$ if the two object positions $x_i$ and $x_j$ result in a collision as follows.
\begin{align}
f_{ij}(x_i, x_j)
\bigg\{	\begin{matrix}
1 & -w \le x_i - x_j  \le  w\\
0 & \rm otherwise
\end{matrix}\label{eq:fij}
\end{align}
where  $w \triangleq l_i + l_j$. From Fubini theorem, the two integrals commute. 
\begin{flalign}
\begin{split}
=&
\int_{-w}^{+w}
\left(
\int_{-\infty}^{\infty}
\mathbf{P}_o^i (x_i) 
	 \mathbf{P}_o^j(x_i+x) 
	 dx_i\,
	 \right)
	dx \\
=&
\int_{-w}^{+w}
\left(
\int_{-\infty}^{\infty}
\mathbf{P}_o^i (x_i) 
	 \mathbf{P}_o^j\left(-(-x-x_i) \right) 
	 dx_i\,
	 \right)
	dx 
\end{split}
\end{flalign}
Defining a new function, ${\mathbf{P}_o^j}^\prime$ such that ${\mathbf{P}_o^j}^\prime(x) \triangleq \mathbf{P}_o^j(-x)$, then the probability becomes the integral of the convolution as follows.
\begin{flalign}
=&
\int_{-w}^{+w}
\Big(
\mathbf{P}_o^i(-x) \ast {\mathbf{P}_o^j}^\prime(-x) 
\Big)
dx 
\end{flalign}
In the case of a one-dimensional scenario, the shape of objects are line segments, which are symmetric, so the integration of their Minkowski sum is also symmetric.
\begin{flalign}
=&
\int_{-w}^{+w}
\Big(
\mathbf{P}_o^i(x) \ast {\mathbf{P}_o^j}^\prime(x) 
\Big)
dx 
\end{flalign}
The convolution of two normal distributions is also a normal distribution, and its mean and covariance are derived from those of the two normal distributions. So, if $\mathbf{P}_o^i$ and $\mathbf{P}_o^j$ are assumed to be normal with the following properties, 
\begin{align}
\begin{split}
\mathbf{P}_o^i(x_i) = \mathcal{N}\left( \mu_i,\ \sigma_i^2 \right) \\
\mathbf{P}_o^j(x_j) = \mathcal{N}\left( \mu_j,\ \sigma_j^2 \right)
\end{split},
\end{align}
the convolution of the two normal distribution functions becomes another normal distribution, $\mathbf{P}_{conv}$ as follows.
\begin{align}
\mathbf{P}_{conv}(x) &= \mathbf{P}_o^i(x) \ast {\mathbf{P}_o^j}^\prime(x) \nonumber\\
&= \mathcal{N} \left( \mu_i - \mu_j,\ \sigma_i^2 + \sigma_j^2 \right)
\end{align}
Therefore, the probability of the collision of the two objects in one-dimensional space is the cumulative probability density function of the new normal distribution, and the instantaneous collision probability between the two objects, $\mathbf{p}_{ic}^{ij}$, becomes a cumulative density function of a normal distribution as follows.
\begin{align}
\mathbf{p}_{ic}^{ij} &= \int_{-\omega}^{\omega} \mathbf{P}_{conv} \left( x \right) dx
\end{align}
Depending on the distance of the two normal distributions and the size of the objects, there are two ways to express the collision probability with respect to the error function, $erf$.
\begin{align}
\mathbf{p}_{ic}^{ij} =
\begin{cases}
\frac{1}{2}\left( erf\left( x^+\right) - erf\left( x^- \right) \right)
& \left| \mu_i - \mu_j \right| > \omega \\
\frac{1}{2}\left( erf\left( x^+ \right) + erf\left( - x^- \right) \right)
& \left| \mu_i - \mu_j \right| \le \omega
\end{cases} 
\end{align}
where $x^+ \triangleq \frac{ \left| \mu_i - \mu_j \right| + \omega}{\sqrt{\sigma_i^2 + \sigma_j^2}}$ and 
 $x^- \triangleq \frac{ \left| \mu_i - \mu_j \right| - \omega}{\sqrt{\sigma_i^2 + \sigma_j^2}}$.

To extend collisions from one-dimensional space to 3D space, the predicate function of Eq. (\ref{eq:fij}) needs to be extended. If it is assumed that the objects are rigid, then the predicate depends only on their relative position and we can define another predication function for the relative position, $F_{ij}$ as follows.
\begin{align}
F_{ij}(x) \triangleq f \left( 0,\, x \right)
\label{eq:Fij}
\end{align}
Because the predicate only depends on the relative position of the two objects, it has the following property.
\begin{align}
\forall x_i.\, 
F_{ij}(x) = f \left( x_i,\, x_i + x \right)
\label{eq:Fij2}
\end{align}
Also, we can define a Minkowski sum of the two objects as $\mathcal{B}$, and the integral of an arbitrary function over the region has the following property.
\begin{align}
\int_{\mathcal{B}} g(x) dx = \int_{-\infty}^{\infty} 
F_{ij}(x)\, g(x) dx
\end{align}
Then, the one-dimensional collision probability of Eq. (\ref{eq:a_pc}) is extended to 3D as follows.
\begin{align}
\begin{split}
&\int_{-\infty}^{\infty}
\mathbf{P}_o^i (x_i) 
\left(\int_{\mathcal{B}}
	 \mathbf{P}_o^j(x_i+x) dx\, \right)dx_i \\
=&
\int_{\mathcal{B}}
\Big(
\mathbf{P}_o^i(-x) \ast {\mathbf{P}_o^j}^\prime(-x) 
\Big)
dx
\end{split}
\end{align}
where $\mathbf{P}_o^i$ and $\mathbf{P}_o^j$ are joint probability density functions in 3D with the following properties.
\begin{align}
\begin{split}
\mathbf{P}_o^i(x_i) = \mathcal{N}\left( \mu_i,\ \Sigma_i \right) \\
\mathbf{P}_o^j(x_j) = \mathcal{N}\left( \mu_j,\ \Sigma_j \right)
\end{split}
\end{align}
The convolution of the normal distribution in 3D is also a joint normal distribution. So, the collision probability in 3D is also a cumulated probability function of a normal distribution as follows.
\begin{align}
\mathbf{p}_{ic}^{ij} &= \int_{\mathcal{B}} \mathbf{P}_{conv} \left( x \right) dx
\end{align}
where $\mathbf{P}_{conv}$ is the convolution of the two normal distributions consisting of another normal distribution with the following properties.
\begin{align}
\mathbf{P}_{conv} &\sim \mathcal{N} \left( 
		\mu_i - \mu_j,\ 
		\Sigma_i + \Sigma_j
\right)
\end{align}
In this study, we assume that all the objects are spheres or non-rotating ellipsoids, so the Minkowski sums are always ellipsoid. Therefore, the collision probability is derived from the integration of the normal distribution around an ellipsoid. A computationally-efficient integration algorithm is described in \cite{sheil1977}, and we exploit it to compute the collision probability.

\subsection{Conditional probability density function of collisions}
\label{sec:xf}
From Eq. (\ref{eq:a_pc}), the probability that the i-th object located at $p^i$ collides with the j-th object in a 3D space can be described as follows.
\begin{align}
\begin{split}
\mathbf{p}_{B}^{ji} (p^i) =&
\int_{-\infty}^{\infty}
f_{ij}(p^i,\, p^j) P_o^j(p^j) dp^j \\
=&
\int_{-\infty}^{\infty}
F_{ij}(p^j- p^i) P_o^j(p^j) dp^j \\
=&
\int_{-\infty}^{\infty}
F_{ij}\big( -\left(p^i - p^j\right)\big) P_o^j(p^j) dp^j \\
=&\,
F_{ij}\left( -p^i\right) \ast P_o^j(p^i) \label{eq:conv}
\end{split}
\end{align}
where $f_{ij}$ and $F_{ij}$ are the predicate functions defined in Eqs. (\ref{eq:fij}) and (\ref{eq:Fij}) .
From that probability, the corresponding random variable, $X_{c,1}$ also can be defined. We can assume that $P_2$ is a normal distribution, and $F_{ij}$ is obviously a function that indicates the Minkowski sum of the two objects. However, based on the central limit theorem that repeated convolutions of two probability density functions converges to a normal distribution, we can assume that the convolution can be approximated to a normal distribution. To derive the properties of the approximated normal distribution, we use the real mean and the variance of the convoluted function as follows.
\begin{align}
\begin{split}
{\rm E} \left( X_{c,1} \right) &= 
\frac{
\int_{-\infty}^{\infty}
p^i
P_c^1(p^i) dp^i
}
{
\int_{-\infty}^{\infty}
P_c^1(p^i) dp^i
	}
\\
&= 
\frac{
\int_{-\infty}^{\infty}
p^i
P_c^1(p^i) dp^i
}
{
\int_{-\infty}^{\infty} F_{ij}(p^i) dp^i \,
\int_{-\infty}^{\infty} P_{2}(p^i) dp^i 
}\\
&=
\frac{1}{V_\mathcal{B}}
\int_{-\infty}^{\infty}
	P_2(p^j) 
\int_{-\infty}^{\infty}
p^i
F_{ij}(p^j- p^i) 
	dp^i dp^j 
\end{split}
\end{align}
The denominator becomes the volume of the Minkowski sum of the two objects, $V_\mathcal{B}$, because the integral of the convolution in Eq. (\ref{eq:conv}) becomes the product of the integrations of the both functions. By defining a new variable $\eta \triangleq p^j - p^i$, it can be simplified further as follows.
\begin{align}
\begin{split}
=&
\frac{1}{V_\mathcal{B}} 
\int_{-\infty}^{\infty}
	P_2(p^j) 
\int_{-\infty}^{\infty}
\left(p^j - \eta\right)
F_{ij}(\eta)
	d\eta\, dp^j  \\
=&
\frac{1}{V_\mathcal{B}} 
\bigg(
\int_{-\infty}^{\infty}
p^j
P_2(p^j) 
dp^j \,
\int_{-\infty}^{\infty}
F_{ij}(\eta)
	d\eta  \\
&
- \int_{-\infty}^{\infty}
	P_2(p^j) 
	dp^j\,
\int_{-\infty}^{\infty}
\eta
F_{ij}(\eta)
	d\eta
	\bigg) \\
=&
\mu_2
- 
\frac{1}{V_\mathcal{B}} 
\int_{-\infty}^{\infty}
\eta
F_{ij}(\eta)
	d\eta
\end{split}
\end{align}
We can define that the center of the Minkowski sum $F_{ij}$ is located at the origin, then the second term becomes zero and the mean of the conditional variable is the same as the mean of the second object position. The covariance of $X_{c,1}$ can be derived from the expected value of $X_{c,1}X_{c,1}^T$ as follows.
\begin{align}
\rm cov (X_{c,1}) =& \rm E (X_{c,1}X_{c,1}^T ) - E(X_{c,1})^2
\end{align}
With the same definition of $\eta$, the expected value of $X_{c,1} X_{c,1}^T$ can be derived as follows.
\begin{align}
\begin{split}
& \rm E (X_{c,1} X_{c,1}^T ) \\ 
=&
\frac{
\int_{-\infty}^{\infty}
p^i\,{p^i}^T
P_c^1(p^i) dp^i
}
{
\int_{-\infty}^{\infty}
P_c^1(p^i) dp^i
	}\\
=&
\frac{1}{V_\mathcal{B}}
\int_{-\infty}^{\infty}
	P_2(p^j) 
\int_{-\infty}^{\infty}
p^i {p^i}^T
F_{ij}(p^j- p^i) 
	dp^i dp^j \\ 
=&
\frac{1}{V_\mathcal{B}}
\int_{-\infty}^{\infty}
	P_2(p^j) 
\int_{-\infty}^{\infty}
(p^j - \eta) (p^j - \eta)^T
F_{ij}(\eta) 
	d\eta\, dp^j \\ 
=&
\frac{1}{V_\mathcal{B}}
\int_{-\infty}^{\infty}
p^j {p^j}^T
	P_2(p^j) 
	dp^j 
\int_{-\infty}^{\infty}
F_{ij}(\eta) 
	d\eta
	\\ 
&- 2\frac{1}{V_\mathcal{B}}
\int_{-\infty}^{\infty}
p^j
	P_2(p^j) 
	dp^j
\int_{-\infty}^{\infty}
\eta^T
F_{ij}(\eta) 
	d\eta
	\\ 
& + \frac{1}{V_\mathcal{B}}
\int_{-\infty}^{\infty}
	P_2(p^j) 
	dp^j
\int_{-\infty}^{\infty}
\eta \eta^T
F_{ij}(\eta) 
	d\eta\\
=&
\left(\Sigma_2 + \mu_2 \mu_2^T\right) - \frac{2}{V_\mathcal{B}} \mu_2 
\int_{-\infty}^{\infty}
\eta^T
F_{ij}(\eta) 
	d\eta \\
& + \frac{1}{V_\mathcal{B}}\int_{-\infty}^{\infty}
\eta \eta^T
F_{ij}(\eta) 
	d\eta\\
\end{split}
\end{align}
The second term becomes zero because the center of $F_{ij}$ is located at the origin as mentioned above. Therefore, the covariance of the conditional variable, $X_{c,1}$ can be simplified as follows.
\begin{align}
\rm cov(X_{c,1}) &= 
\Sigma_2 
 + \frac{1}{V_\mathcal{B}}\int_{-\infty}^{\infty}
\eta \eta^T
F_{ij}(\eta) 
	d\eta
\end{align}
Then, we can define a new random variable, $\tilde{X}_c$, having a normal distribution and its statistic properties are the same as those of $X_c$ as follows.
\begin{align}
\tilde{X}_{c,1} \sim \mathcal{N} \left( \mu_2,\, \Sigma_2 + C_{\mathcal{B}} / V_{\mathcal{B}} \right)
\end{align}